\documentclass[10pt,twocolumn,letterpaper]{article}

\usepackage{cvpr}              

\usepackage[accsupp]{axessibility}  

\usepackage{graphicx}
\usepackage{amsmath}
\usepackage{amssymb}
\usepackage{booktabs}

\usepackage{multirow}
\usepackage{subcaption}
\usepackage{calc}

\makeatletter
\@namedef{ver@everyshi.sty}{}
\makeatother
\usepackage{tikz}
\usetikzlibrary{calc,spy}

\usepackage[title]{appendix}

\usepackage[pagebackref,breaklinks,colorlinks]{hyperref}

\usepackage[capitalize]{cleveref}
\crefname{section}{Sec.}{Secs.}
\Crefname{section}{Section}{Sections}
\Crefname{table}{Table}{Tables}
\crefname{table}{Tab.}{Tabs.}
\newcommand{\Tref}[1]{Table~\ref{#1}}
\newcommand{\eref}[1]{Eq.~(\ref{#1})}

\newcommand{\fref}[1]{Fig.~\ref{#1}}

\newcommand{\argmax}[1]{\underset{#1}{\operatorname{arg}\,\operatorname{max}}\;}

\newcommand{\symbolImageIn}{x}
\newcommand{\symbolImageGT}{y}
\newcommand{\imageIn}{\mathbf{\symbolImageIn}}
\newcommand{\imageGT}{\mathbf{\symbolImageGT}}
\newcommand{\imageOut}{\hat{\mathbf{\symbolImageGT}}}
\newcommand{\imageOutMeta}{\tilde{\mathbf{\symbolImageGT}}}
\newcommand{\reconDNN}{f}
\newcommand{\samplerDNN}{g}
\newcommand{\sample}{\mathbf{s}}
\newcommand{\sampleMeta}{\tilde{\mathbf{s}}}
\newcommand{\sampleMask}{\mathbf{m}}
\newcommand{\sampleMaskMeta}{\tilde{\mathbf{m}}}
\newcommand{\xy}{\mathbf{p}}
\newcommand{\association}{q}
\newcommand{\cell}{c}
\newcommand{\loss}{L}
\newcommand{\lossSP}{\loss_{S}}
\newcommand{\lossR}{\loss_{R}}
\newcommand{\lossO}{\loss_{O}}
\newcommand{\lossMeta}{\loss_{M}}
\newcommand{\lossTotal}{\loss_{Total}}
\newcommand{\lambdaSP}{\lambda_{S}}
\newcommand{\lambdaMeta}{\lambda_{M}}
\newcommand{\param}{\theta}

\def\cvprPaperID{4883} 
\def\confName{CVPR}
\def\confYear{2022}

\begin{document}

\title{Learning sRGB-to-Raw-RGB De-rendering with Content-Aware Metadata}

\author{Seonghyeon Nam$^1$\thanks{Work done while an intern at the Samsung AI Center -- Toronto.} \quad \quad Abhijith Punnappurath$^2$ \quad \quad Marcus A. Brubaker$^{1,2}$ \quad \quad Michael S. Brown$^{2}$\\
$^1$York University \quad \quad \quad \quad $^2$Samsung AI Center -- Toronto\\
{\tt\small  \{shnnam,mab\}@eecs.yorku.ca, \{abhijith.p,michael.b1\}@samsung.com}
}
\maketitle

\begin{abstract}
Most camera images are rendered and saved in the standard RGB (sRGB) format by the camera's hardware.  Due to the in-camera photo-finishing routines, nonlinear sRGB images are undesirable for computer vision tasks that assume a direct relationship between pixel values and scene radiance.  For such applications,  linear raw-RGB sensor images are preferred.  Saving images in their raw-RGB format is still uncommon due to the large storage requirement and lack of support by many imaging applications.  Several ``raw reconstruction'' methods have been proposed that utilize specialized metadata sampled from the raw-RGB image at capture time and embedded in the sRGB image.  This metadata is used to parameterize a mapping function to de-render the sRGB image back to its original raw-RGB format when needed.  Existing raw reconstruction methods rely on simple sampling strategies and global mapping to perform the de-rendering. This paper shows how to improve the de-rendering results by jointly learning sampling and reconstruction. Our experiments show that our learned sampling can adapt to the image content to produce better raw reconstructions than existing methods. We also describe an online fine-tuning strategy for the reconstruction network to improve results further.
\end{abstract}

%
%
\section{Introduction}
\begin{figure}
    \centering
    \setlength{\tabcolsep}{1pt}
    \begin{tabular}{cccc}
        \includegraphics[width=0.3\linewidth]{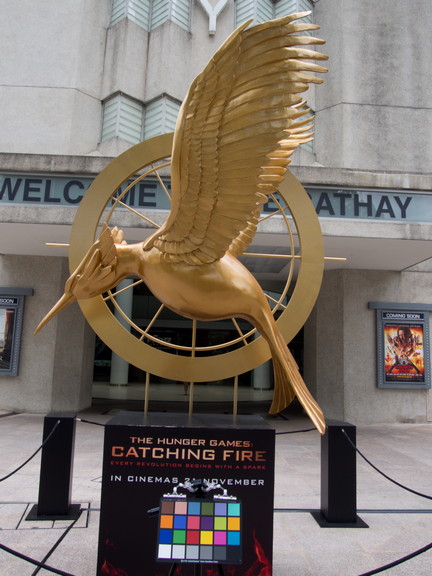} &  \includegraphics[width=0.3\linewidth]{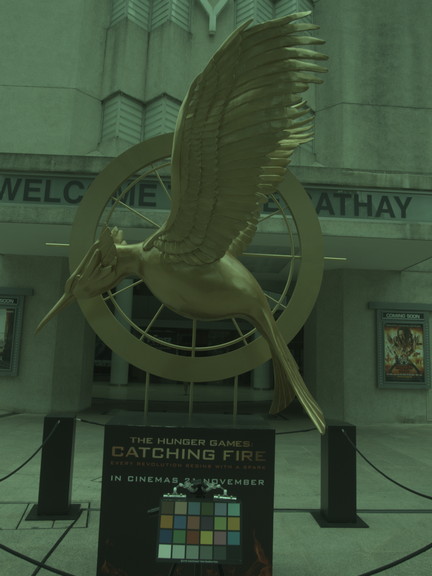} & \settototalheight{\dimen0}{\includegraphics[width=0.3\linewidth]{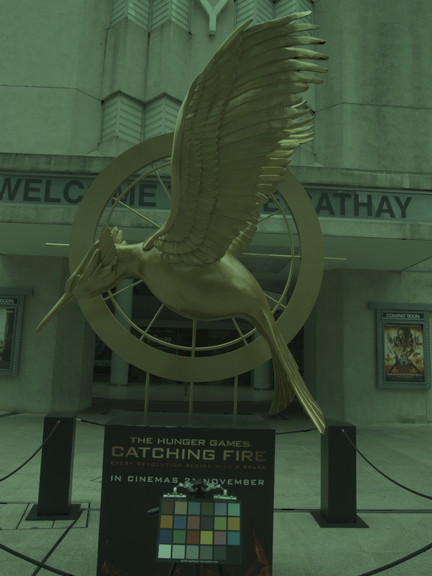}}\includegraphics[width=0.3\linewidth]{figures_arxiv/teaser/OlympusEPL6_0111_st4_wacv.jpg}\llap{{\setlength{\fboxsep}{0pt}\setlength{\fboxrule}{2pt}\fcolorbox{red}{yellow}{\includegraphics[width=0.23\linewidth]{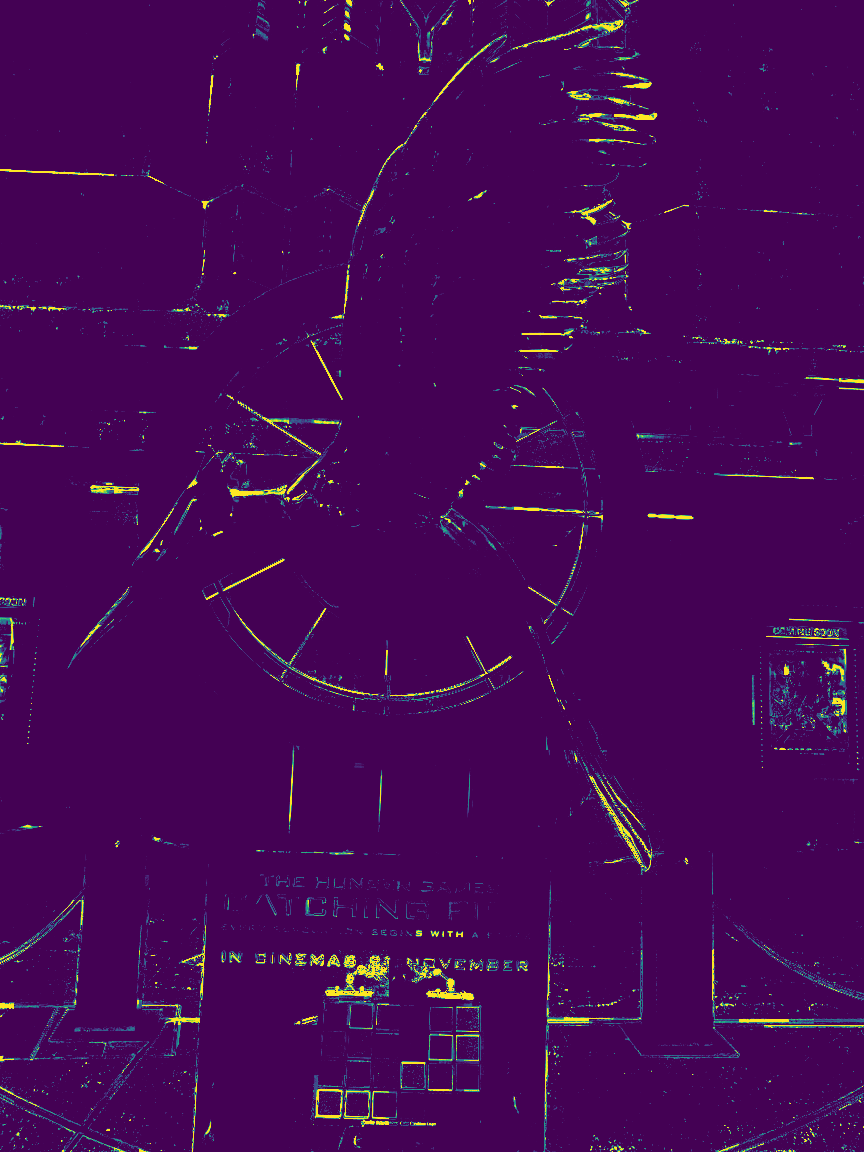}}}}\llap{\raisebox{\dimen0-7pt}{\setlength{\fboxsep}{2pt}\colorbox{white}{\scriptsize	 PSNR: 49.43dB}}} & \includegraphics[width=0.05\linewidth]{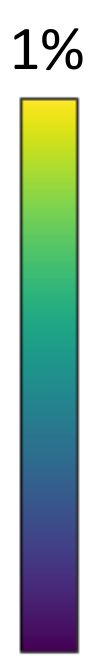} \\
        {\small sRGB (input)} & {\small Raw-RGB (GT)} & {\small SAM~\cite{wacv}} & \\
        {\setlength{\fboxsep}{0pt}\setlength{\fboxrule}{0.5pt}\fcolorbox{black}{yellow}{\includegraphics[width=0.3\linewidth]{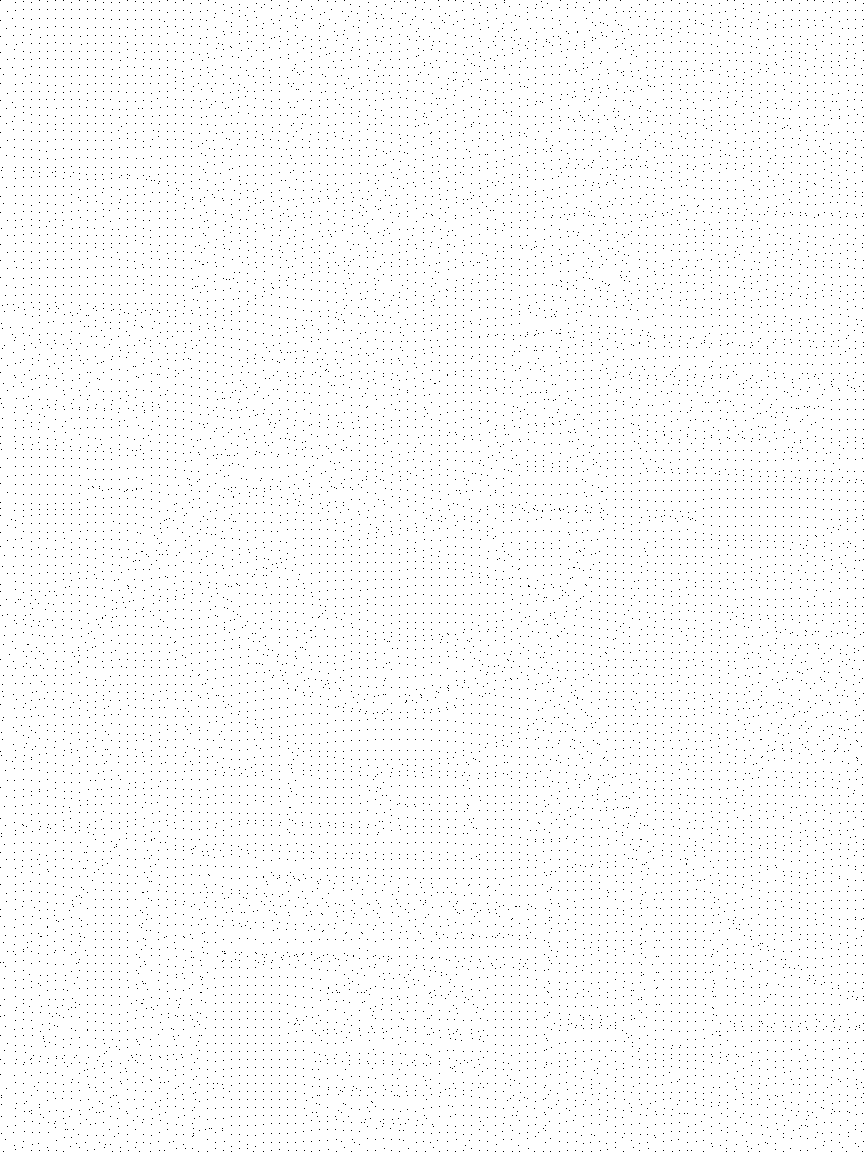}}}\llap{{\setlength{\fboxsep}{0pt}\setlength{\fboxrule}{2pt}\fcolorbox{red}{yellow}{\includegraphics[width=0.23\linewidth,clip,trim=120 470 570 450]{figures_arxiv/teaser/0000012_kmap.png}}}} & \settototalheight{\dimen0}{\includegraphics[width=0.3\linewidth]{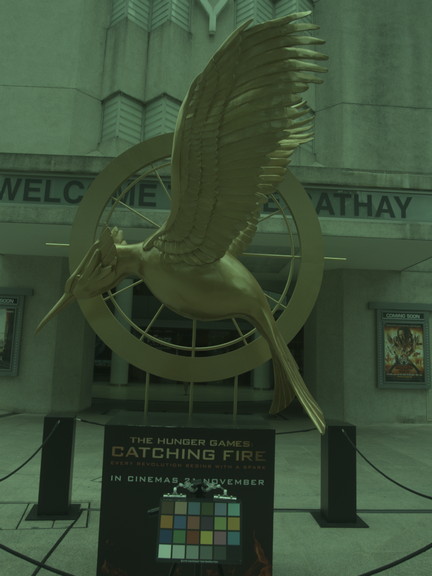}}\includegraphics[width=0.3\linewidth]{figures_arxiv/teaser/0000012_out_ours.jpg}\llap{{\setlength{\fboxsep}{0pt}\setlength{\fboxrule}{2pt}\fcolorbox{red}{yellow}{\includegraphics[width=0.23\linewidth]{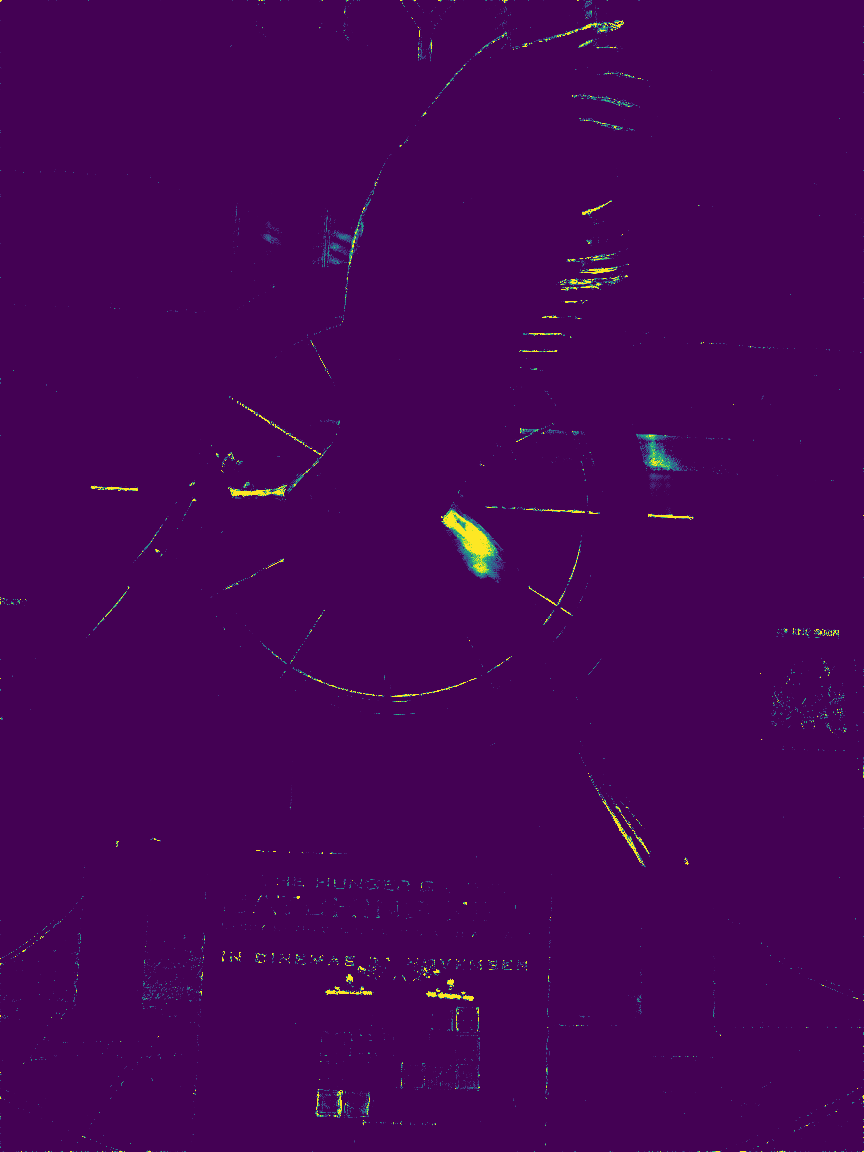}}}}\llap{\raisebox{\dimen0-7pt}{\setlength{\fboxsep}{2pt}\colorbox{white}{\scriptsize	PSNR: 57.45dB}}} & \settototalheight{\dimen0}{\includegraphics[width=0.3\linewidth]{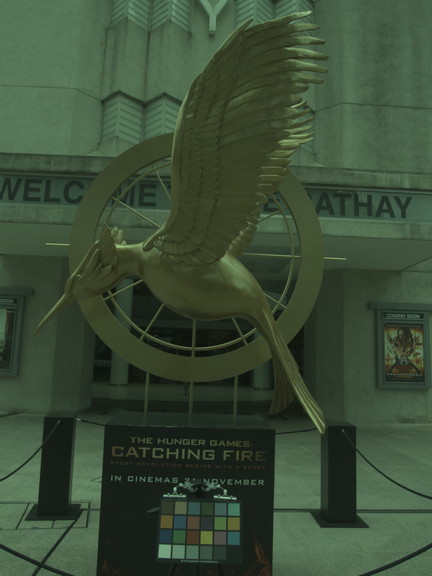}}\includegraphics[width=0.3\linewidth]{figures_arxiv/teaser/0000012_out_ours_ft.jpg}\llap{{\setlength{\fboxsep}{0pt}\setlength{\fboxrule}{2pt}\fcolorbox{red}{yellow}{\includegraphics[width=0.23\linewidth]{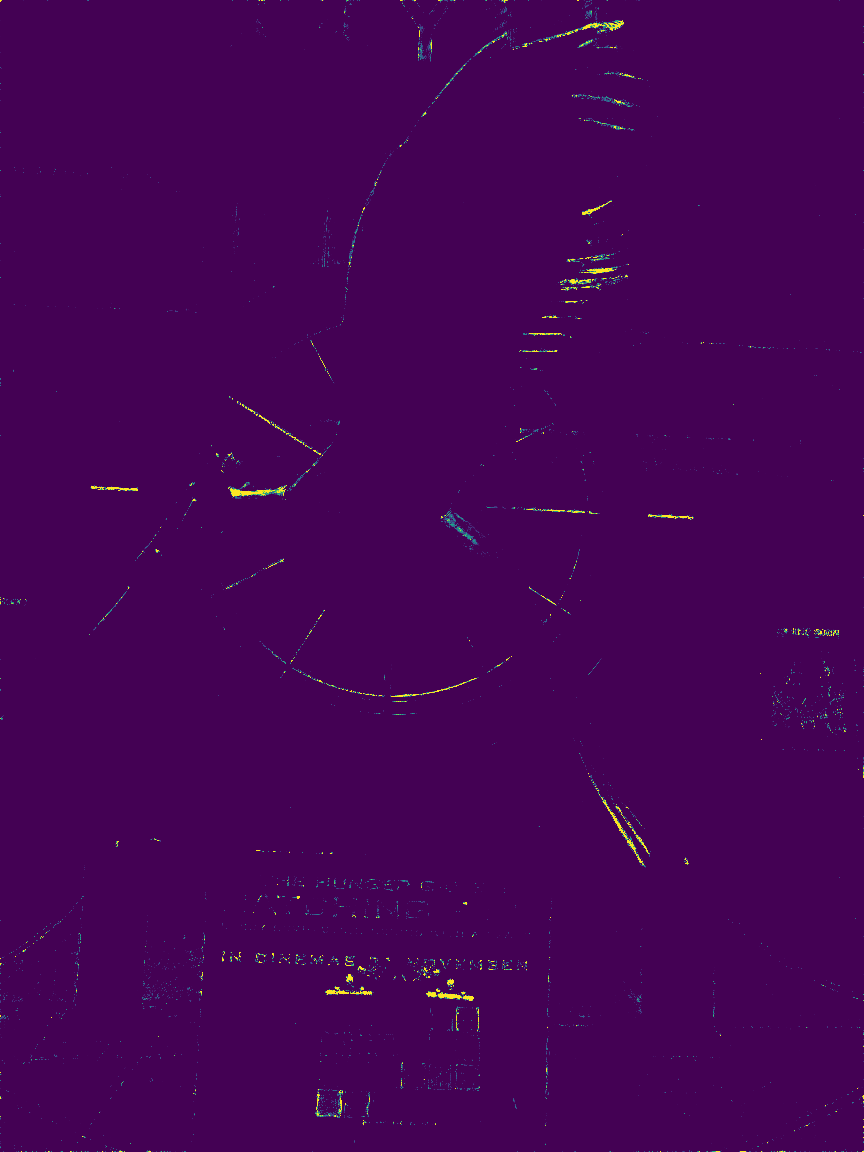}}}}\llap{\raisebox{\dimen0-7pt}{\setlength{\fboxsep}{2pt}\colorbox{white}{\scriptsize	PSNR: 57.89dB}}} & \includegraphics[width=0.05\linewidth]{figures_arxiv/colorbar.pdf}  \\
        {\small Sampling mask} & {\small Ours} & {\small Ours + fine-tuning} & \\
    \end{tabular}
    \caption{Overview of our paper. We address the problem of de-rendering an sRGB image to a raw-RGB image using a metadata saved along with the sRGB image. At capture time, we sample the raw-RGB values at the locations in a sampling mask, and save them as a metadata. When the raw-RGB image is needed, we reconstruct the full raw-RGB image from the sRGB image with the metadata. We propose an end-to-end deep learning framework to achieve it. With our approach, the reconstruction can be further improved by online fine-tuning.}
    \label{fig:teaser}
\end{figure}
For many low-level computer vision tasks, it is desirable to have access to the camera's raw-RGB sensor image whose pixel values have a linear relationship with scene radiance~\cite{nguyen2018raw, upsampling,chakrabarti2014modeling}. In addition, photo-editing operations, such as white-balance adjustment or color manipulation, are more accurate when applied on raw-RGB images~\cite{kim2012new}.  However, most images are still saved in the standard RGB (sRGB) format.  sRGB images are raw-RGB images that have been rendered by the camera's image signal processor (ISP).  The nonlinear photo-finishing routines applied by the ISP break the well-behaved relationship to scene radiance present in the original raw-RGB image.  A solution to recover the raw-RGB values is to ``de-render'' the sRGB image back to its raw-RGB format~\cite{brooks2019unprocessing}.  Among the various methods to de-render an image, the most accurate are those that collect samples from the original raw-RGB image at capture time and embed these samples inside the sRGB image as specialized metadata~\cite{rang,wacv}.  These prior methods rely on uniform sampling of the raw-RGB image and simple mapping functions to de-render from sRGB to raw-RGB.

\noindent{\textbf{Contribution.}} We propose a deep-learning framework to address the sRGB de-rendering task using metadata sampled from the raw-RGB at capture time.  In particular, we demonstrate how sampling and reconstruction can both be learned in an end-to-end framework.  Sampling is performed in a content-aware manner based on superpixel-based max-pooling and subsequently used by the reconstruction network.  In addition, the reconstruction network incorporates an online fine-tuning approach to improve the performance at inference time. An example is shown in Fig.~\ref{fig:teaser}. We demonstrate the effectiveness of our method on the raw reconstruction task using ~1.5\% of raw-RGB pixels saved in the metadata and show we can achieve state-of-the-art performance.  Additionally, we show the applicability of our method to other image recovery tasks by applying our sampling/reconstruction framework to bit-depth recovery.

\section{Related Work}
Algorithms intended to de-render sRGB images can be categorized into those that save specialized metadata along with the sRGB file at capture time and blind methods that require no additional information. We examine the metadata-based approaches in greater detail since these are more closely connected to our work. We also briefly survey algorithms for bit-depth recovery.

\noindent \textbf{Blind raw reconstruction.}~Raw reconstruction is closely connected to the problem of radiometric calibration.  Early digital cameras did not provide access to the sensor raw-RGB image.  As a result, early radiometric calibration methods did not attempt to recover the raw-RGB values accurately. Instead, they focused on linearizing the sRGB data such that the digital values had a linear relationship to scene radiance.  Radiometric calibration methods (e.g.,~\cite{debevec2008recovering,grossberg2003determining,mitsunaga1999radiometric}) employed simplistic models, such as a simple 1D response function per color channel.

As access to the raw sensor image became more commonplace, radiometric calibration was replaced by raw reconstruction, where the goal was to recover the original raw-RGB sensor data. The simple camera response function was replaced with more complex models~\cite{chakrabarti2014modeling,Chakrabarti2009empirical,kim2012new,gongcic}  to describe the various processing stages of the ISP. However, these methods are based on careful calibration procedures that must be repeated per camera and sometimes even per camera setting. Even recent deep learning methods (e.g.,~\cite{nam2017modelling,hdrisp}) are faced with similar issues in that large amounts of training data has to be captured for each camera, and the trained models are specific to that camera. Generic de-rendering methods, such as~\cite{brooks2019unprocessing,koskinen2019reverse}, that assume a standard set of ISP operations are not very accurate because they cannot model the camera-specific operations.

\noindent \textbf{Raw reconstruction with metadata.} Another strategy for de-rendering is methods that save additional metadata in the sRGB to assist with the de-rendering process.  For example, Yuan and Sun~\cite{upsampling} propose storing a small raw image as additional metadata.  The small raw image could be upsampled to full resolution at edit time using the sRGB as a guide image.  Work by Nguyen and Brown~\cite{rang,nguyen2018raw} computed and stored metadata in the form of estimated parameters used to model the typical operations performed by the ISP. These estimated parameters added a 64 KB overhead and can reconstruct the raw image from the sRGB image. However, they assume that the mapping from sRGB to raw is global and ignore important ISP operations, such as local tone mapping. Punnappurath and Brown~\cite{wacv} recently proposed to use a small set of uniformly sampled raw values as metadata. They propose a spatially aware recovery algorithm that makes their method robust to local tone mapping and other non-global ISP manipulation. However, their raw reconstruction function is based on interpolation using a 5D radial basis function, which is slow in practice.  Our method is closely related to the work in~\cite{wacv}; however, we do not restrict sampling to a uniform grid.  Moreover, we learn both sampling and reconstruction in an end-to-end manner.

\noindent \textbf{Bit-depth recovery.} The bit-depth recovery problem shares similarities to the de-rendering problem and is discussed here. In particular, the camera has applied a nonlinear process (in this case, bit quantization), and the goal is to recover the original pixel values in their full precision.   Traditional bit-depth recovery algorithms (e.g., ~\cite{CRR,Wan:2012:CA,Akira,Wan:2016:ACDC, Liu:2018:IPAD} rely on heuristics for estimating missing bit precision based on the input image's structure.  More recently, deep learning-based approaches learn how to recover missing bits~\cite{BE-CNN,Liu:2019:BECALF,Byun:2018:BitNet,BDEN,Hou,GG-DCNN,jing:2018:bde,bitmore} based on training data before and after bit quantization.  To the best of our knowledge, the use of metadata has not been explored for bit-depth recovery.  As an extension of our work, we show that our proposed de-rendering framework can be adapted to bit-depth recovery with no changes in network architecture.

\section{De-rendering Framework}
We begin with a high-level description of our framework followed by details on the sampling and reconstruction components.

\begin{figure*}
\centering
\includegraphics[width=\textwidth]{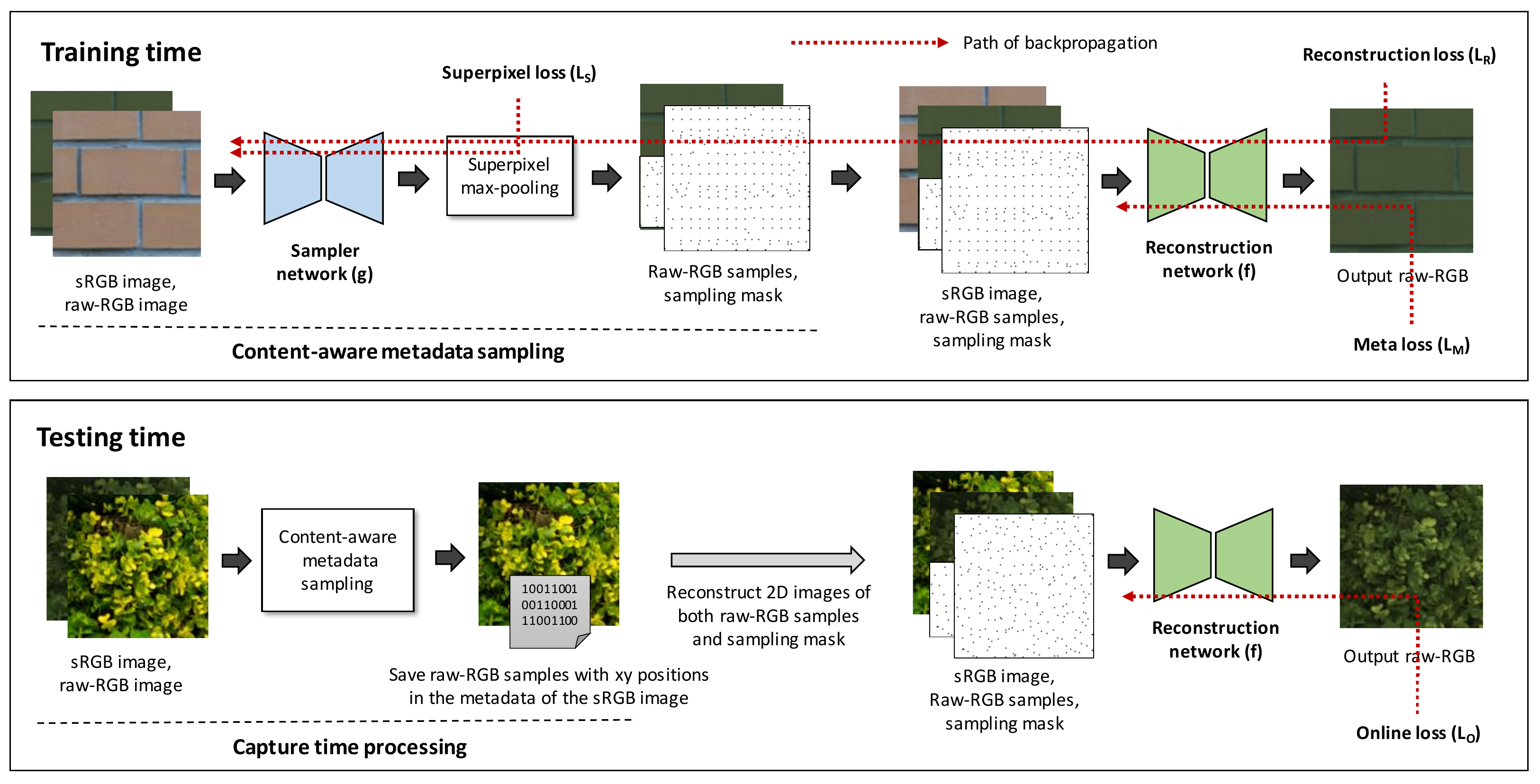}
\caption{Overview of our sRGB-to-raw de-rendering framework. At training time, we train a sampler network $\samplerDNN$ and reconstruction network $\reconDNN$ in an end-to-end manner. $\samplerDNN$ predicts a binary sampling mask used to sample raw-RGB values, while $\reconDNN$ recovers the full raw-RGB image from a full sRGB image with the sampled raw-RGB values. Particularly, our sampling mask is generated by a superpixel-based max-pooling. At testing time, $\samplerDNN$ is used to save content-aware metadata along with an sRGB image. When needed, the full raw-RGB image is reconstructed by $\reconDNN$. We fine-tune $\reconDNN$ using the sparse raw-RGB samples in the metadata on the fly to further improve the performance.}
\label{fig:overview}
\end{figure*}
Let $\imageIn$ and $\imageGT$ denote an sRGB image and a raw-RGB image, respectively.
Conventionally, the raw reconstruction problem has been mostly formulated by finding the mapping $\imageGT = \reconDNN(\imageIn)$ using only the sRGB image as input.  For metadata approaches, the mapping $\imageGT = \reconDNN(\imageIn; \sample_\symbolImageGT)$ is inferred by exploiting a small number of pixels $\sample_\symbolImageGT$ sampled from the raw-RGB image and saved in the metadata of the sRGB image.
The pixels are usually sampled by a pre-defined method, such as uniform sampling applied to all images globally.
Our goal is to learn both the sRGB-to-raw-RGB mapping and the sampling function, which is formally described as $\hat{\imageGT} = \reconDNN(\imageIn; \hat{\sample}_\symbolImageGT = \samplerDNN(\imageIn, \imageGT))$, where $\samplerDNN(\imageIn, \imageGT)$ is a learnable sampling function.

\begin{figure}
    \centering
    \includegraphics[width=0.98\linewidth]{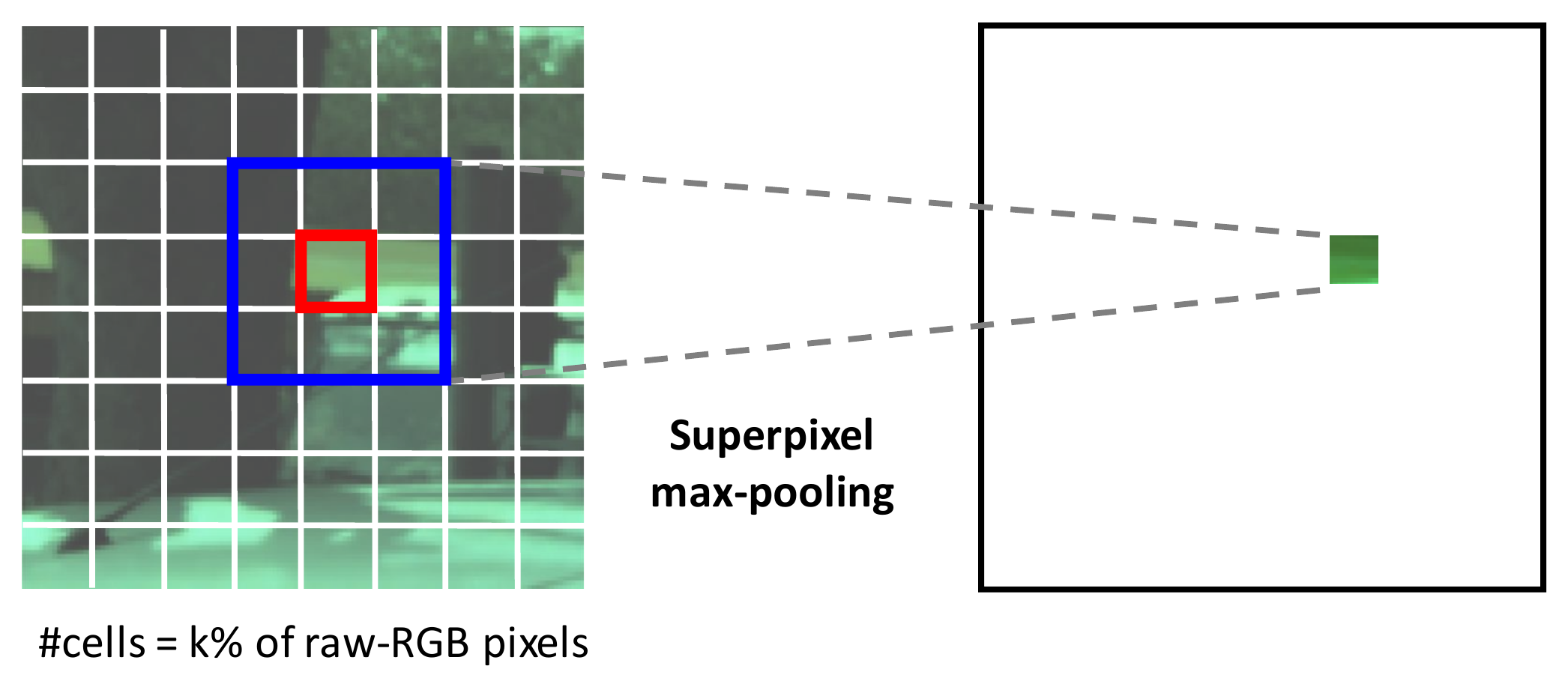}
    \caption{Illustration of the superpixel max-pooling. For all pixels in the blue box, their association to the red cell is learned by the superpixel loss. The superpixel max-pooling is to sample the highest association score among the pixels in the blue box. Even though we sample a pixel from a regular box, the sampling heavily relies on the association score learned by the superpixel loss.}
    \label{fig:method_superpixel}
\end{figure}
\fref{fig:overview} shows an overview of our framework.
We model two functions $\reconDNN$ and $\samplerDNN$ as U-Net-based~\cite{Ronneberger:2015:UNet} deep neural networks, and train them in an end-to-end manner.
At training time, we first sample $k\%$ of pixels from a raw-RGB image.
Specifically, the sampler network $\samplerDNN$ takes both a raw-RGB and an sRGB image as input and predicts a binary sample map $\sample$ in which sampled pixels are assigned to $1$.
To effectively compute samples, the sampler network also learns to divide the raw-RGB image into superpixels and selects samples by per-superpixel max-pooling.
In the reconstruction network, both the sRGB image and sampled raw-RGB pixels with their corresponding mask are fed into the network to recover the full raw-RGB image.
The two networks are jointly trained by minimizing the pixel-wise distance between the output raw-RGB image and ground truth.

The inference time scenario is composed of two stages.  At capture time, we use the sampler network $\samplerDNN$ to sample the raw-RGB image. These samples are stored as metadata in the sRGB image as comments.
To save memory, we only save the sampled RGB values with their pixel positions in the metadata.  When needed, the raw-RGB image is reconstructed by the reconstruction network $\reconDNN$ with the saved metadata in the sRGB image.
Even though the pre-trained reconstruction network produces high-quality raw-RGB images, it can be fine-tuned by the sparse raw-RGB samples to further improve the performance on the test data.

\subsection{Content-Aware Metadata Sampling}
The goal of our content-aware metadata sampling is to find the optimal raw-RGB samples based on the content of the image.  To this end, our key idea is to divide a raw-RGB image into superpixels and select the best pixel in each superpixel as a sample for metadata.  We found that it is beneficial for the reconstruction network to choose the raw-RGB samples that are well distributed over the space of a raw-RGB image.  As a result, we split the xy-RGB space of raw-RGB images into multiple subspaces using a superpixel segmentation and collect the representative pixels.

Specifically, inspired by~\cite{Yang:2020:SuperpixelFCN}, our sampler network first computes superpixels directly from the input.  As shown in~\fref{fig:method_superpixel}, we divide a raw-RGB image into uniform grid cells.  The network predicts the association scores $\association_\cell(\xy)$ for each pixel $\xy$ that indicate how likely a pixel $\xy$ belongs to a grid cell $\cell$.  For computational efficiency, only the nine neighboring cells in the blue box are considered for computing the association to the red-highlighted cell.
The association map is learned by optimizing the following superpixel segmentation loss:
\begin{equation}
\begin{split}
   \lossSP =~&\alpha \sum_\xy \lVert \imageIn(\xy) - \hat{\imageIn}(\xy) \rVert^2_2\\
   &+ (1 - \alpha) \sum_\xy \lVert \imageGT(\xy) - \hat{\imageGT}(\xy) \rVert^2_2\\
   &+ \frac{m^2}{S^2} \sum_\xy \lVert \xy - \hat{\xy} \rVert^2_2,
\end{split}
\label{eq:loss_sp}
\end{equation}
where $\hat{\imageIn}(\xy)$ is a reconstructed RGB value of $\imageIn(\xy)$ computed by the following equations:
\begin{equation}
   \mathbf{u}_\cell = \frac{\sum_{\xy \in \mathcal{N}_\cell} \imageIn(\xy) \cdot \association_\cell(\xy)}{\sum_{\xy \in \mathcal{N}_\cell} \association_\cell(\xy)},~\hat{\imageIn}(\xy) = \sum_\cell \mathbf{u}_\cell \cdot \association_\cell(\xy).
\end{equation}
In the equation, $\mathbf{u}_\cell$ is the feature vector of the center of the superpixel $\cell$ and $\mathcal{N}_\cell$ is a set of all pixels in the nine surrounding cells of a cell $\cell$.
Similar to SLIC~\cite{Achanta:2012:SLIC}, the loss in~\eref{eq:loss_sp} enforces that the pixels in each superpixel are not deviated much from the center $\mathbf{u}_\cell$.
$\hat{\imageGT}(\xy)$ and $\hat{\xy}$ are computed by the same equations.
In~\eref{eq:loss_sp}, $m$ and $S$ are the weight parameters in~\cite{Yang:2020:SuperpixelFCN}.
Unlike the loss in~\cite{Yang:2020:SuperpixelFCN} that uses semantic segmentation labels, our loss optimizes the RGB color distance in both the sRGB and raw-RGB images.  Our loss forces the network to consider how to sample the sRGB and raw-RGB images jointly.
We add a hyperparameter $\alpha$ to balance the sRGB and raw-RGB terms.

To select $k\%$ of samples, we set the number of uniform grid cells to $k\%$ of the number of raw-RGB pixels.
We then choose the representative pixel for each grid cell that provides the maximum $\association$ in the pixels of the nine neighborhood cells.
The per-cell max-pooling is formulated as
\begin{equation}
   \xy^{*}_\cell = \argmax{\xy \in \mathcal{N}_\cell} \association_\cell(\xy).
\end{equation}
We compute a binary sampling mask $\sampleMask$ to feed the samples to the reconstruction network in the form of a 2D image, which is formally described as
\begin{equation}
   \sampleMask(\xy) =
   \begin{cases}
      1, & \text{if } \xy \in \{\xy^{*}_0, \xy^{*}_1, ..., \xy^{*}_\cell\}\\
      0, & \text{otherwise}.
  \end{cases}
\end{equation}
The sample map $\sample_\symbolImageGT$ is simply computed by multiplying $\sampleMask$ and $\imageGT$, which is described as $\sample_\symbolImageGT = \sampleMask \otimes \imageGT$ where $\otimes$ is a Hadamard product.
The gradient is back-propagated only through $\association(\xy)$ such that $\sampleMask(\xy) = 1$ using a straight-through estimator~\cite{Bengio:2013:StraightThrough}.
Note that our method does not guarantee that the number of samples equals the $k\%$ of the number of pixels because multiple cells may choose the same pixel after max-pooling.

\subsection{sRGB-to-Raw-RGB De-rendering Using Metadata}
The sRGB-to-raw-RGB de-rendering task is formulated as an image-to-image transform learned by the reconstruction network $\reconDNN$.  To exploit the sparse raw-RGB samples in the metadata for reconstruction, we concatenate $\imageIn$, $\sample_\symbolImageGT$, and $\sampleMask$ followed by feeding them to $\reconDNN$.
With these inputs, the network can infer the sRGB-to-raw-RGB mapping for all pixels based on the sparse sRGB-raw-RGB pairs and the full sRGB image.
Both the sampler and reconstruction networks are jointly trained by the pixel distance loss described as
\begin{equation}
   \lossR = \sum_\xy \lVert \imageOut(\xy) - \imageGT(\xy) \rVert_1,
\end{equation}
where $\hat{\imageGT} = \reconDNN(\imageIn, \sample_\symbolImageGT, \sampleMask)$.

\subsection{Online Fine-Tuning at Inference Time}
\begin{figure}
    \centering
    \includegraphics[width=\linewidth]{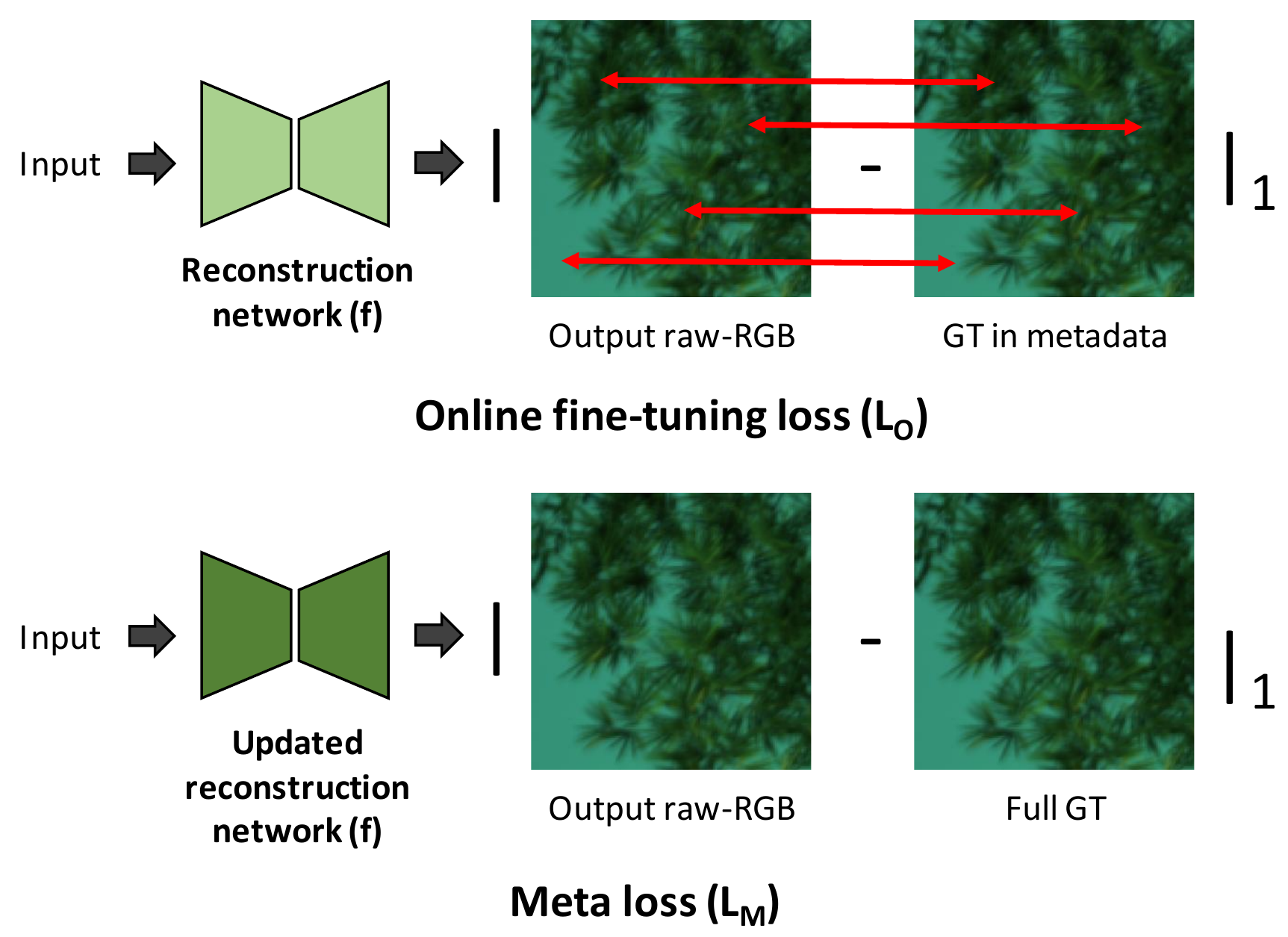}
    \caption{Illustration of the online fine-tuning loss $\lossO$ and meta loss $\lossMeta$. The $\lossO$ is computed only at the positions that have ground-truth raw-RGB samples in metadata. To compute $\lossMeta$, $\reconDNN$ is first updated by $\lossO$, and then the loss for the full output image is computed.}
    \label{fig:method_online_meta}
\end{figure}
Another benefit of storing raw-RGB samples in the metadata is that they can be used to fine-tune the pre-trained reconstruction network on the fly for a test image to improve the performance further.
As shown in~\fref{fig:overview} and~\fref{fig:method_online_meta}, we minimize the pixel-wise distance loss only for the pixels that have their corresponding ground truth at inference time, which is formally described as
\begin{equation}
   \lossO = \sum_\xy \sampleMask(\xy) \cdot \lVert \imageOut(\xy) - \imageGT(\xy) \rVert_1.
   \label{eq:online_loss}
\end{equation}

\paragraph{Meta-learning for optimizing fine-tuning.}
Our reconstruction network can be further optimized at training time to be generalized to fine-tuning on test-time examples.  Intuitively, we expect that the overall error is minimized after the network is fine-tuned to fit the sparse samples.  We can encourage our network to be receptive to fine-tuning by adding another loss term at training time.  Specifically, we first compute the reconstruction output using the metadata samples $\sampleMeta$ and $\sampleMaskMeta$, which is described as $\imageOutMeta_\param = \reconDNN_\param(\imageIn, \sampleMeta_\symbolImageGT, \sampleMaskMeta)$, where $\param$ is the parameters of $\reconDNN$.
Then, we update $\lossO$ in~\eref{eq:online_loss} by several gradient descent steps using the following update rule: $\param^{'} = \param - \beta \nabla_\param \lossO(\sampleMaskMeta, \imageOutMeta_\param)$, where $\beta$ is a learning rate.
We finally compute the pixel distance loss between the output of the updated model and ground truth for all pixels as shown in~\fref{fig:method_online_meta}, which is formally described as
\begin{equation}
   \lossMeta = \sum_\xy \lVert \imageOutMeta_{\param^{'}}(\xy) - \imageGT(\xy) \rVert_1.
\end{equation}
We have found that feeding the learned samples from the sampler network to this loss degrades performance.
The reason is that the goal of two losses conflicts: the main loss $\lossR$ is optimized to overfit training batches while $\lossMeta$ seeks generalization.  Therefore, we use different data to compute the loss.
Specifically, we use random samples for $\sampleMeta$ and $\sampleMaskMeta$.
This strategy alleviates the performance degradation when optimizing both losses and forces the network to cope with a variety of sampling maps, which is also helpful for generalization.
Our formulation shares a similar spirit with MAML~\cite{Finn:2017:MAML}, a meta-learning approach, in that both are to find generalizable parameters that result in better performance at fine-tuning.
Our goal is to improve the overall image quality even after overfitting the network to the $k\%$ of pixels.
We use FOMAML~\cite{Finn:2017:MAML}, a first-order approximation of gradients, to compute the gradients of the meta loss.

\subsection{Training Objective}
The final training objective at training time is composed of the reconstruction loss, superpixel loss, and meta loss, which is described as
\begin{equation}
   \lossTotal = \lossR + \lambdaSP \lossSP + \lambdaMeta \lossMeta,
\end{equation}
where $\lambdaSP$ and $\lambdaMeta$ are hyperparameters.
At testing time, the online fine-tuning is performed by optimizing the online optimization loss in~\eref{eq:online_loss}.

~~

\section{Experiments}
\label{sec:expts}
\subsection{Experimental Settings}
\noindent \textbf{Dataset.}
To test the effectiveness of our method, we use the NUS dataset~\cite{Cheng:2014:NUS}, which contains raw images from several different cameras. For our experiments, we use three cameras---Samsung NX2000, Olympus E-PL6, and Sony SLT-A57---containing 202, 208, and 268 raw images, respectively. We demosaic the raw Bayer image using standard bi-linear interpolation to obtain a 3-channel raw-RGB image. We then process the raw-RGB image using a software ISP emulator~\cite{Karaimer:2016:SoftwareISP} to render the corresponding sRGB image. This rendering mimics the photo-finishing applied by the camera. We randomly split images from each camera into training, validation, and test sets. In addition, we crop all images into overlapping 128$\times$128 patches.

In addition to the raw images, the NUS dataset also contains sRGB-JPEG images rendered by each individual camera's ISP. We also performed experiments where we used these sRGB images instead of the software ISP emulator~\cite{Karaimer:2016:SoftwareISP}. These results are reported in Section~\ref{sec:nus_jpg}.

\noindent \textbf{Baselines.}
We compare our method with two metadata-based raw reconstruction methods: RIR~\cite{rang} and SAM~\cite{wacv}.
The RIR  method stores the parameters of global operations of an ISP as metadata. The SAM method, which is most similar to our approach, saves uniformly sampled raw-RGB values along with an sRGB image.
Since the source codes of the methods are not publicly available, we implement them to reproduce results.
For SAM, we use the same sampling ratio used in our method.

\noindent \textbf{Implementation details.}
As a backbone of both the sampler and reconstruction networks, we use a U-Net~\cite{Ronneberger:2015:UNet} architecture.
We train our networks using the Adam optimizer~\cite{Kingma:2014:Adam} with a learning rate of 0.001 and a batch size of 128 for 120 epochs.
In the superpixel loss, we use 0.2 and 10 for $\alpha$ and $m$, respectively.
In the meta loss, we use five gradient descent steps for the inner update with a learning rate of 0.001.
We also set $\lambdaSP$ and $\lambdaMeta$ to 0.0001 and 0.01 for all cameras except for the Sony; we use $\lambdaMeta$ of 0.001 for the Sony set.
At testing time, we fine-tune the network for ten iterations with a learning rate of 0.0001.
In all experiments, we sample ~1.5\% of pixels from a raw-RGB image, which are 256 pixels in a 128$\times$128 patch. We train a model per camera since raw images are in a sensor-dependent color space. Our code and pre-trained models are available at \url{https://github.com/SamsungLabs/content-aware-metadata}.

\subsection{Experimental Results}
\begin{table*}
\begin{center}
\setlength{\tabcolsep}{15pt}
\begin{tabular}{c|c|c|c|c|c|c|c}
\toprule
\multirow{2}{*}{Method} & \multirow{2}{*}{Fine-tuning} & \multicolumn{2}{c|}{Samsung NX2000} & \multicolumn{2}{c|}{Olympus E-PL6} & \multicolumn{2}{c}{Sony SLT-A57} \\
\cline{3-8}
& & PSNR & SSIM & PSNR & SSIM & PSNR & SSIM \\
\hline
RIR~\cite{rang} & N/A & 45.66 & 0.9939 & 48.42 & 0.9924 & 51.26 & 0.9982 \\
SAM~\cite{wacv} & N/A & 47.03 & 0.9962 & 49.35 & 0.9978 & 50.44 & 0.9982 \\
\hline
Ours & No & 48.08 & 0.9968 & 50.71 & 0.9975 & 50.49 & 0.9973 \\
Ours & Yes & \textbf{49.57} & \textbf{0.9975} & \textbf{51.54} & \textbf{0.9980} & \textbf{53.11} & \textbf{0.9985} \\
\bottomrule
\end{tabular}
\end{center}
\caption{Quantitative evaluation on raw reconstruction.}
\label{table:quant_raw}
\end{table*}
\begin{figure*}
    \centering
    \setlength{\tabcolsep}{1pt}
    \begin{tabular}{ccccccc}
        \includegraphics[width=0.15\linewidth]{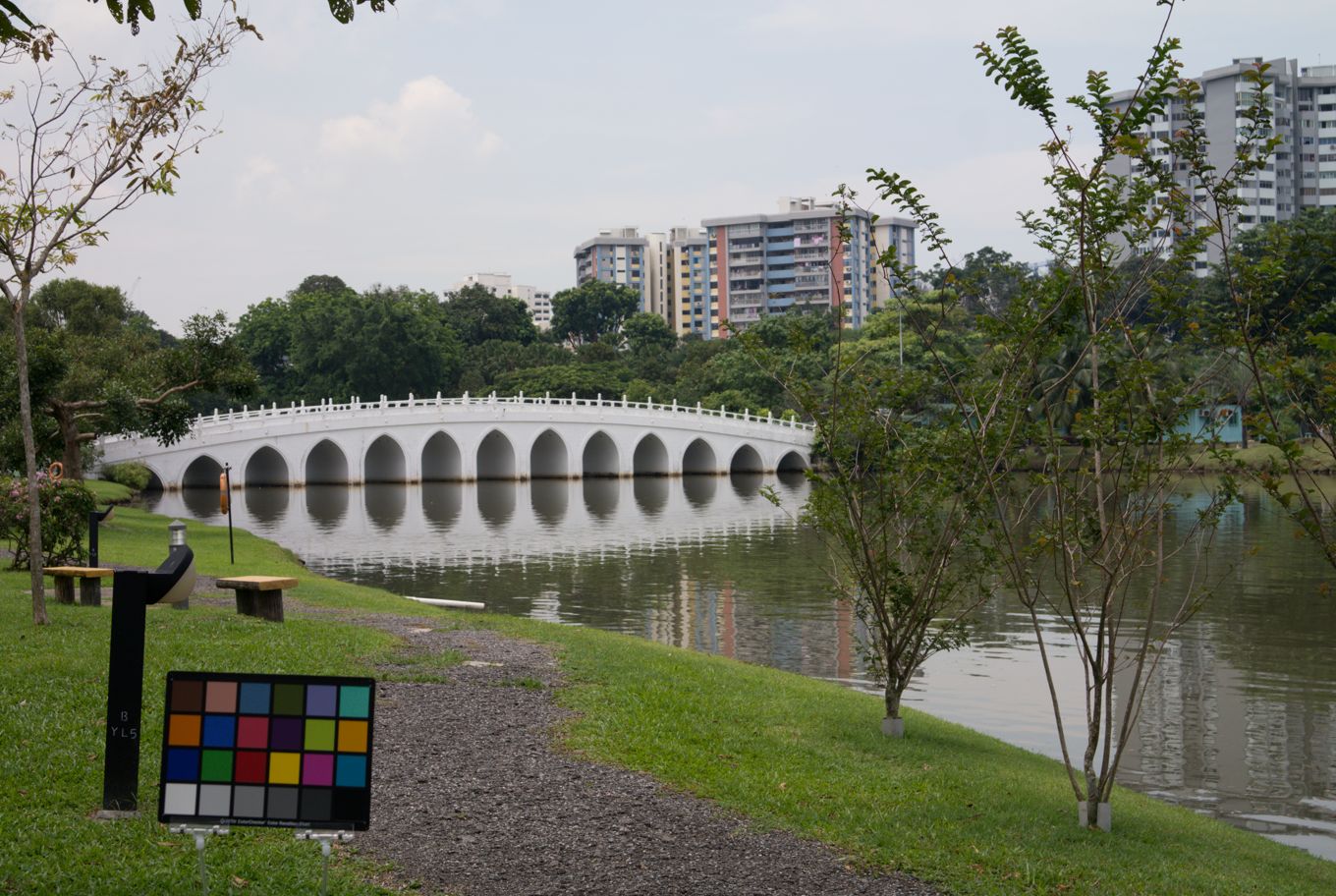} & \includegraphics[width=0.15\linewidth]{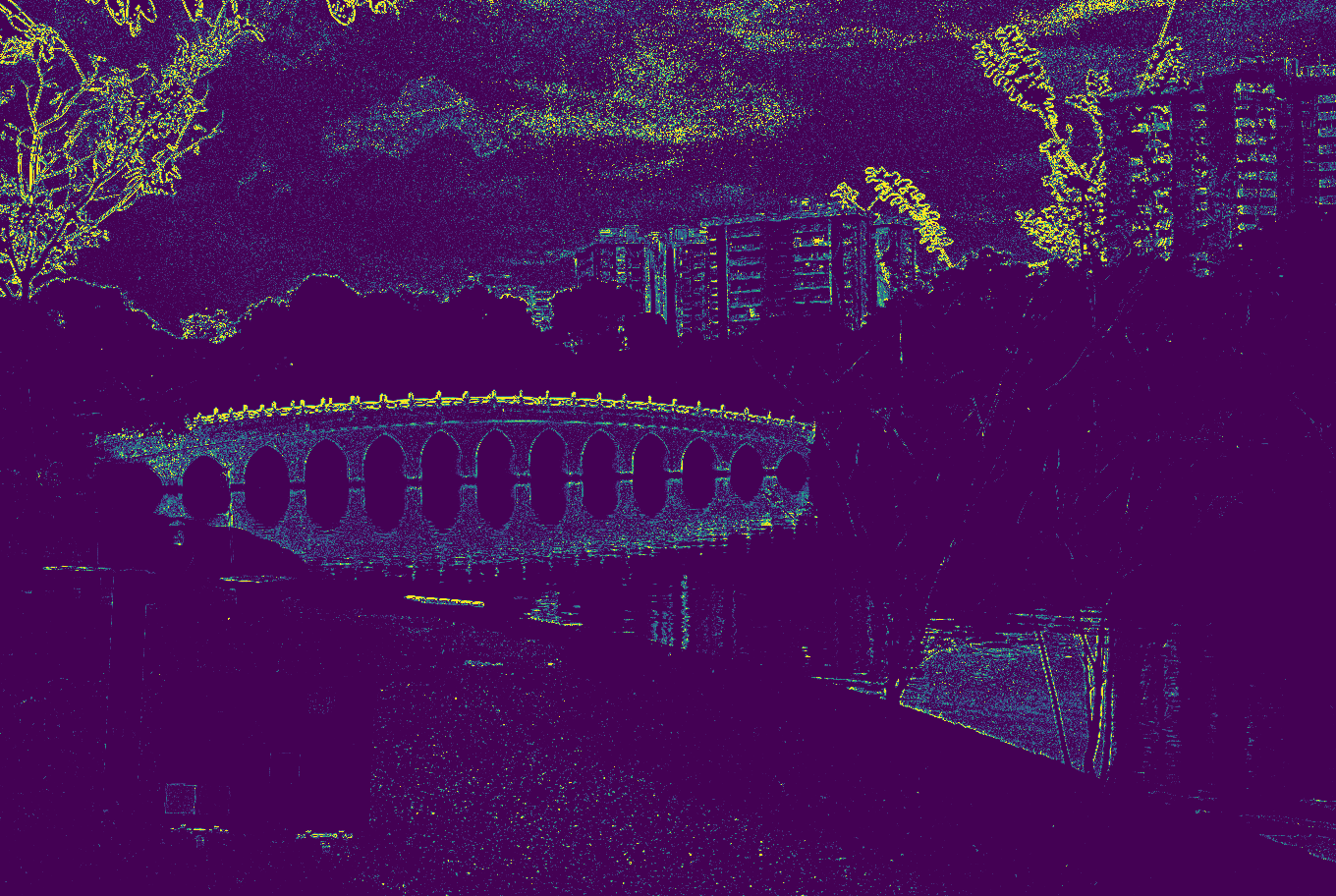}\llap{{\setlength{\fboxsep}{0pt}\setlength{\fboxrule}{2pt}\fcolorbox{red}{yellow}{\includegraphics[width=0.11\linewidth,clip,trim=400 400 400 150]{figures_arxiv/comparison/samsung/SamsungNX2000_0010_err_rang.png}}}} & \includegraphics[width=0.15\linewidth]{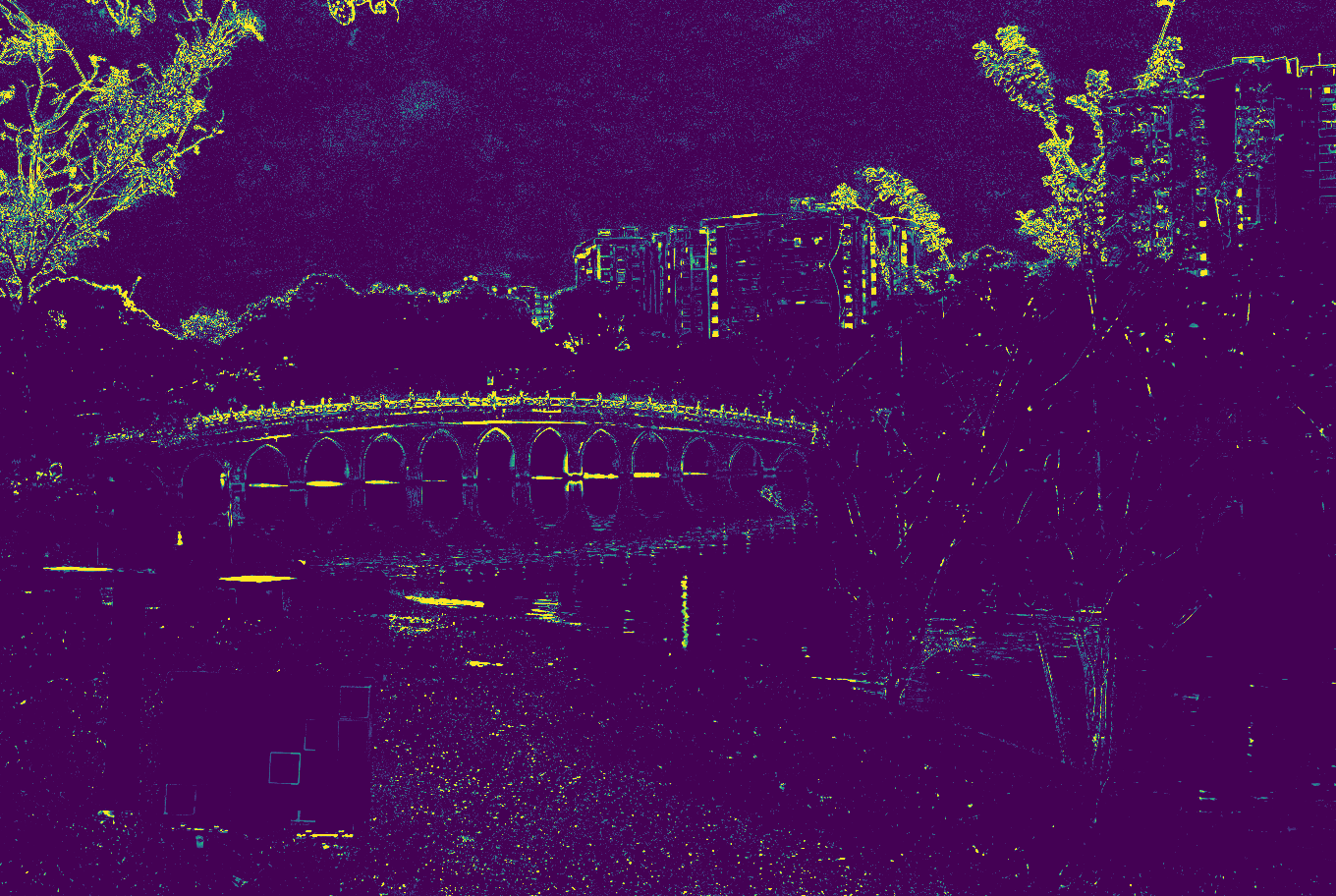}\llap{{\setlength{\fboxsep}{0pt}\setlength{\fboxrule}{2pt}\fcolorbox{red}{yellow}{\includegraphics[width=0.11\linewidth,clip,trim=400 400 400 150]{figures_arxiv/comparison/samsung/SamsungNX2000_0010_err_wacv.png}}}} & \includegraphics[width=0.15\linewidth]{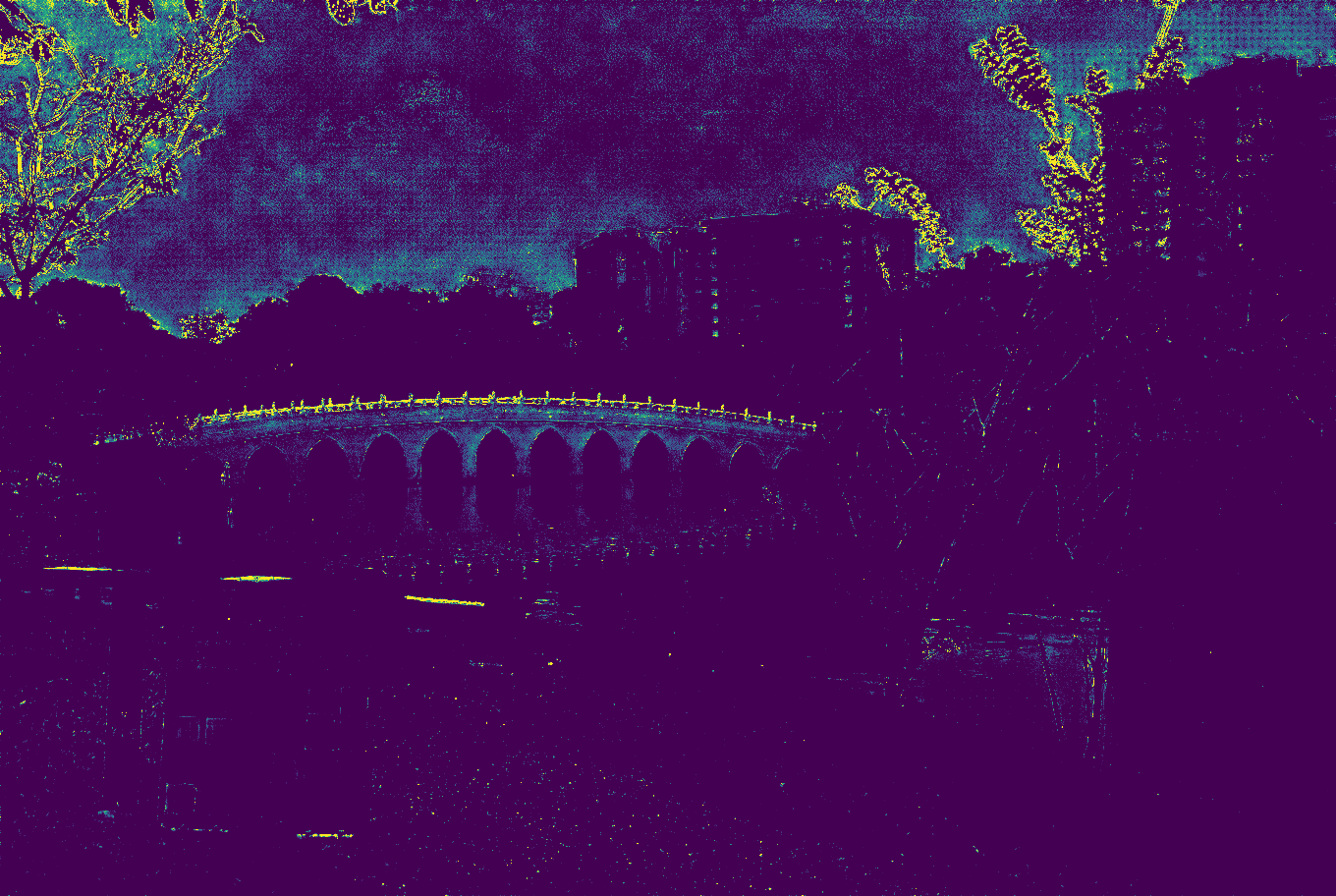}\llap{{\setlength{\fboxsep}{0pt}\setlength{\fboxrule}{2pt}\fcolorbox{red}{yellow}{\includegraphics[width=0.11\linewidth,clip,trim=400 400 400 150]{figures_arxiv/comparison/samsung/0000001_err_ours.png}}}} & \includegraphics[width=0.15\linewidth]{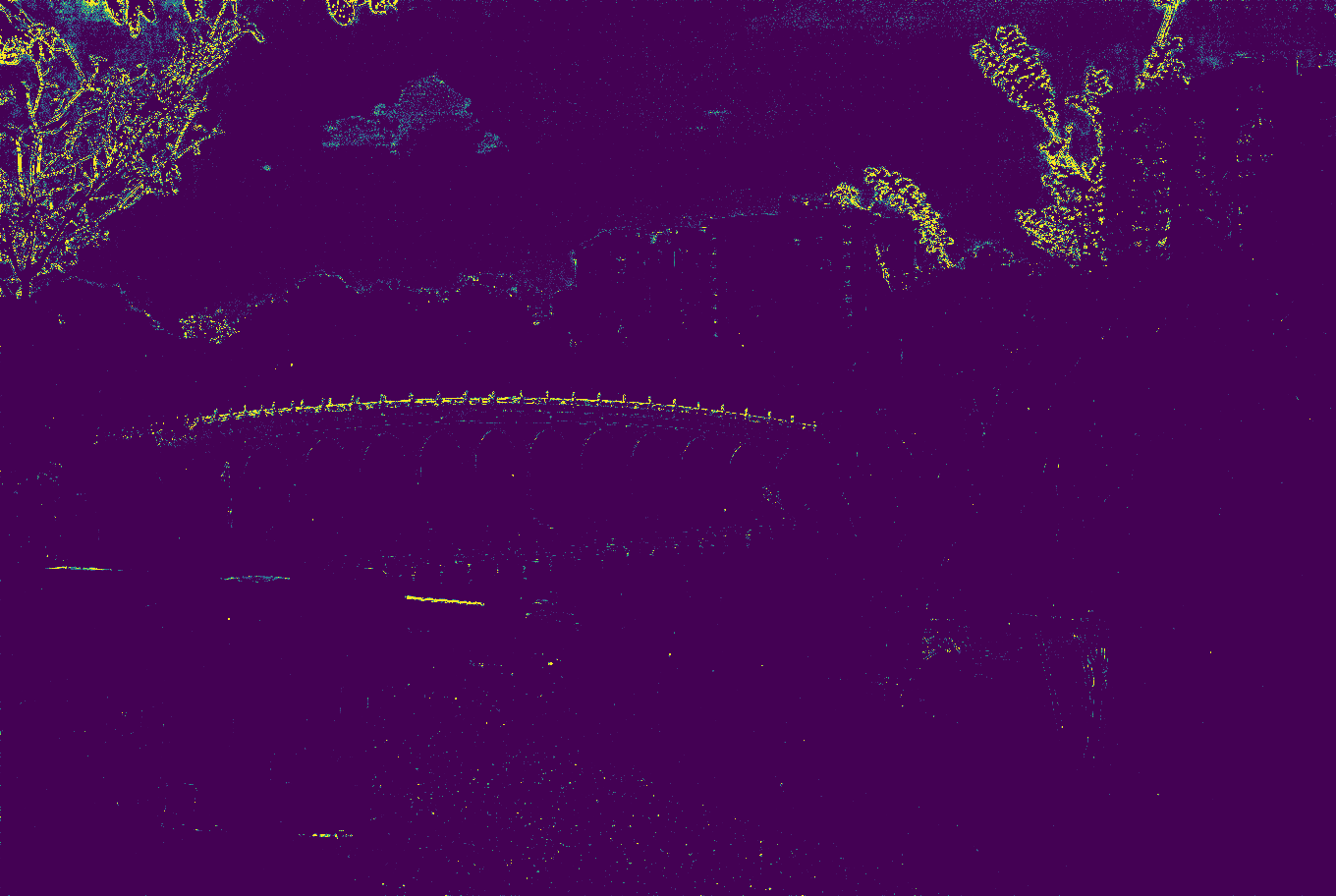}\llap{{\setlength{\fboxsep}{0pt}\setlength{\fboxrule}{2pt}\fcolorbox{red}{yellow}{\includegraphics[width=0.11\linewidth,clip,trim=400 400 400 150]{figures_arxiv/comparison/samsung/0000001_err_ours_ft.png}}}} & \includegraphics[width=0.016\linewidth]{figures_arxiv/colorbar.pdf} & \includegraphics[width=0.15\linewidth]{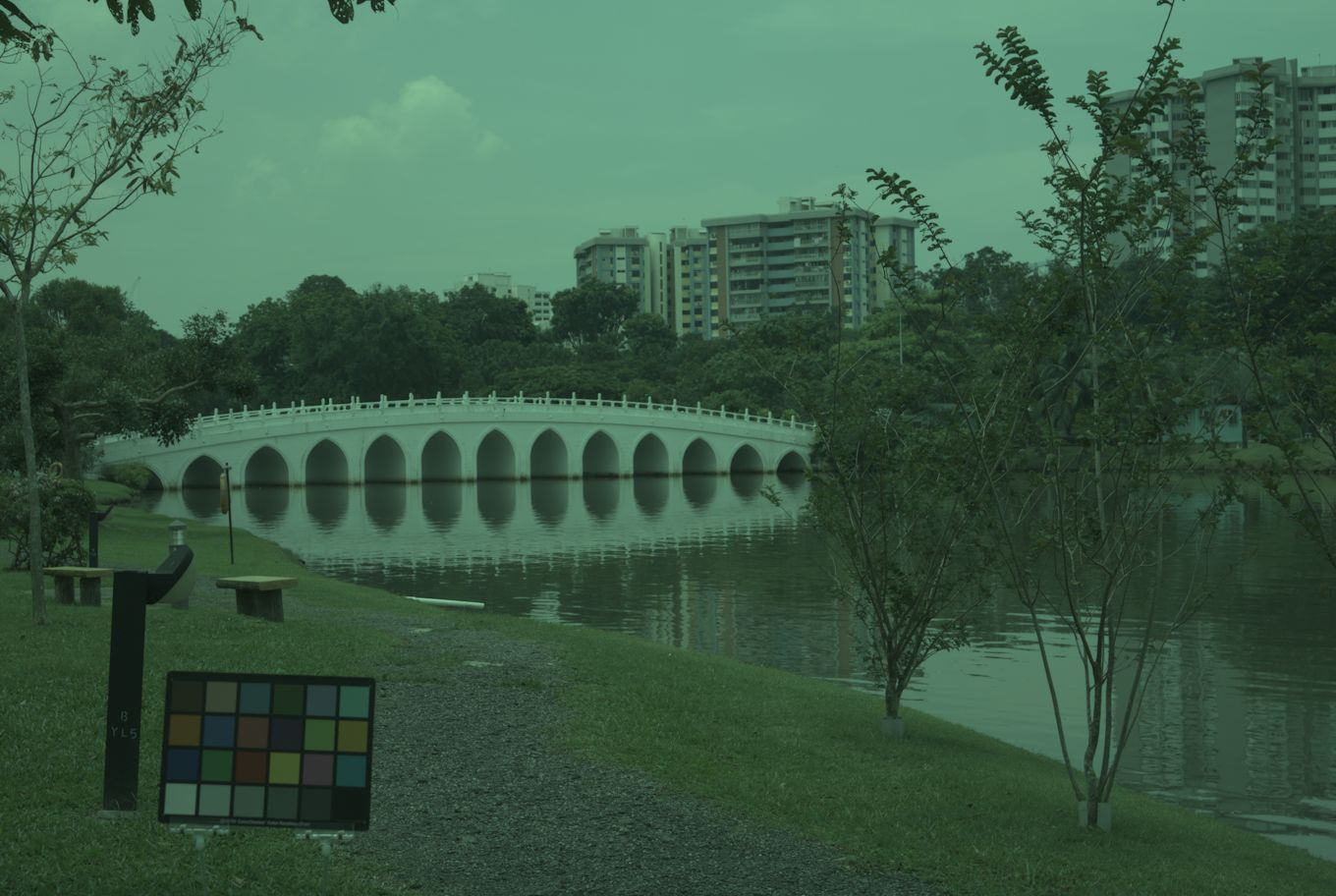} \\

        \includegraphics[width=0.15\linewidth]{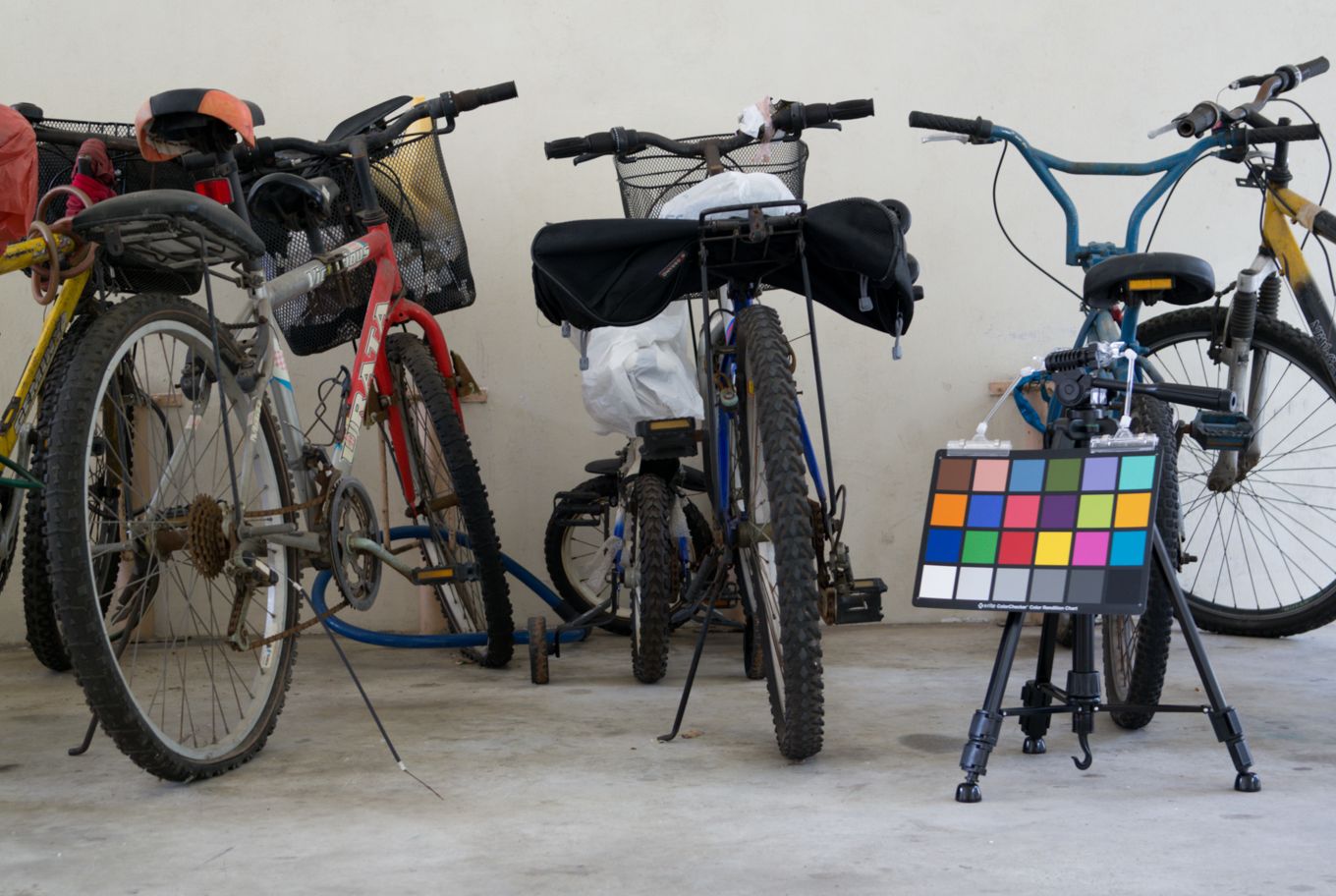} & \includegraphics[width=0.15\linewidth]{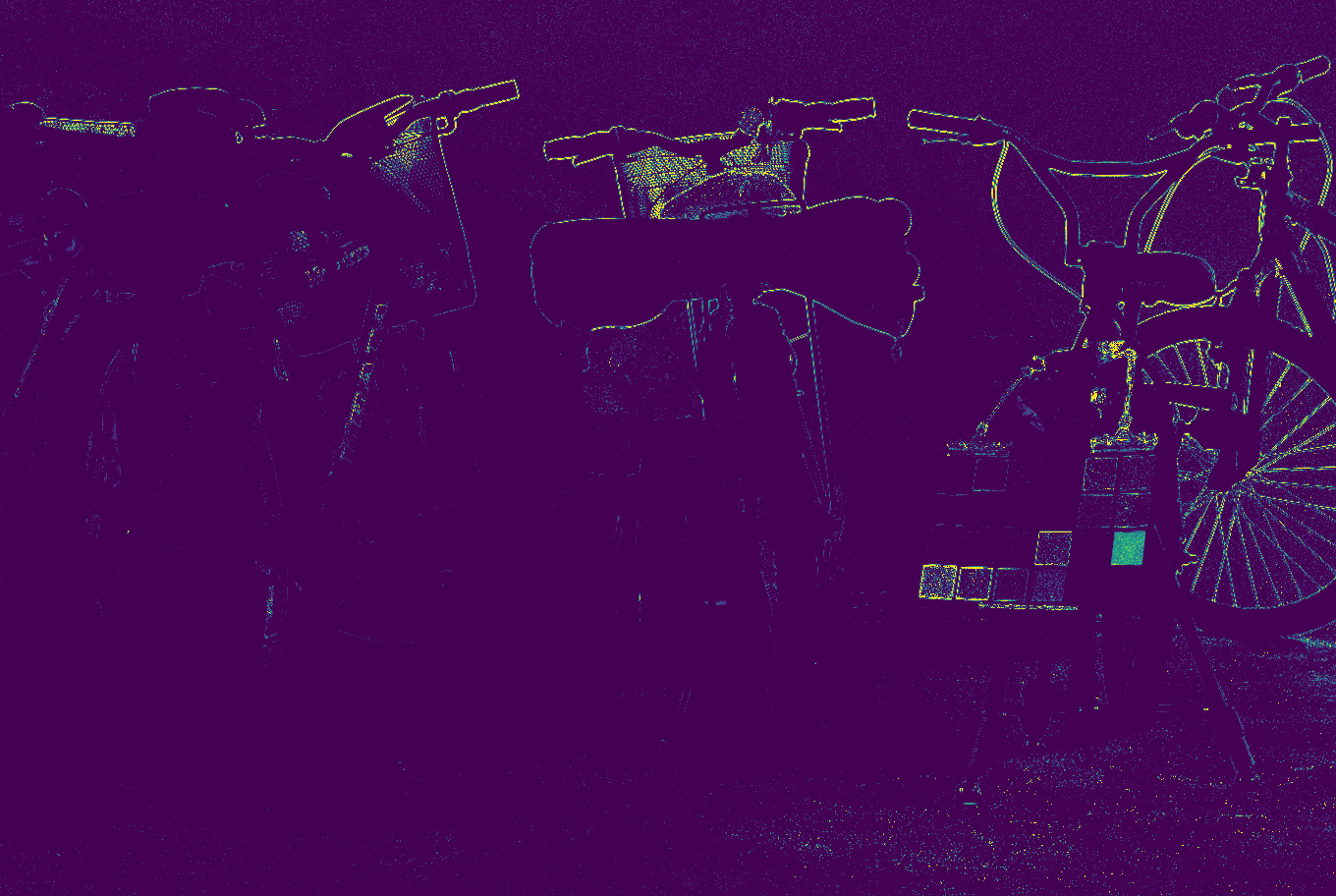}\llap{{\setlength{\fboxsep}{0pt}\setlength{\fboxrule}{2pt}\fcolorbox{red}{yellow}{\includegraphics[width=0.11\linewidth,clip,trim=400 500 400 50]{figures_arxiv/comparison/samsung/SamsungNX2000_0061_err_rang.png}}}} & \includegraphics[width=0.15\linewidth]{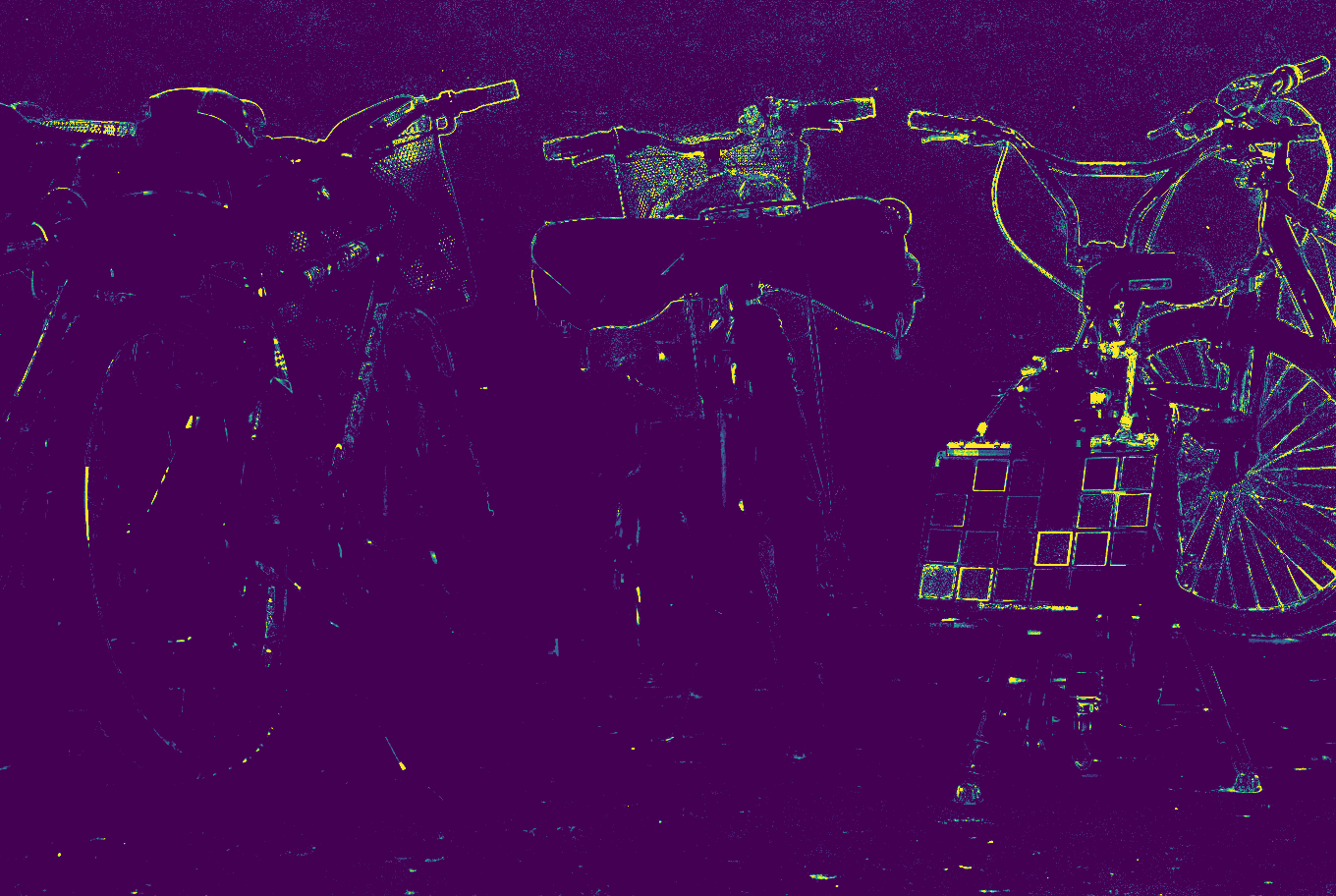}\llap{{\setlength{\fboxsep}{0pt}\setlength{\fboxrule}{2pt}\fcolorbox{red}{yellow}{\includegraphics[width=0.11\linewidth,clip,trim=400 500 400 50]{figures_arxiv/comparison/samsung/SamsungNX2000_0061_err_wacv.png}}}} & \includegraphics[width=0.15\linewidth]{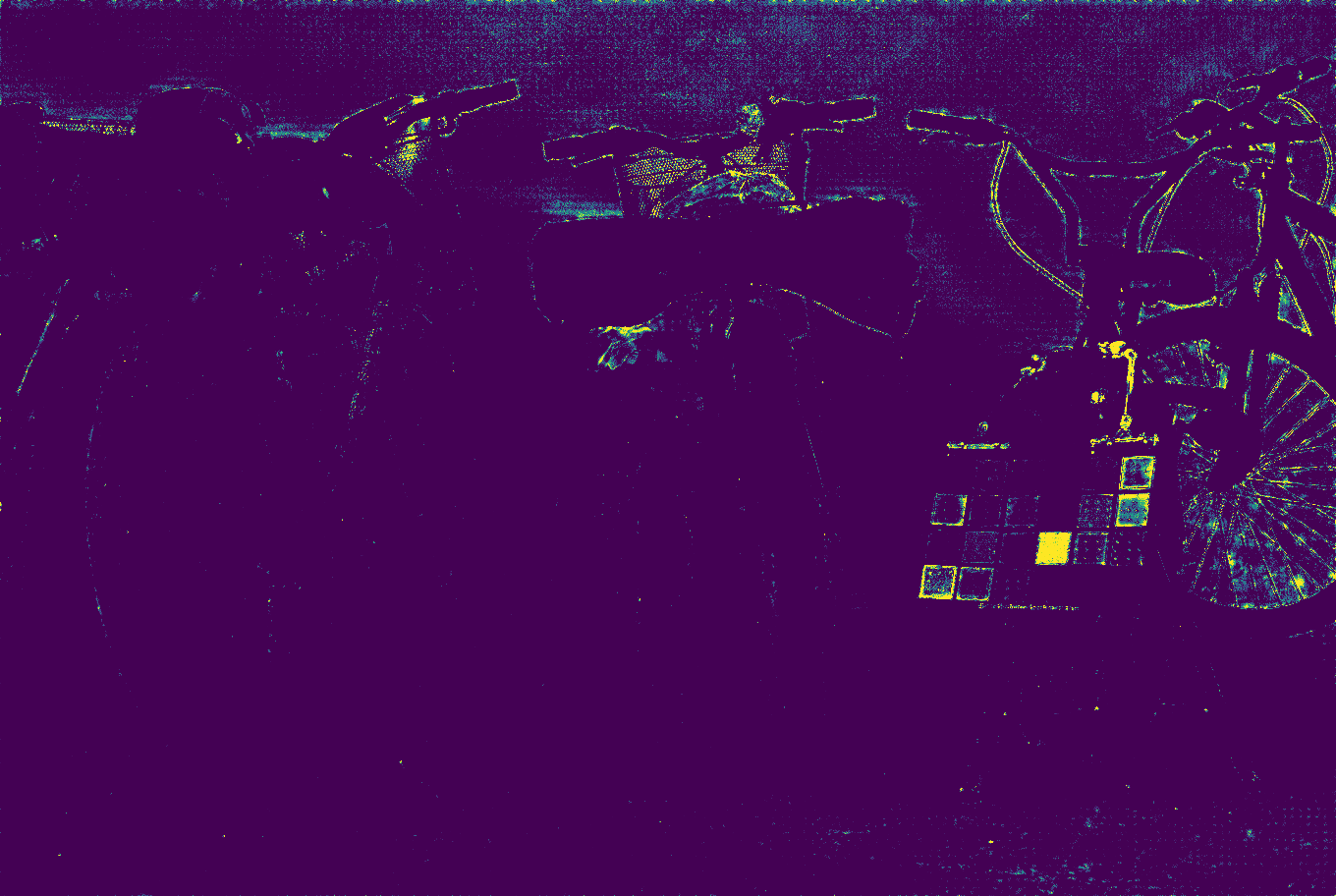}\llap{{\setlength{\fboxsep}{0pt}\setlength{\fboxrule}{2pt}\fcolorbox{red}{yellow}{\includegraphics[width=0.11\linewidth,clip,trim=400 500 400 50]{figures_arxiv/comparison/samsung/0000007_err_ours.png}}}} & \includegraphics[width=0.15\linewidth]{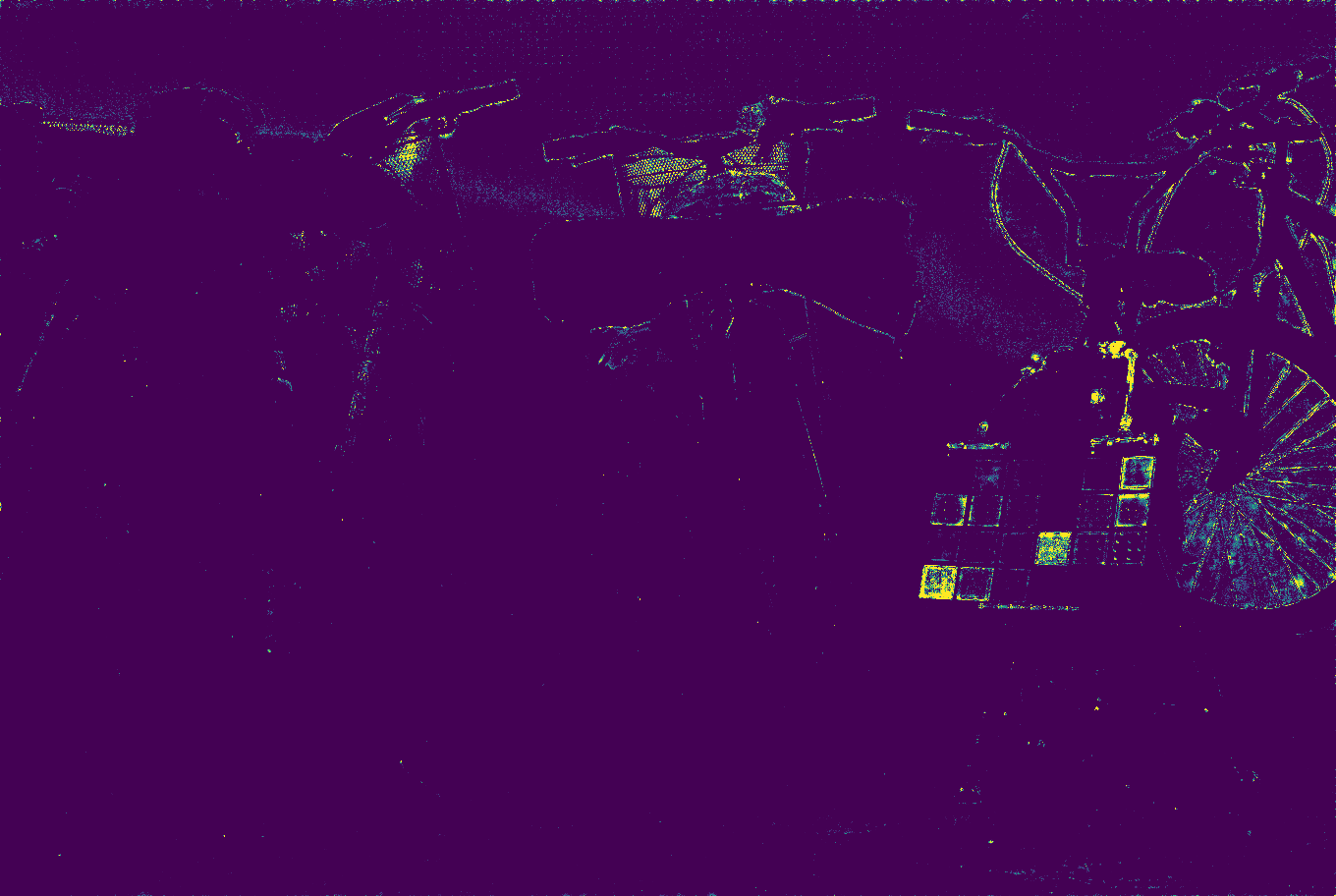}\llap{{\setlength{\fboxsep}{0pt}\setlength{\fboxrule}{2pt}\fcolorbox{red}{yellow}{\includegraphics[width=0.11\linewidth,clip,trim=400 500 400 50]{figures_arxiv/comparison/samsung/0000007_err_ours_ft.png}}}} & \includegraphics[width=0.016\linewidth]{figures_arxiv/colorbar.pdf} & \includegraphics[width=0.15\linewidth]{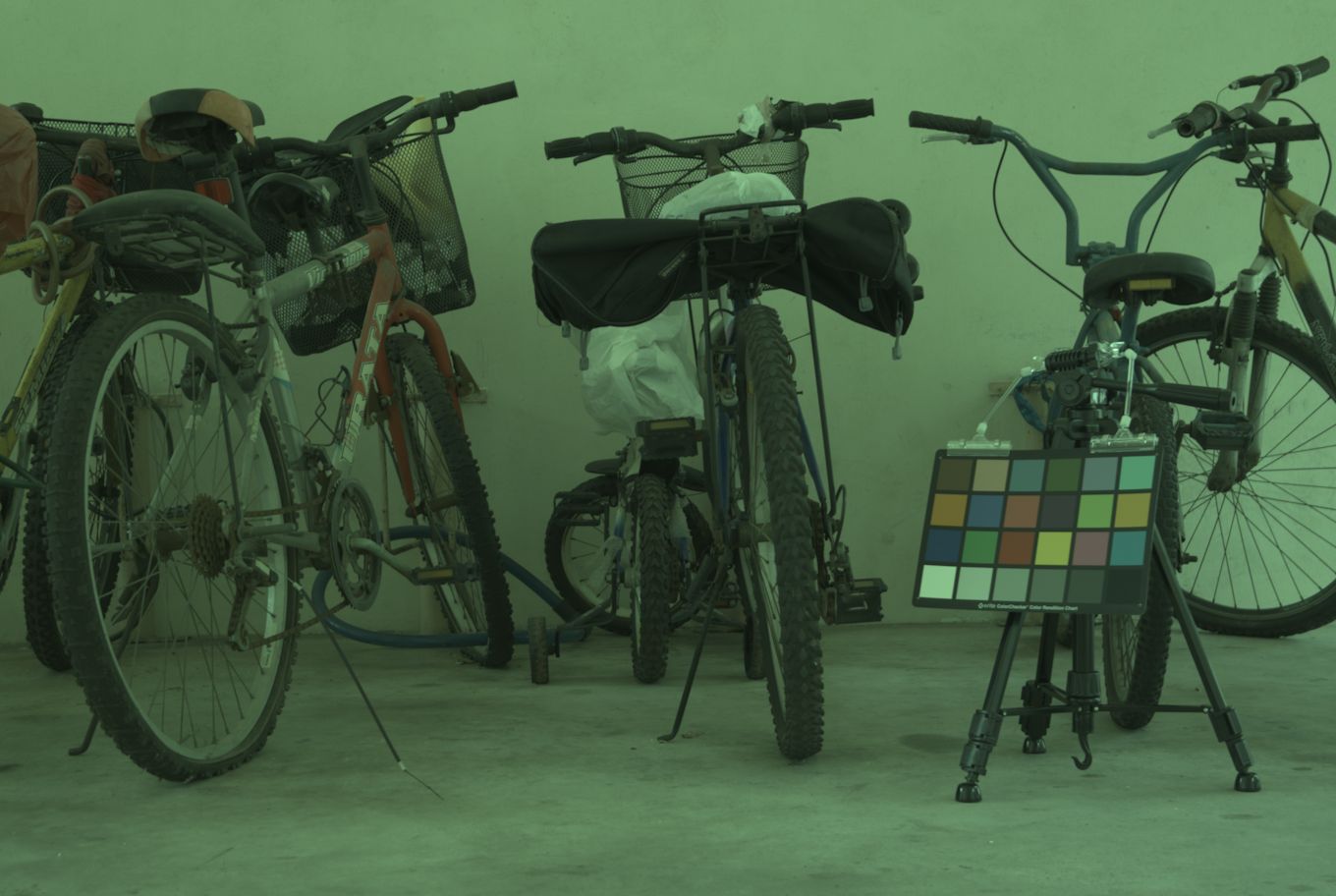} \\

        \includegraphics[width=0.15\linewidth]{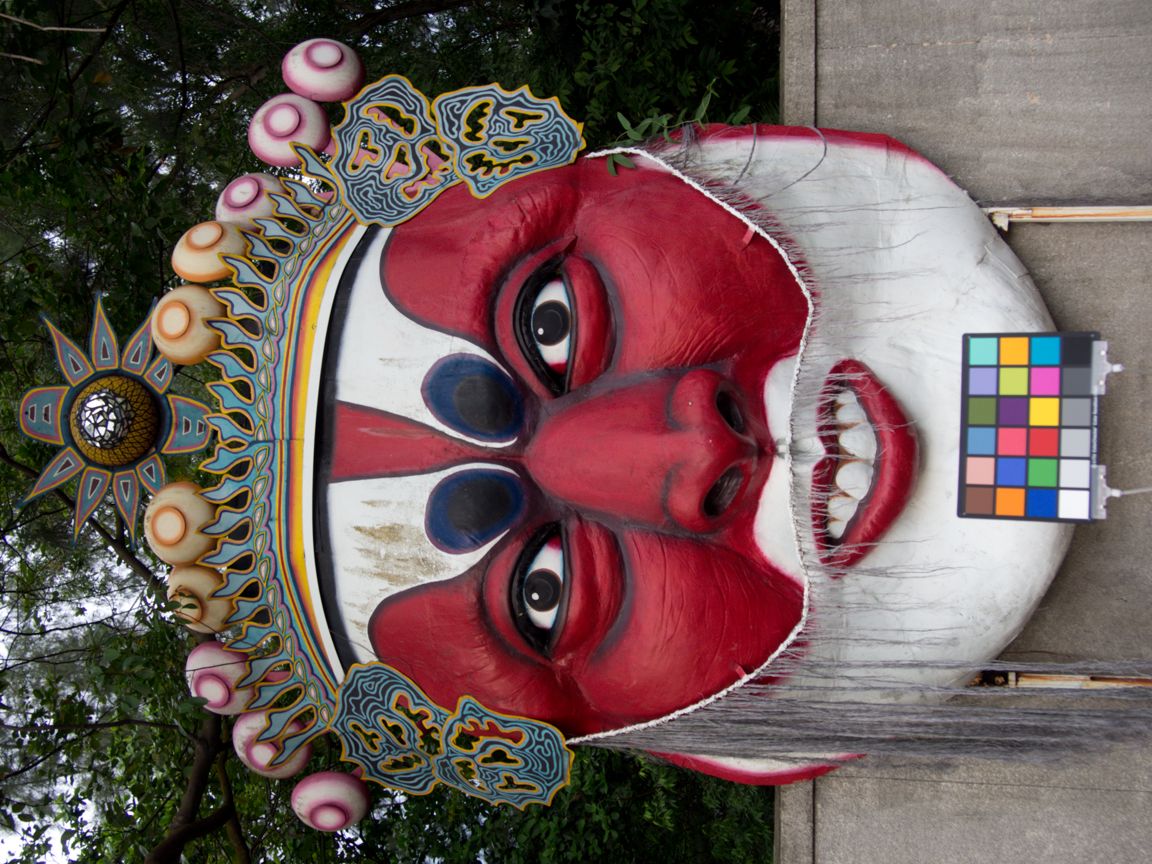} & \includegraphics[width=0.15\linewidth]{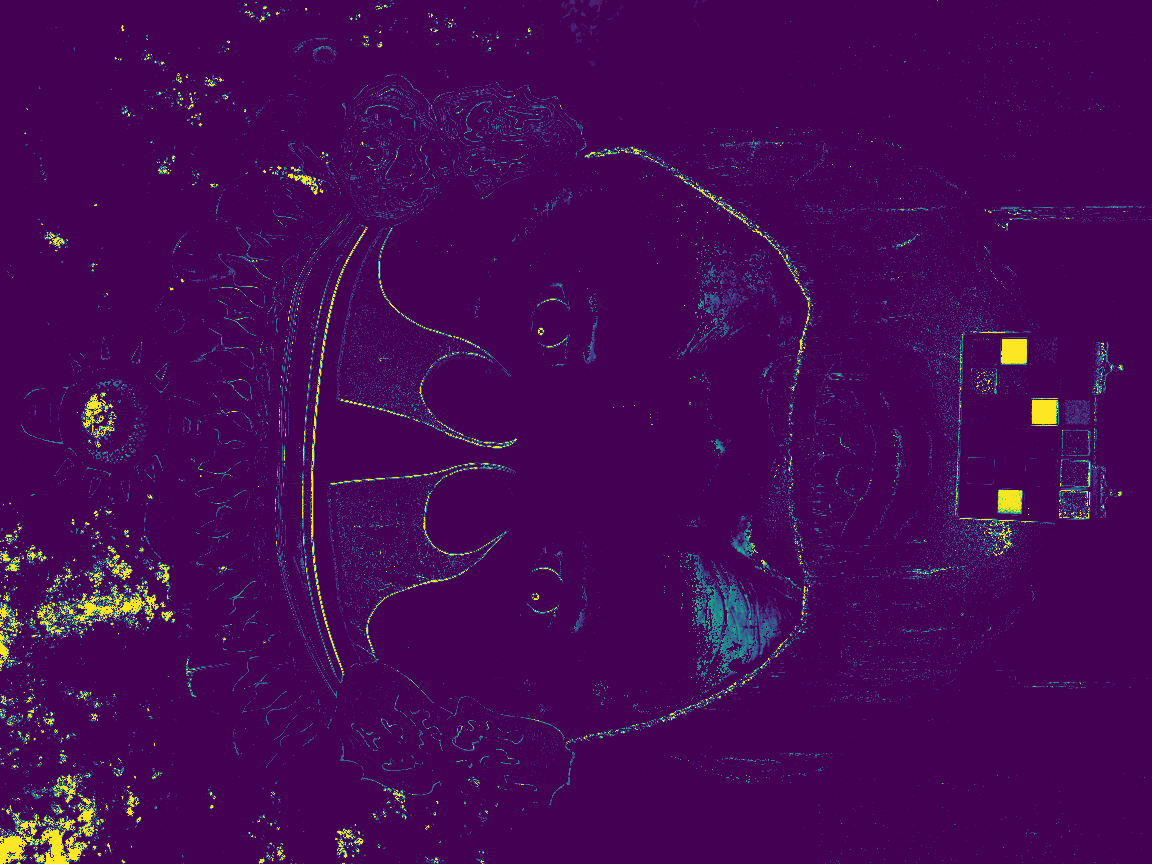}\llap{{\setlength{\fboxsep}{0pt}\setlength{\fboxrule}{2pt}\fcolorbox{red}{yellow}{\includegraphics[width=0.11\linewidth,clip,trim=200 150 200 150]{figures_arxiv/comparison/olympus/OlympusEPL6_0158_st4_err_rang.png}}}} & \includegraphics[width=0.15\linewidth]{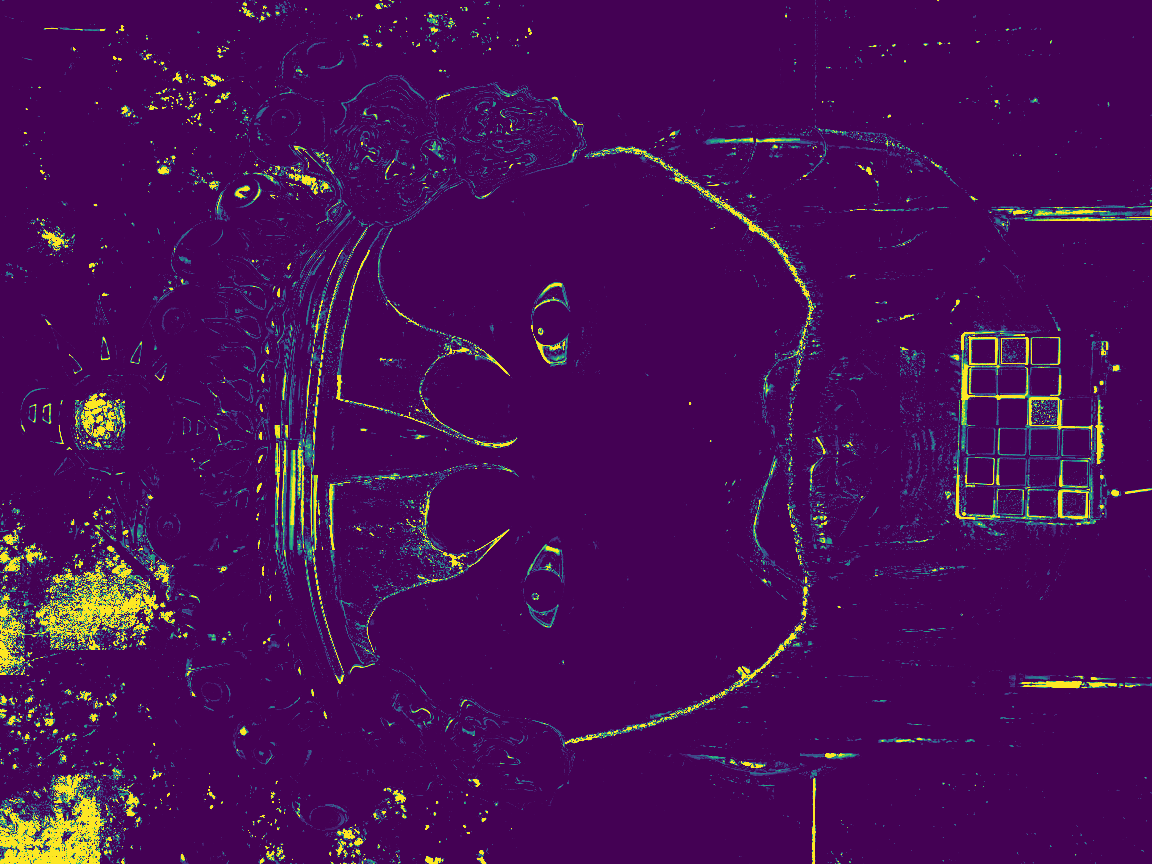}\llap{{\setlength{\fboxsep}{0pt}\setlength{\fboxrule}{2pt}\fcolorbox{red}{yellow}{\includegraphics[width=0.11\linewidth,clip,trim=200 150 200 150]{figures_arxiv/comparison/olympus/OlympusEPL6_0158_st4_err_wacv.png}}}} & \includegraphics[width=0.15\linewidth]{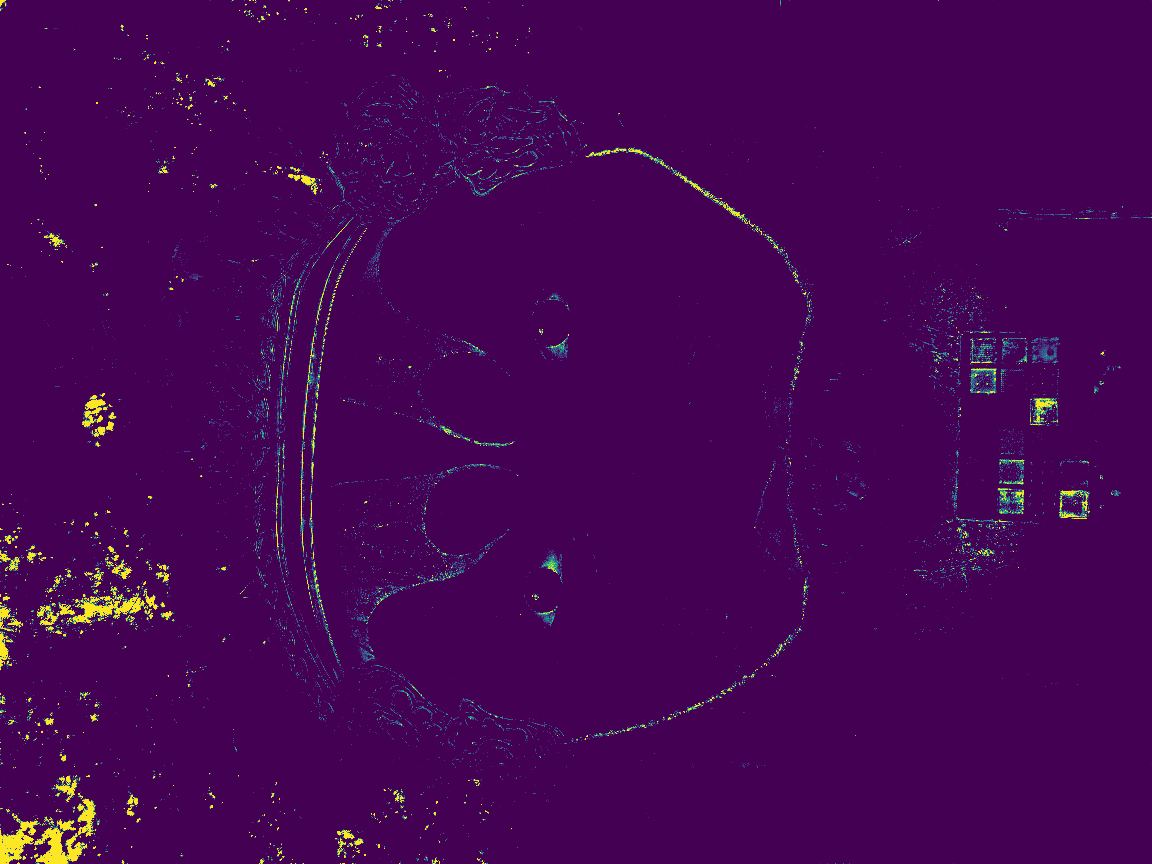}\llap{{\setlength{\fboxsep}{0pt}\setlength{\fboxrule}{2pt}\fcolorbox{red}{yellow}{\includegraphics[width=0.11\linewidth,clip,trim=200 150 200 150]{figures_arxiv/comparison/olympus/0000019_err_ours.png}}}} & \includegraphics[width=0.15\linewidth]{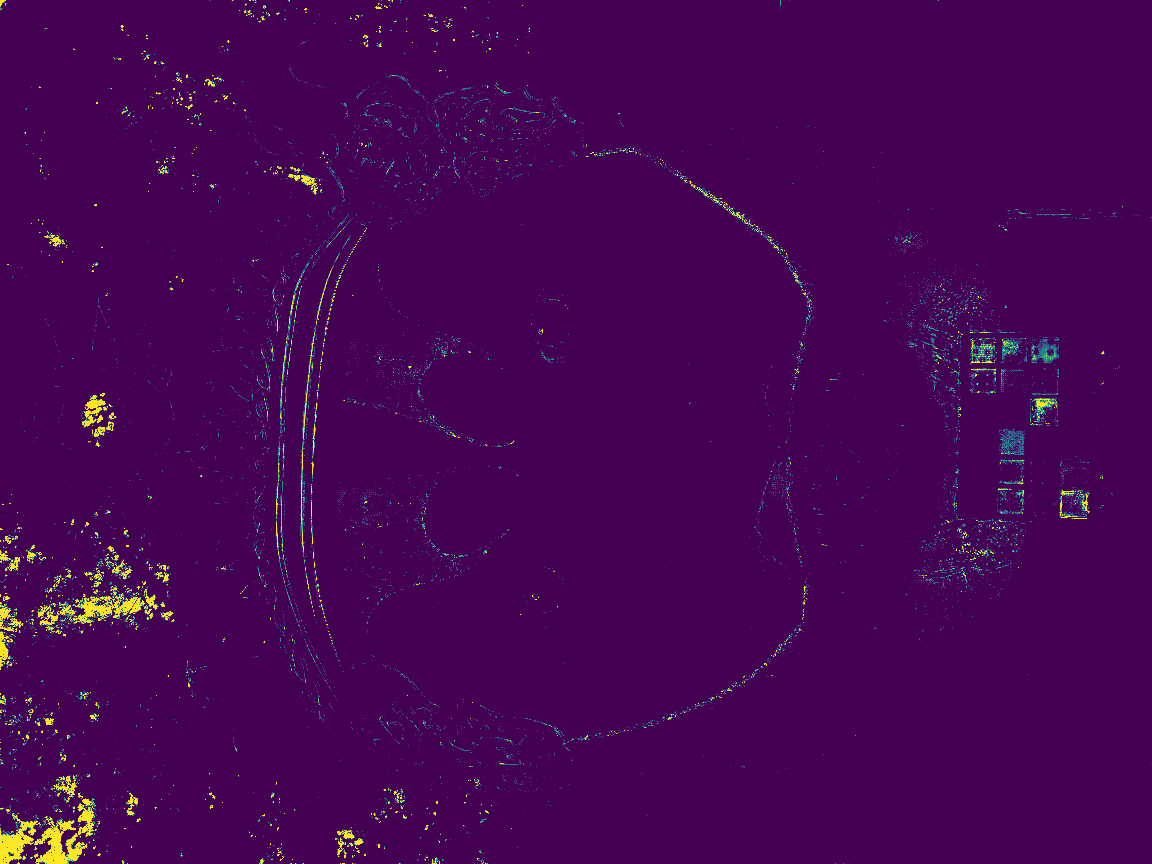}\llap{{\setlength{\fboxsep}{0pt}\setlength{\fboxrule}{2pt}\fcolorbox{red}{yellow}{\includegraphics[width=0.11\linewidth,clip,trim=200 150 200 150]{figures_arxiv/comparison/olympus/0000019_err_ours_ft.png}}}} & \includegraphics[width=0.016\linewidth]{figures_arxiv/colorbar.pdf} & \includegraphics[width=0.15\linewidth]{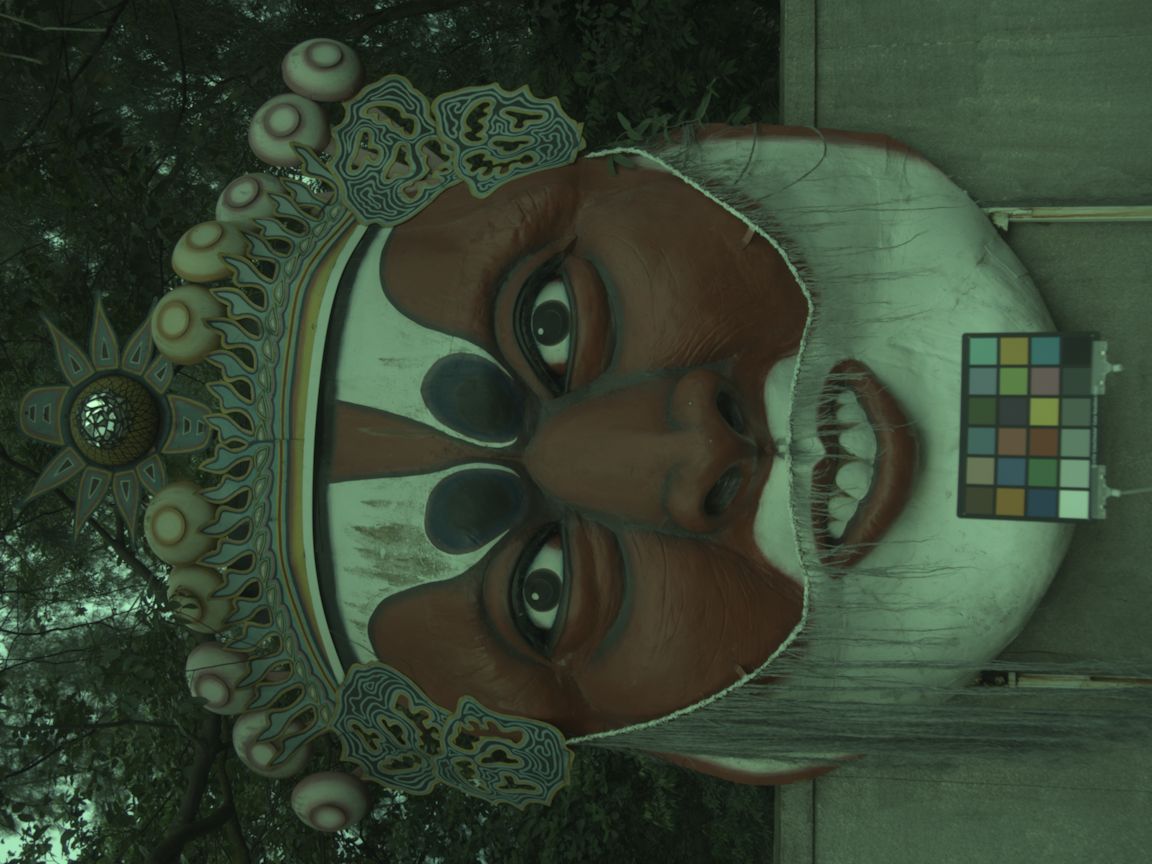} \\

        \includegraphics[width=0.15\linewidth]{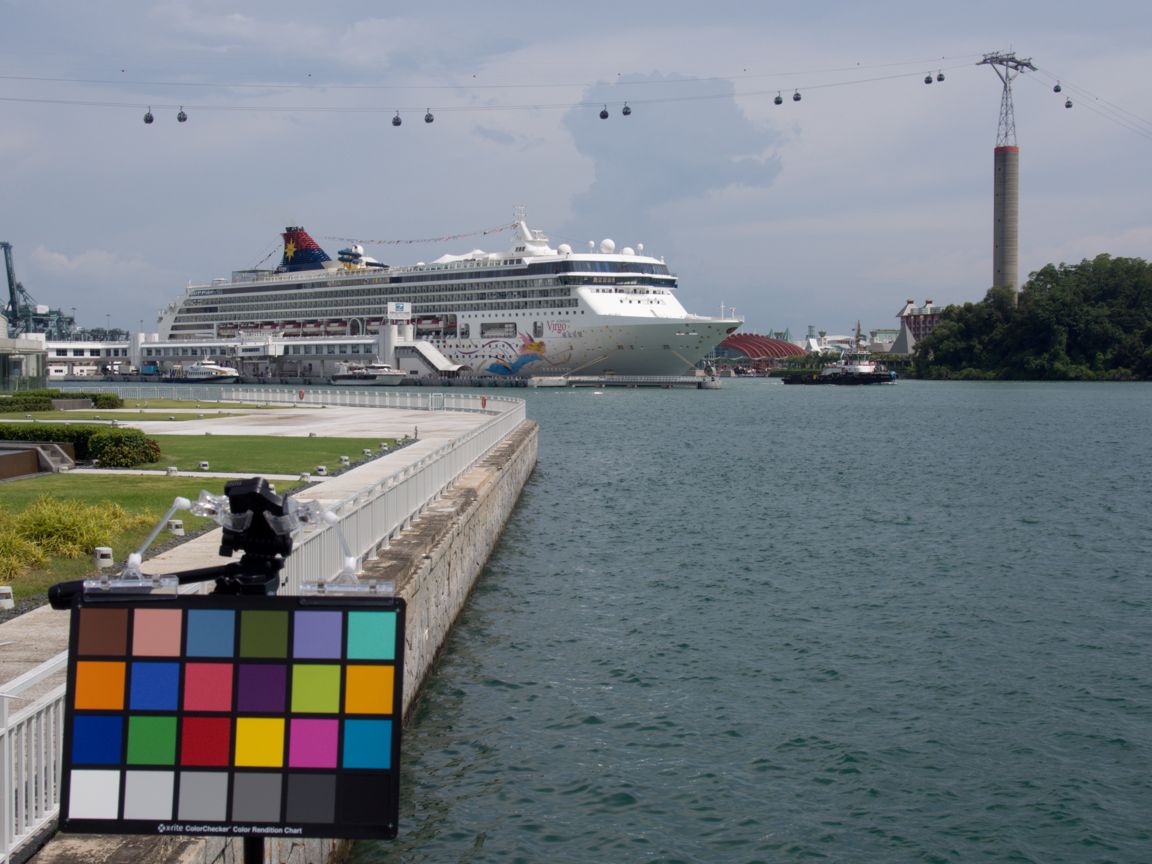} & \includegraphics[width=0.15\linewidth]{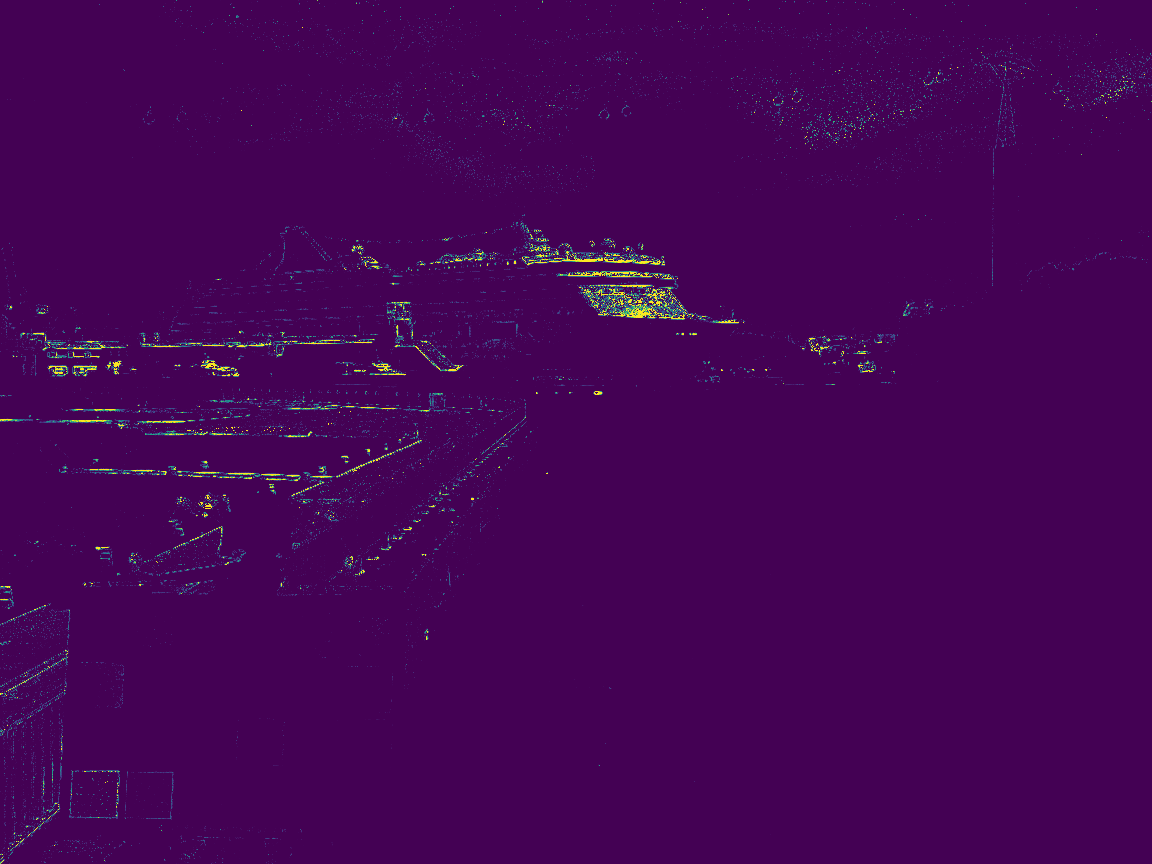}\llap{{\setlength{\fboxsep}{0pt}\setlength{\fboxrule}{2pt}\fcolorbox{red}{yellow}{\includegraphics[width=0.11\linewidth,clip,trim=0 200 550 200]{figures_arxiv/comparison/olympus/OlympusEPL6_0186_st4_err_rang.png}}}} & \includegraphics[width=0.15\linewidth]{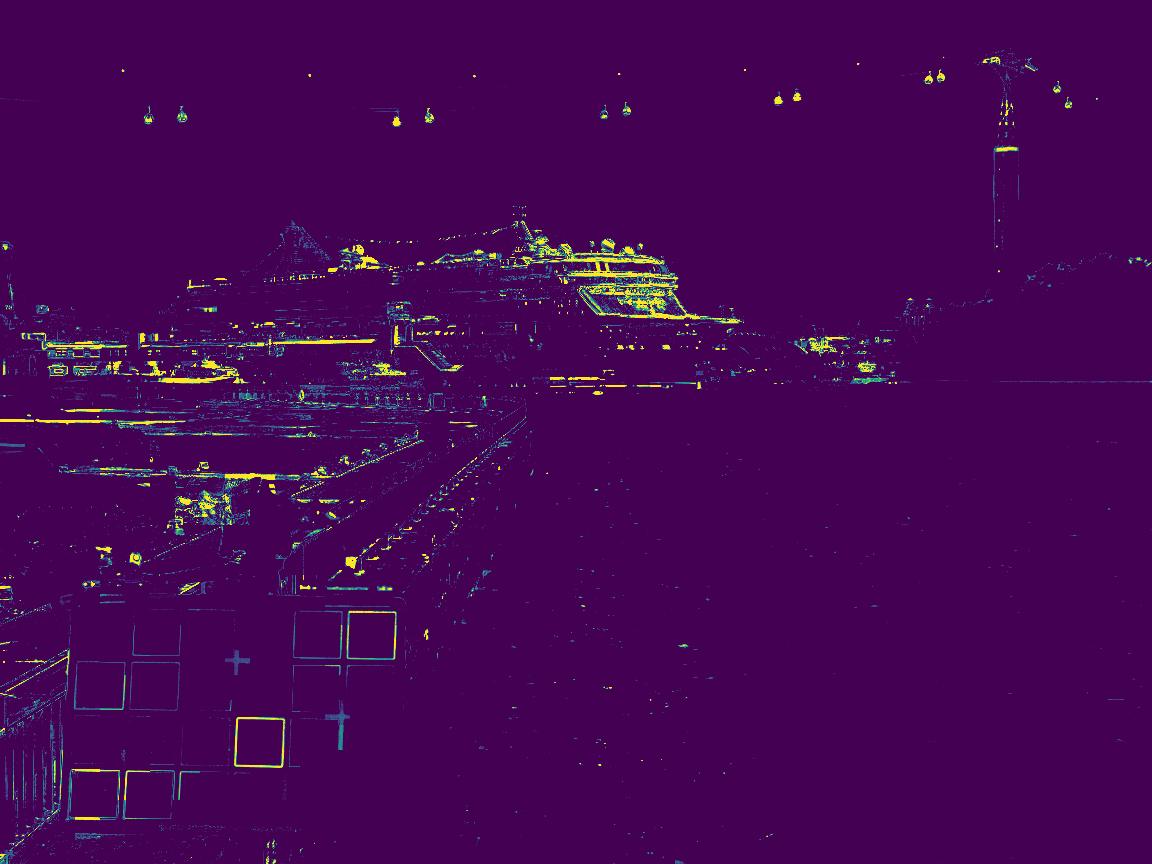}\llap{{\setlength{\fboxsep}{0pt}\setlength{\fboxrule}{2pt}\fcolorbox{red}{yellow}{\includegraphics[width=0.11\linewidth,clip,trim=0 200 550 200]{figures_arxiv/comparison/olympus/OlympusEPL6_0186_st4_err_wacv.png}}}} & \includegraphics[width=0.15\linewidth]{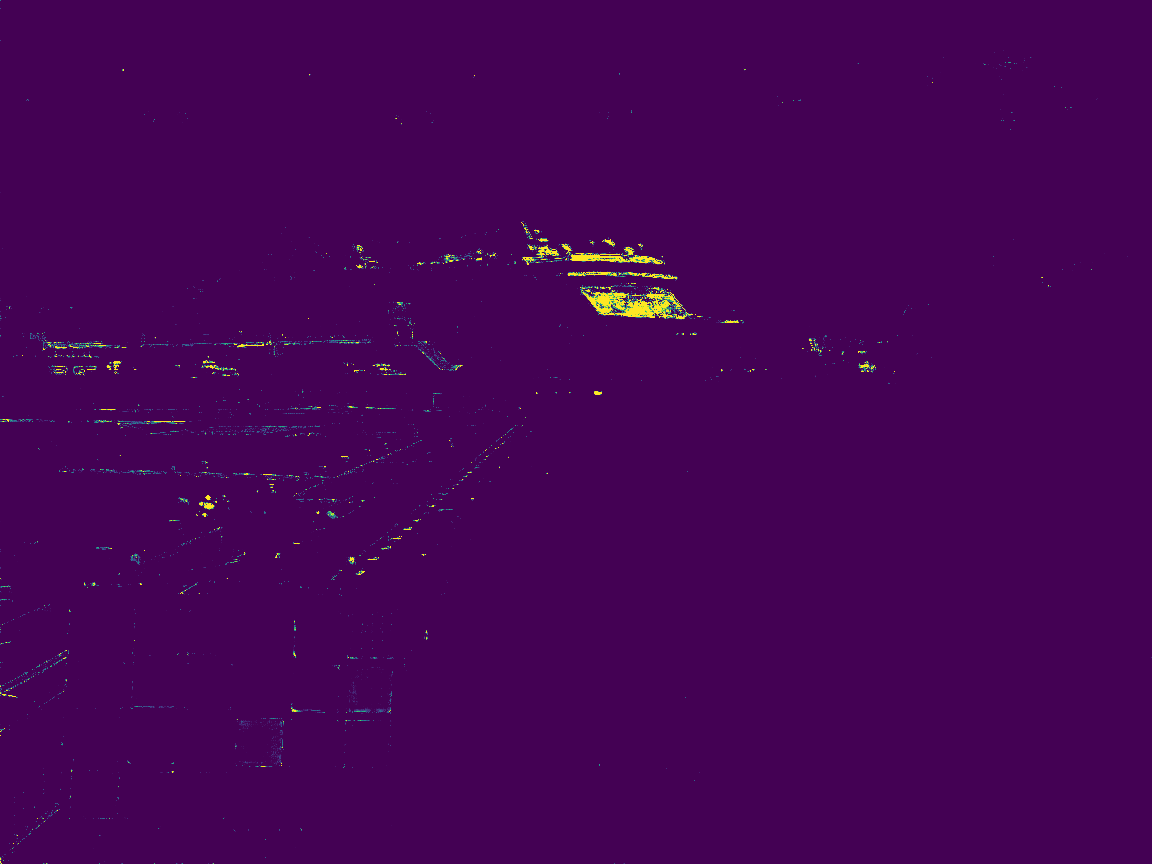}\llap{{\setlength{\fboxsep}{0pt}\setlength{\fboxrule}{2pt}\fcolorbox{red}{yellow}{\includegraphics[width=0.11\linewidth,clip,trim=0 200 550 200]{figures_arxiv/comparison/olympus/0000021_err_ours.png}}}} & \includegraphics[width=0.15\linewidth]{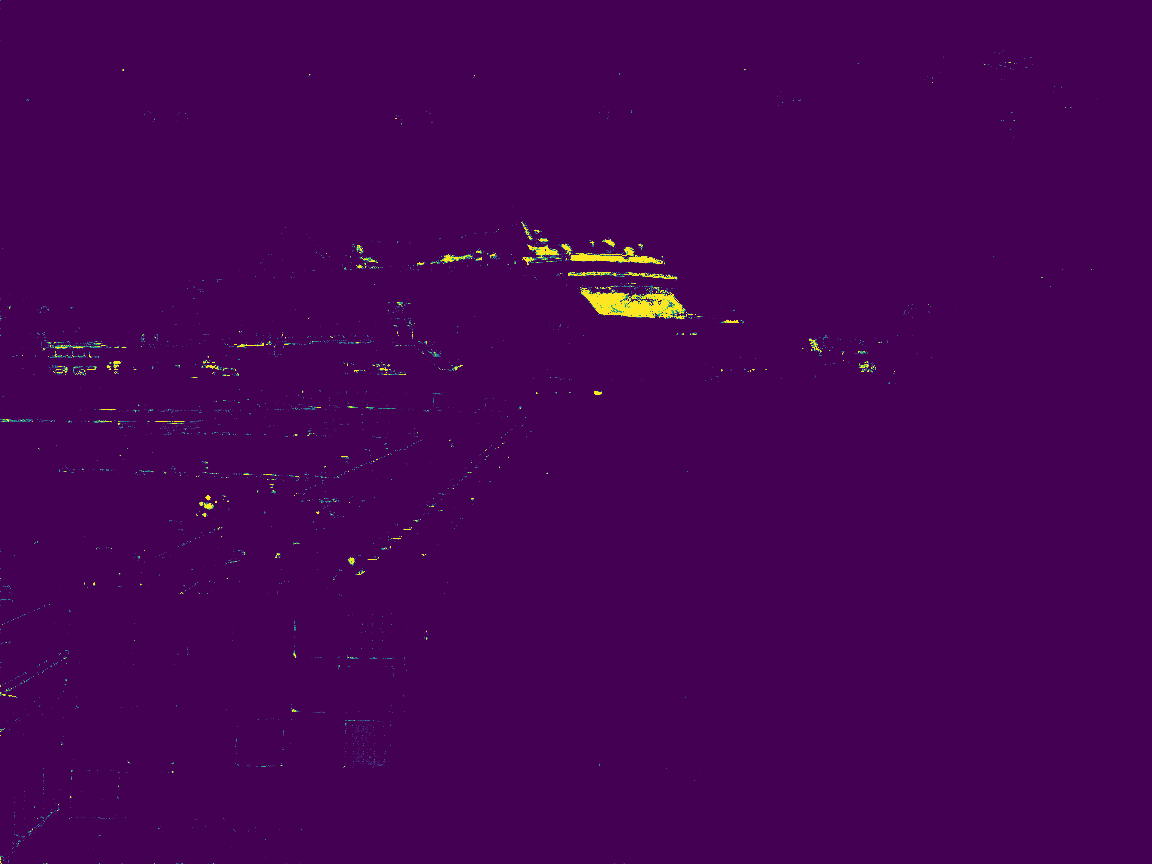}\llap{{\setlength{\fboxsep}{0pt}\setlength{\fboxrule}{2pt}\fcolorbox{red}{yellow}{\includegraphics[width=0.11\linewidth,clip,trim=0 200 550 200]{figures_arxiv/comparison/olympus/0000021_err_ours_ft.png}}}} & \includegraphics[width=0.016\linewidth]{figures_arxiv/colorbar.pdf} & \includegraphics[width=0.15\linewidth]{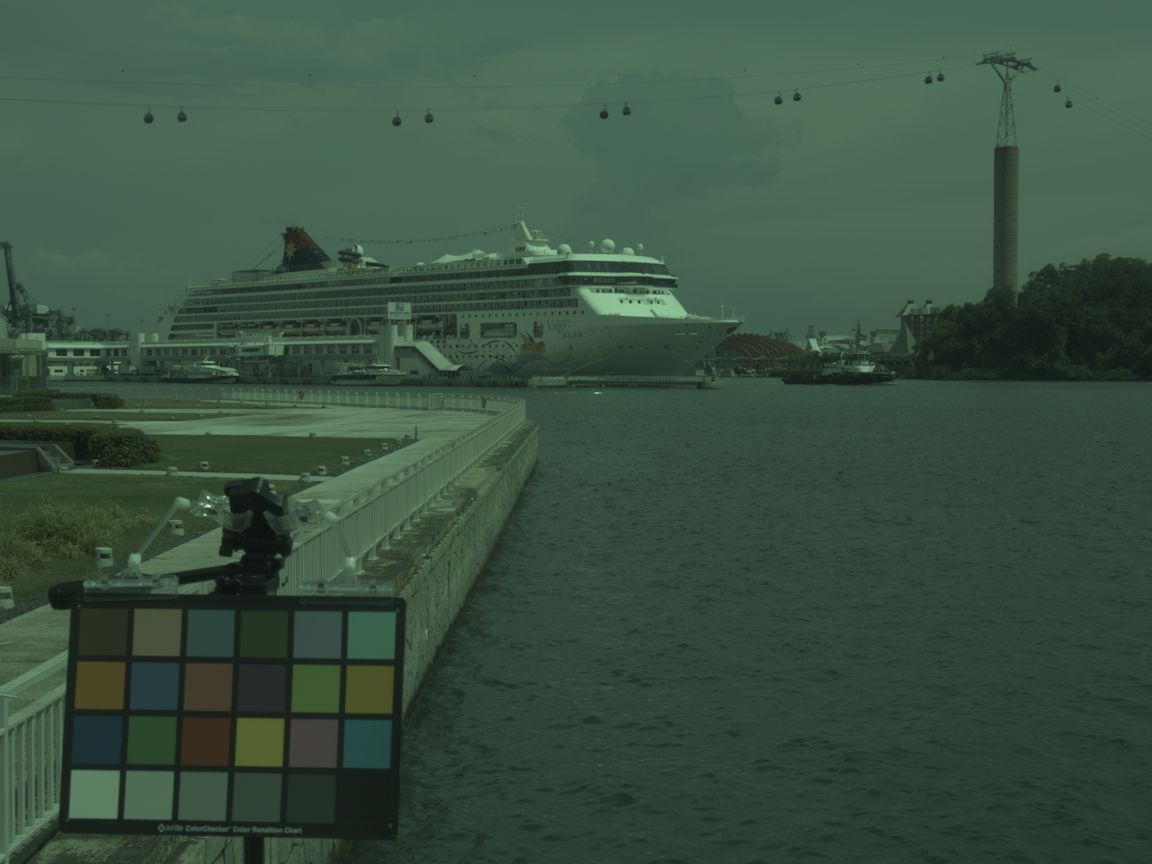} \\
        
        \includegraphics[width=0.15\linewidth]{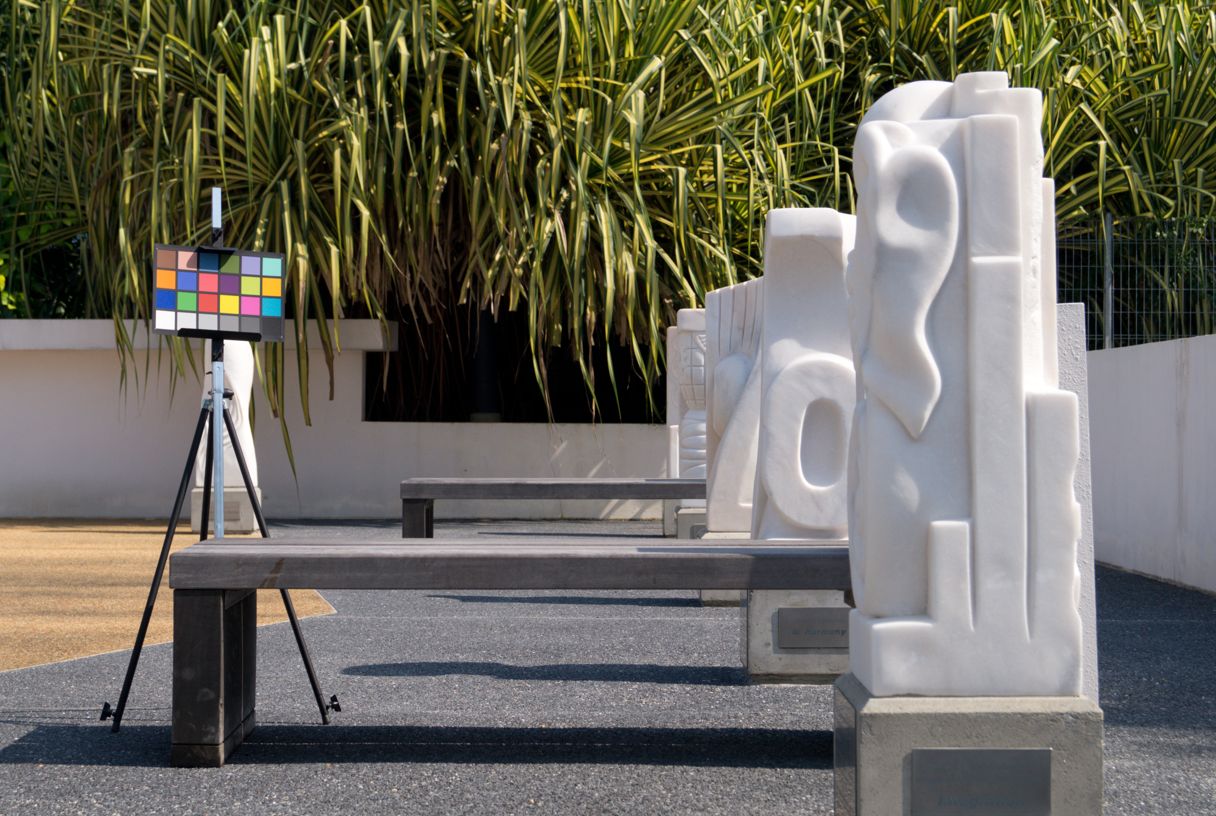} & \includegraphics[width=0.15\linewidth]{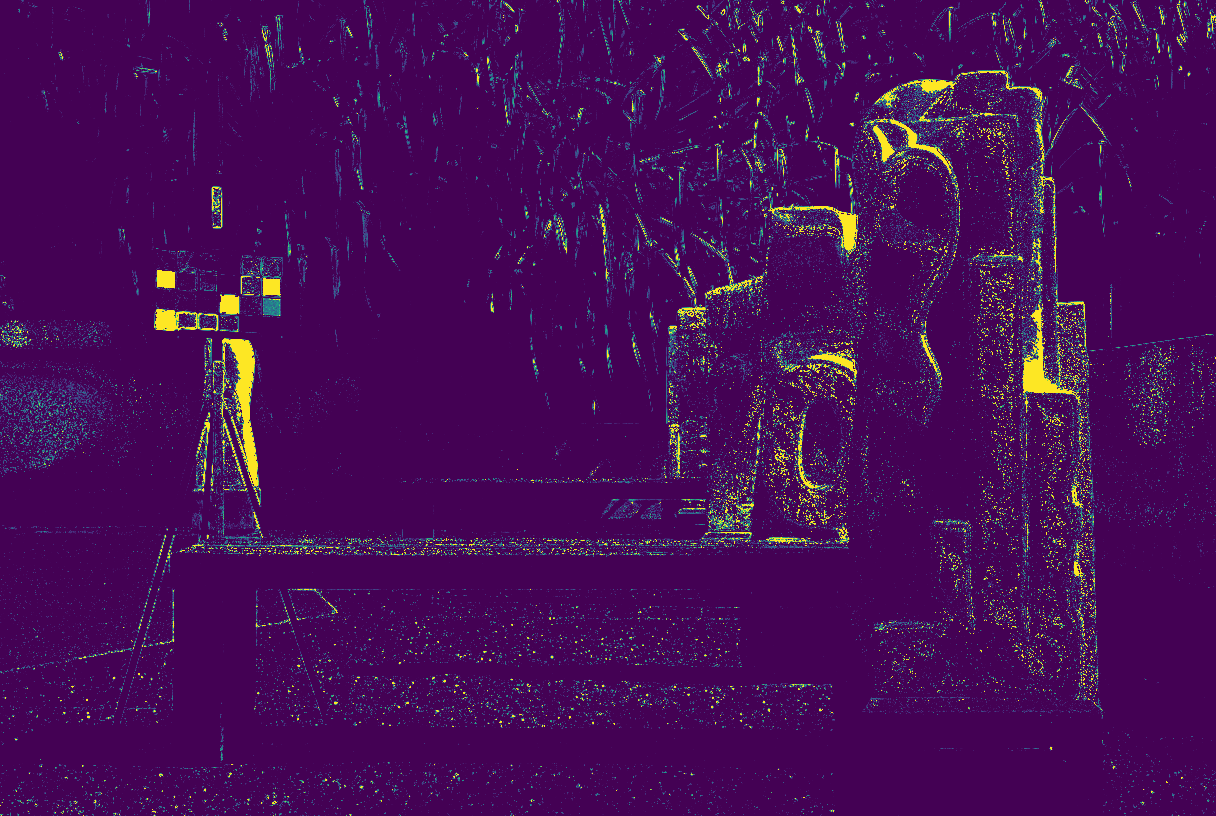}\llap{{\setlength{\fboxsep}{0pt}\setlength{\fboxrule}{2pt}\fcolorbox{red}{yellow}{\includegraphics[width=0.11\linewidth,clip,trim=500 530 300 0]{figures_arxiv/comparison/sony/SonyA57_0006_st4_err_rang.png}}}} & \includegraphics[width=0.15\linewidth]{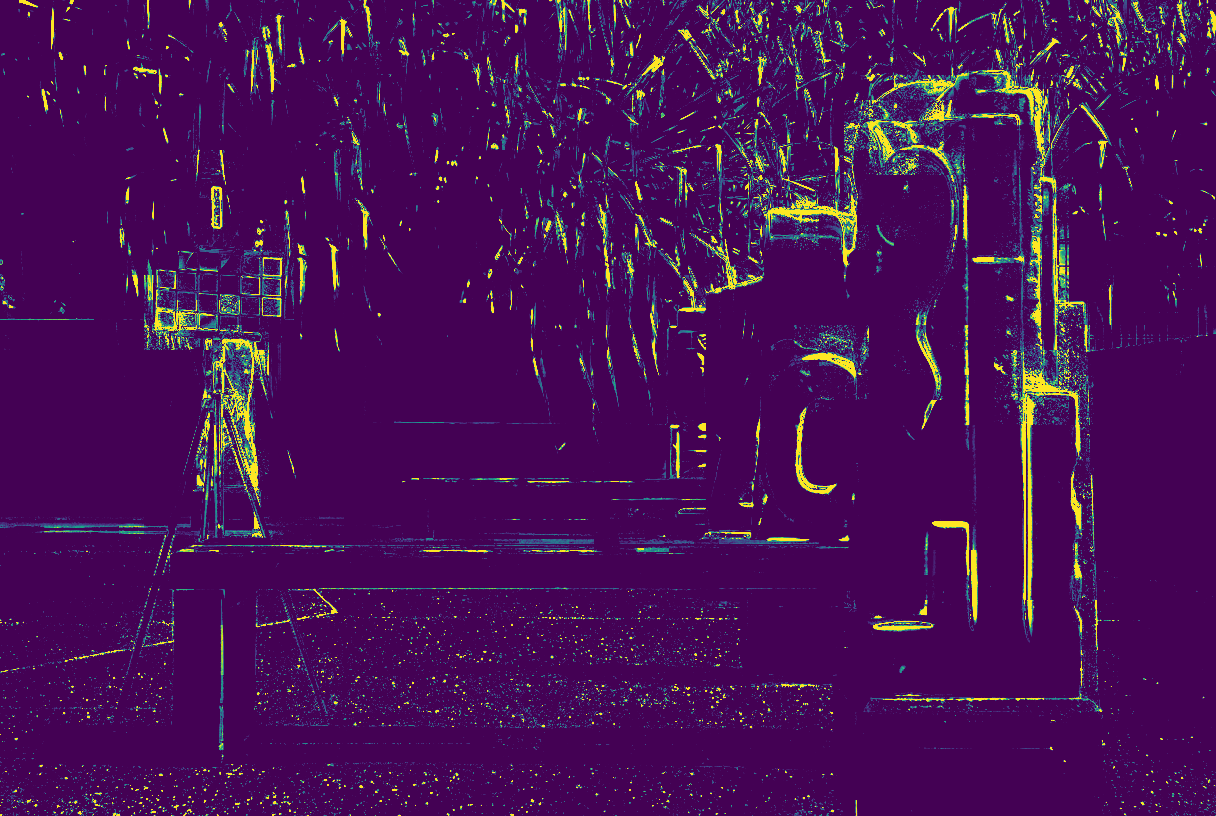}\llap{{\setlength{\fboxsep}{0pt}\setlength{\fboxrule}{2pt}\fcolorbox{red}{yellow}{\includegraphics[width=0.11\linewidth,clip,trim=500 530 300 0]{figures_arxiv/comparison/sony/SonyA57_0006_st4_err_wacv.png}}}} & \includegraphics[width=0.15\linewidth]{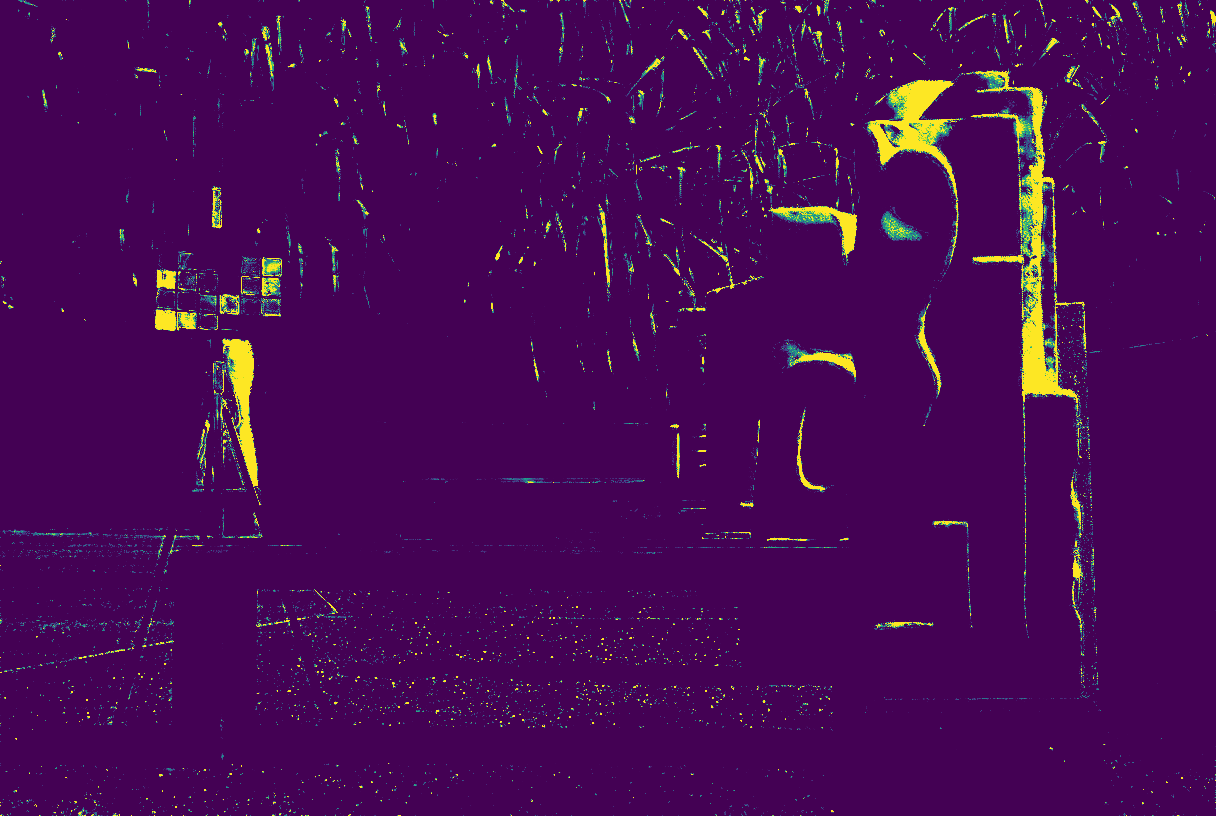}\llap{{\setlength{\fboxsep}{0pt}\setlength{\fboxrule}{2pt}\fcolorbox{red}{yellow}{\includegraphics[width=0.11\linewidth,clip,trim=500 530 300 0]{figures_arxiv/comparison/sony/0000000_err_ours.png}}}} & \includegraphics[width=0.15\linewidth]{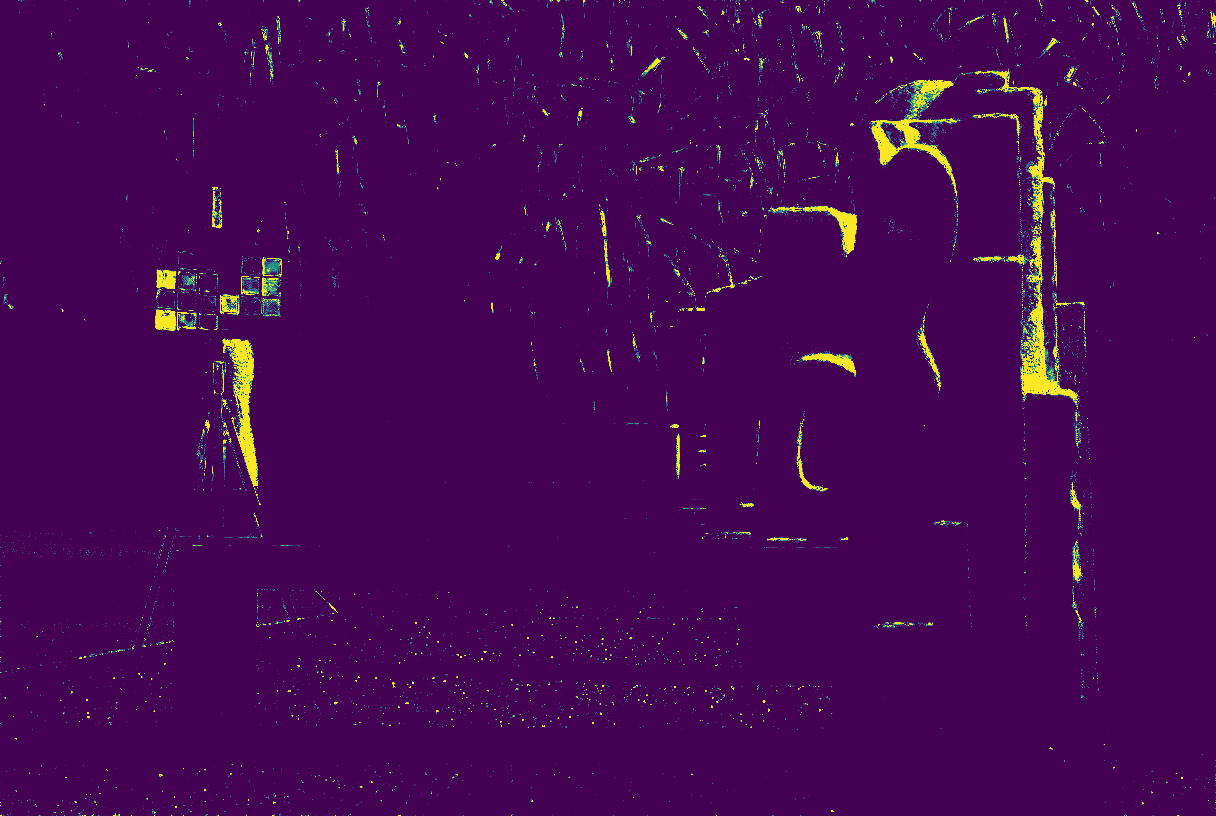}\llap{{\setlength{\fboxsep}{0pt}\setlength{\fboxrule}{2pt}\fcolorbox{red}{yellow}{\includegraphics[width=0.11\linewidth,clip,trim=500 530 300 0]{figures_arxiv/comparison/sony/0000000_err_ours_ft.png}}}} & \includegraphics[width=0.016\linewidth]{figures_arxiv/colorbar.pdf} & \includegraphics[width=0.15\linewidth]{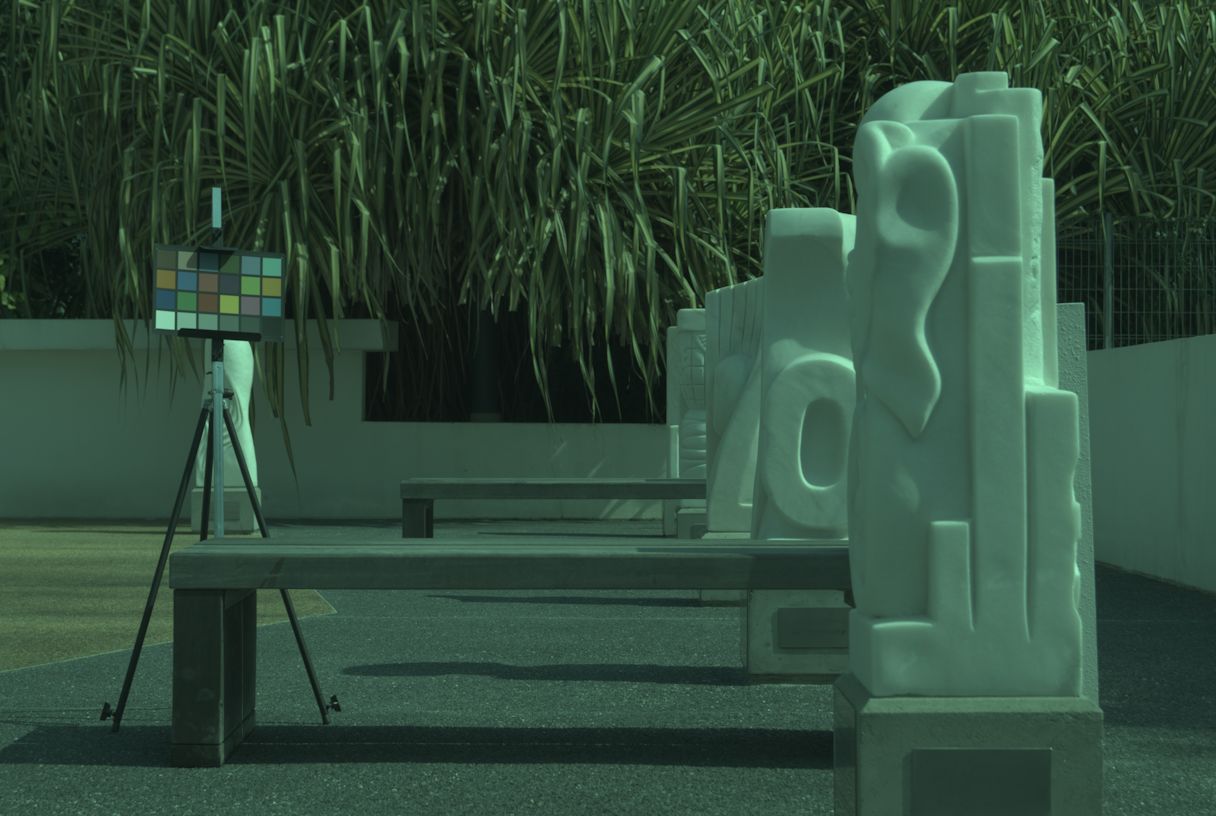} \\

        \includegraphics[width=0.15\linewidth]{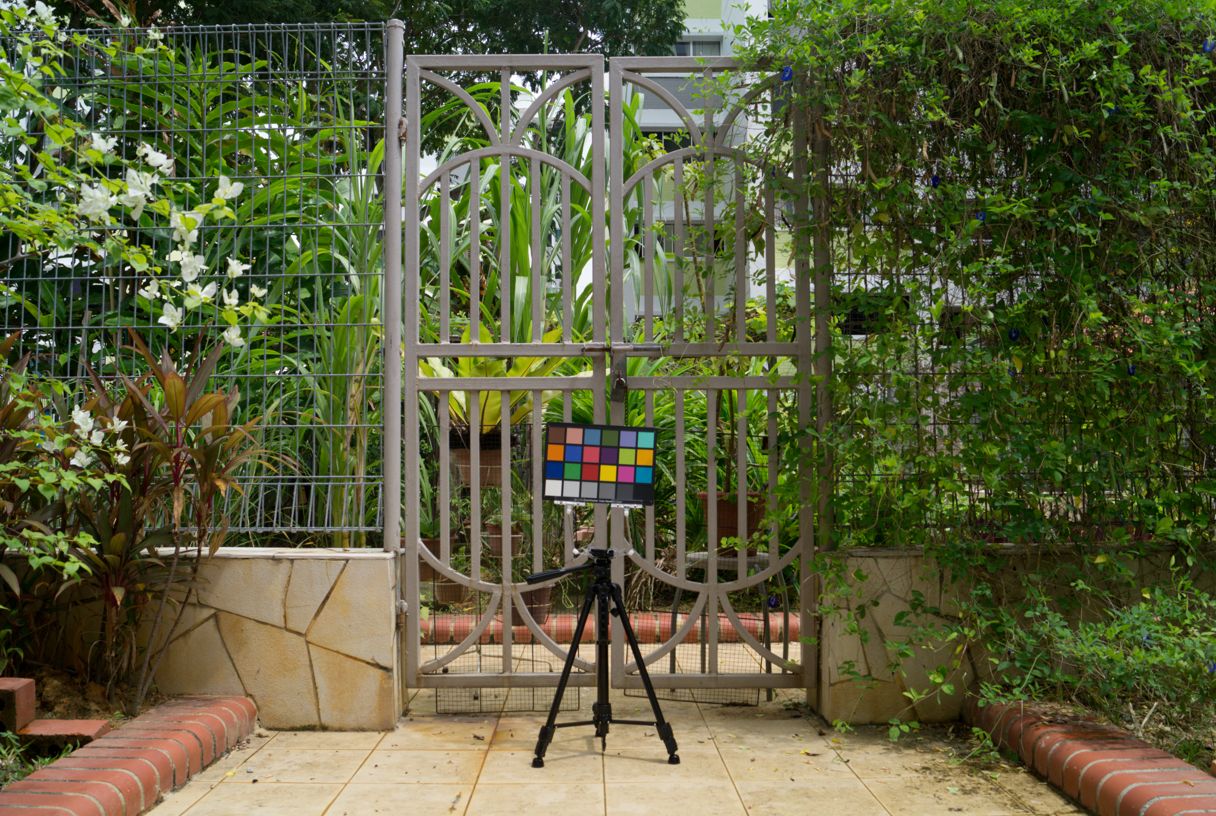} & \includegraphics[width=0.15\linewidth]{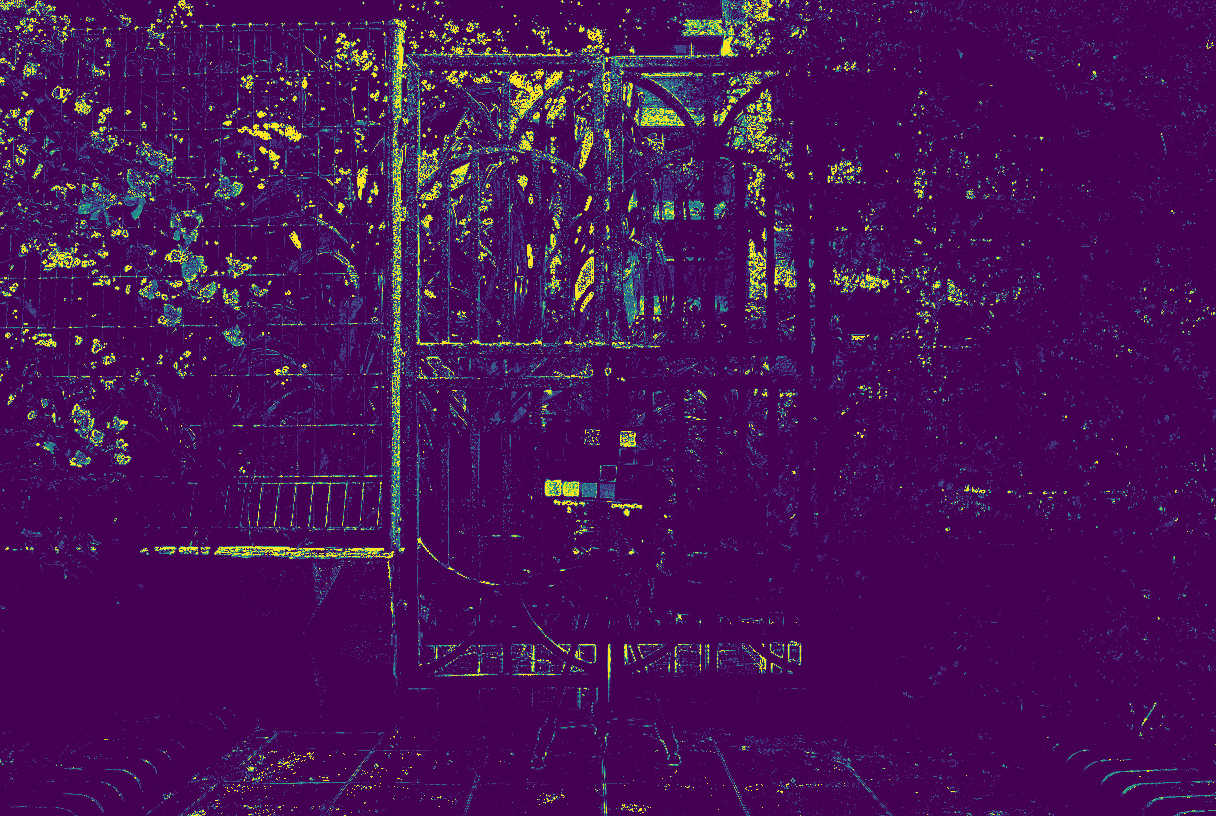}\llap{{\setlength{\fboxsep}{0pt}\setlength{\fboxrule}{2pt}\fcolorbox{red}{yellow}{\includegraphics[width=0.11\linewidth,clip,trim=0 380 800 150]{figures_arxiv/comparison/sony/SonyA57_0139_st4_err_rang.png}}}} & \includegraphics[width=0.15\linewidth]{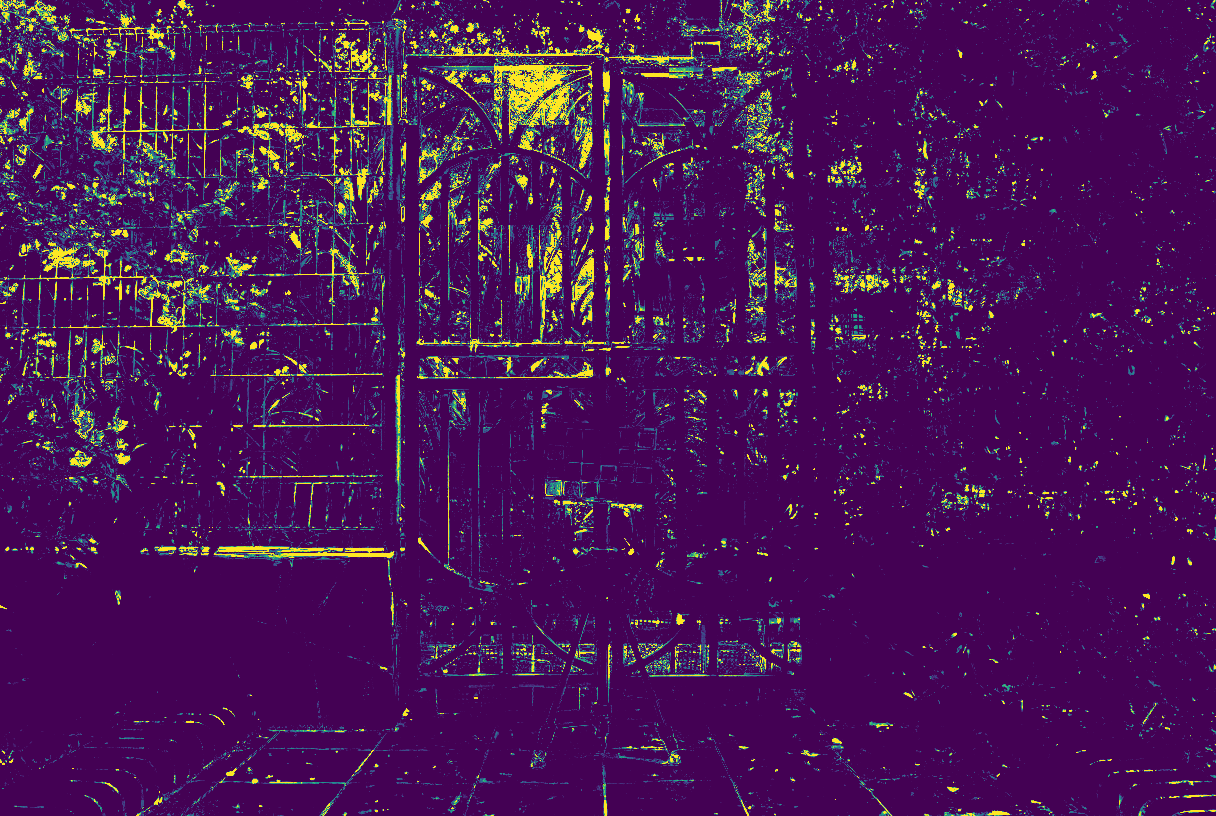}\llap{{\setlength{\fboxsep}{0pt}\setlength{\fboxrule}{2pt}\fcolorbox{red}{yellow}{\includegraphics[width=0.11\linewidth,clip,trim=0 380 800 150]{figures_arxiv/comparison/sony/SonyA57_0139_st4_err_wacv.png}}}} & \includegraphics[width=0.15\linewidth]{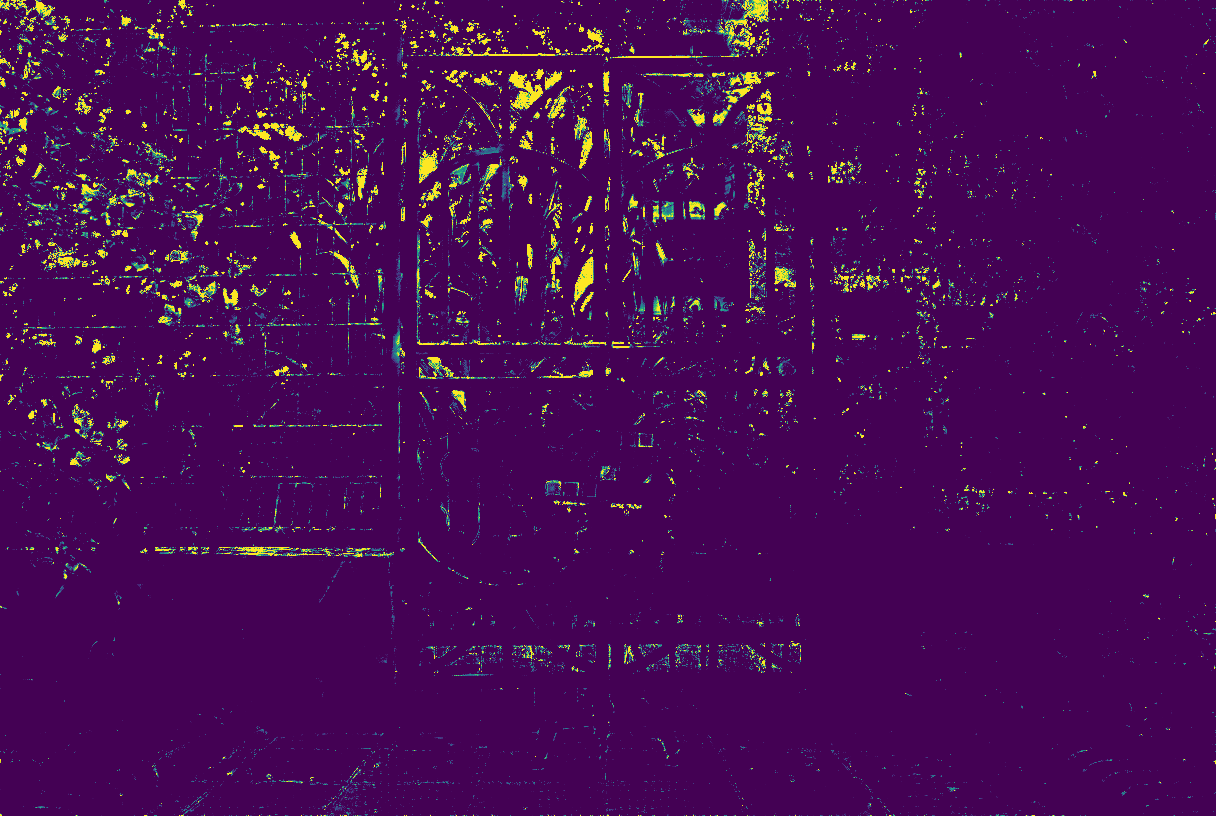}\llap{{\setlength{\fboxsep}{0pt}\setlength{\fboxrule}{2pt}\fcolorbox{red}{yellow}{\includegraphics[width=0.11\linewidth,clip,trim=0 380 800 150]{figures_arxiv/comparison/sony/0000019_err_ours.png}}}} & \includegraphics[width=0.15\linewidth]{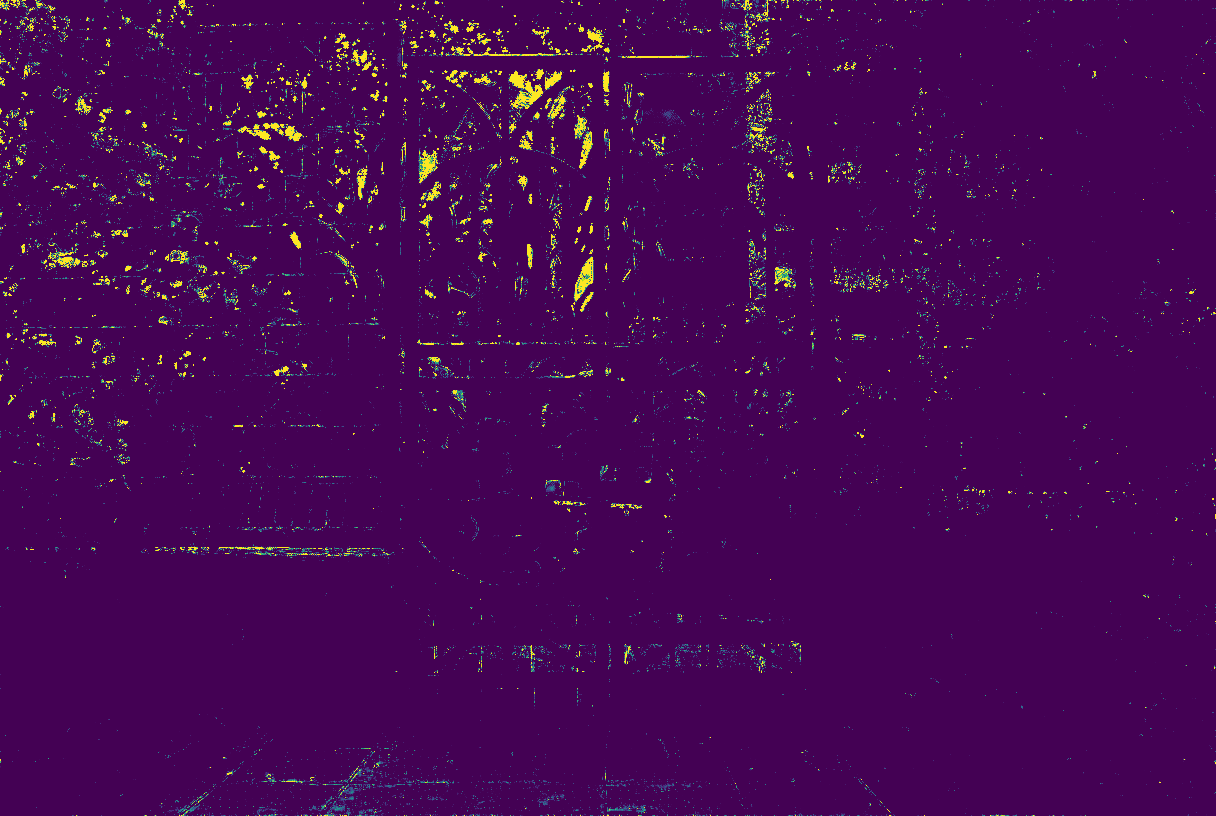}\llap{{\setlength{\fboxsep}{0pt}\setlength{\fboxrule}{2pt}\fcolorbox{red}{yellow}{\includegraphics[width=0.11\linewidth,clip,trim=0 380 800 150]{figures_arxiv/comparison/sony/0000019_err_ours_ft.png}}}} & \includegraphics[width=0.016\linewidth]{figures_arxiv/colorbar.pdf} & \includegraphics[width=0.15\linewidth]{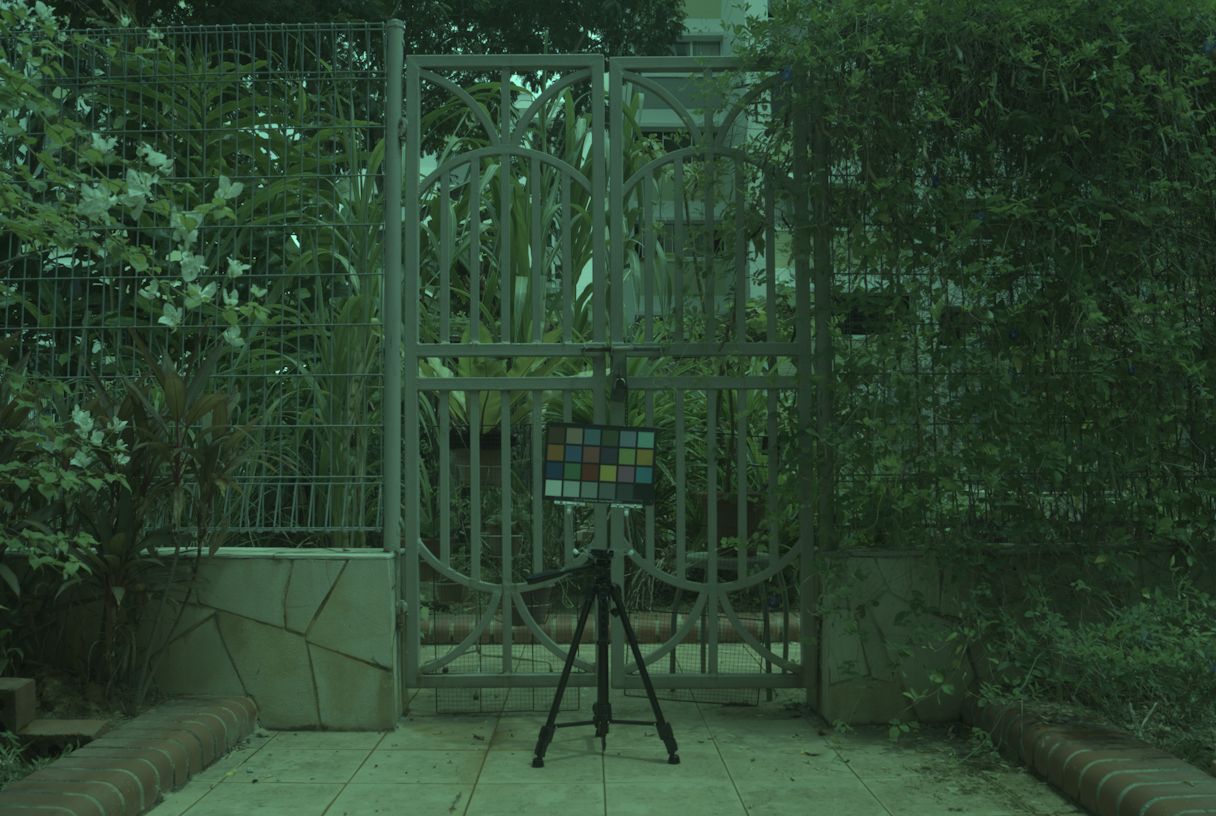} \\

        {\small Input} & {\small RIR~\cite{rang}} & {\small SAM~\cite{wacv}} & {\small Ours} & {\small Ours + FT} & & {\small Ground truth}\\
    \end{tabular}
    \caption{Qualitative comparison. Each two rows show results on Samsung NX2000, Olympus E-PL6, and Sony SLT-A57, respectively.}
    \label{fig:qualitative}
\end{figure*}
\Tref{table:quant_raw} shows a quantitative comparison of the three cameras in the NUS dataset.
For a fair comparison, we evaluate the performance of our method on the fully reconstructed raw-RGB images.
As can be seen, our method outperforms the baseline methods by a large margin after fine-tuning.
The PSNRs of our method without fine-tuning are also higher than those of the baselines on the Samsung and Olympus data.
The RIR achieved high performance on the Sony data, while the method is the worst on the Samsung and Olympus data.
We speculate that the Sony camera's raw-RGB-to-sRGB mapping has little effect on local processing.
Thus, the de-rendering is modeled well by a global approach.
Nevertheless, our method achieves the best after fine-tuning.
Even though the fine-tuning heavily relies on a very small subset of raw-RGB pixels, the training is generally effective.
We attribute this to the inductive bias of self-similarity in CNNs~\cite{Ulyanov:2018:DIP}.
As convolution filters are shared in all spatial locations, the training signals at sparse locations can be propagated to neighborhood pixels.

\fref{fig:qualitative} shows a qualitative comparison.
From top to bottom, we show two results each on Samsung NX2000, Olympus E-PL6, and Sony SLT-A57, respectively.
The baseline results have high errors, mostly on edges, as their models are not sophisticated.
The RIR relies on global operators of an ISP, but the edges are usually processed further by local tone mappings.
Even though the SAM can model spatially varying color mappings, its uniform sampling is not enough to store complex information around edges.
In contrast, our method adaptively samples raw-RGB values according to the scene structure, improving the overall performance.

\subsection{Discussion}
\paragraph{Uniform vs. random vs. content-aware sampling.}
\begin{table}
    \begin{center}
    \begin{tabular}{c|c|c|c}
    \toprule
    Method & Fine-tuning & PSNR & SSIM\\
    \hline
    No metadata & N/A & 47.67 & 0.9913 \\
    \hline
    Uniform & \multirow{3}{*}{No} & 49.58 & 0.9940 \\
    Random & & 49.68 & 0.9940 \\
    Ours & & \textbf{50.64} & \textbf{0.9942} \\
    \hline
    Uniform & \multirow{3}{*}{Yes} & 52.59 & 0.9960 \\ 
    Random & & 52.55 & 0.9957 \\
    Ours & & \textbf{53.32} & \textbf{0.9961}\\
    \bottomrule
    \end{tabular}
    \end{center}
    \caption{Comparison of different sampling methods. All methods share the same reconstruction network.}
    \label{table:comparison_sampling}
\end{table}
\begin{figure}
    \centering
    \setlength{\tabcolsep}{1pt}
    \begin{tabular}{ccccc}
        \includegraphics[width=0.22\linewidth]{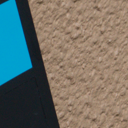} & \includegraphics[width=0.22\linewidth]{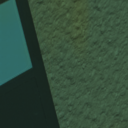}\llap{{\setlength{\fboxsep}{0pt}\setlength{\fboxrule}{2pt}\fcolorbox{red}{yellow}{\includegraphics[width=0.18\linewidth]{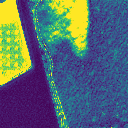}}}} & \includegraphics[width=0.22\linewidth]{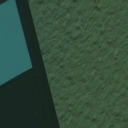}\llap{{\setlength{\fboxsep}{0pt}\setlength{\fboxrule}{2pt}\fcolorbox{red}{yellow}{\includegraphics[width=0.18\linewidth]{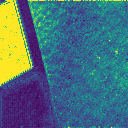}}}} & \includegraphics[width=0.22\linewidth]{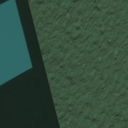}\llap{{\setlength{\fboxsep}{0pt}\setlength{\fboxrule}{2pt}\fcolorbox{red}{yellow}{\includegraphics[width=0.18\linewidth]{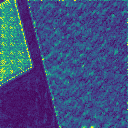}}}} & \includegraphics[width=0.035\linewidth]{figures_arxiv/colorbar.pdf} \\
        & {\setlength{\fboxsep}{0pt}\setlength{\fboxrule}{0.5pt}\fcolorbox{black}{yellow}{\includegraphics[width=0.22\linewidth]{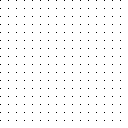}}} & {\setlength{\fboxsep}{0pt}\setlength{\fboxrule}{0.5pt}\fcolorbox{black}{yellow}{\includegraphics[width=0.22\linewidth]{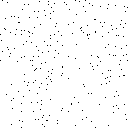}}} & {\setlength{\fboxsep}{0pt}\setlength{\fboxrule}{0.5pt}\fcolorbox{black}{yellow}{\includegraphics[width=0.22\linewidth]{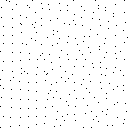}}} & \\
        {\small Input} & {\small Uniform} & {\small Random} & {\small Ours} & \\
    \end{tabular}
    \caption{Comparison of different samplings. The top and bottom rows show the error maps and sampling maps, respectively.}
    \label{fig:comp_sampling}
\end{figure}
In~\Tref{table:comparison_sampling}, we compare different sampling approaches, including uniform and random sampling. We also compare against a baseline where there is no sampling i.e., the reconstruction network is provided no additional metadata.
We average the PSNR and SSIM results of patches from the three cameras.
Uniform sampling chooses samples from a 2D grid of the image, while random sampling samples $k\%$ of pixels randomly in an image.
Both the samplings are independent of the image content.
As shown in the table, exploiting metadata significantly improves the quality of raw reconstruction regardless of sampling methods, which demonstrates that it is beneficial to save a small number of raw-RGB pixel values in the metadata of an sRGB image.
The uniform and random samplings are simple but effective approaches as they choose samples evenly in the entire space of pixels in an image.
As shown in~\fref{fig:comp_sampling}, however, the performance is limited since they are unable to select the samples particularly useful for reconstruction.
On the other hand, our content-aware sampling outperforms the simple approaches by a large margin because our method not only samples pixels evenly but also considers their effectiveness on reconstruction. An ablation on the sampling rate $k$ is provided in Section~\ref{sec:other_sampling}.

\noindent \textbf{Ablation study.}
\begin{table}
    \begin{center}
    \setlength{\tabcolsep}{10pt}
    \begin{tabular}{c|c|c}
    \toprule
    Method & PSNR & SSIM\\
    \hline
    sRGB & 49.10 & 0.9927 \\
    RAW & 49.72 & 0.9933 \\
    sRGB + RAW & \textbf{50.15} & \textbf{0.9943} \\
    \hline
    Free-form max-pooling & 47.75 & 0.9911 \\
    Superpixel max-pooling & \textbf{50.15} & \textbf{0.9943} \\
    \hline
    W/o meta loss & 50.15 & 0.9943 \\ 
    W/o meta loss + fine-tuning & 53.07 & 0.9959 \\ 
    W/ meta loss & 50.64 & 0.9942 \\
    W/ meta loss + fine-tuning & \textbf{53.32} & \textbf{0.9961} \\
    \bottomrule
    \end{tabular}
    \end{center}
    \caption{Ablation study on our method. We compare different inputs fed into the sampling process and different pooling approaches without using a meta loss. We also analyze the effectiveness of the meta loss.}
    \label{table:ablation_study}
\end{table}
\begin{figure}
    \centering
    \setlength{\tabcolsep}{1pt}
    \begin{tabular}{cccc}
        \includegraphics[width=0.26\linewidth]{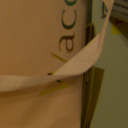} & \includegraphics[width=0.26\linewidth]{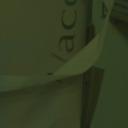}\llap{{\setlength{\fboxsep}{0pt}\setlength{\fboxrule}{2pt}\fcolorbox{red}{yellow}{\includegraphics[width=0.16\linewidth]{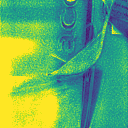}}}} & \includegraphics[width=0.26\linewidth]{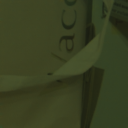}\llap{{\setlength{\fboxsep}{0pt}\setlength{\fboxrule}{2pt}\fcolorbox{red}{yellow}{\includegraphics[width=0.16\linewidth]{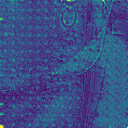}}}} & \includegraphics[width=0.03\linewidth]{figures_arxiv/colorbar.pdf} \\
        & {\setlength{\fboxsep}{0pt}\setlength{\fboxrule}{0.5pt}\fcolorbox{black}{yellow}{\includegraphics[width=0.26\linewidth]{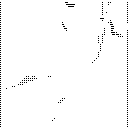}}} & {\setlength{\fboxsep}{0pt}\setlength{\fboxrule}{0.5pt}\fcolorbox{black}{yellow}{\includegraphics[width=0.26\linewidth]{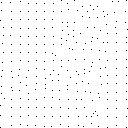}}} & \\
        {\small Input} & {\small Free-form} & {\small Superpixel} &  \\
    \end{tabular}
    \caption{Ablation study on the superpixel max-pooling. The two rows show error maps and sampling masks, respectively.}
    \label{fig:ablation_superpixel}
\end{figure}
To demonstrate individual components in our method, we conduct an ablation study, as shown in~\Tref{table:ablation_study}.
We first try different inputs to the sampler network and superpixel loss: sRGB image, raw-RGB image, and both images.
As can be seen, it is beneficial to use raw-RGB images as input and then use sRGB images to find samples from the raw-RGB images.  However, the network achieves the highest scores when using both images, indicating that the sRGB images still provide useful information for better sampling when both images are jointly used.

We also compare our superpixel-based sampling with a naive sampling approach.
For the free-form max-pooling in the table, we train a sampler network with a single-channel sigmoid output and extract the top $k\%$ of pixels that have large sigmoid values.
As shown in the table, the performance of the free-form sampling significantly degrades compared with the superpixel-based sampling.
\fref{fig:ablation_superpixel} shows a qualitative comparison.
Since the free-form sampling is unconstrained, most samples are clustered in the same region.   In contrast, our superpixel-based sampling enables the network to sample pixels at various locations while covering the full spatial range of the image.

Lastly, we conduct an ablation study on the meta loss. As shown in~\Tref{table:ablation_study}, the meta loss improves the performance after fine-tuning and the direct result from the network, which demonstrates that the loss forces the reconstruction network to learn its weights generalizable to unseen test cases.
With the loss, the network can improve its performance over 3dB compared to the direct network output without the loss using sparse raw-RGB samples.

\noindent \textbf{Limitations.}
Unlike uniform and random sampling, our method is required to run a deep neural network to sample pixels at capture time, which is an additional computational cost on the device. We have not tested edge cases, such as high compression, very noisy low-light images, or under/overexposure issues in raw.  Since convolutional neural networks generally fit well with natural images that are spatially smooth, it is still unclear how well our reconstruction network can process sparse sampling masks.
Investigating efficient deep architectures for sparse samples is an interesting direction of research in the future.

\subsection{Other application: Bit-Depth Recovery}
\begin{table}
    \begin{center}
    \setlength{\tabcolsep}{10pt}
    \begin{tabular}{c|c|c}
    \toprule
    Method & PSNR & SSIM\\
    \hline
    CA~\cite{Wan:2012:CA} & 34.74 & 0.9317 \\
    ACDC~\cite{Wan:2016:ACDC} & 34.68 & 0.9152 \\
    IPAD~\cite{Liu:2018:IPAD} & 34.91 & 0.9345 \\
    BitNet~\cite{Byun:2018:BitNet} & 38.48 & 0.9657 \\
    BE-CALF~\cite{Liu:2019:BECALF} & 38.94 & 0.9680 \\
    \hline
    Ours & 39.57 & 0.9719 \\
    Ours + fine-tuning & \textbf{39.73} & \textbf{0.9721} \\
    \bottomrule
    \end{tabular}
    \end{center}
    \caption{Quantitative comparison of bit-depth recovery (4-to-8-bit) methods on the Kodak dataset~\cite{Kodak:1999}.}
    \label{table:bitdepth}
\end{table}
\begin{figure}
    \centering
    \setlength{\tabcolsep}{1pt}
    \begin{tabular}{cccc}
        \includegraphics[width=0.24\linewidth]{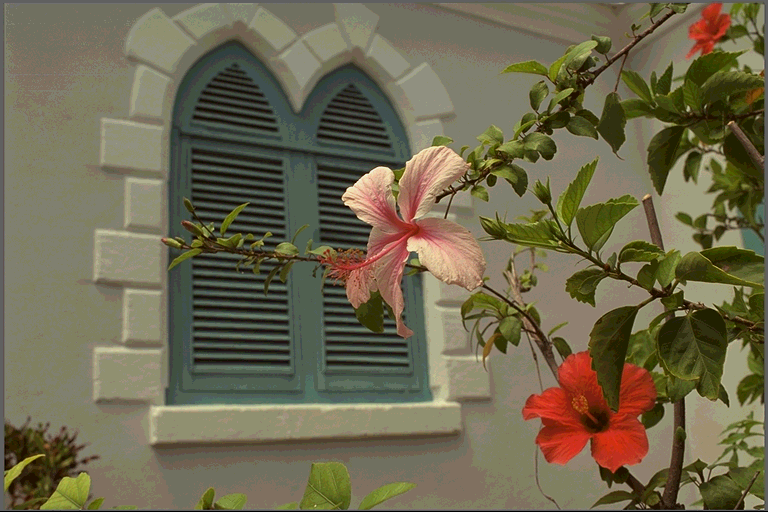}\llap{{\setlength{\fboxsep}{0pt}\setlength{\fboxrule}{2pt}\fcolorbox{red}{yellow}{\includegraphics[width=0.17\linewidth,clip,trim=360 400 300 30]{figures_arxiv/bitdepth/0000006_in.png}}}} & \includegraphics[width=0.24\linewidth]{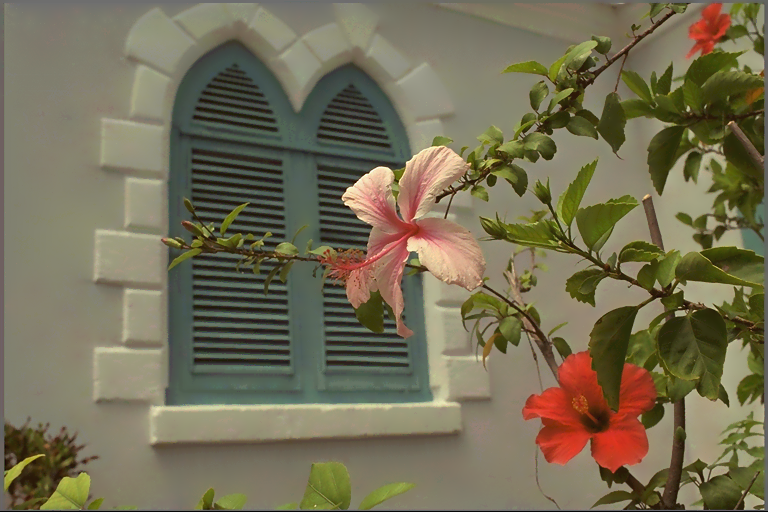}\llap{{\setlength{\fboxsep}{0pt}\setlength{\fboxrule}{2pt}\fcolorbox{red}{yellow}{\includegraphics[width=0.17\linewidth,clip,trim=360 400 300 30]{figures_arxiv/bitdepth/kodim07_4_8_ca.png}}}} & \includegraphics[width=0.24\linewidth]{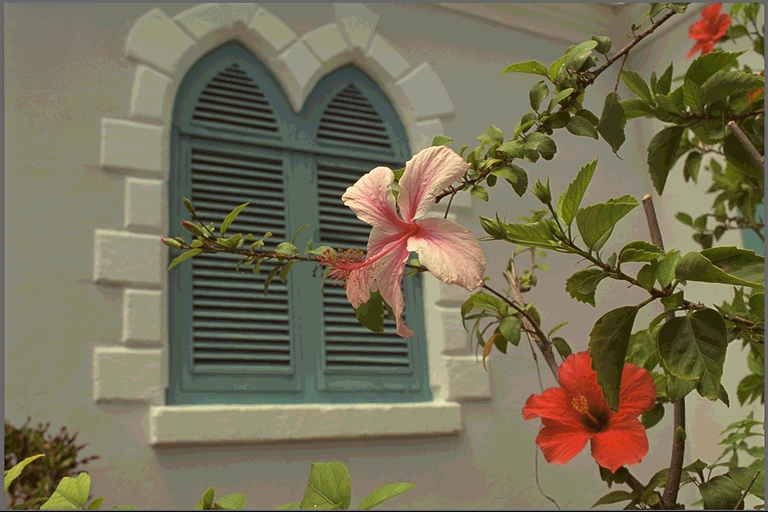}\llap{{\setlength{\fboxsep}{0pt}\setlength{\fboxrule}{2pt}\fcolorbox{red}{yellow}{\includegraphics[width=0.17\linewidth,clip,trim=360 400 300 30]{figures_arxiv/bitdepth/kodim07_4_8_acdc.png}}}} & \includegraphics[width=0.24\linewidth]{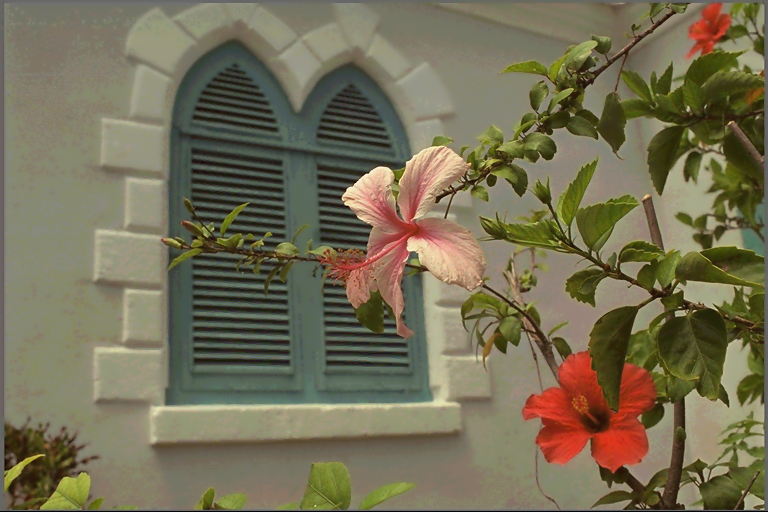}\llap{{\setlength{\fboxsep}{0pt}\setlength{\fboxrule}{2pt}\fcolorbox{red}{yellow}{\includegraphics[width=0.17\linewidth,clip,trim=360 400 300 30]{figures_arxiv/bitdepth/kodim07_4_8_ipad.png}}}} \\
        {\small Input} & {\small CA~\cite{Wan:2012:CA}} & {\small ACDC~\cite{Wan:2016:ACDC}} & {\small IPAD~\cite{Liu:2018:IPAD}} \\
        \includegraphics[width=0.24\linewidth]{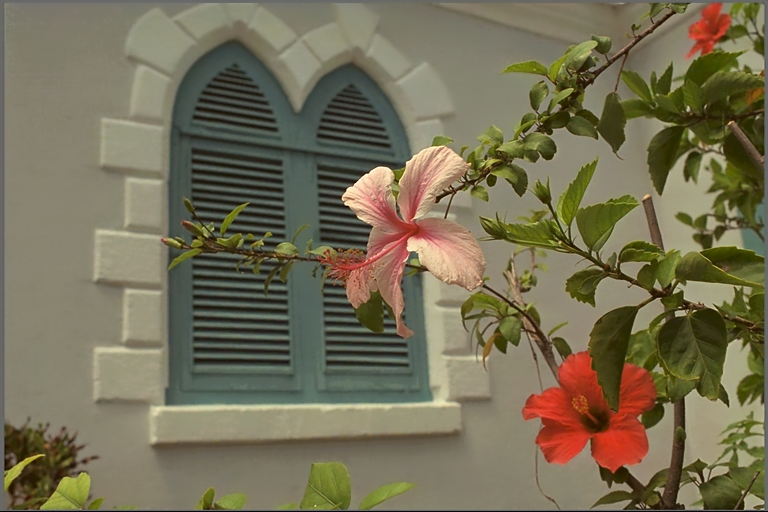}\llap{{\setlength{\fboxsep}{0pt}\setlength{\fboxrule}{2pt}\fcolorbox{red}{yellow}{\includegraphics[width=0.17\linewidth,clip,trim=360 400 300 30]{figures_arxiv/bitdepth/kodim07_4_8_bitnet.png}}}} & \includegraphics[width=0.24\linewidth]{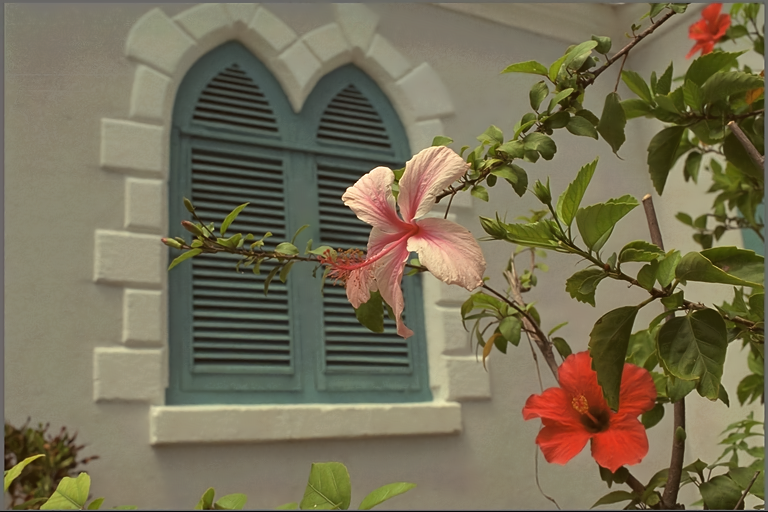}\llap{{\setlength{\fboxsep}{0pt}\setlength{\fboxrule}{2pt}\fcolorbox{red}{yellow}{\includegraphics[width=0.17\linewidth,clip,trim=360 400 300 30]{figures_arxiv/bitdepth/kodim07_4_8_becalf.png}}}} & \includegraphics[width=0.24\linewidth]{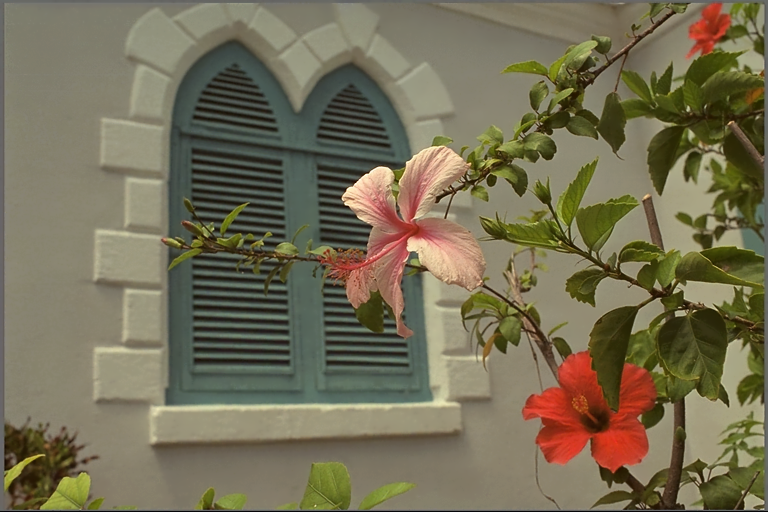}\llap{{\setlength{\fboxsep}{0pt}\setlength{\fboxrule}{2pt}\fcolorbox{red}{yellow}{\includegraphics[width=0.17\linewidth,clip,trim=360 400 300 30]{figures_arxiv/bitdepth/0000006_out_finetune.png}}}} & \includegraphics[width=0.24\linewidth]{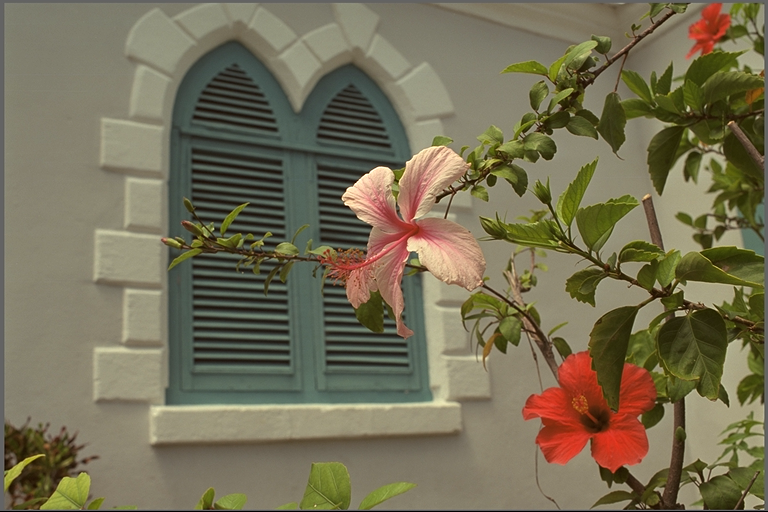}\llap{{\setlength{\fboxsep}{0pt}\setlength{\fboxrule}{2pt}\fcolorbox{red}{yellow}{\includegraphics[width=0.17\linewidth,clip,trim=360 400 300 30]{figures_arxiv/bitdepth/0000006_gt.png}}}} \\
        {\small BitNet~\cite{Byun:2018:BitNet}} & {\small BE-CALF~\cite{Liu:2019:BECALF}} & {\small Ours + FT} & {\small Ground truth} \\

        \includegraphics[width=0.24\linewidth]{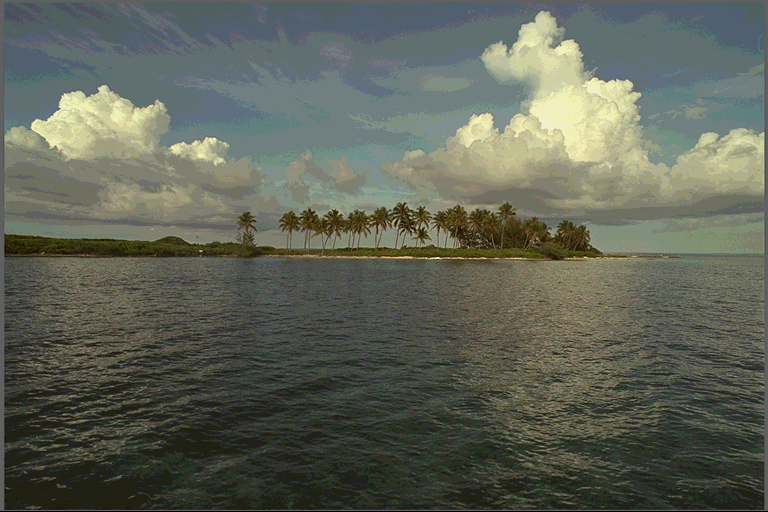}\llap{{\setlength{\fboxsep}{0pt}\setlength{\fboxrule}{2pt}\fcolorbox{red}{yellow}{\includegraphics[width=0.17\linewidth,clip,trim=360 365 250 30]{figures_arxiv/bitdepth/0000015_in.png}}}} & \includegraphics[width=0.24\linewidth]{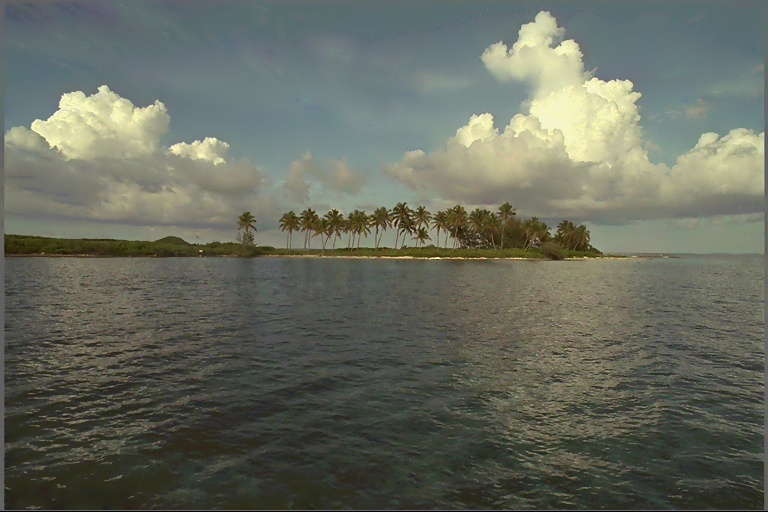}\llap{{\setlength{\fboxsep}{0pt}\setlength{\fboxrule}{2pt}\fcolorbox{red}{yellow}{\includegraphics[width=0.17\linewidth,clip,trim=360 365 250 30]{figures_arxiv/bitdepth/kodim16_4_8_ca.png}}}} & \includegraphics[width=0.24\linewidth]{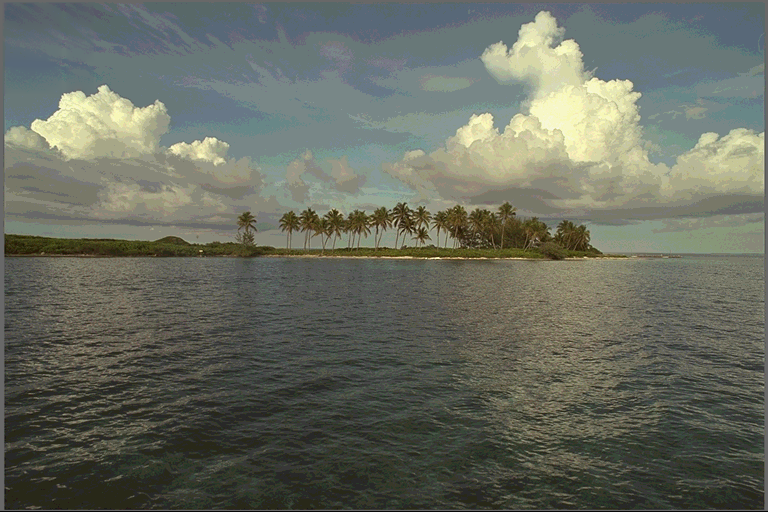}\llap{{\setlength{\fboxsep}{0pt}\setlength{\fboxrule}{2pt}\fcolorbox{red}{yellow}{\includegraphics[width=0.17\linewidth,clip,trim=360 365 250 30]{figures_arxiv/bitdepth/kodim16_4_8_acdc.png}}}} & \includegraphics[width=0.24\linewidth]{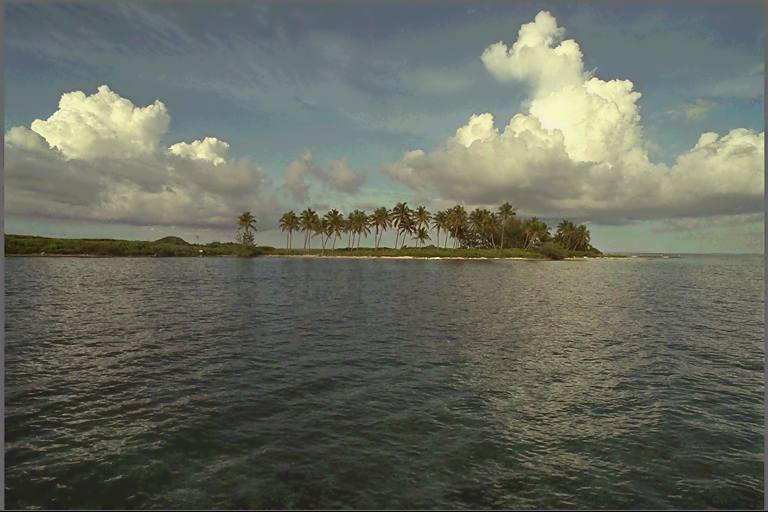}\llap{{\setlength{\fboxsep}{0pt}\setlength{\fboxrule}{2pt}\fcolorbox{red}{yellow}{\includegraphics[width=0.17\linewidth,clip,trim=360 365 250 30]{figures_arxiv/bitdepth/kodim16_4_8_ipad.png}}}} \\
        {\small Input} & {\small CA~\cite{Wan:2012:CA}} & {\small ACDC~\cite{Wan:2016:ACDC}} & {\small IPAD~\cite{Liu:2018:IPAD}} \\
        \includegraphics[width=0.24\linewidth]{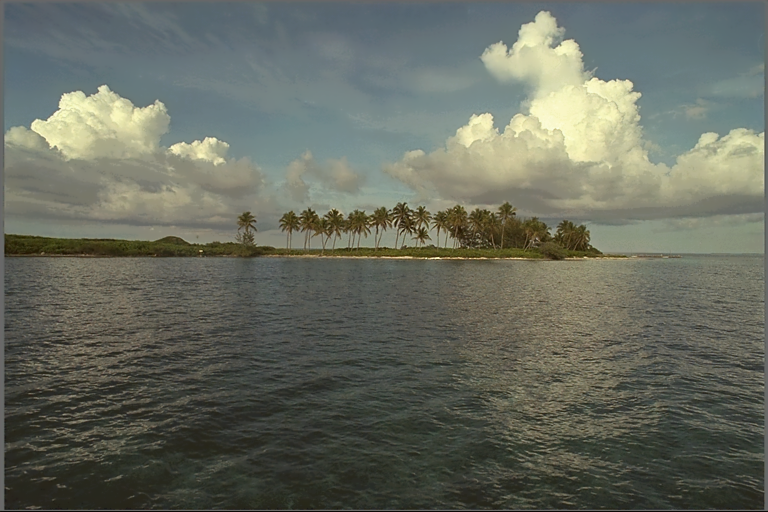}\llap{{\setlength{\fboxsep}{0pt}\setlength{\fboxrule}{2pt}\fcolorbox{red}{yellow}{\includegraphics[width=0.17\linewidth,clip,trim=360 365 250 30]{figures_arxiv/bitdepth/kodim16_4_8_bitnet.png}}}} & \includegraphics[width=0.24\linewidth]{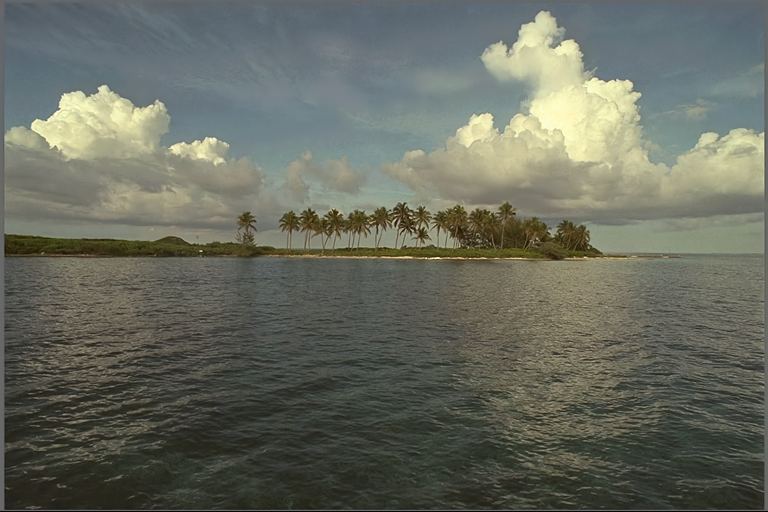}\llap{{\setlength{\fboxsep}{0pt}\setlength{\fboxrule}{2pt}\fcolorbox{red}{yellow}{\includegraphics[width=0.17\linewidth,clip,trim=360 365 250 30]{figures_arxiv/bitdepth/kodim16_4_8_becalf.png}}}} & \includegraphics[width=0.24\linewidth]{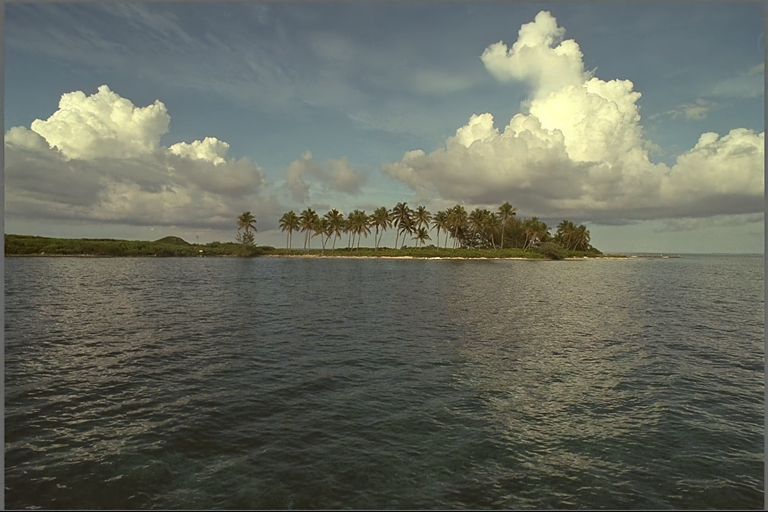}\llap{{\setlength{\fboxsep}{0pt}\setlength{\fboxrule}{2pt}\fcolorbox{red}{yellow}{\includegraphics[width=0.17\linewidth,clip,trim=360 365 250 30]{figures_arxiv/bitdepth/0000015_out_finetune.png}}}} & \includegraphics[width=0.24\linewidth]{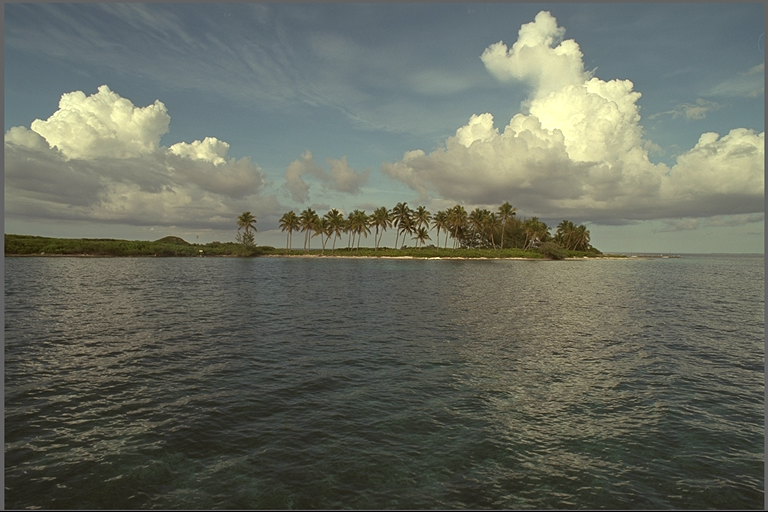}\llap{{\setlength{\fboxsep}{0pt}\setlength{\fboxrule}{2pt}\fcolorbox{red}{yellow}{\includegraphics[width=0.17\linewidth,clip,trim=360 365 250 30]{figures_arxiv/bitdepth/0000015_gt.png}}}} \\
        {\small BitNet~\cite{Byun:2018:BitNet}} & {\small BE-CALF~\cite{Liu:2019:BECALF}} & {\small Ours + FT} & {\small Ground truth} \\
    \end{tabular}
    \caption{Qualitative comparison of bit-depth recovery algorithms on the Kodak dataset~\cite{Kodak:1999}. Zoom in for better visibility.}
    \label{fig:qualitative_bitdepth}
\end{figure}
Our framework for sampling and reconstruction can also be applied to the problem of bit-depth recovery. To test on this task, we use two publicly available datasets for training: MIT-Adobe 5K~\cite{Bychkovsky:2011:5K} and Sintel~\cite{Butler:2012:Sintel}.
We synthesize pairs of 4-bit and 8-bit images to train a 4-to-8-bit recovery task.
We use the ~1.5\% of pixels of an 8-bit image as metadata for our method and run the update steps for fine-tuning.

\Tref{table:bitdepth} and \fref{fig:qualitative_bitdepth} show a quantitative and qualitative comparison with blind bit-depth recovery approaches.
We evaluate all methods on the Kodak dataset~\cite{Kodak:1999}.
As expected, our method outperforms the baselines using a small number of pixels in the metadata.
Note that our method is not explicitly designed for bit-depth recovery.
Our reconstruction network is a generic U-Net, while BitNet~\cite{Byun:2018:BitNet} and BE-CALF~\cite{Liu:2019:BECALF} use networks particularly designed for the task.
This experiment demonstrates the applicability of our metadata framework to other image processing tasks.

\section{Conclusion}
We have presented a method for sRGB image de-rendering that recovers the original raw-RGB images with the assistance of a small amount of metadata that is sampled from the raw-RGB image at capture time.  Our approach learns both the sampling and reconstruction network in an end-to-end manner. Moreover, we train the reconstruction network to lend itself for further fine-tuning from the sparse metadata samples.   We show significant improvements over the existing state-of-the-art approaches that also use metadata.  Finally, we use our framework for the related task of bit-depth recovery and show compelling results.

\vspace*{-0.25cm}
\section*{Acknowledgments}
{\footnotesize
This work was done as part of an internship at the Samsung AI Center in Toronto, Canada.  Seonghyeon's internship was funded by a Mitacs Accelerate. Seonghyeon's research fellow funding came in part from the Canada First Research Excellence Fund for the Vision: Science to Applications (VISTA) programme and an NSERC Discovery Grant.
}

{\small
\bibliographystyle{ieee_fullname}
\bibliography{reference}
}

\clearpage


\setcounter{table}{0}
\setcounter{figure}{0}
\renewcommand{\thetable}{A\arabic{table}}
\renewcommand{\thefigure}{A\arabic{figure}}

\begin{appendices}

\onecolumn

\section{Results on camera ISP images}
\label{sec:nus_jpg}
In Section~\ref{sec:expts}, we had evaluated our raw reconstruction accuracy using the NUS~\cite{Cheng:2014:NUS} dataset with the sRGB images rendered using a software ISP emulator~\cite{Karaimer:2016:SoftwareISP}. The NUS dataset also contains the sRGB-JPEG images rendered by each individual camera's hardware ISP. We used these sRGB images instead of the software ISP emulator~\cite{Karaimer:2016:SoftwareISP}, and the results are presented in Table~\ref{table:jpg}. Our method generalizes well to different ISPs, and outperforms competitors. We do note that compared to Table~\ref{table:quant_raw}, there is a drop in performance for all methods due to the more complex ISPs. 

\section{Other sampling rates} 
\label{sec:other_sampling} We report PSNR (dB) for our method (with fine-tuning) at different sampling rates in Table~\ref{table:sampling}.
The results for $k=1.5\%$ are reproduced from Table~\ref{table:quant_raw}.
There is a significant improvement from 0\% i.e., no metadata, to 0.4\%. Performance improves with higher $k$ values but at the expense of larger metadata size.

\section{Comparison with SLIC}
We performed an experiment where we replaced our learned superpixel with SLIC~\cite{Achanta:2012:SLIC} for sampling, and trained our reconstruction network under the same settings. On the Samsung camera, we obtained PSNR/SSIM values of 45.94 / 0.9958 as against 49.57 / 0.9975 produced by our method, demonstrating the superiority of an end-to-end learnable superpixel sampler.

\section{Additional experiments}
In~\fref{fig:finetuning_supp}, we compare the error maps of our outputs before and after fine-tuning.
\fref{fig:supp_samsung}, \fref{fig:supp_olympus}, and \fref{fig:supp_sony} show additional qualitative results on three cameras.
For visibility, we omit output raw-RGB images and ground truth.
\fref{fig:vis_supp_1} to \fref{fig:vis_supp_3} show visualizations of learned superpixels and sampling masks.

\begin{table*}
   \begin{center}
   \setlength{\tabcolsep}{15pt}
   \begin{tabular}{c|c|c|c|c|c|c|c}
   \toprule
   \multirow{2}{*}{Method} & \multirow{2}{*}{Fine-tuning} & \multicolumn{2}{c|}{Samsung NX2000} & \multicolumn{2}{c|}{Olympus E-PL6} & \multicolumn{2}{c}{Sony SLT-A57} \\
   \cline{3-8}
   & & PSNR & SSIM & PSNR & SSIM & PSNR & SSIM \\
   \hline
   RIR~\cite{rang} & N/A & 37.62 & 0.9696 & 42.19 & 0.9865 & 45.22 & 0.9916 \\
   SAM~\cite{wacv} & N/A & 38.80 & 0.9725 & 43.15 & 0.9881 & 46.02 & 0.9921 \\
   \hline
   Ours & No & 40.80 & 0.9812 & 46.89 & 0.9938 & 48.51 & 0.9947 \\
   Ours & Yes & \textbf{41.59} & \textbf{0.9818} & \textbf{47.76} & \textbf{0.9944} & \textbf{49.58} & \textbf{0.9954} \\
   \bottomrule
   \end{tabular}
   \end{center}
   \caption{Quantitative evaluation on raw reconstruction using the camera ISP sRGB images.}
   \label{table:jpg}
\end{table*}

\begin{table}
   \begin{center}
   \setlength{\tabcolsep}{5pt}
   \begin{tabular}{c|c|c|c|c}
   \toprule
   Percentage Samples $k$ & 0\% & 0.4\% & 1.5\% & 6.25\% \\
   \hline
   Samsung NX2000 & 38.86 & 48.56 & 49.57 & 50.31 \\
   Olympus E-PL6 & 42.30 & 50.62 & 51.54 & 52.20  \\
   Sony SLT-A57 & 44.79 & 51.09 & 53.11 & 53.44 \\
   \bottomrule
   \end{tabular}
   \end{center}
   \caption{An ablation on the sampling rate $k$.}
   \label{table:sampling}
\end{table}

\begin{figure*}
    \centering
    \setlength{\tabcolsep}{1pt}
    \begin{tabular}{ccccc}
        \includegraphics[width=0.23\linewidth]{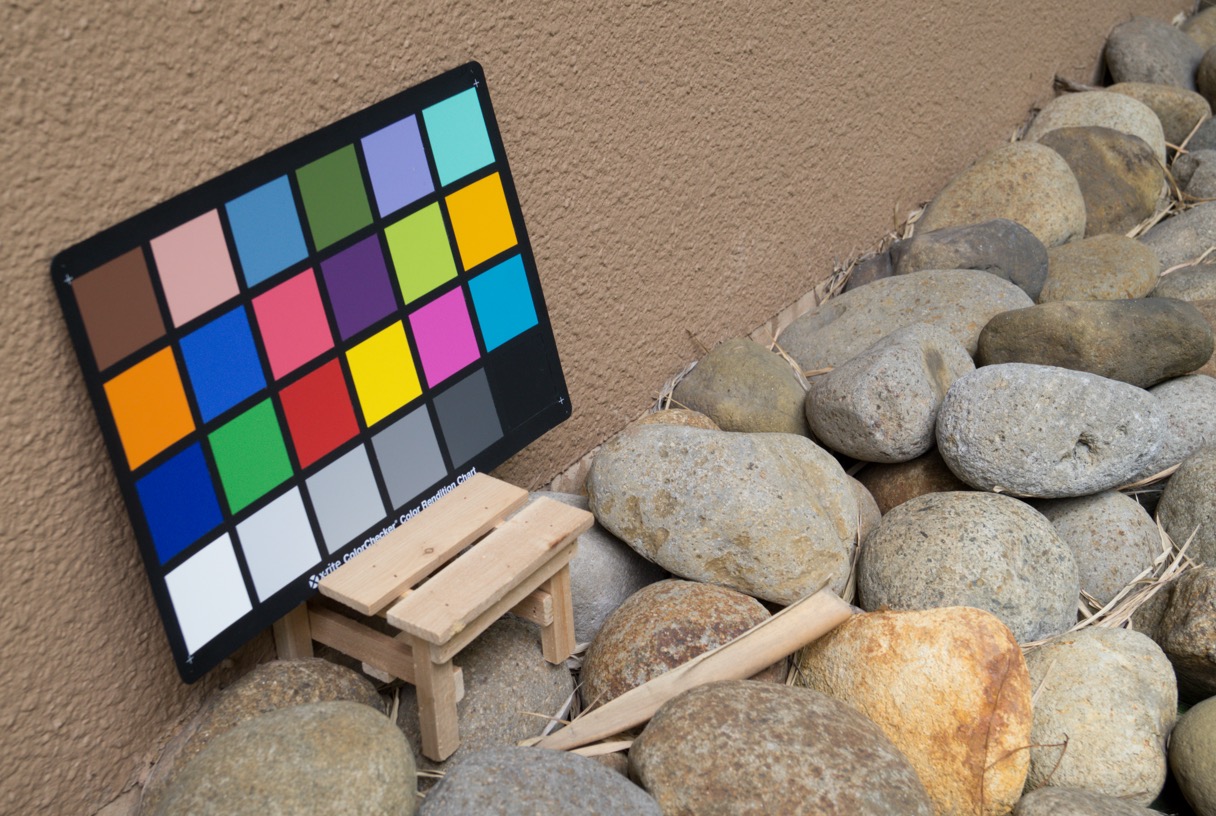} & \settototalheight{\dimen0}{\includegraphics[width=0.23\linewidth]{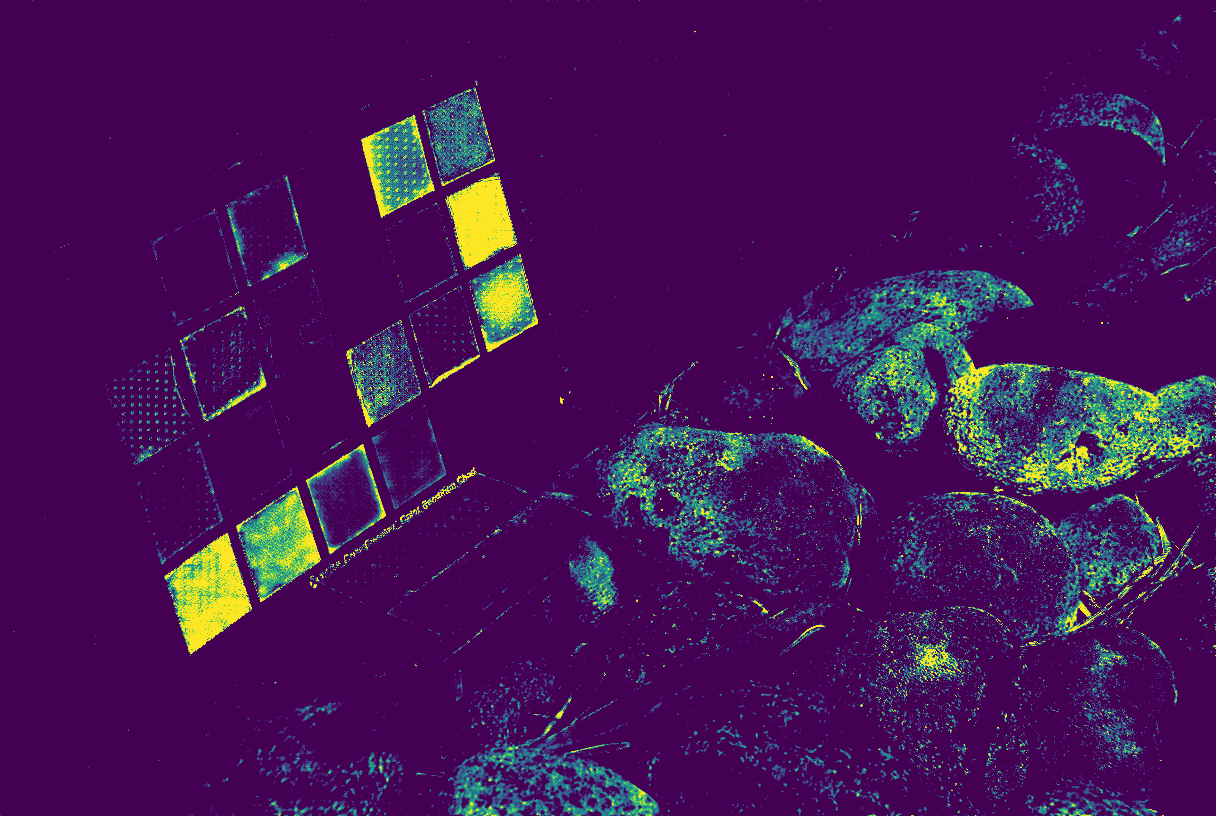}}\includegraphics[width=0.23\linewidth]{figures_arxiv/supp/finetuning/0000007_err_ours.png}\llap{\raisebox{\dimen0-7pt}{\setlength{\fboxsep}{2pt}\colorbox{white}{\scriptsize	 PSNR: 53.30dB}}} & \settototalheight{\dimen0}{\includegraphics[width=0.23\linewidth]{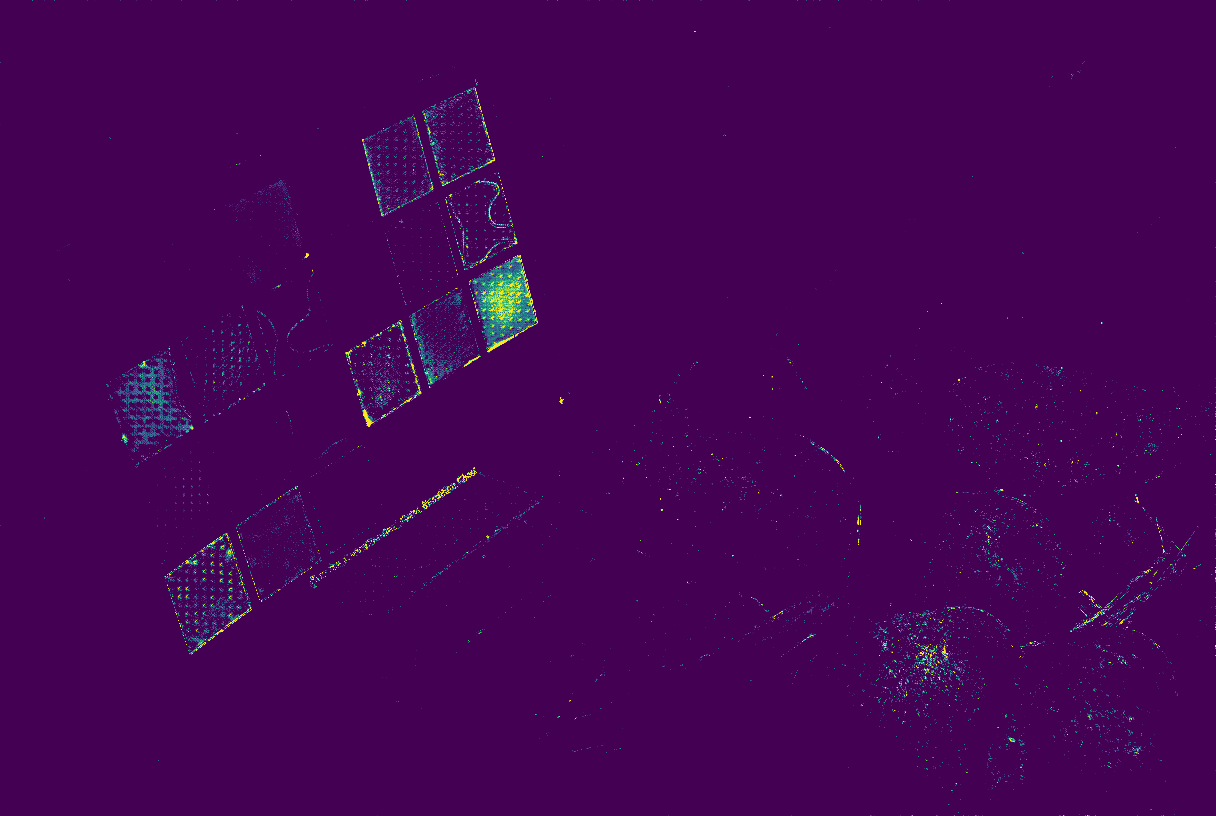}}\includegraphics[width=0.23\linewidth]{figures_arxiv/supp/finetuning/0000007_err_ours_finetuning.png}\llap{\raisebox{\dimen0-7pt}{\setlength{\fboxsep}{2pt}\colorbox{white}{\scriptsize	 PSNR: 58.31dB}}} & \includegraphics[width=0.025\linewidth]{figures_arxiv/colorbar.pdf} & \includegraphics[width=0.23\linewidth]{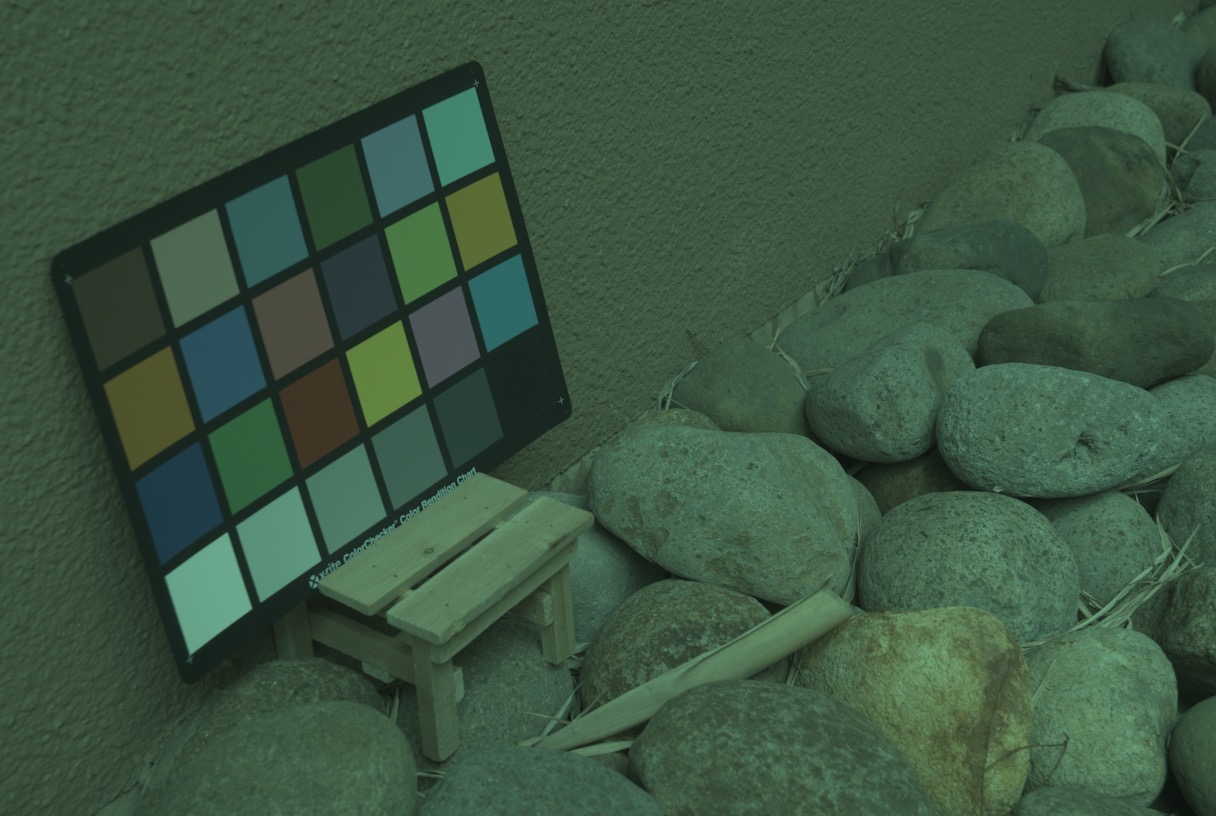}\\

        \includegraphics[width=0.23\linewidth]{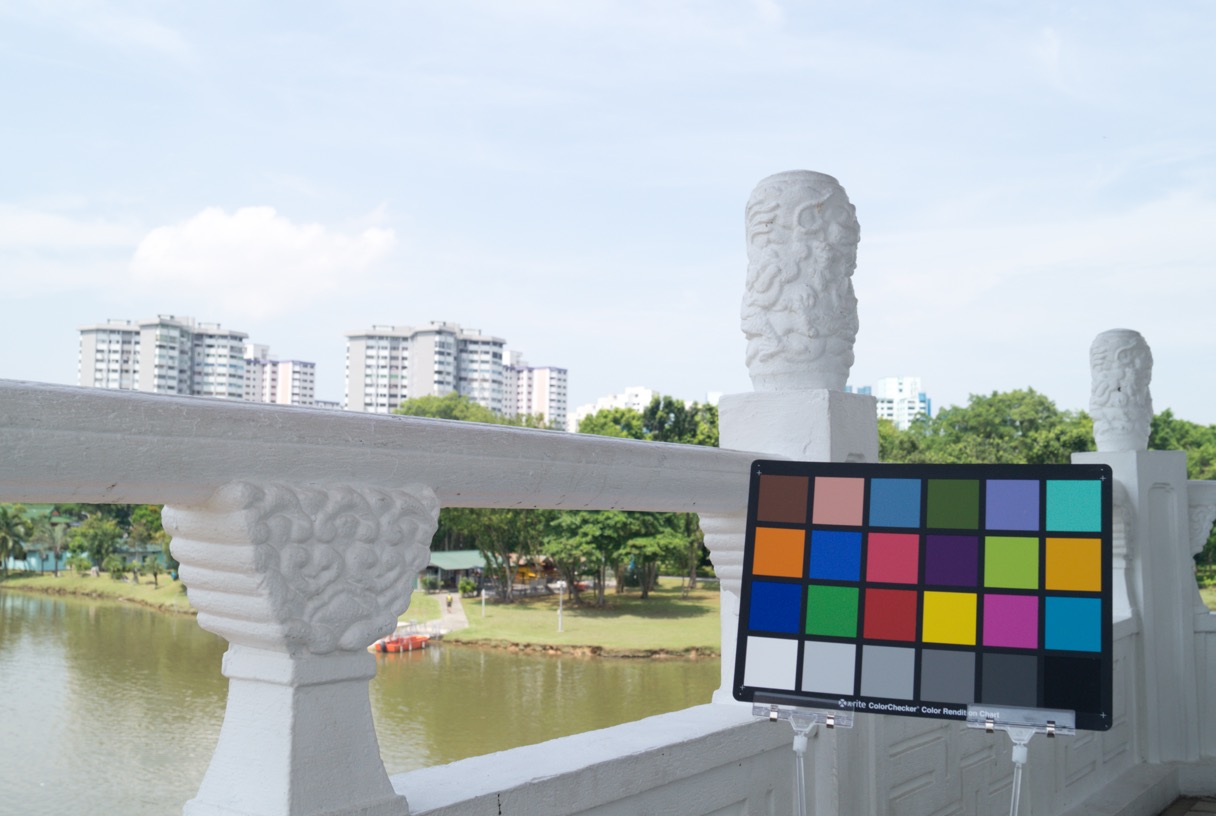} & \settototalheight{\dimen0}{\includegraphics[width=0.23\linewidth]{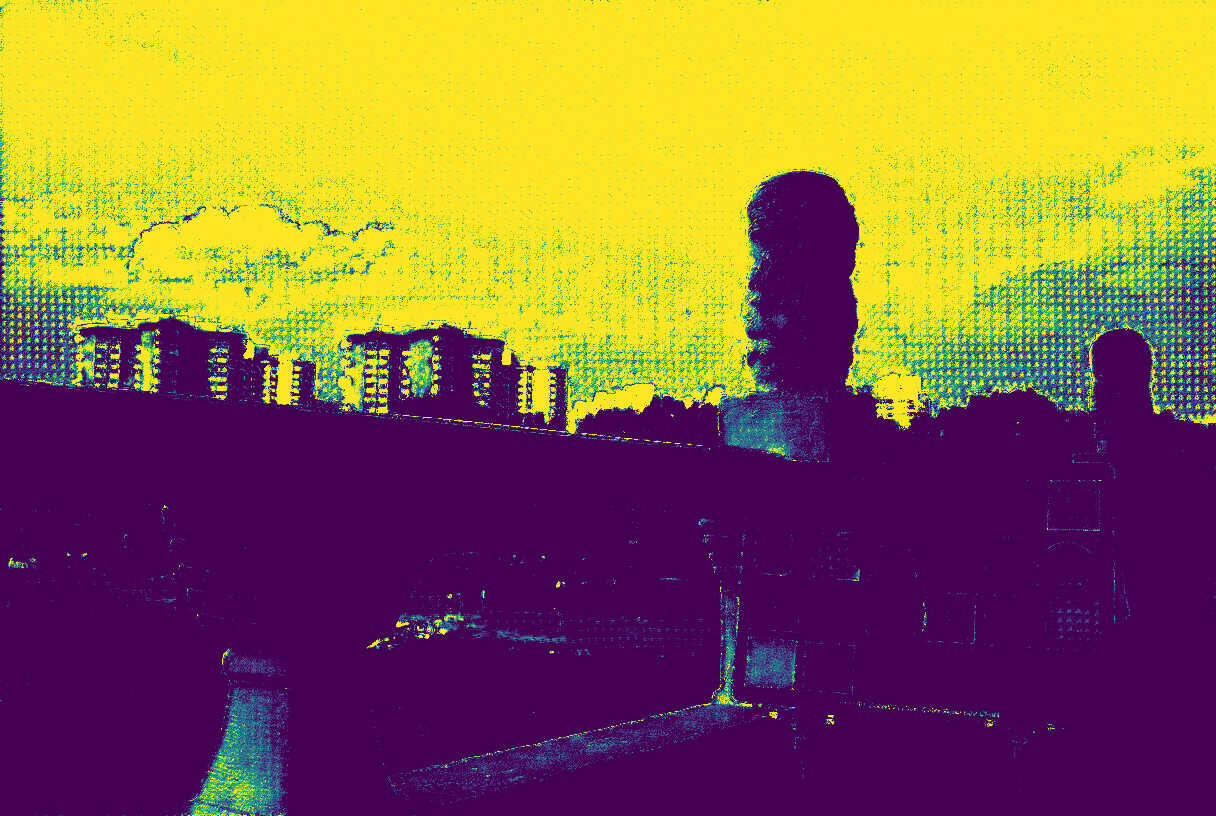}}\includegraphics[width=0.23\linewidth]{figures_arxiv/supp/finetuning/0000008_err_ours.png}\llap{\raisebox{\dimen0-7pt}{\setlength{\fboxsep}{2pt}\colorbox{white}{\scriptsize	 PSNR: 44.71dB}}} & \settototalheight{\dimen0}{\includegraphics[width=0.23\linewidth]{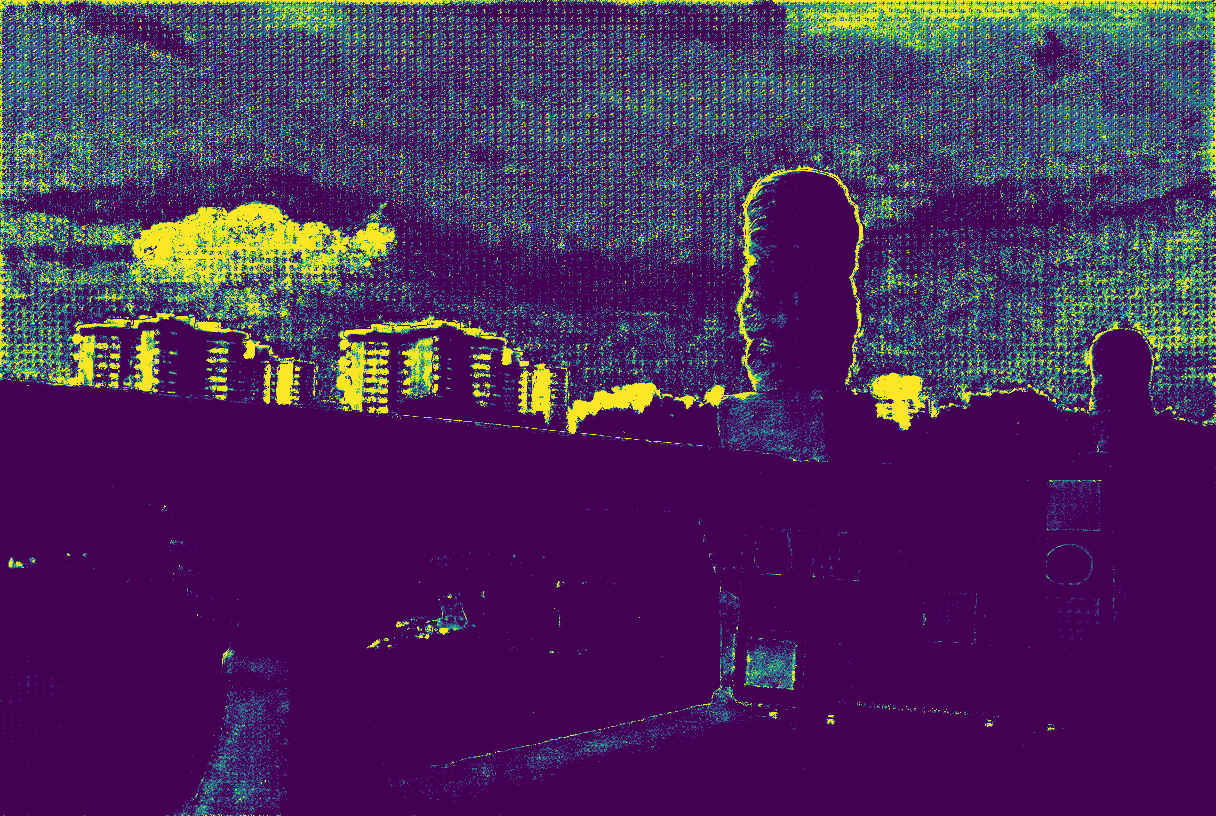}}\includegraphics[width=0.23\linewidth]{figures_arxiv/supp/finetuning/0000008_err_ours_finetuning.png}\llap{\raisebox{\dimen0-7pt}{\setlength{\fboxsep}{2pt}\colorbox{white}{\scriptsize	 PSNR: 48.39dB}}} & \includegraphics[width=0.025\linewidth]{figures_arxiv/colorbar.pdf} & \includegraphics[width=0.23\linewidth]{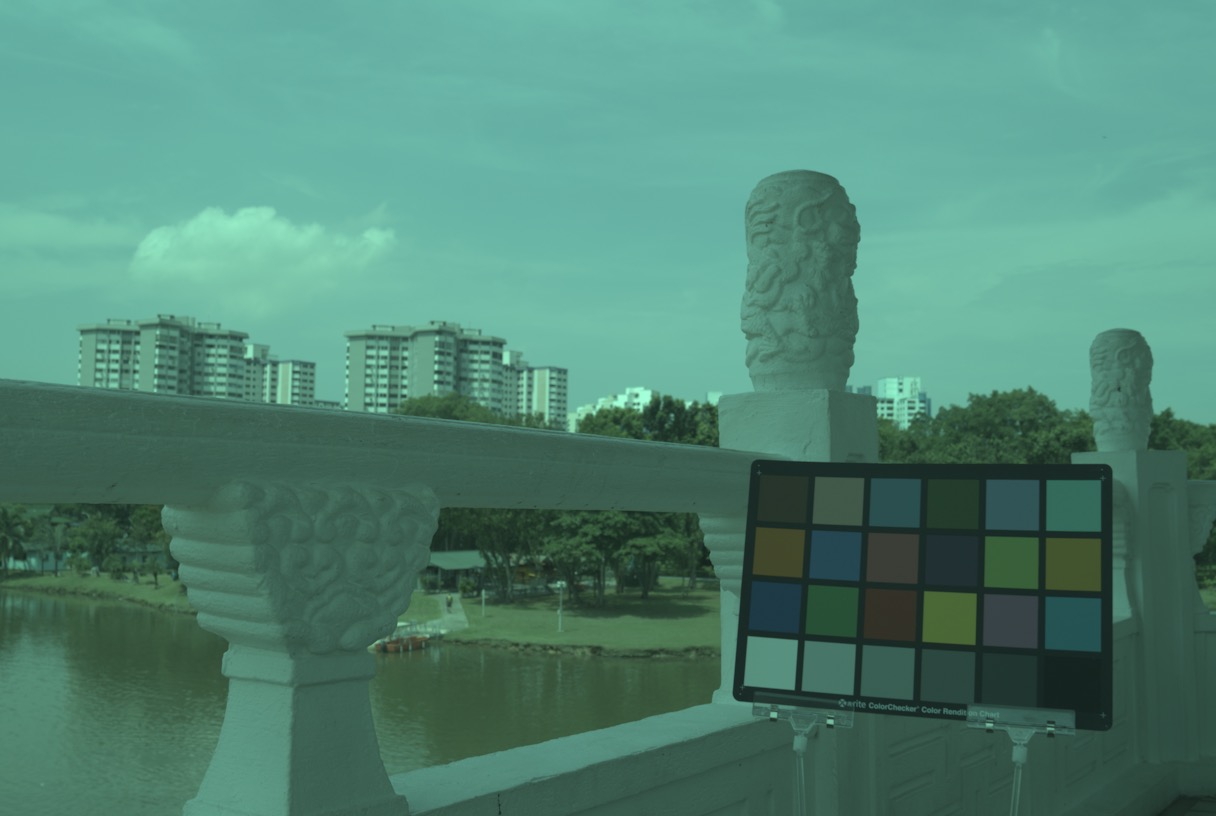}\\

        \includegraphics[width=0.23\linewidth]{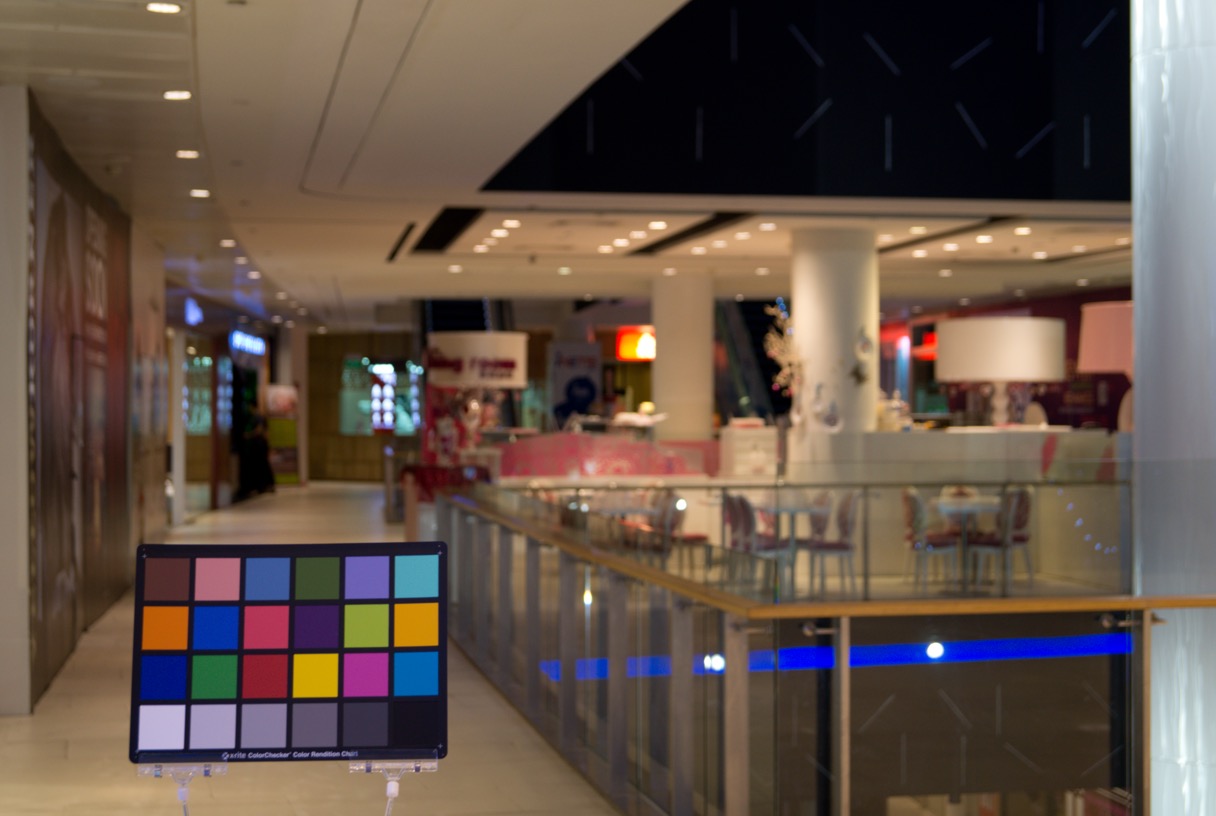} & \settototalheight{\dimen0}{\includegraphics[width=0.23\linewidth]{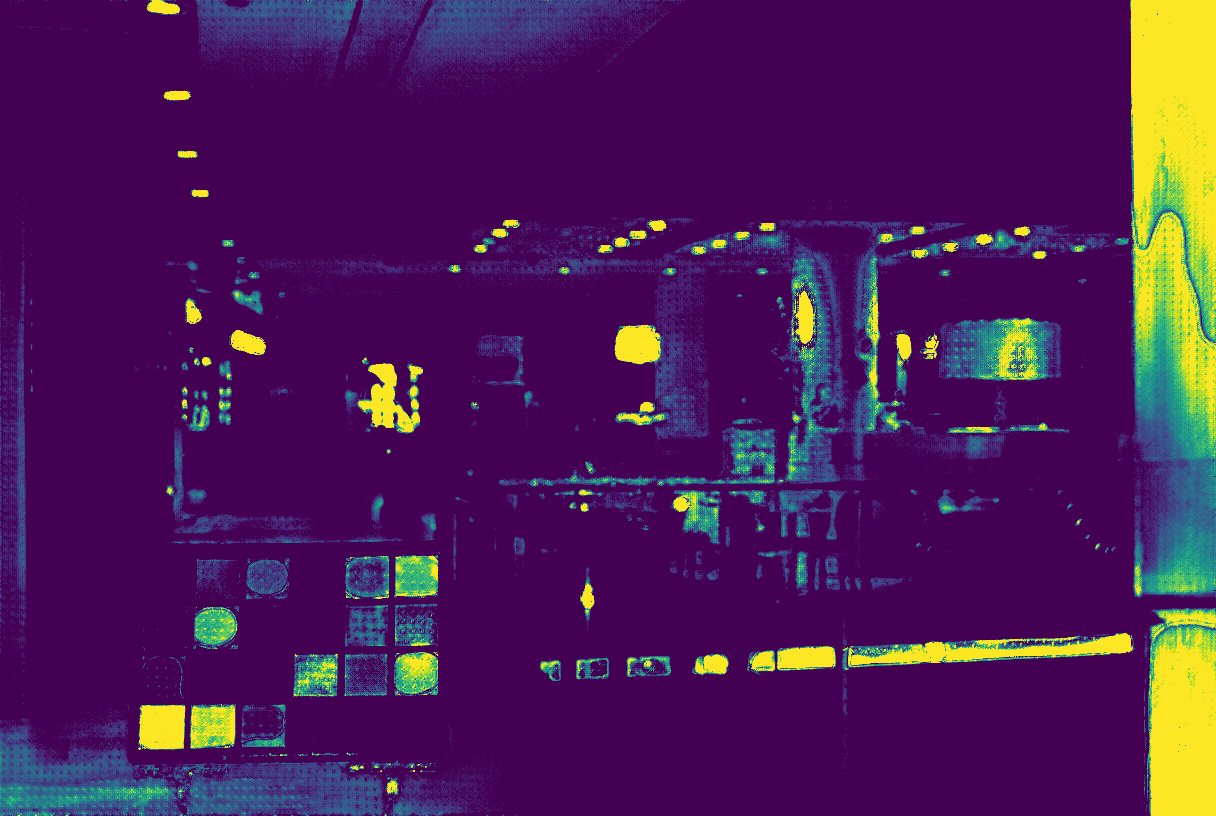}}\includegraphics[width=0.23\linewidth]{figures_arxiv/supp/finetuning/0000024_err_ours.png}\llap{\raisebox{\dimen0-7pt}{\setlength{\fboxsep}{2pt}\colorbox{white}{\scriptsize	 PSNR: 43.21dB}}} & \settototalheight{\dimen0}{\includegraphics[width=0.23\linewidth]{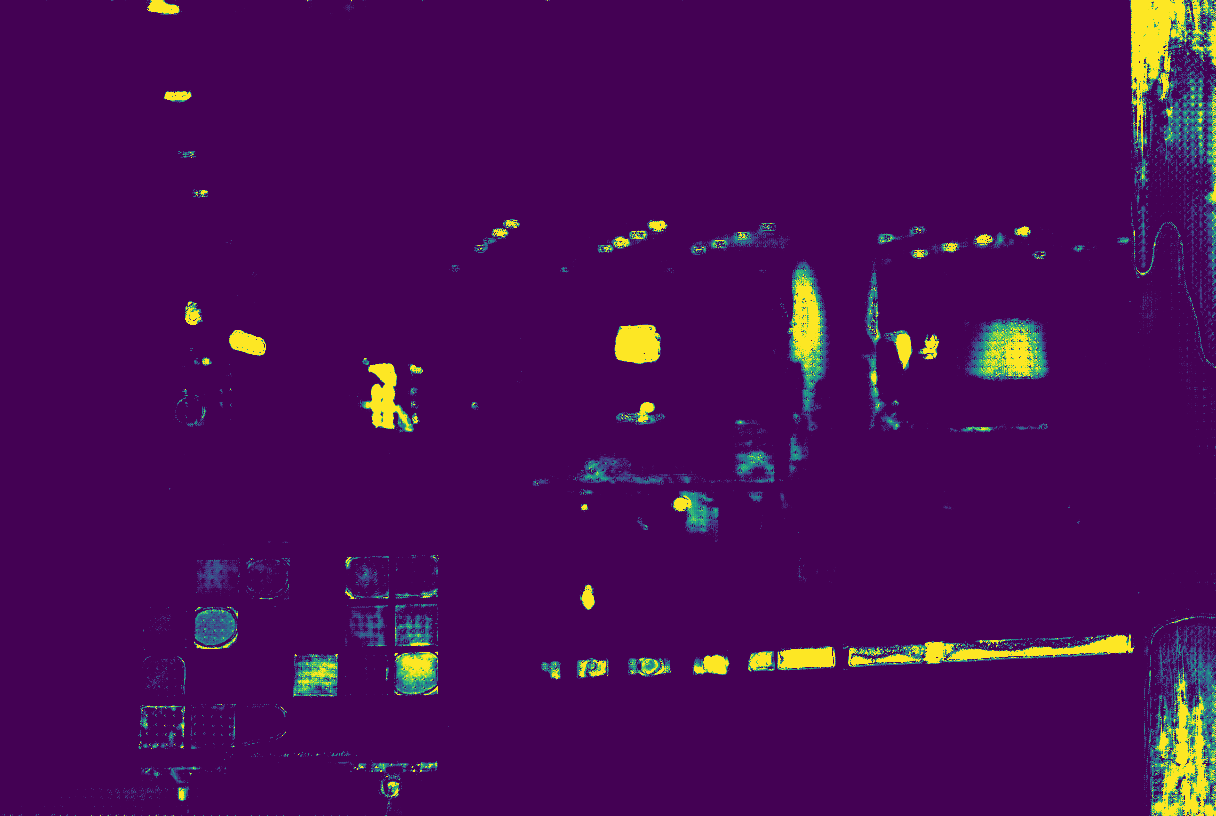}}\includegraphics[width=0.23\linewidth]{figures_arxiv/supp/finetuning/0000024_err_ours_finetuning.png}\llap{\raisebox{\dimen0-7pt}{\setlength{\fboxsep}{2pt}\colorbox{white}{\scriptsize	 PSNR: 44.27dB}}} & \includegraphics[width=0.025\linewidth]{figures_arxiv/colorbar.pdf} & \includegraphics[width=0.23\linewidth]{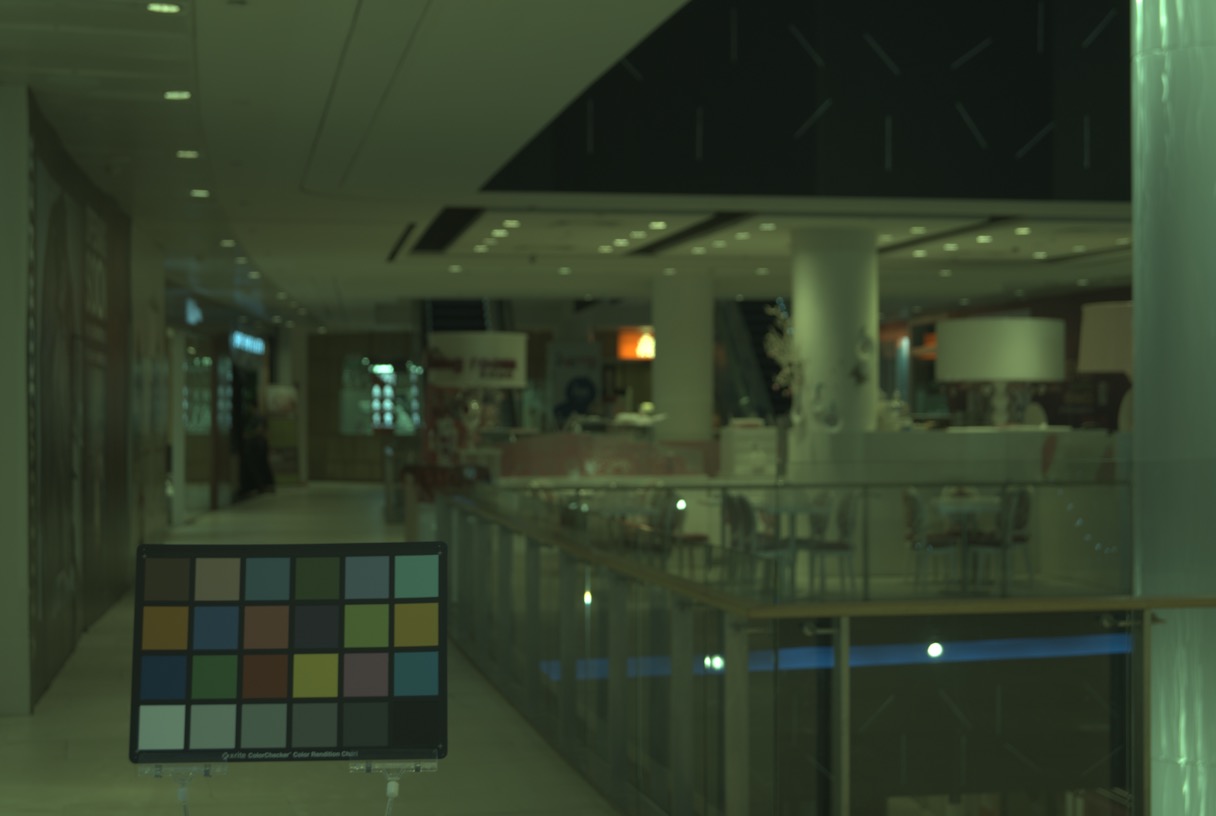}\\

        {\small Input} & {\small Ours} & {\small Ours + fine-tuning} & & {\small Ground truth} \\
    \end{tabular}
    \caption{Qualitative evaluation of our online fine-tuning.}
    \label{fig:finetuning_supp}
\end{figure*}
\begin{figure*}
    \centering
    \setlength{\tabcolsep}{1pt}
    \begin{tabular}{ccccc}
        \includegraphics[width=0.23\linewidth]{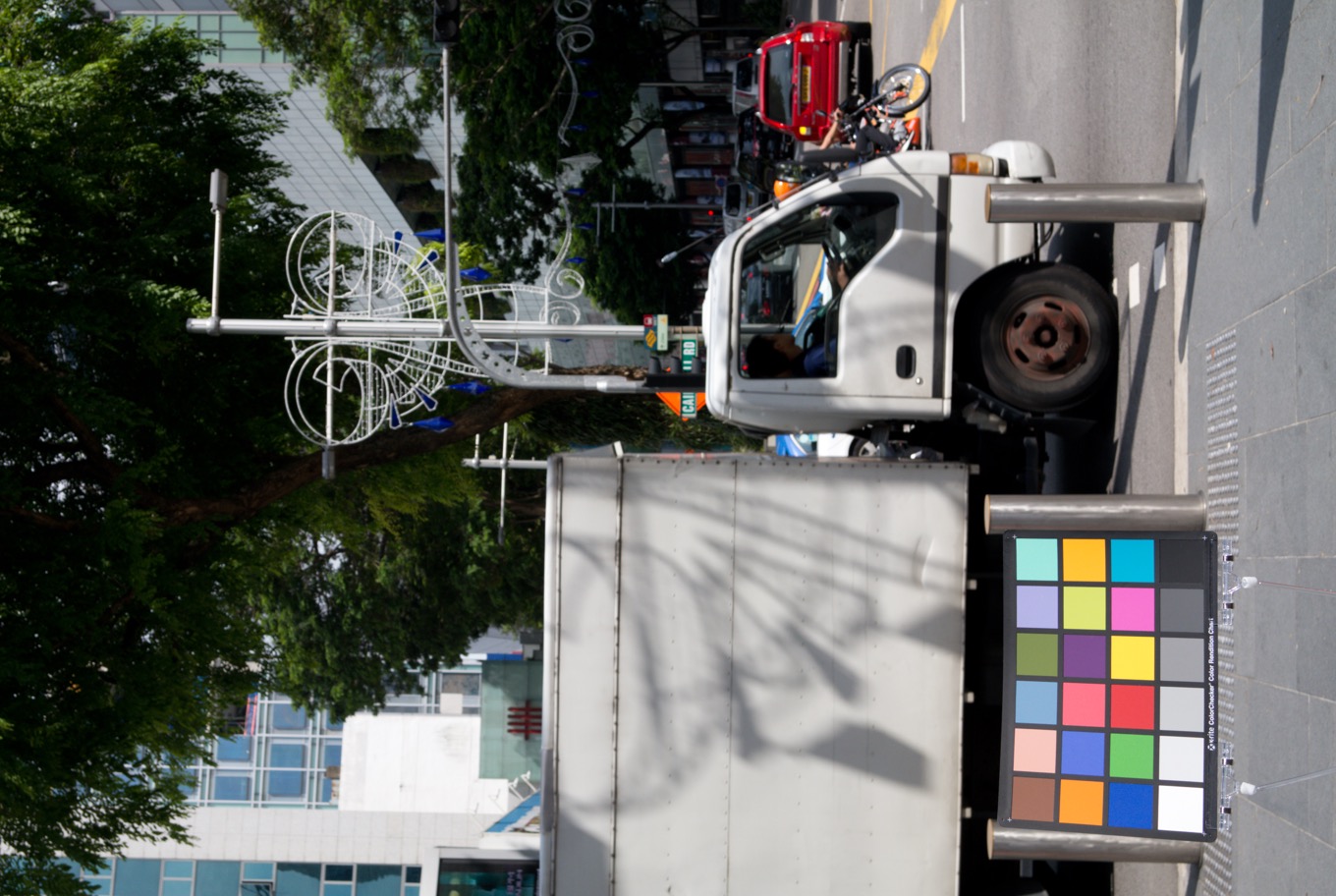} & \settototalheight{\dimen0}{\includegraphics[width=0.23\linewidth]{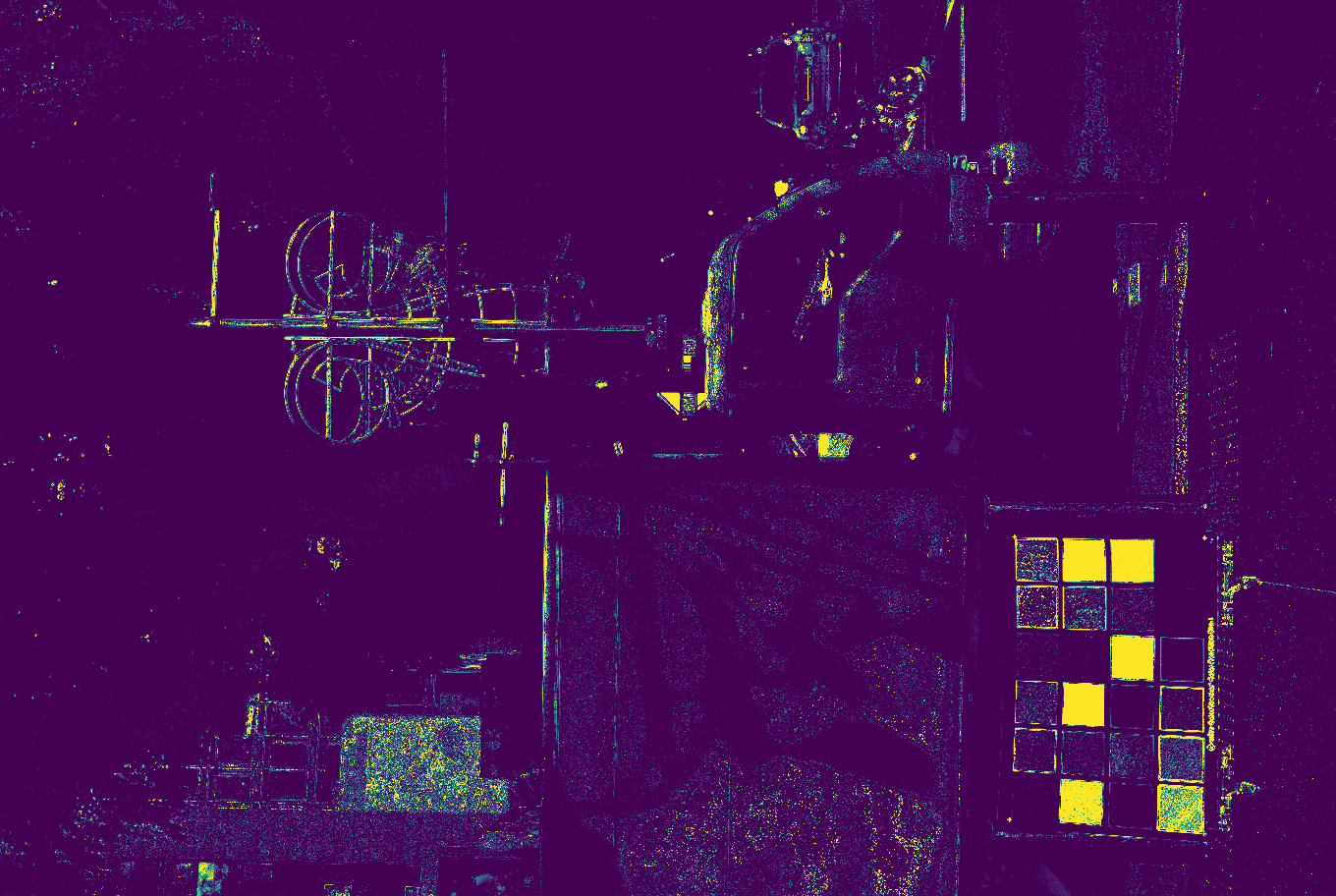}}\includegraphics[width=0.23\linewidth]{figures_arxiv/supp/samsung/SamsungNX2000_0165_err_rang.png}\llap{\raisebox{\dimen0-7pt}{\setlength{\fboxsep}{2pt}\colorbox{white}{\scriptsize	 PSNR: 49.08dB}}} & \settototalheight{\dimen0}{\includegraphics[width=0.23\linewidth]{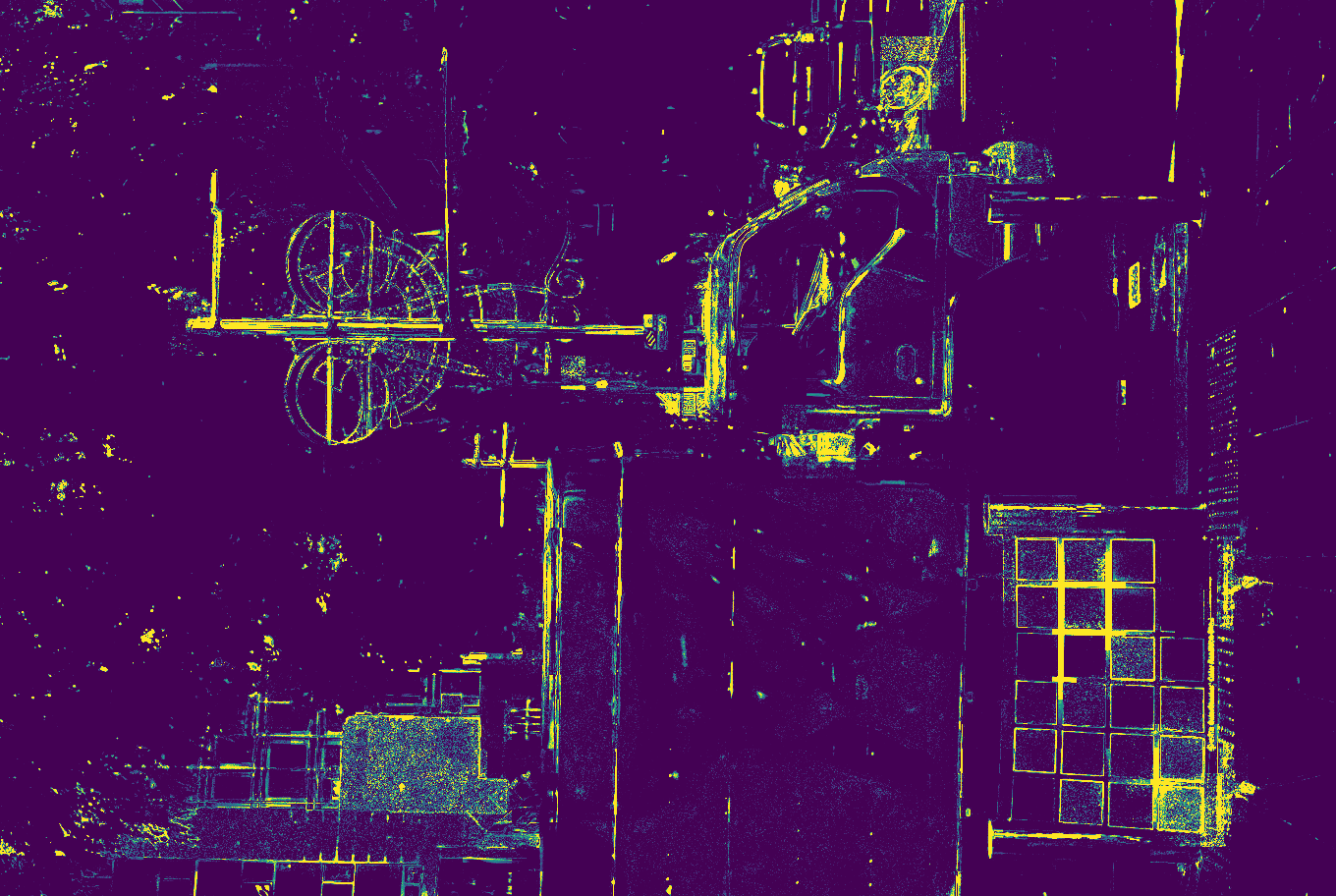}}\includegraphics[width=0.23\linewidth]{figures_arxiv/supp/samsung/SamsungNX2000_0165_err_wacv.png}\llap{\raisebox{\dimen0-7pt}{\setlength{\fboxsep}{2pt}\colorbox{white}{\scriptsize	 PSNR: 44.05dB}}} & \settototalheight{\dimen0}{\includegraphics[width=0.23\linewidth]{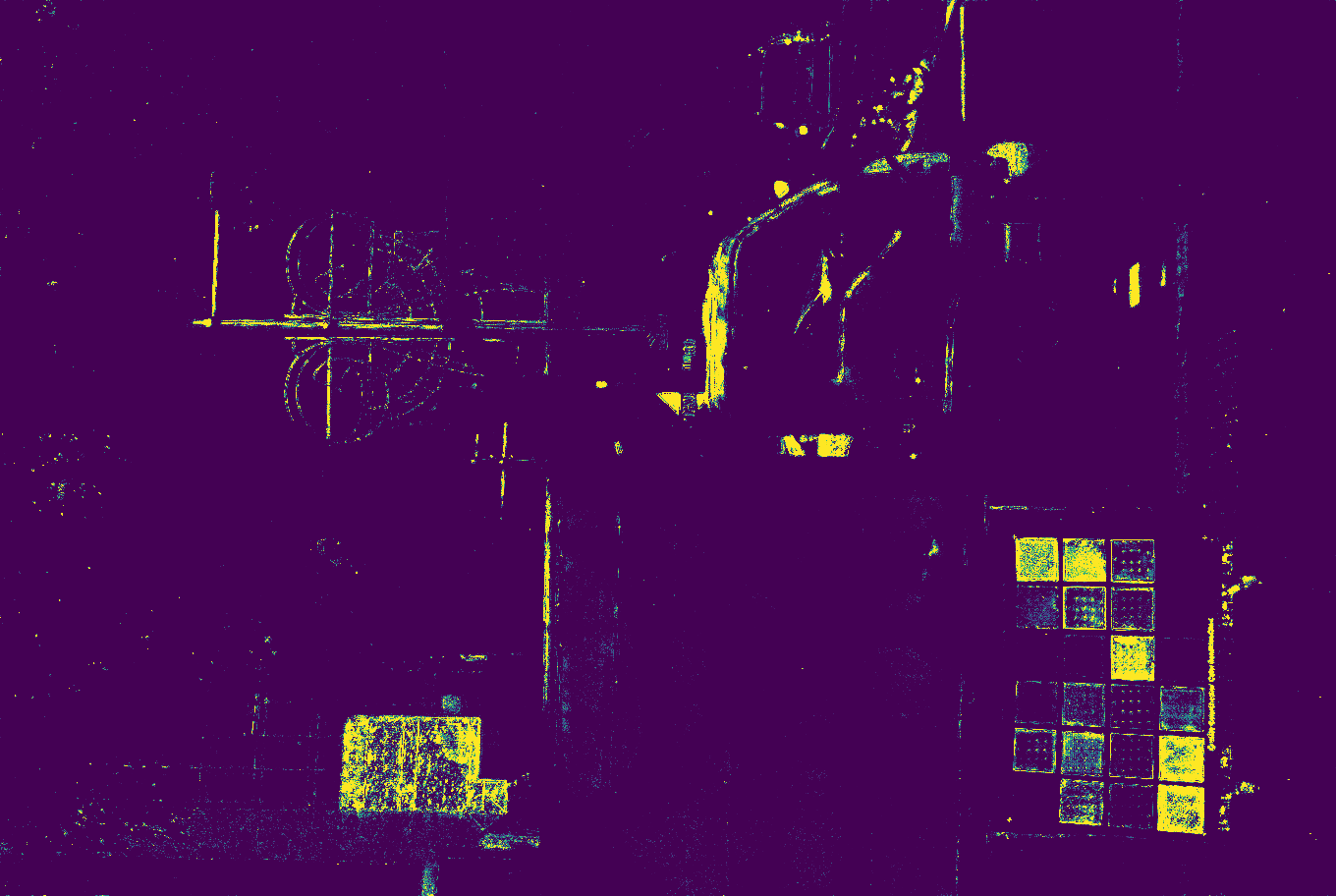}}\includegraphics[width=0.23\linewidth]{figures_arxiv/supp/samsung/0000022_err.png}\llap{\raisebox{\dimen0-7pt}{\setlength{\fboxsep}{2pt}\colorbox{white}{\scriptsize	 PSNR: 52.47dB}}} & \includegraphics[width=0.025\linewidth]{figures_arxiv/colorbar.pdf} \\

        \includegraphics[width=0.23\linewidth]{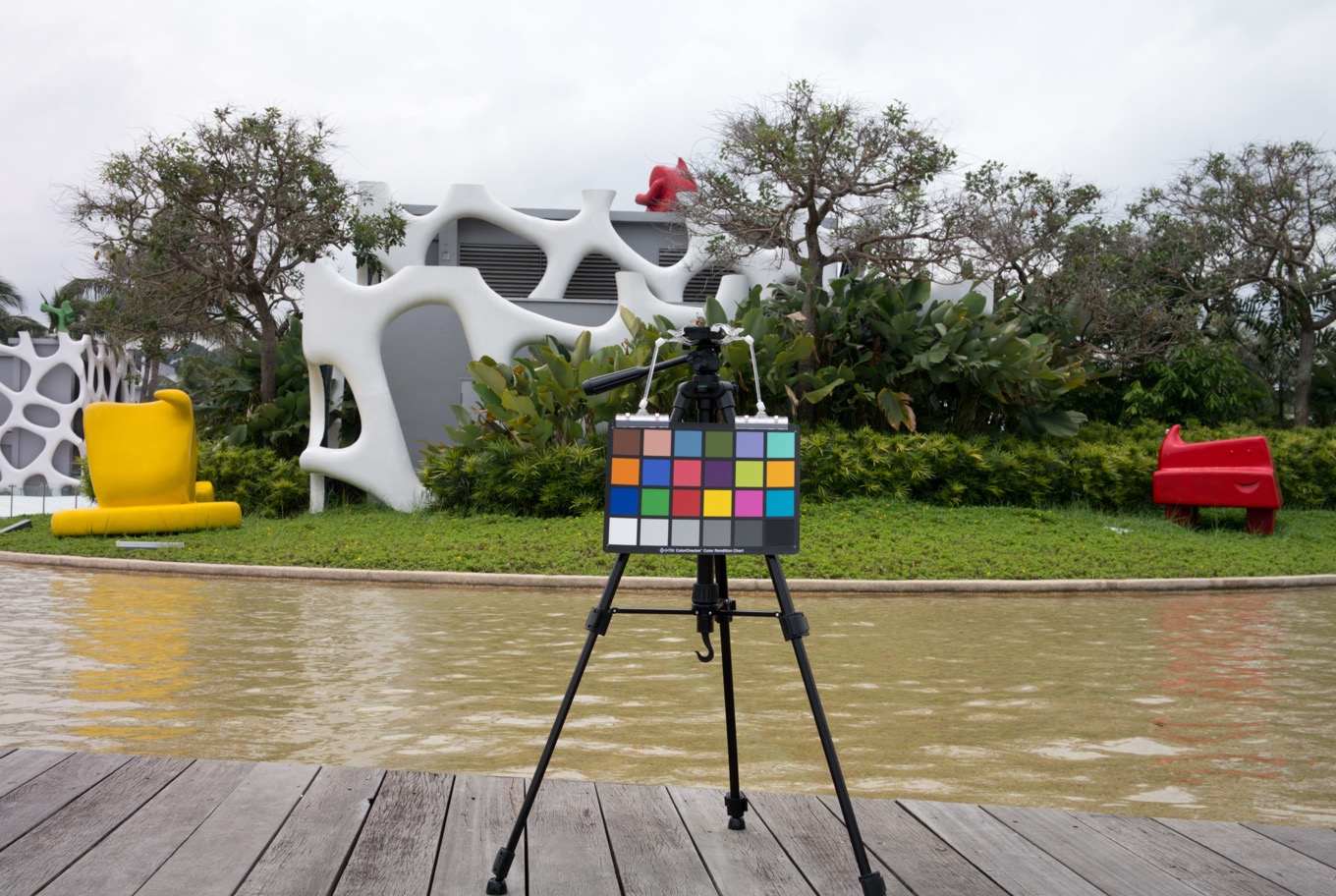} & \settototalheight{\dimen0}{\includegraphics[width=0.23\linewidth]{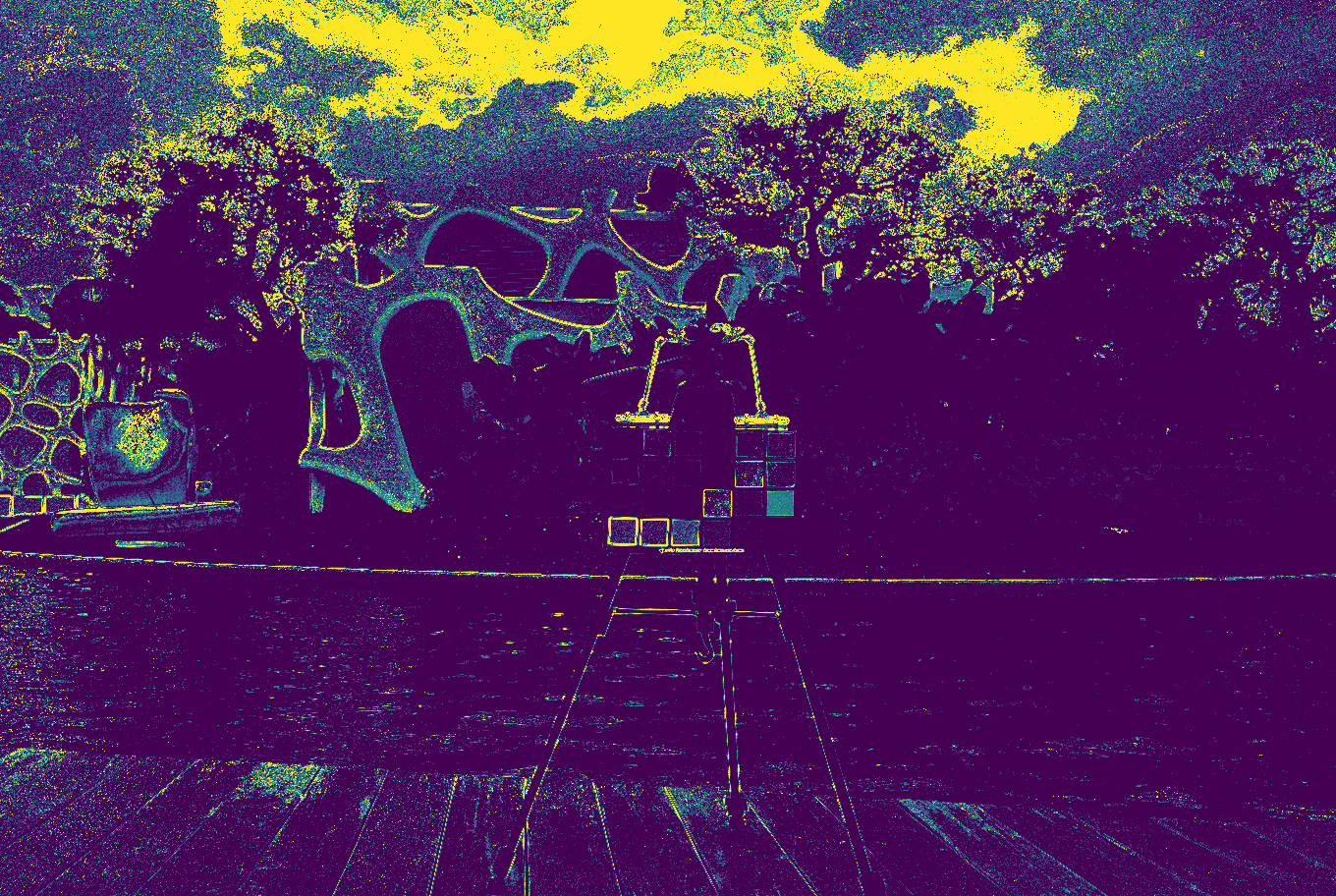}}\includegraphics[width=0.23\linewidth]{figures_arxiv/supp/samsung/SamsungNX2000_0054_err_rang.png}\llap{\raisebox{\dimen0-7pt}{\setlength{\fboxsep}{2pt}\colorbox{white}{\scriptsize	 PSNR: 43.91dB}}} & \settototalheight{\dimen0}{\includegraphics[width=0.23\linewidth]{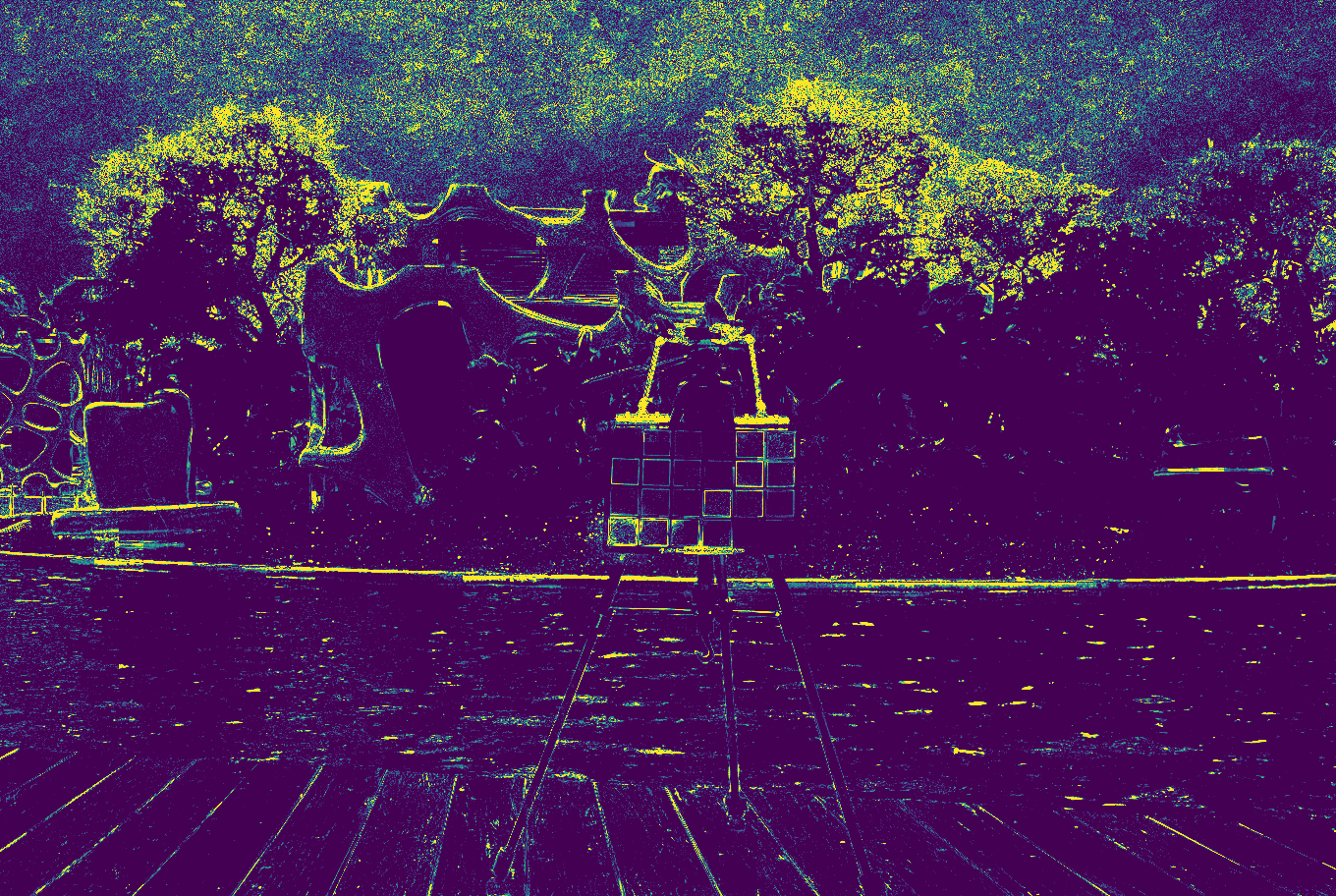}}\includegraphics[width=0.23\linewidth]{figures_arxiv/supp/samsung/SamsungNX2000_0054_err_wacv.png}\llap{\raisebox{\dimen0-7pt}{\setlength{\fboxsep}{2pt}\colorbox{white}{\scriptsize	 PSNR: 46.89dB}}} & \settototalheight{\dimen0}{\includegraphics[width=0.23\linewidth]{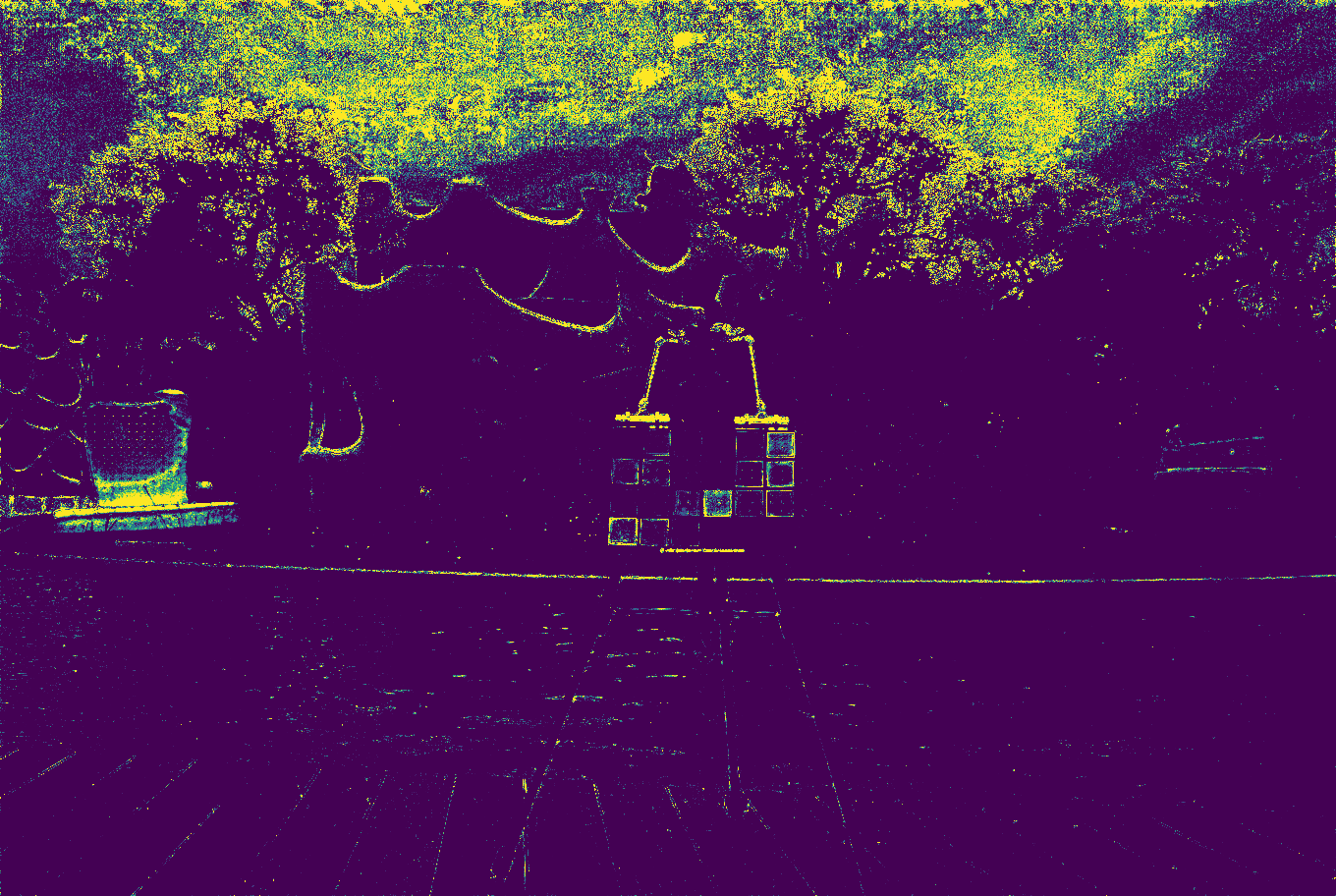}}\includegraphics[width=0.23\linewidth]{figures_arxiv/supp/samsung/0000006_err.png}\llap{\raisebox{\dimen0-7pt}{\setlength{\fboxsep}{2pt}\colorbox{white}{\scriptsize	 PSNR: 50.93dB}}} & \includegraphics[width=0.025\linewidth]{figures_arxiv/colorbar.pdf} \\

        \includegraphics[width=0.23\linewidth]{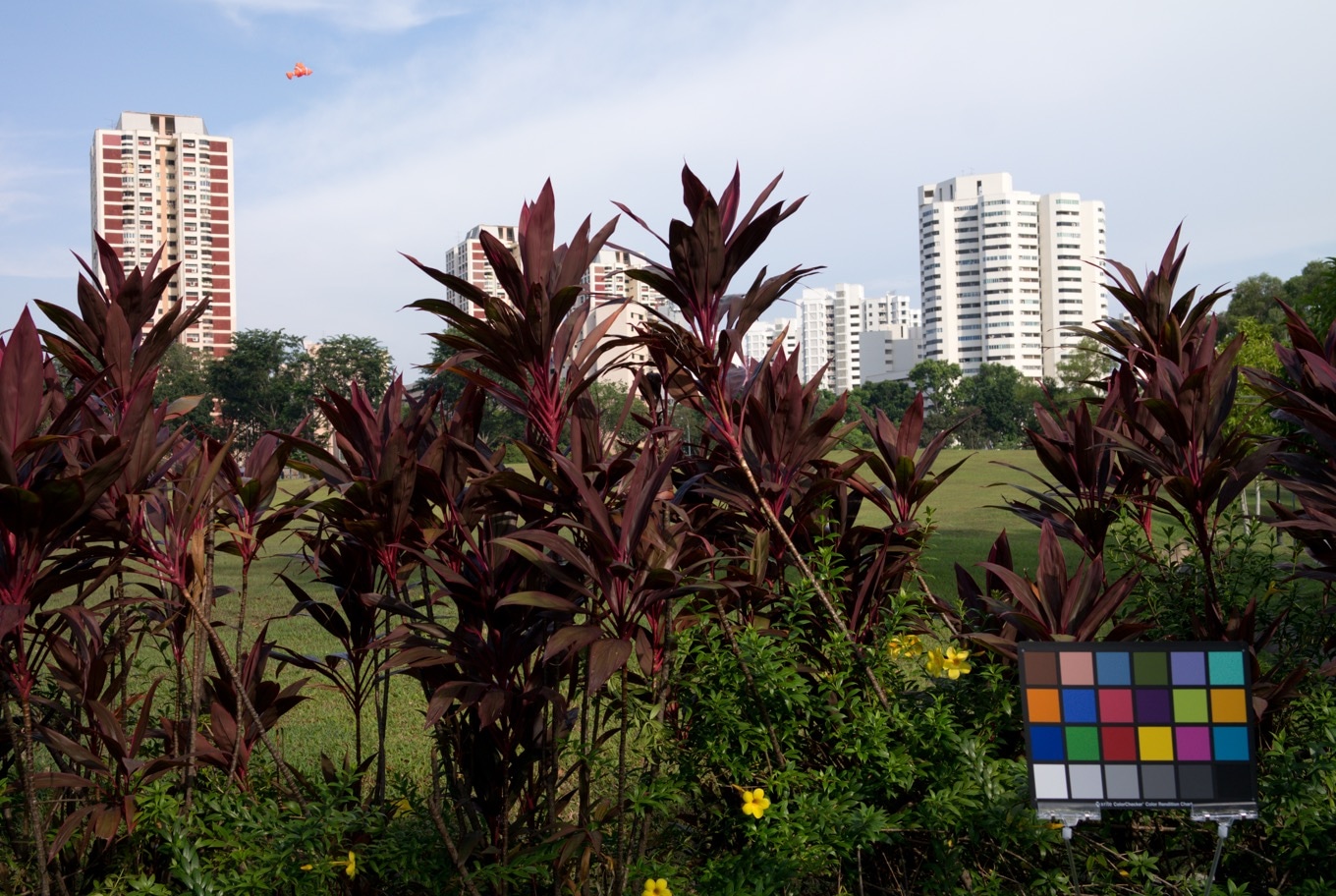} & \settototalheight{\dimen0}{\includegraphics[width=0.23\linewidth]{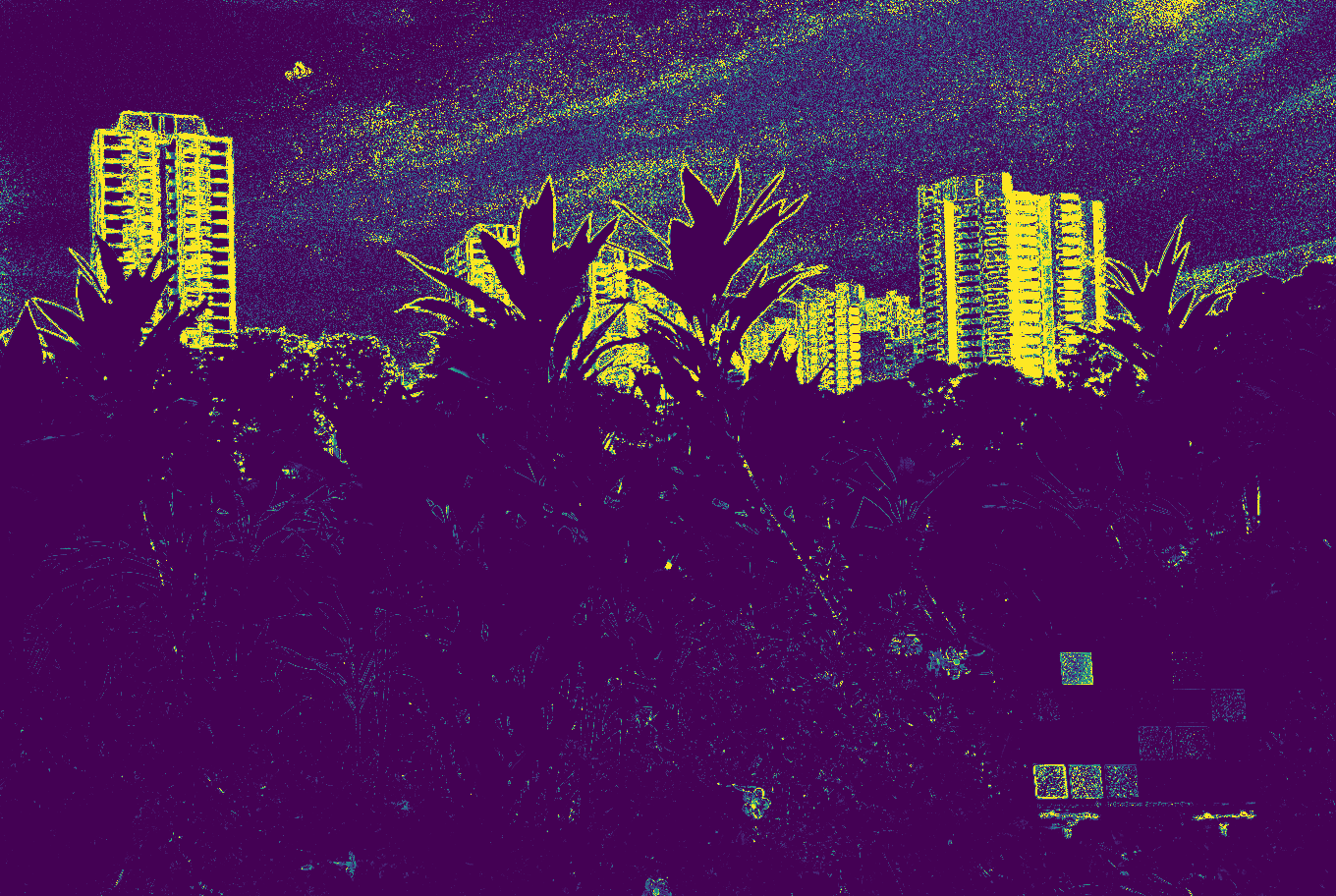}}\includegraphics[width=0.23\linewidth]{figures_arxiv/supp/samsung/SamsungNX2000_0004_err_rang.png}\llap{\raisebox{\dimen0-7pt}{\setlength{\fboxsep}{2pt}\colorbox{white}{\scriptsize	 PSNR: 35.93dB}}} & \settototalheight{\dimen0}{\includegraphics[width=0.23\linewidth]{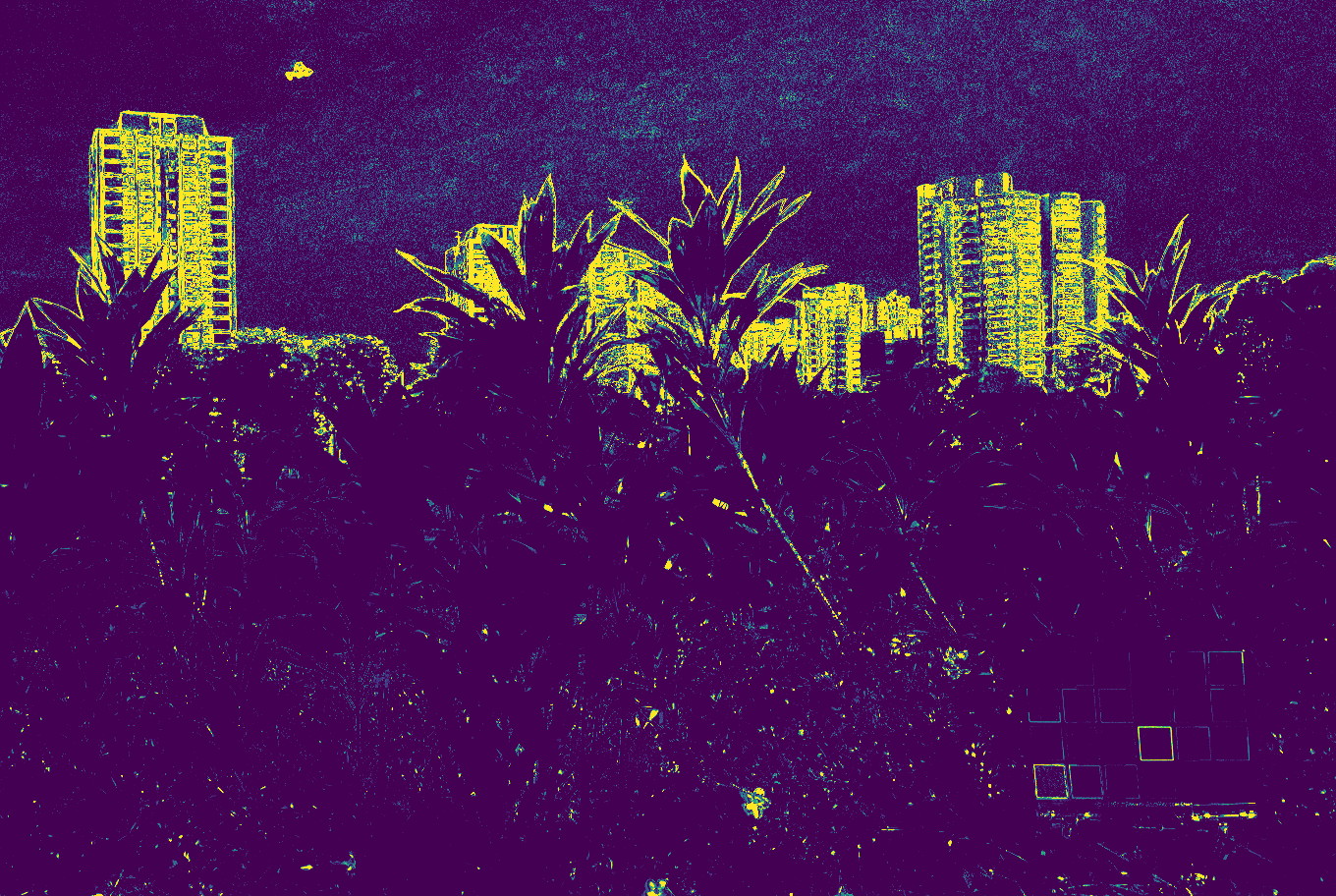}}\includegraphics[width=0.23\linewidth]{figures_arxiv/supp/samsung/SamsungNX2000_0004_err_wacv.png}\llap{\raisebox{\dimen0-7pt}{\setlength{\fboxsep}{2pt}\colorbox{white}{\scriptsize	 PSNR: 46.96dB}}} & \settototalheight{\dimen0}{\includegraphics[width=0.23\linewidth]{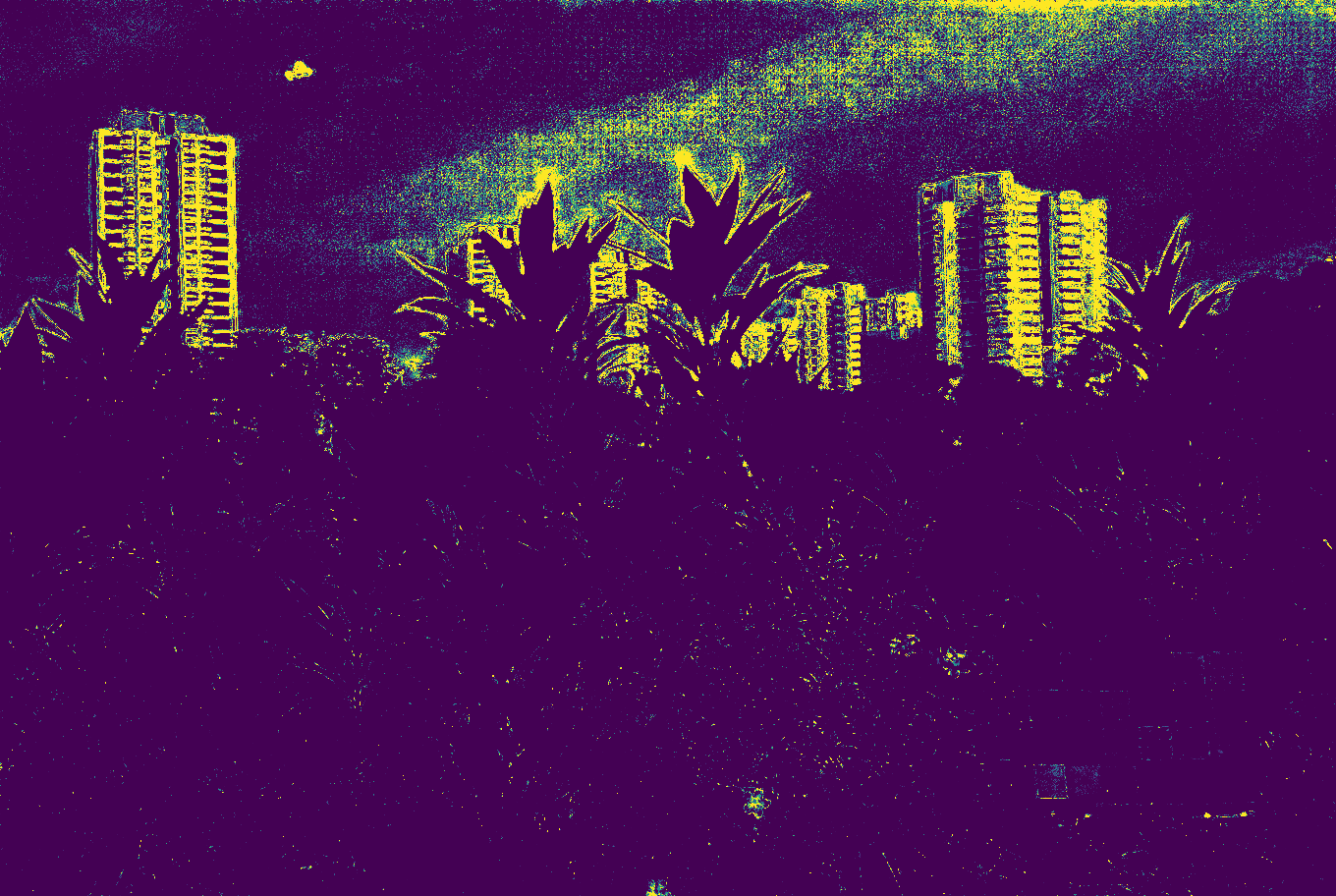}}\includegraphics[width=0.23\linewidth]{figures_arxiv/supp/samsung/0000000_err.png}\llap{\raisebox{\dimen0-7pt}{\setlength{\fboxsep}{2pt}\colorbox{white}{\scriptsize	 PSNR: 50.59dB}}} & \includegraphics[width=0.025\linewidth]{figures_arxiv/colorbar.pdf} \\

        \includegraphics[width=0.23\linewidth]{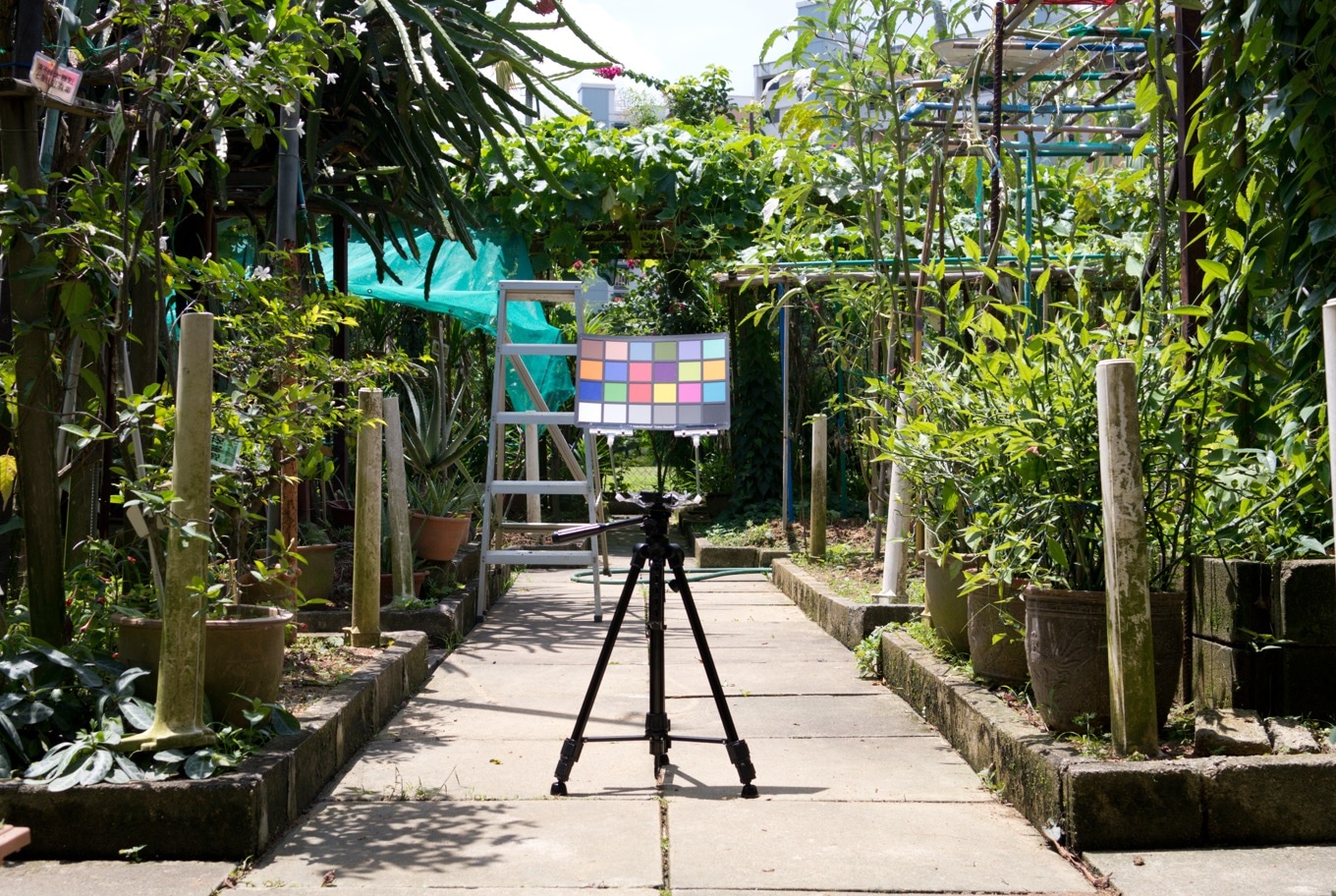} & \settototalheight{\dimen0}{\includegraphics[width=0.23\linewidth]{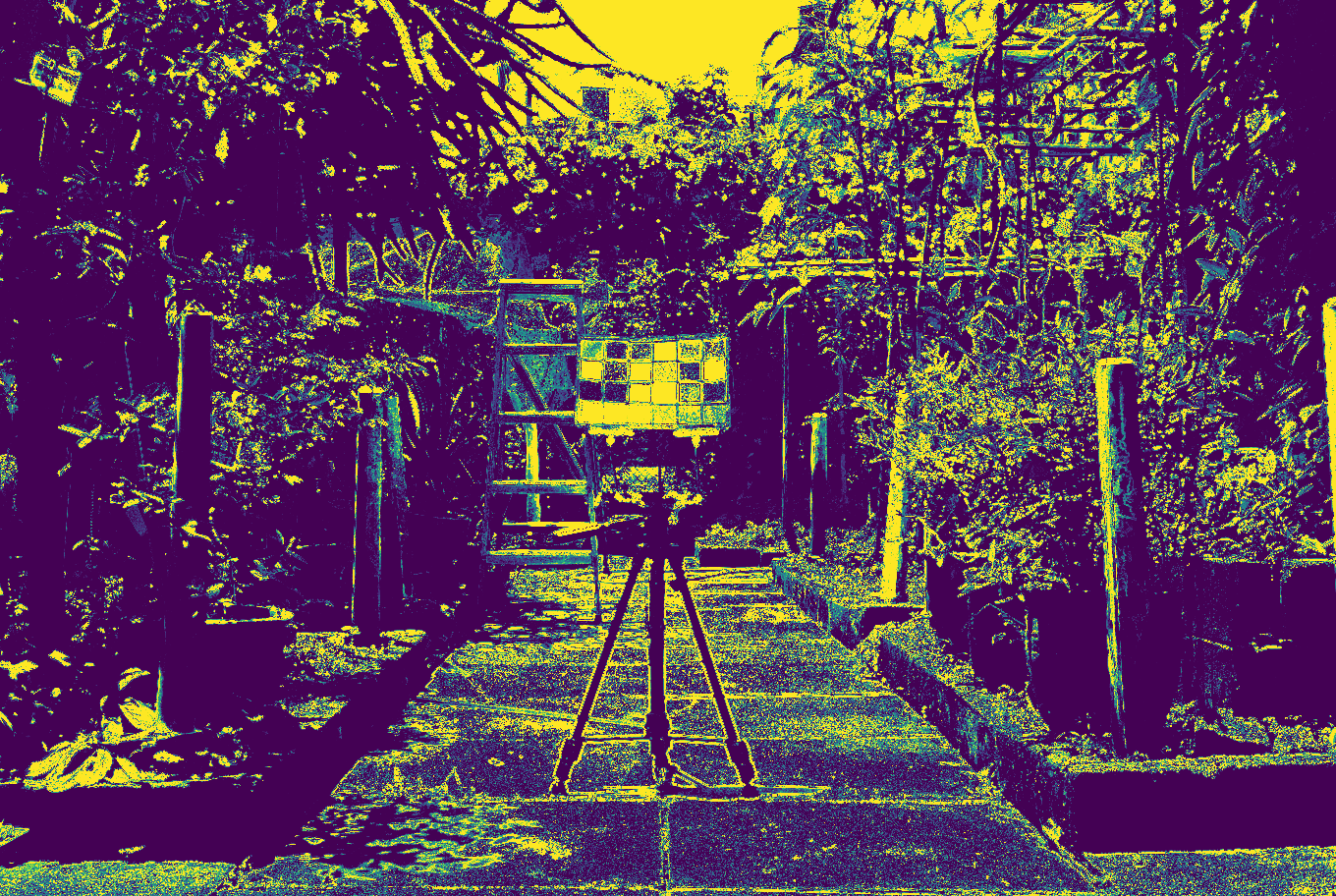}}\includegraphics[width=0.23\linewidth]{figures_arxiv/supp/samsung/SamsungNX2000_0077_err_rang.png}\llap{\raisebox{\dimen0-7pt}{\setlength{\fboxsep}{2pt}\colorbox{white}{\scriptsize	 PSNR: 31.05dB}}} & \settototalheight{\dimen0}{\includegraphics[width=0.23\linewidth]{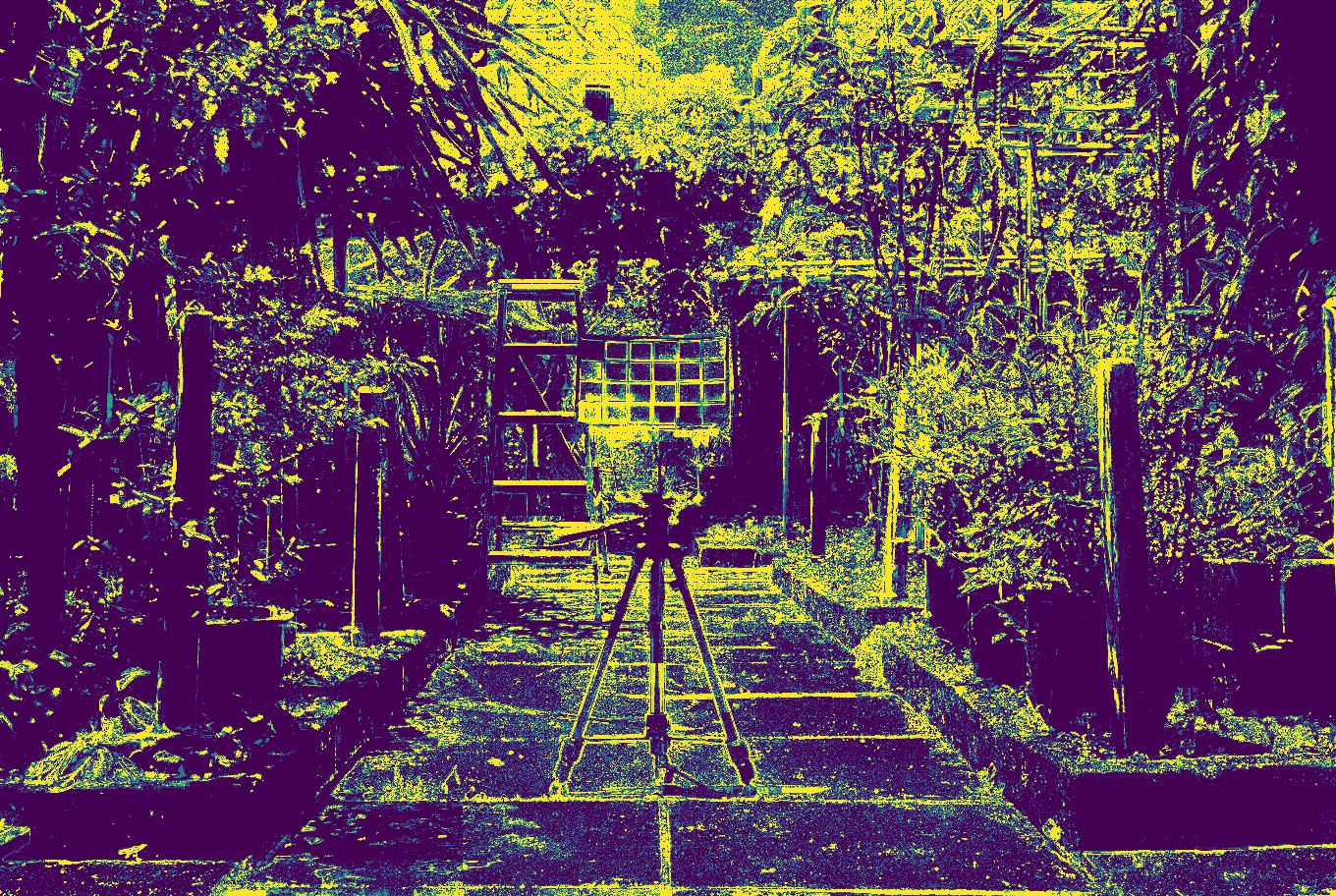}}\includegraphics[width=0.23\linewidth]{figures_arxiv/supp/samsung/SamsungNX2000_0077_err_wacv.png}\llap{\raisebox{\dimen0-7pt}{\setlength{\fboxsep}{2pt}\colorbox{white}{\scriptsize	 PSNR: 38.34dB}}} & \settototalheight{\dimen0}{\includegraphics[width=0.23\linewidth]{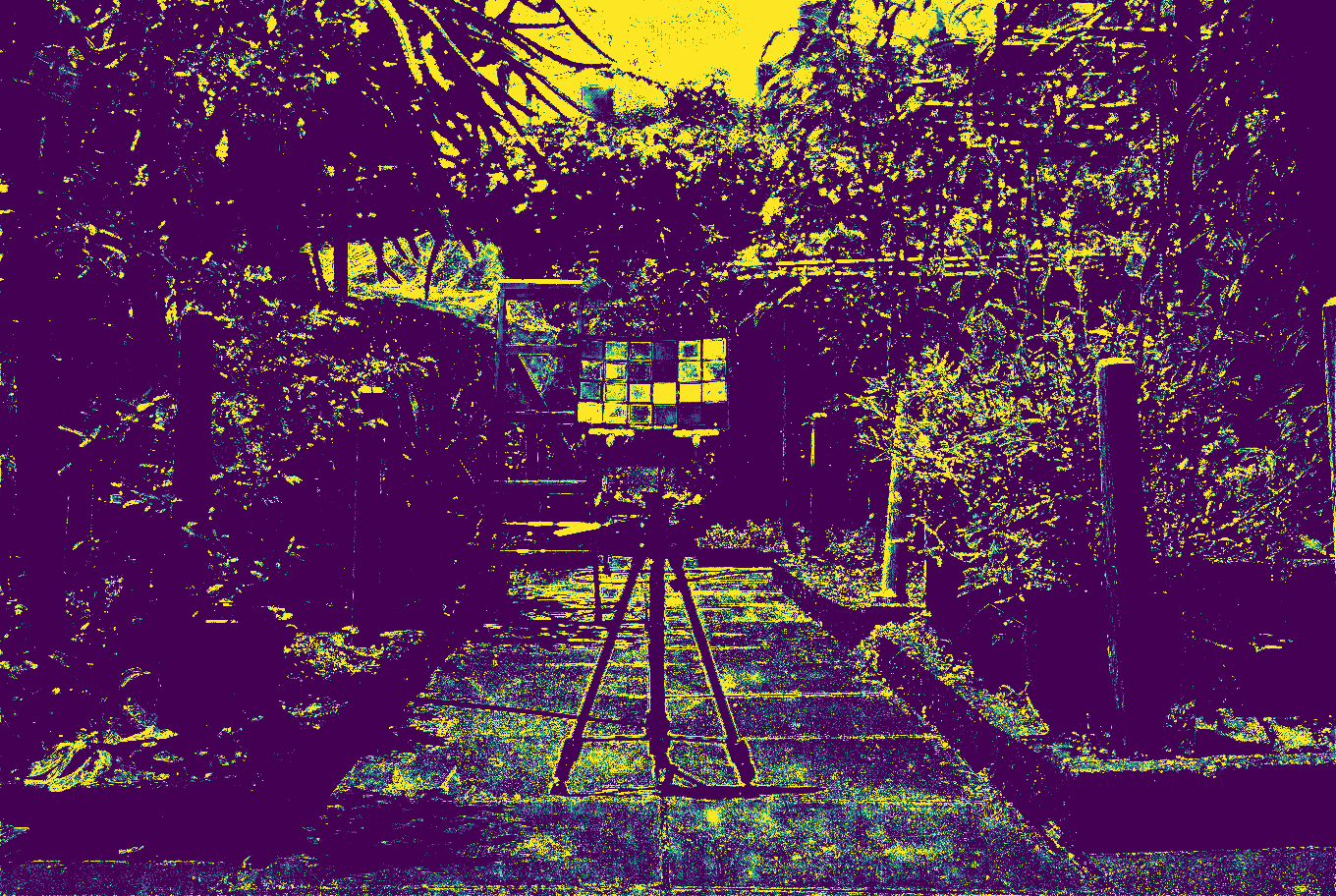}}\includegraphics[width=0.23\linewidth]{figures_arxiv/supp/samsung/0000010_err.png}\llap{\raisebox{\dimen0-7pt}{\setlength{\fboxsep}{2pt}\colorbox{white}{\scriptsize	 PSNR: 42.97dB}}} & \includegraphics[width=0.025\linewidth]{figures_arxiv/colorbar.pdf} \\

        \includegraphics[width=0.23\linewidth]{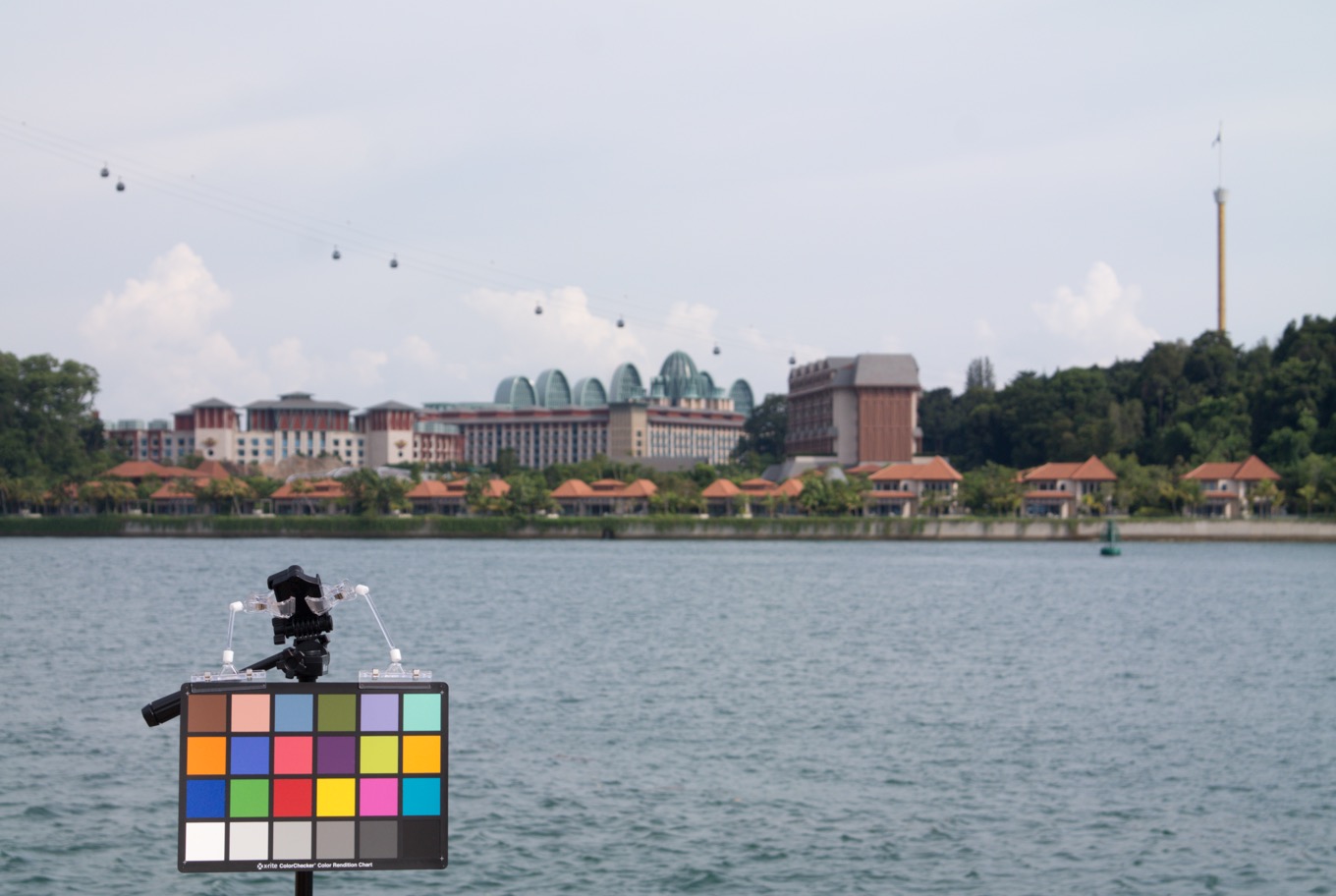} & \settototalheight{\dimen0}{\includegraphics[width=0.23\linewidth]{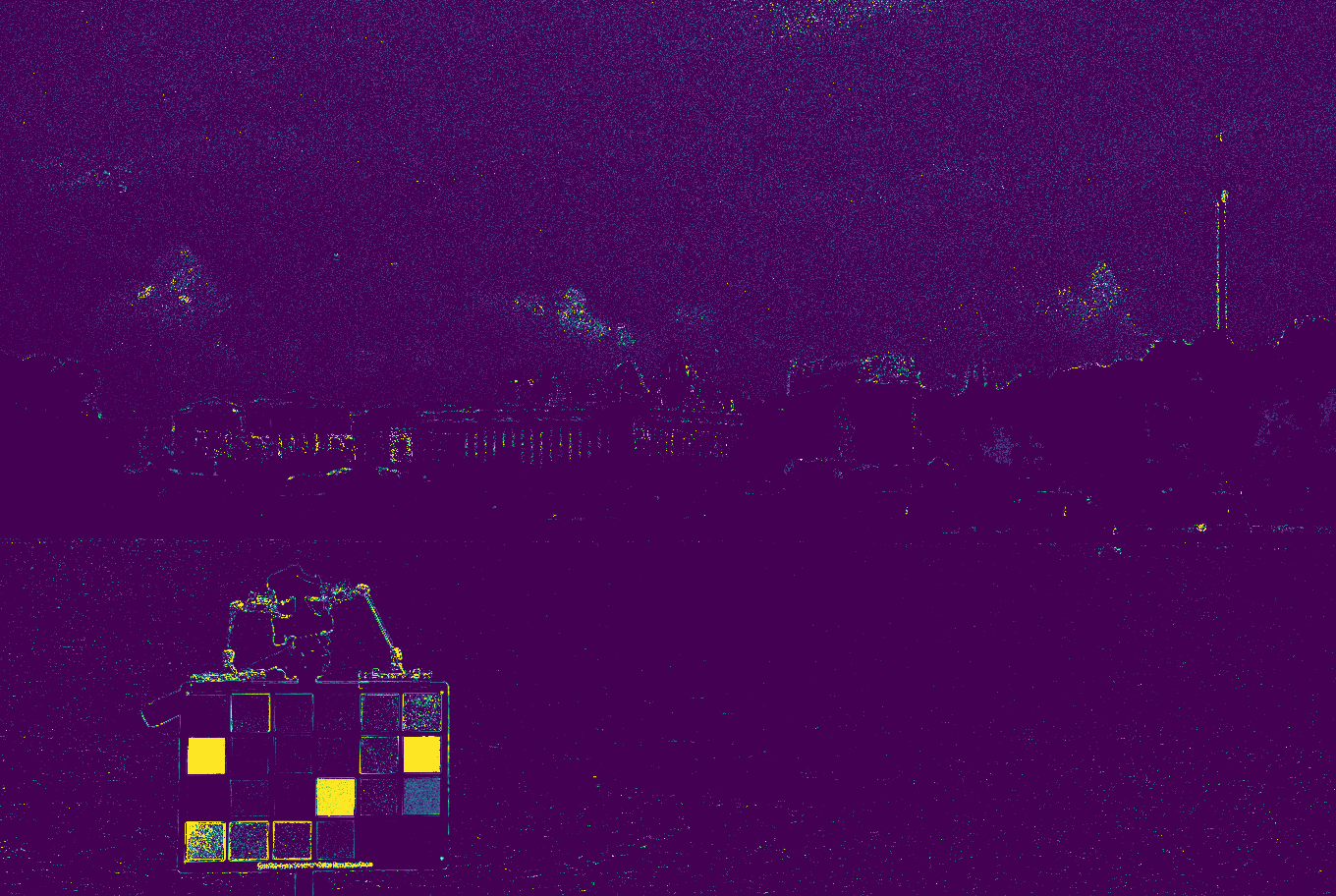}}\includegraphics[width=0.23\linewidth]{figures_arxiv/supp/samsung/SamsungNX2000_0183_err_rang.png}\llap{\raisebox{\dimen0-7pt}{\setlength{\fboxsep}{2pt}\colorbox{white}{\scriptsize	 PSNR: 52.43dB}}} & \settototalheight{\dimen0}{\includegraphics[width=0.23\linewidth]{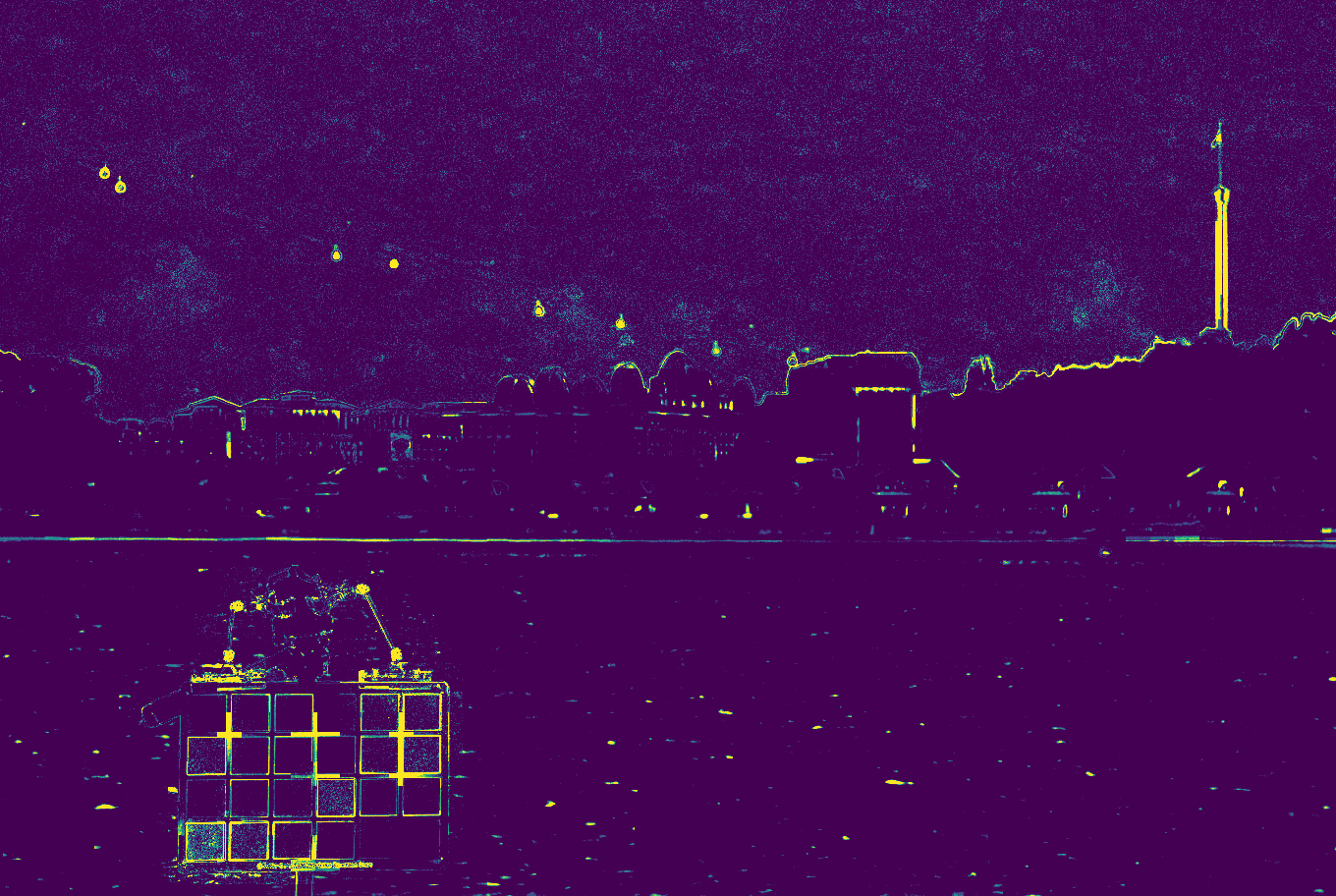}}\includegraphics[width=0.23\linewidth]{figures_arxiv/supp/samsung/SamsungNX2000_0183_err_wacv.png}\llap{\raisebox{\dimen0-7pt}{\setlength{\fboxsep}{2pt}\colorbox{white}{\scriptsize	 PSNR: 54.04dB}}} & \settototalheight{\dimen0}{\includegraphics[width=0.23\linewidth]{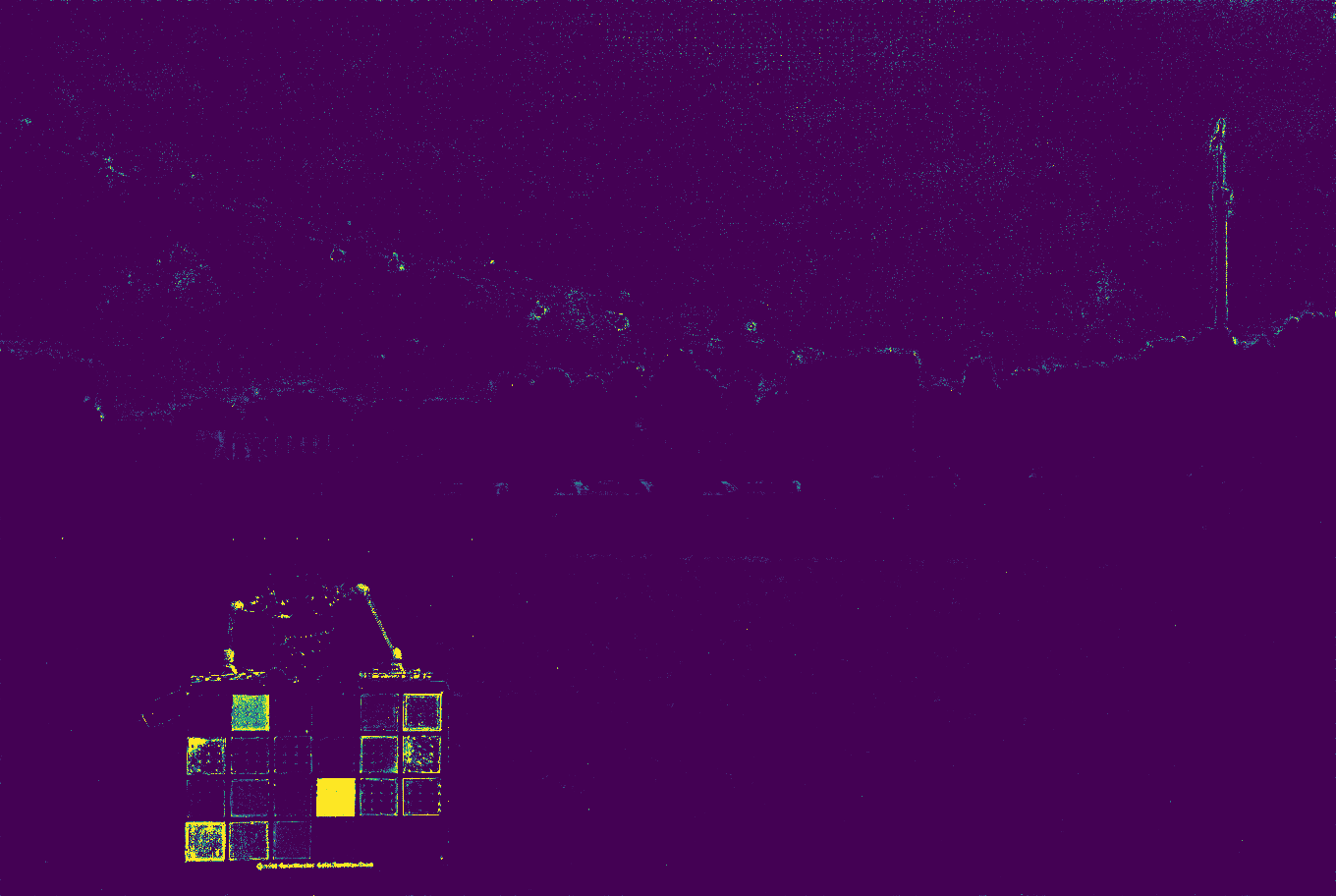}}\includegraphics[width=0.23\linewidth]{figures_arxiv/supp/samsung/0000024_err.png}\llap{\raisebox{\dimen0-7pt}{\setlength{\fboxsep}{2pt}\colorbox{white}{\scriptsize	 PSNR: 55.42dB}}} & \includegraphics[width=0.025\linewidth]{figures_arxiv/colorbar.pdf} \\

        {\small Input} & {\small RIR~\cite{rang}} & {\small SAM~\cite{wacv}} & {\small Ours + fine-tuning} & \\
    \end{tabular}
    \caption{Qualitative comparison on Samsung NX2000.}
    \label{fig:supp_samsung}
\end{figure*}
\begin{figure*}
    \centering
    \setlength{\tabcolsep}{1pt}
    \begin{tabular}{ccccc}
        \includegraphics[width=0.23\linewidth]{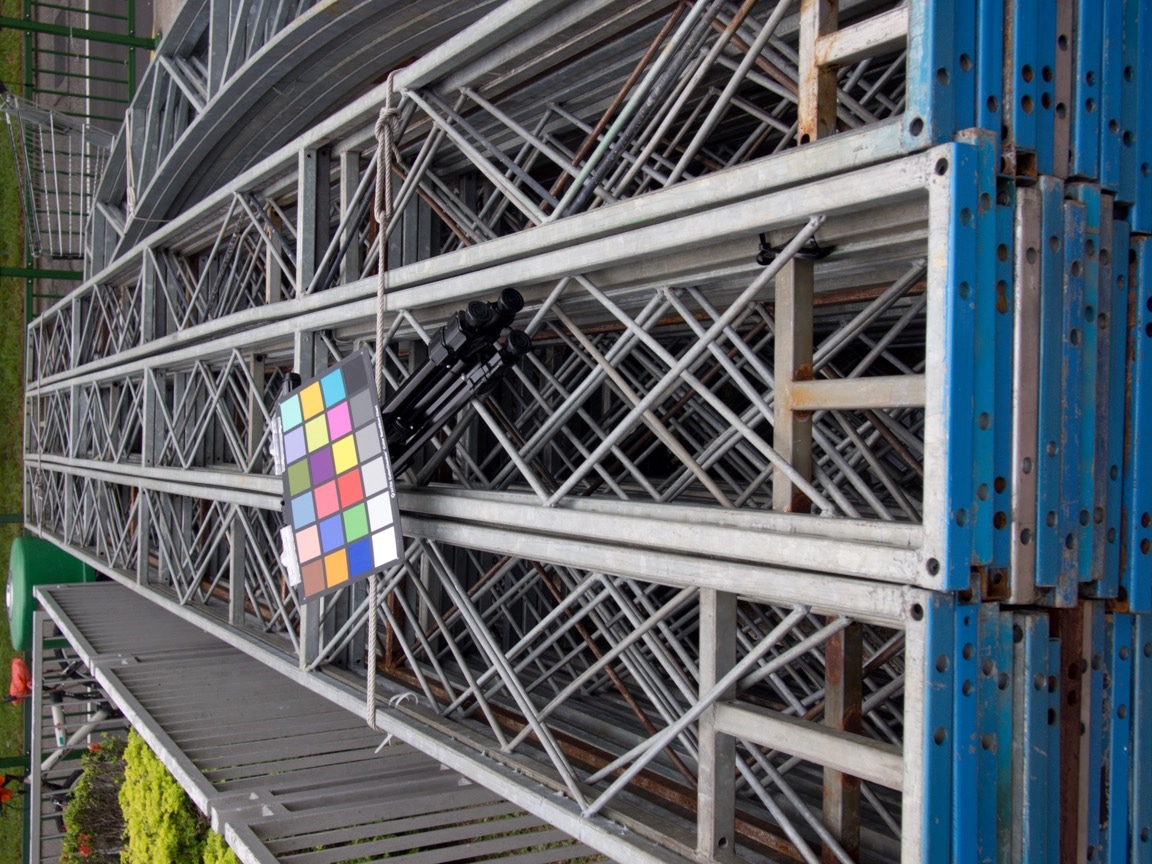} & \settototalheight{\dimen0}{\includegraphics[width=0.23\linewidth]{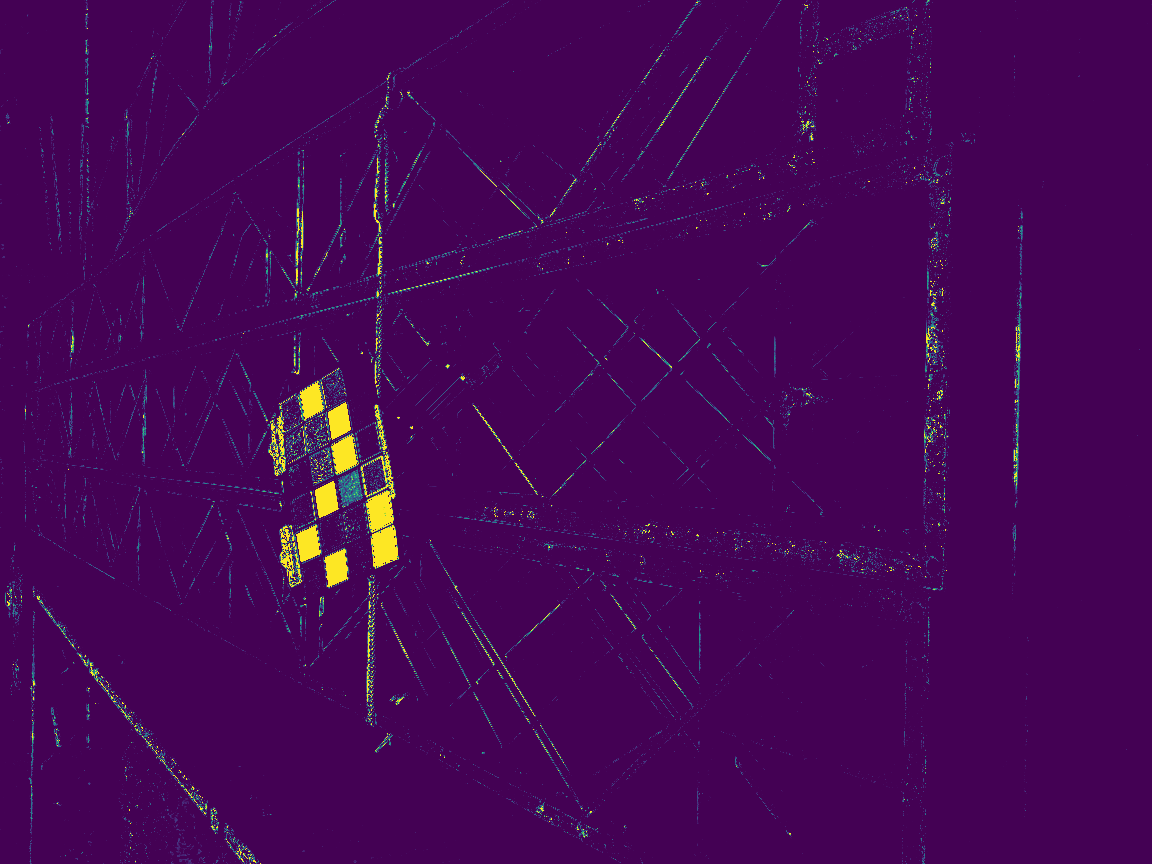}}\includegraphics[width=0.23\linewidth]{figures_arxiv/supp/olympus/OlympusEPL6_0041_st4_err_rang.png}\llap{\raisebox{\dimen0-7pt}{\setlength{\fboxsep}{2pt}\colorbox{white}{\scriptsize	 PSNR: 47.25dB}}} & \settototalheight{\dimen0}{\includegraphics[width=0.23\linewidth]{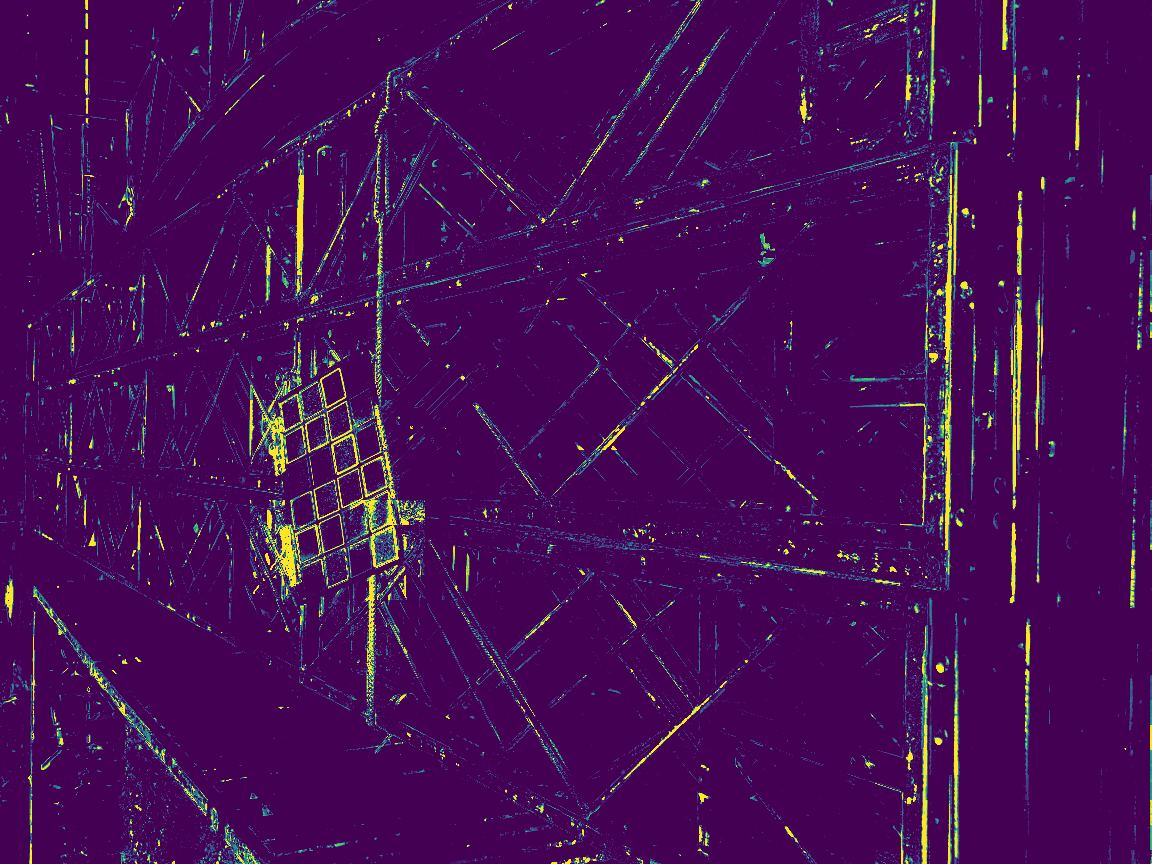}}\includegraphics[width=0.23\linewidth]{figures_arxiv/supp/olympus/OlympusEPL6_0041_st4_err_wacv.png}\llap{\raisebox{\dimen0-7pt}{\setlength{\fboxsep}{2pt}\colorbox{white}{\scriptsize	 PSNR: 51.08dB}}} & \settototalheight{\dimen0}{\includegraphics[width=0.23\linewidth]{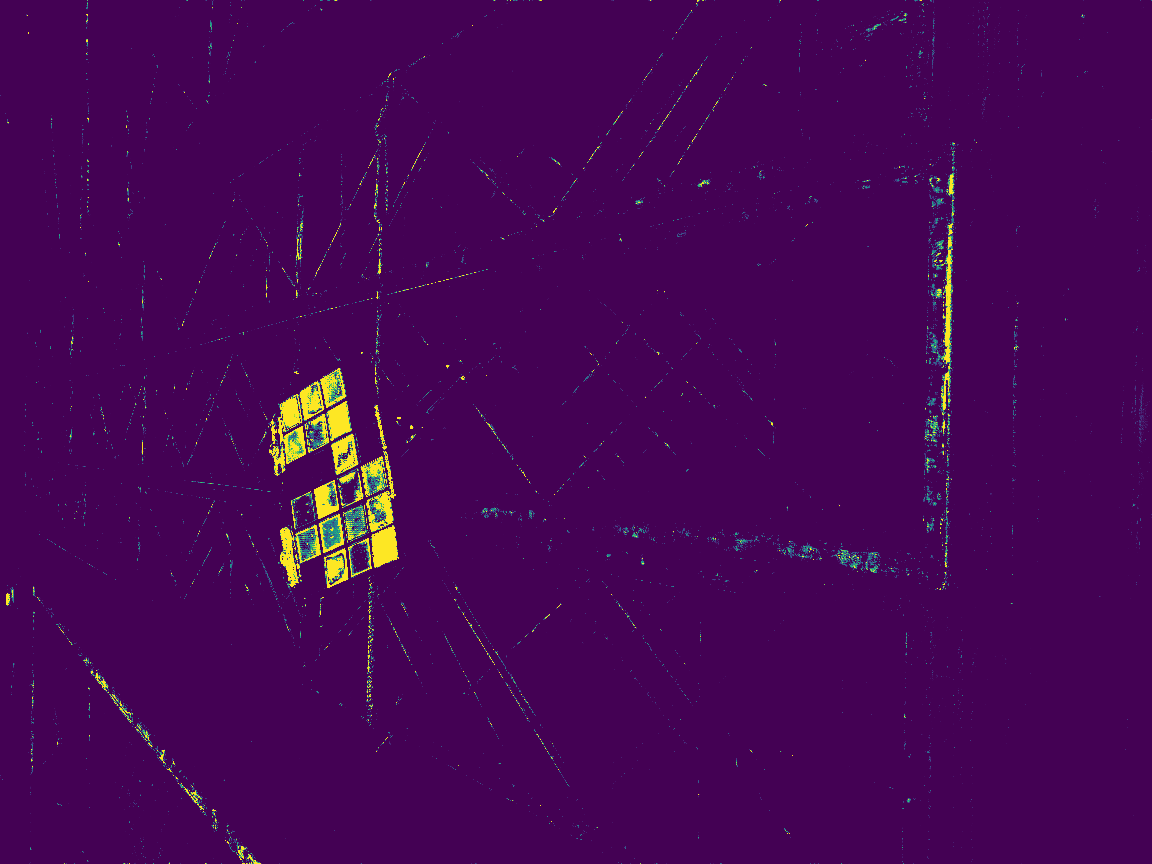}}\includegraphics[width=0.23\linewidth]{figures_arxiv/supp/olympus/0000002_err.png}\llap{\raisebox{\dimen0-7pt}{\setlength{\fboxsep}{2pt}\colorbox{white}{\scriptsize	 PSNR: 53.45.dB}}} & \includegraphics[width=0.027\linewidth]{figures_arxiv/colorbar.pdf} \\

        \includegraphics[width=0.23\linewidth]{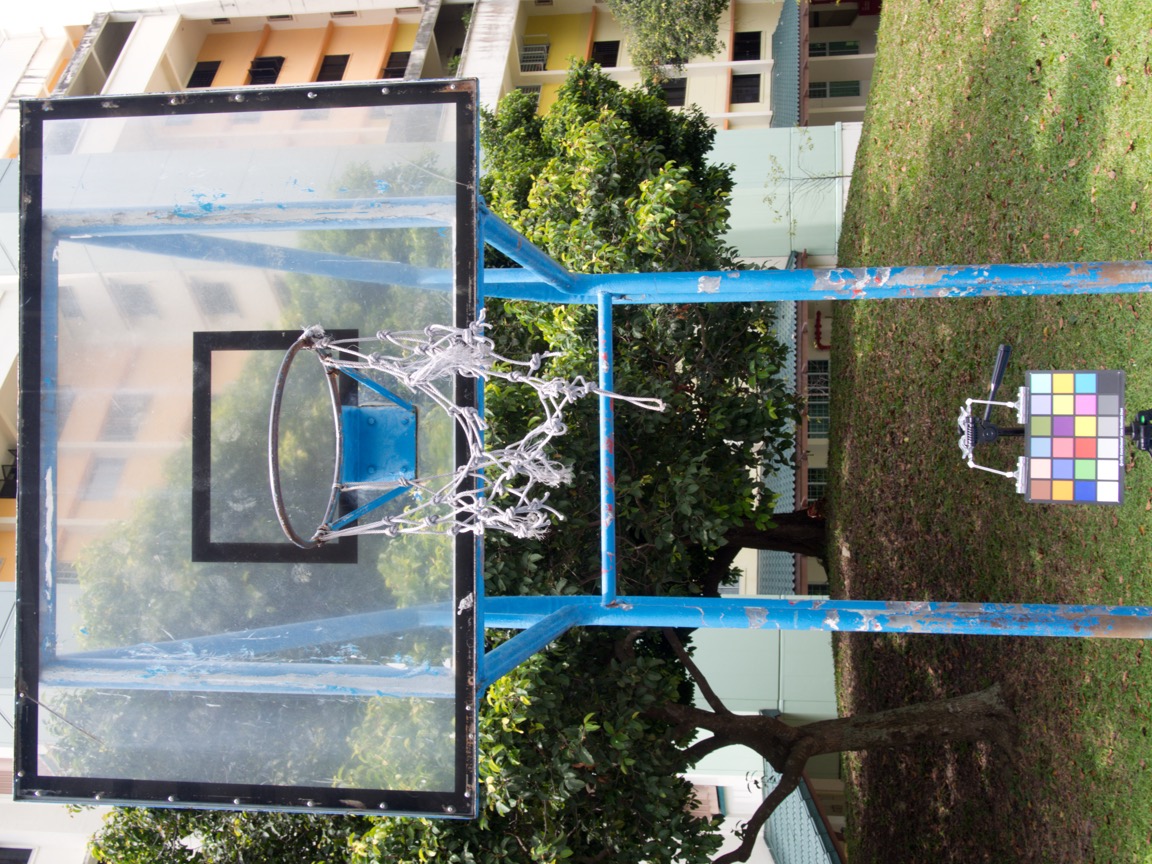} & \settototalheight{\dimen0}{\includegraphics[width=0.23\linewidth]{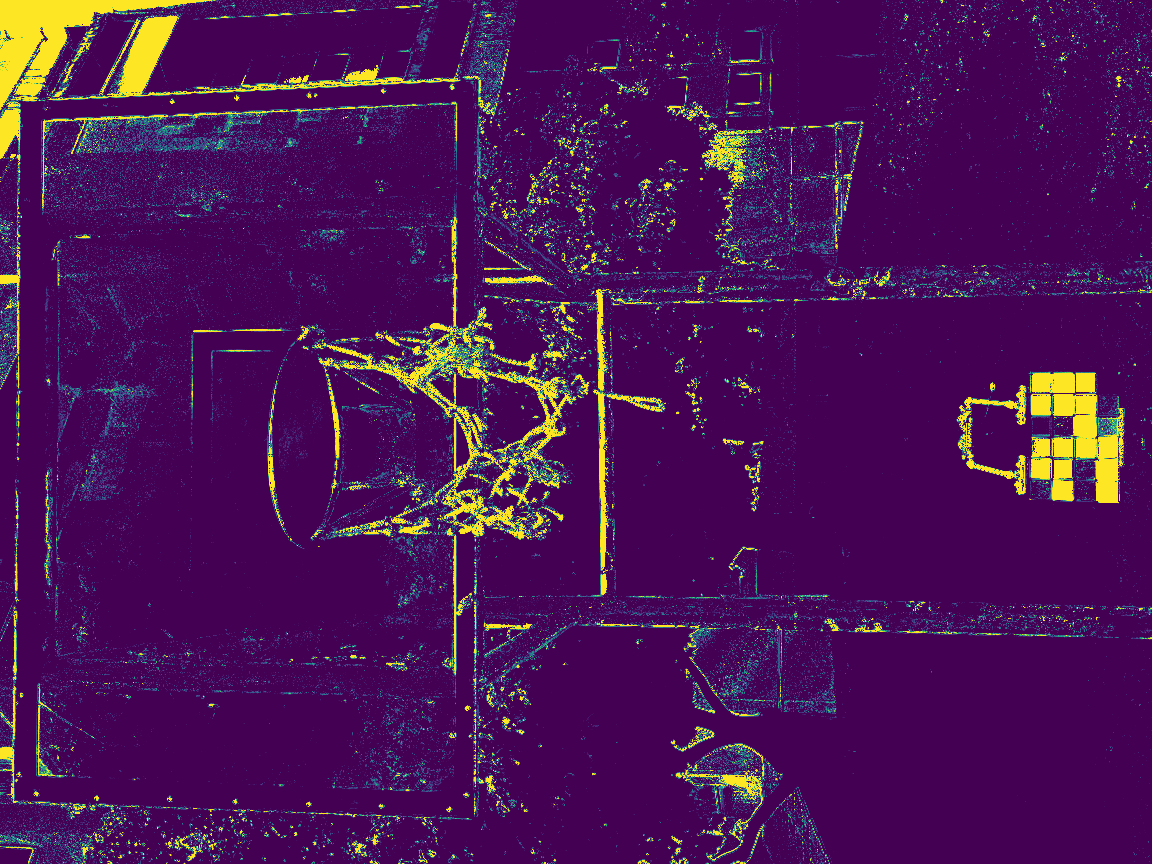}}\includegraphics[width=0.23\linewidth]{figures_arxiv/supp/olympus/OlympusEPL6_0082_st4_err_rang.png}\llap{\raisebox{\dimen0-7pt}{\setlength{\fboxsep}{2pt}\colorbox{white}{\scriptsize	 PSNR: 33.96dB}}} & \settototalheight{\dimen0}{\includegraphics[width=0.23\linewidth]{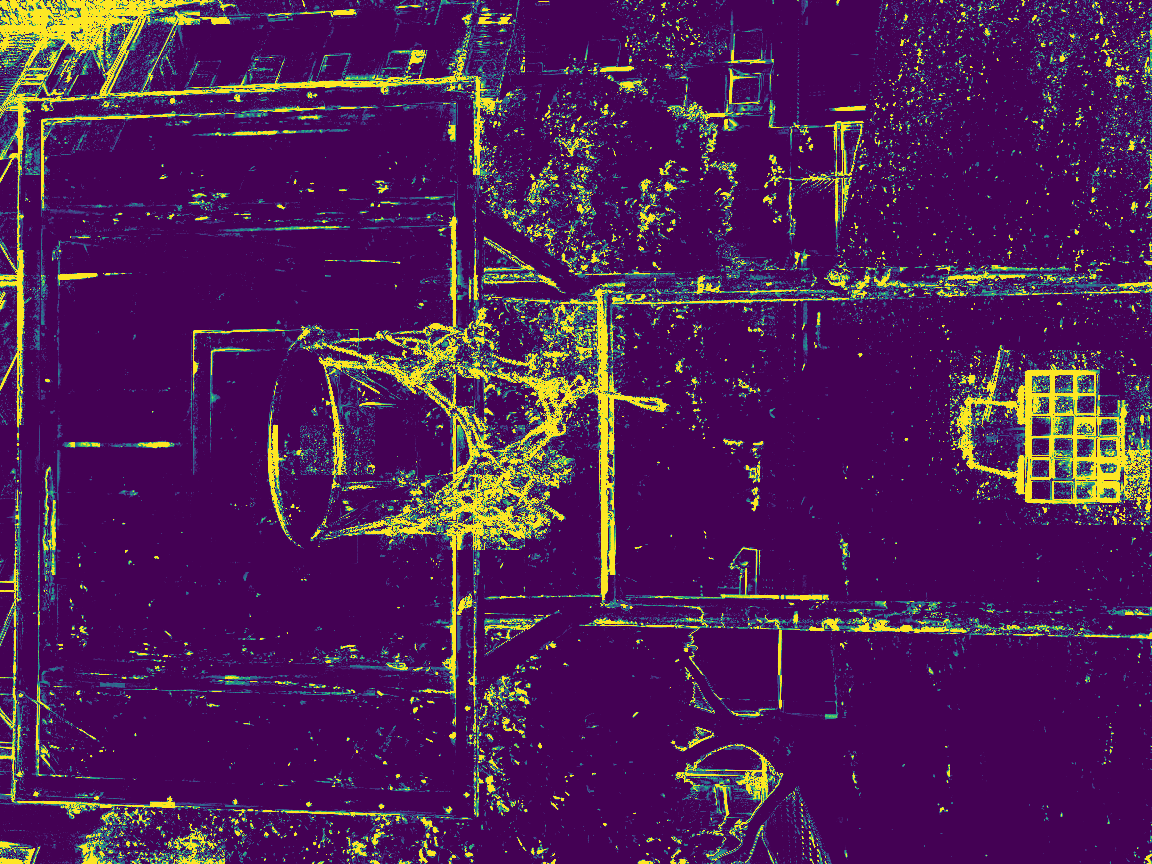}}\includegraphics[width=0.23\linewidth]{figures_arxiv/supp/olympus/OlympusEPL6_0082_st4_err_wacv.png}\llap{\raisebox{\dimen0-7pt}{\setlength{\fboxsep}{2pt}\colorbox{white}{\scriptsize	 PSNR: 40.44dB}}} & \settototalheight{\dimen0}{\includegraphics[width=0.23\linewidth]{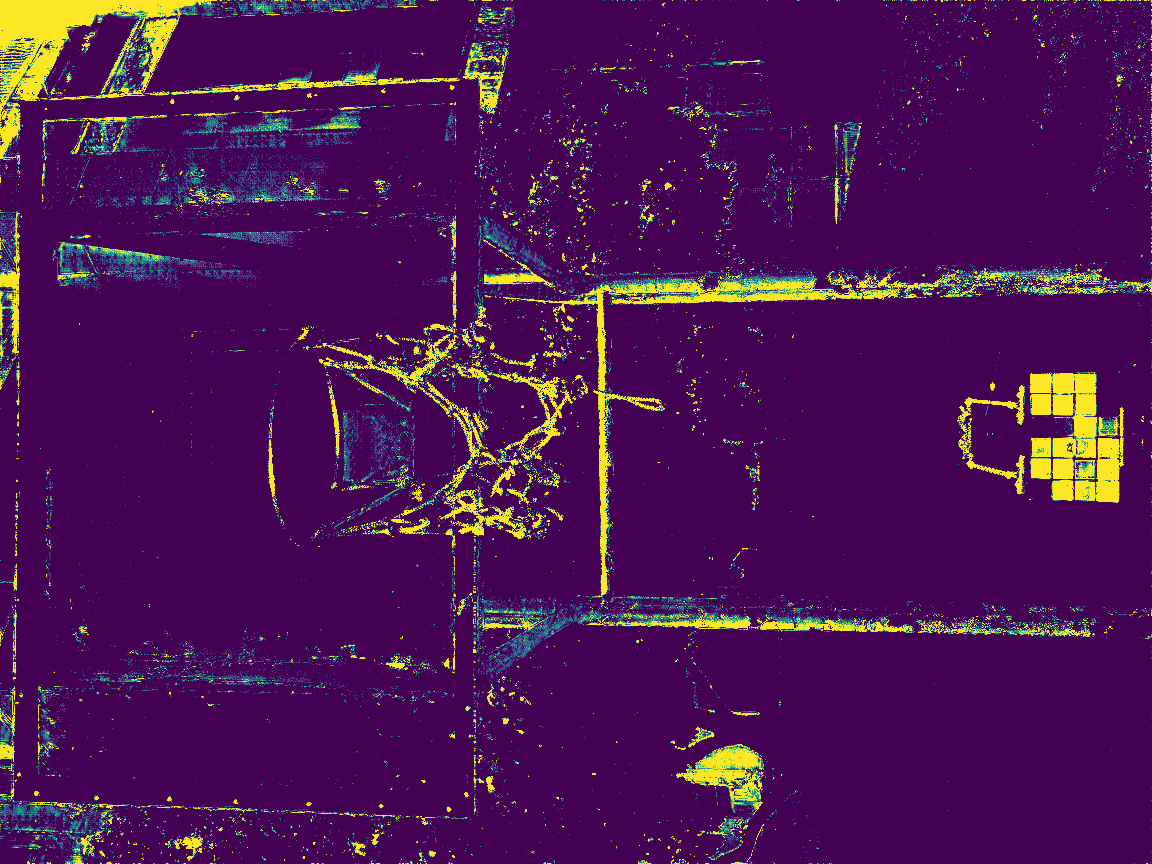}}\includegraphics[width=0.23\linewidth]{figures_arxiv/supp/olympus/0000008_err.png}\llap{\raisebox{\dimen0-7pt}{\setlength{\fboxsep}{2pt}\colorbox{white}{\scriptsize	 PSNR: 43.53.dB}}} & \includegraphics[width=0.027\linewidth]{figures_arxiv/colorbar.pdf} \\

        \includegraphics[width=0.23\linewidth]{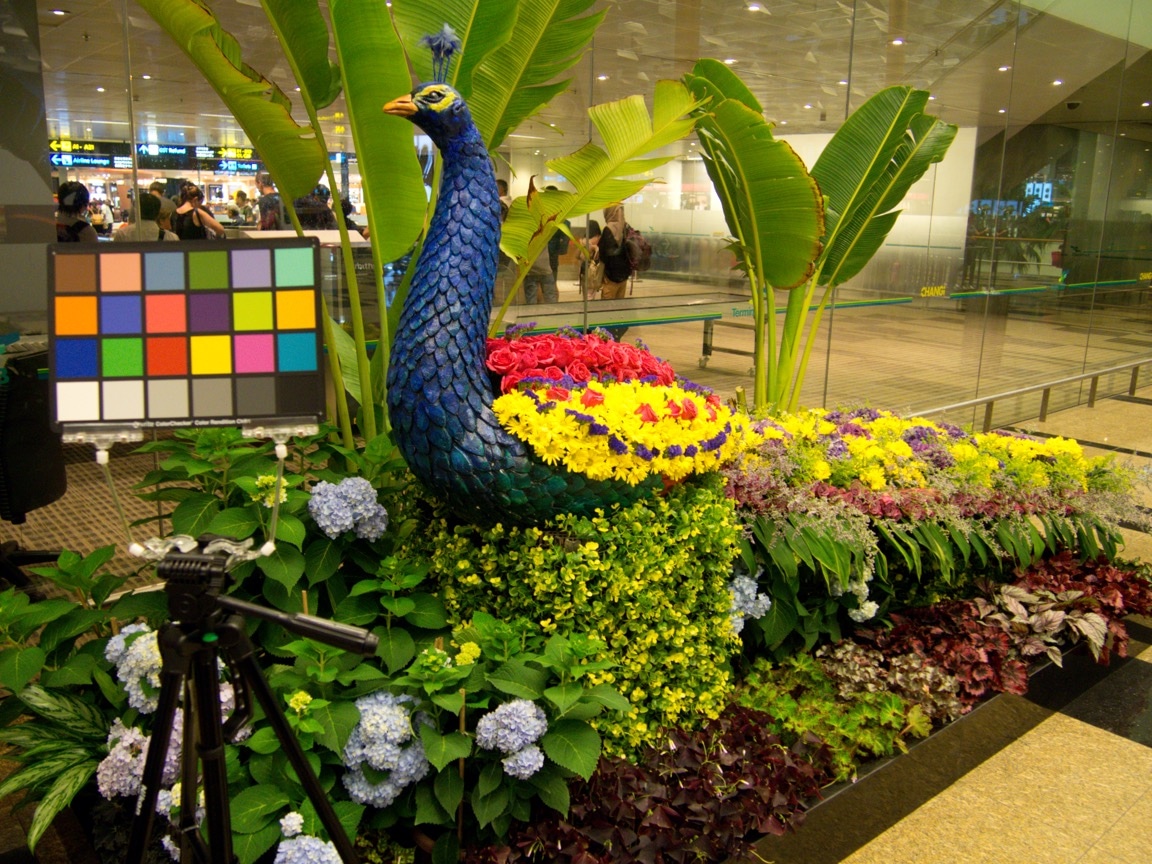} & \settototalheight{\dimen0}{\includegraphics[width=0.23\linewidth]{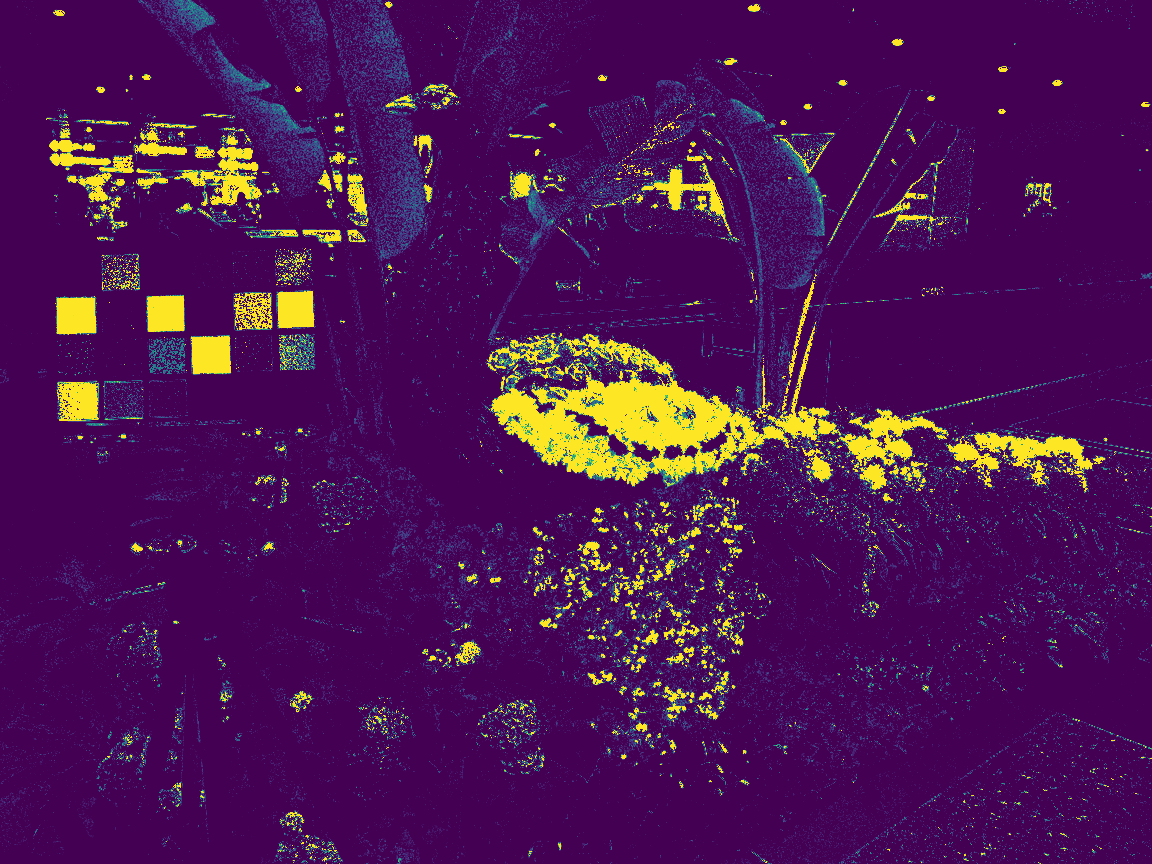}}\includegraphics[width=0.23\linewidth]{figures_arxiv/supp/olympus/OlympusEPL6_0133_st4_err_rang.png}\llap{\raisebox{\dimen0-7pt}{\setlength{\fboxsep}{2pt}\colorbox{white}{\scriptsize	 PSNR: 31.73dB}}} & \settototalheight{\dimen0}{\includegraphics[width=0.23\linewidth]{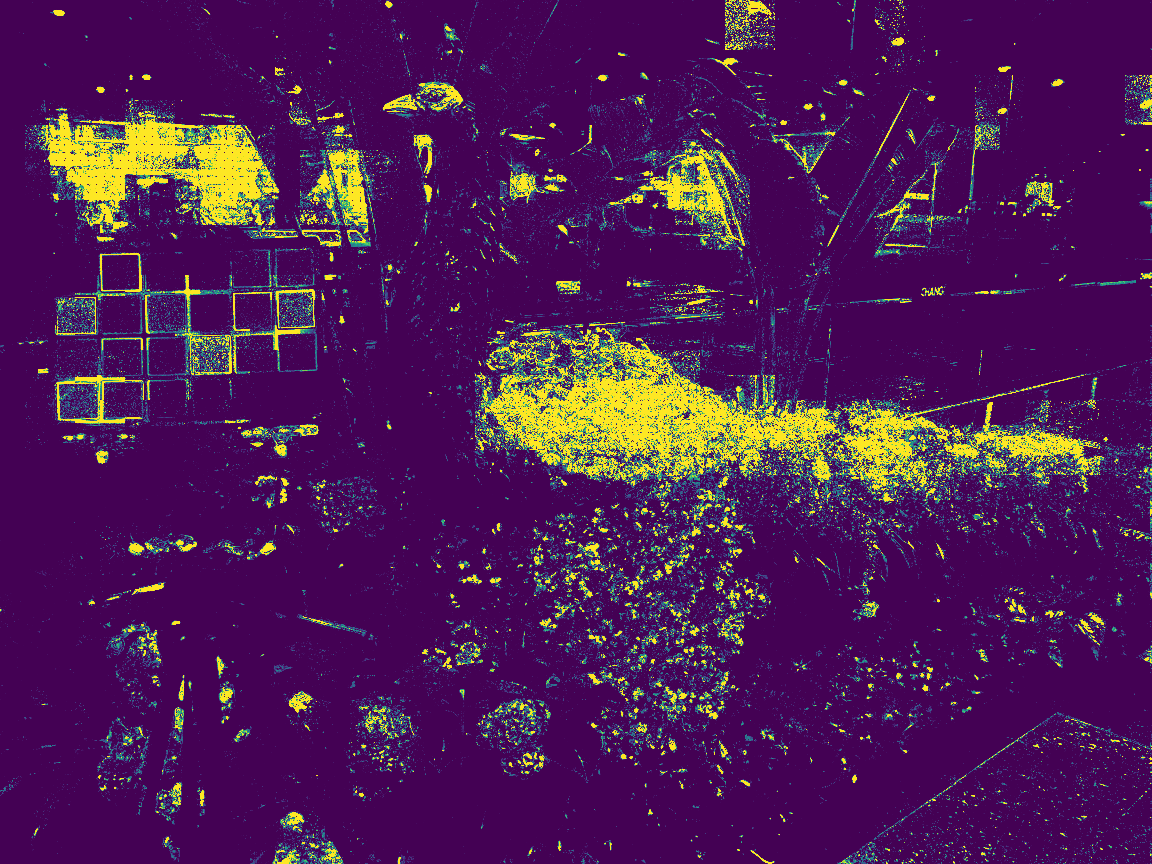}}\includegraphics[width=0.23\linewidth]{figures_arxiv/supp/olympus/OlympusEPL6_0133_st4_err_wacv.png}\llap{\raisebox{\dimen0-7pt}{\setlength{\fboxsep}{2pt}\colorbox{white}{\scriptsize	 PSNR: 36.05dB}}} & \settototalheight{\dimen0}{\includegraphics[width=0.23\linewidth]{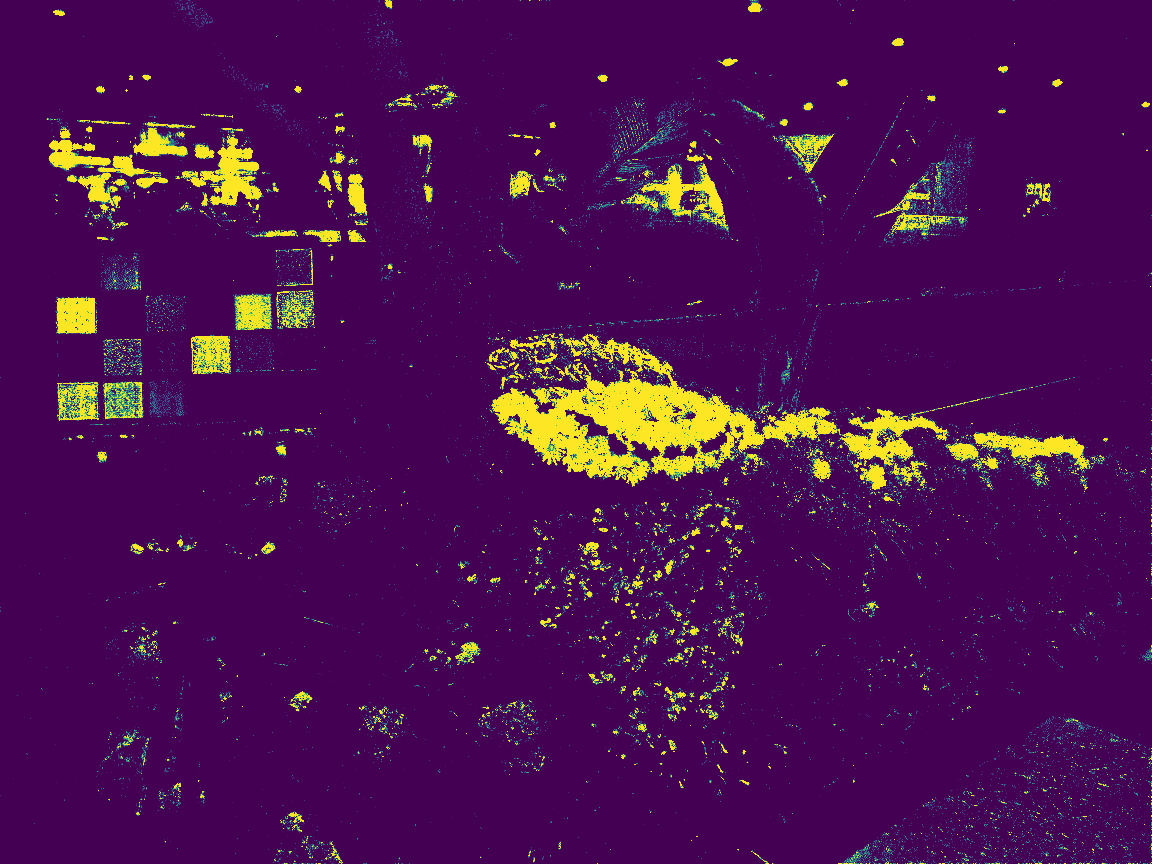}}\includegraphics[width=0.23\linewidth]{figures_arxiv/supp/olympus/0000015_err.png}\llap{\raisebox{\dimen0-7pt}{\setlength{\fboxsep}{2pt}\colorbox{white}{\scriptsize	 PSNR: 40.05.dB}}} & \includegraphics[width=0.027\linewidth]{figures_arxiv/colorbar.pdf} \\

        \includegraphics[width=0.23\linewidth]{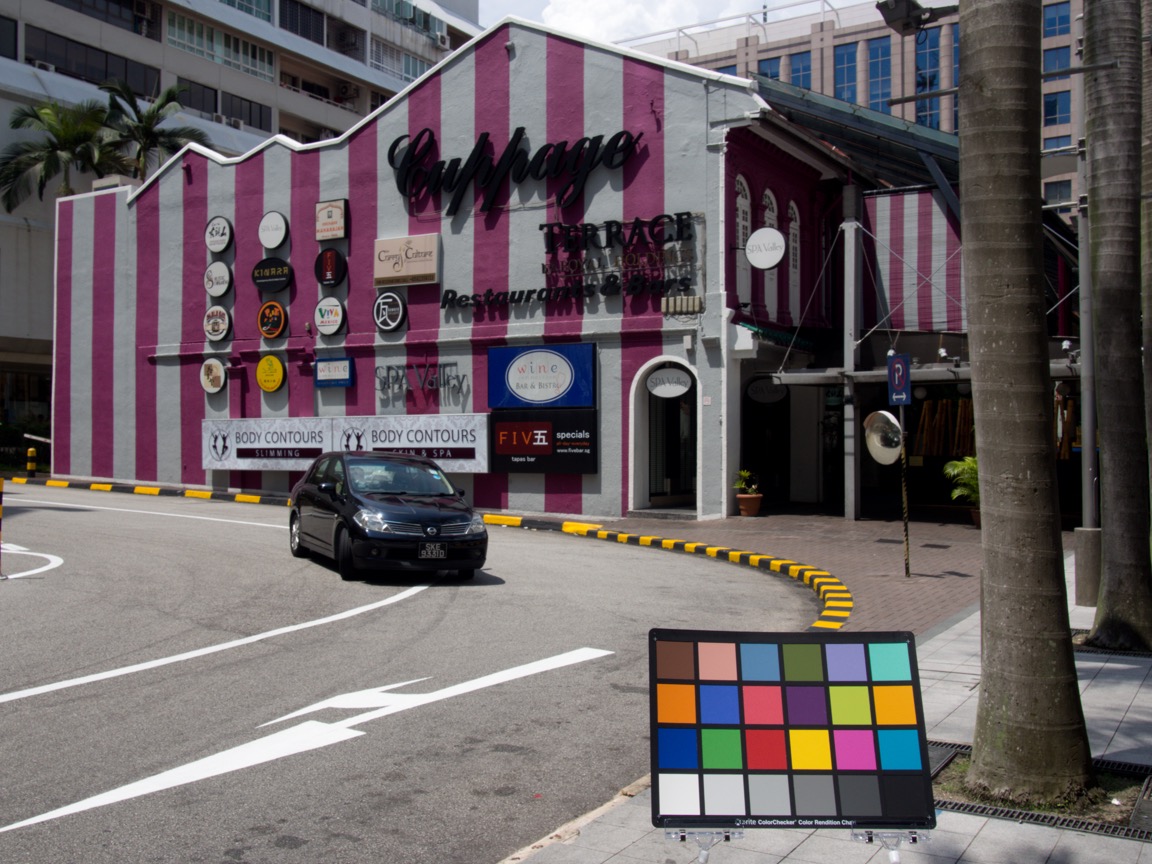} & \settototalheight{\dimen0}{\includegraphics[width=0.23\linewidth]{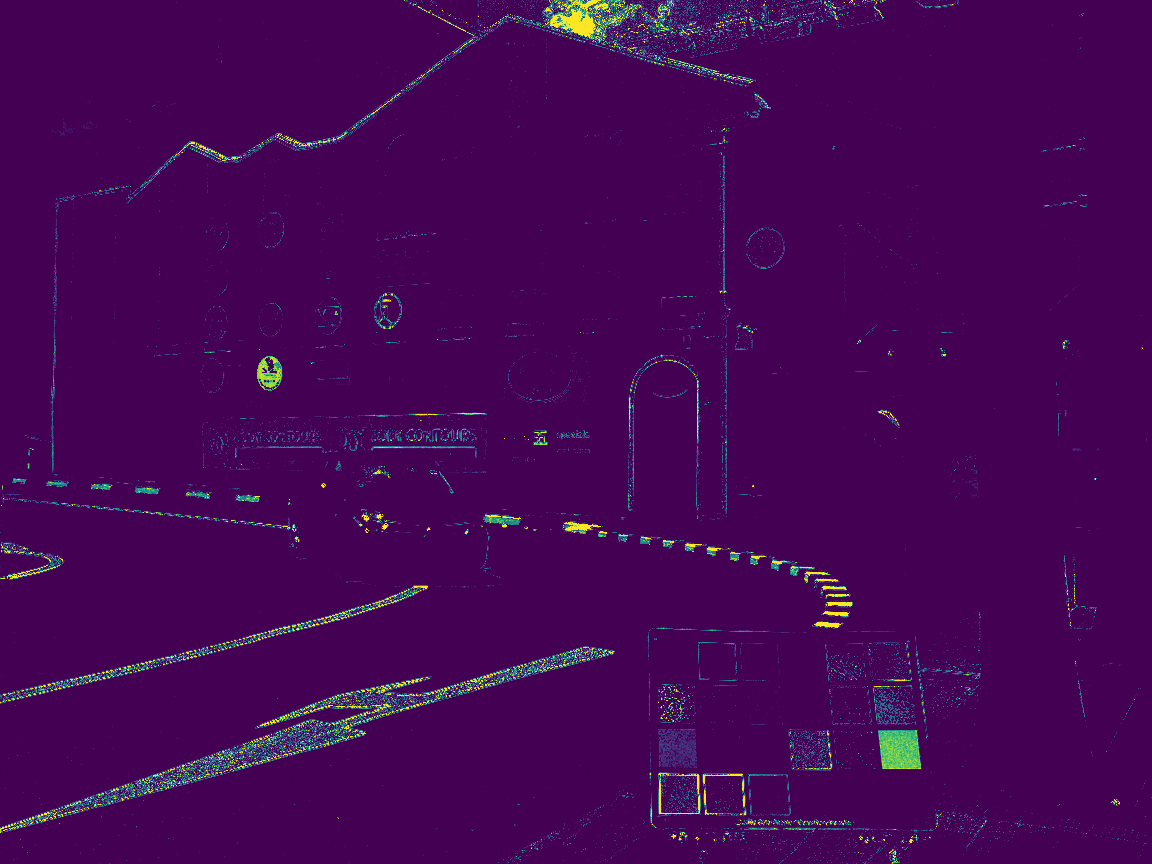}}\includegraphics[width=0.23\linewidth]{figures_arxiv/supp/olympus/OlympusEPL6_0169_st4_err_rang.png}\llap{\raisebox{\dimen0-7pt}{\setlength{\fboxsep}{2pt}\colorbox{white}{\scriptsize	 PSNR: 54.24dB}}} & \settototalheight{\dimen0}{\includegraphics[width=0.23\linewidth]{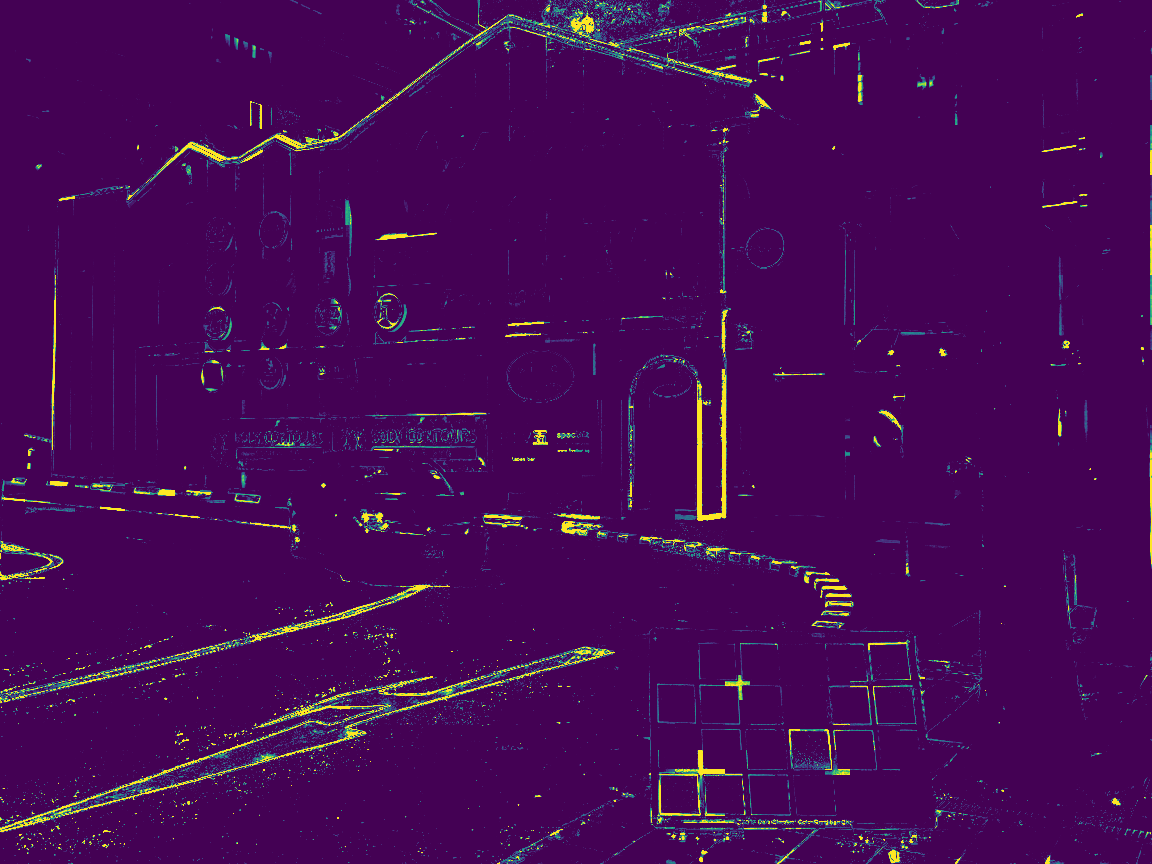}}\includegraphics[width=0.23\linewidth]{figures_arxiv/supp/olympus/OlympusEPL6_0169_st4_err_wacv.png}\llap{\raisebox{\dimen0-7pt}{\setlength{\fboxsep}{2pt}\colorbox{white}{\scriptsize	 PSNR: 48.84dB}}} & \settototalheight{\dimen0}{\includegraphics[width=0.23\linewidth]{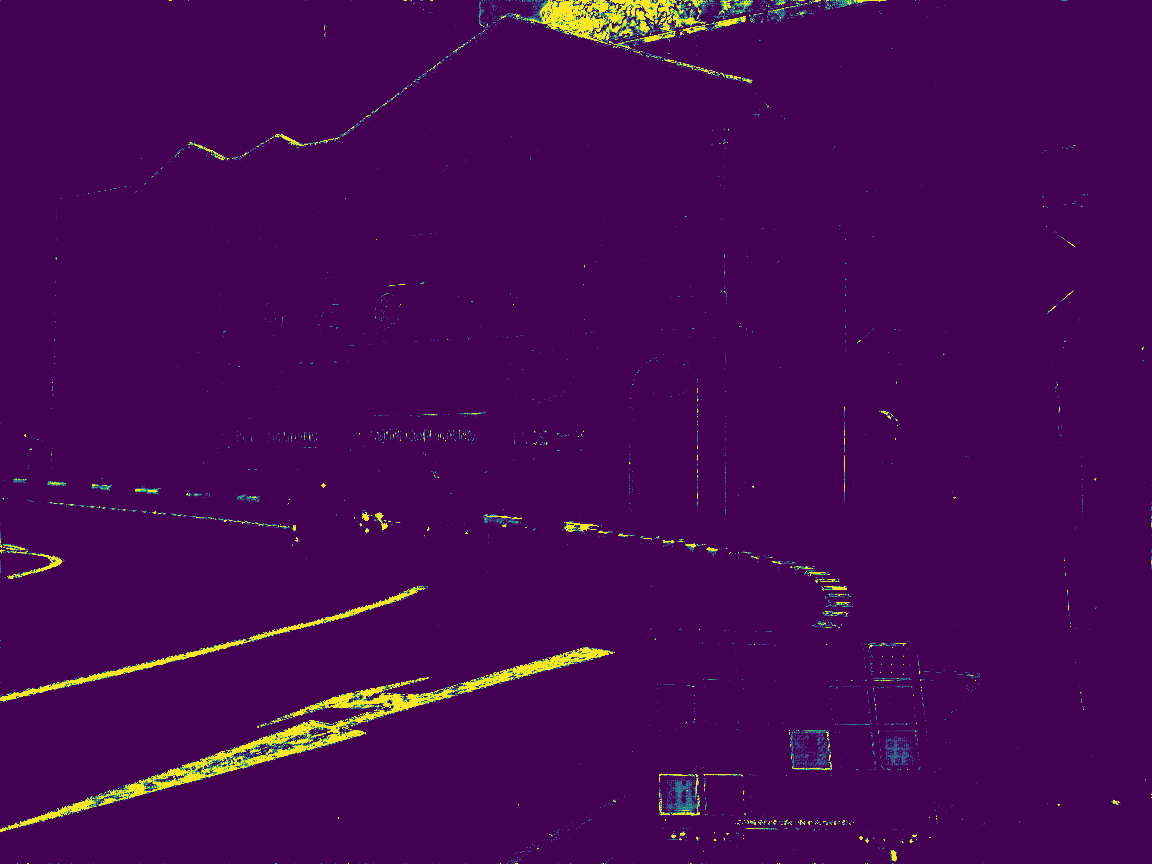}}\includegraphics[width=0.23\linewidth]{figures_arxiv/supp/olympus/0000020_err.png}\llap{\raisebox{\dimen0-7pt}{\setlength{\fboxsep}{2pt}\colorbox{white}{\scriptsize	 PSNR: 55.60.dB}}} & \includegraphics[width=0.027\linewidth]{figures_arxiv/colorbar.pdf} \\

        \includegraphics[width=0.23\linewidth]{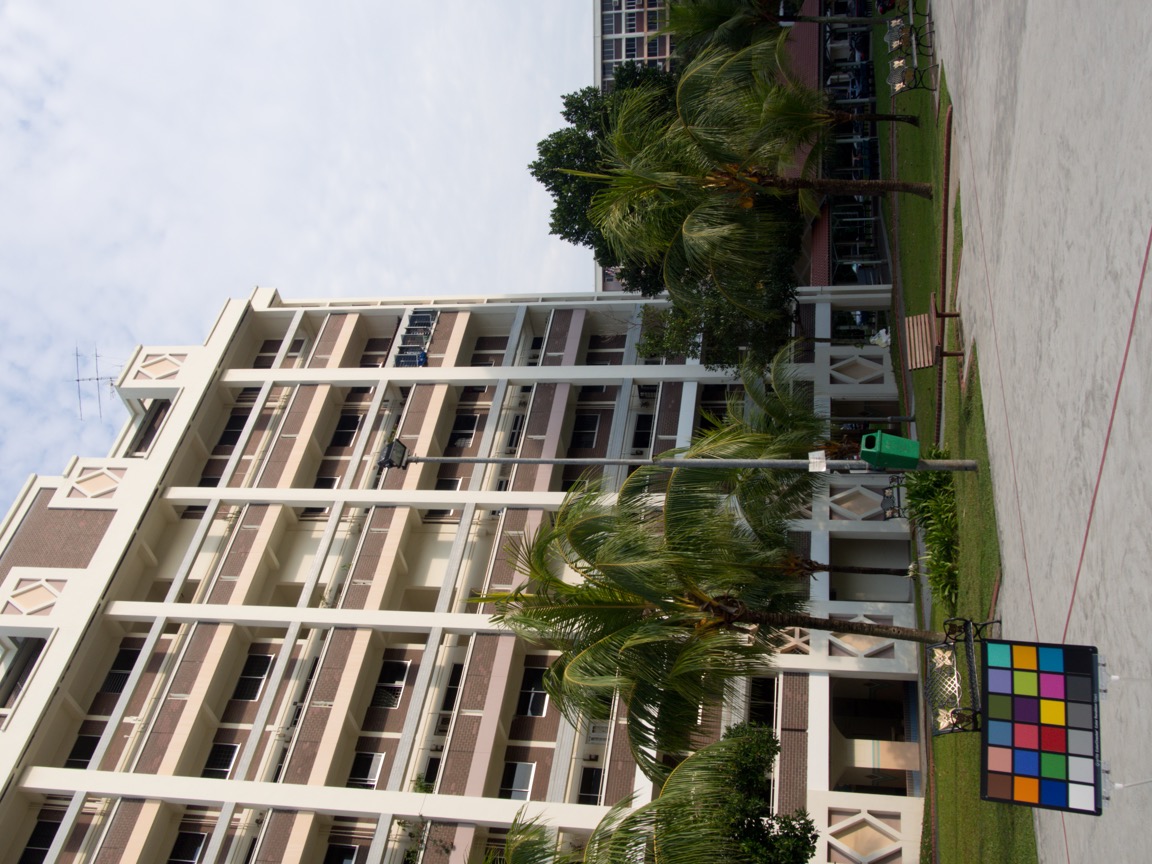} & \settototalheight{\dimen0}{\includegraphics[width=0.23\linewidth]{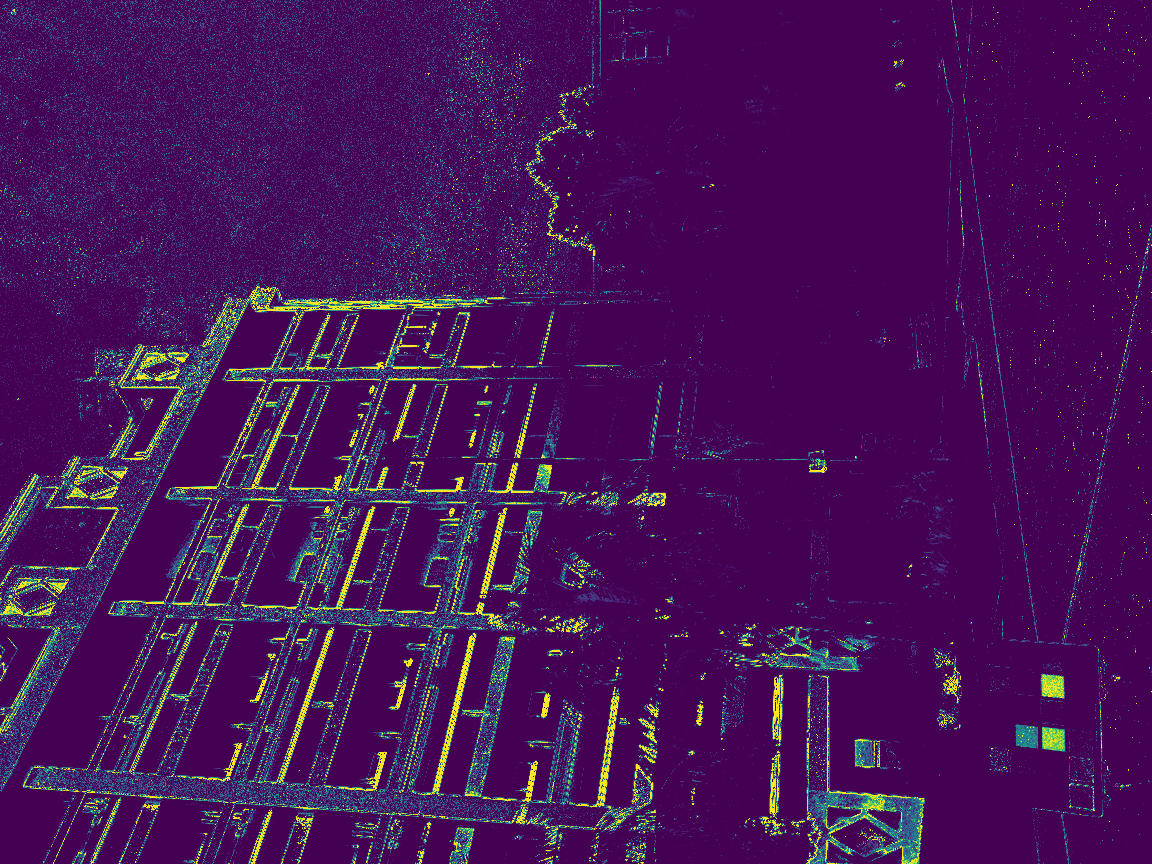}}\includegraphics[width=0.23\linewidth]{figures_arxiv/supp/olympus/OlympusEPL6_0200_st4_err_rang.png}\llap{\raisebox{\dimen0-7pt}{\setlength{\fboxsep}{2pt}\colorbox{white}{\scriptsize	 PSNR: 49.15dB}}} & \settototalheight{\dimen0}{\includegraphics[width=0.23\linewidth]{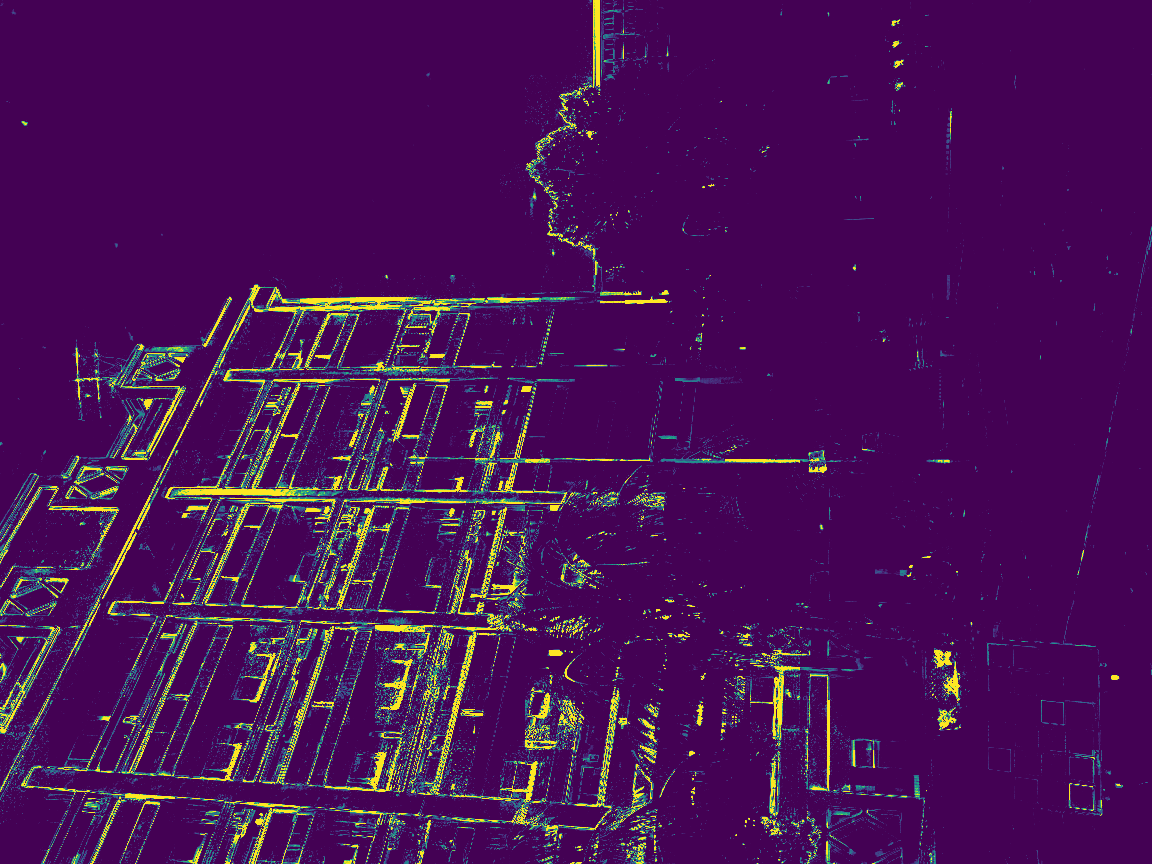}}\includegraphics[width=0.23\linewidth]{figures_arxiv/supp/olympus/OlympusEPL6_0200_st4_err_wacv.png}\llap{\raisebox{\dimen0-7pt}{\setlength{\fboxsep}{2pt}\colorbox{white}{\scriptsize	 PSNR: 46.72dB}}} & \settototalheight{\dimen0}{\includegraphics[width=0.23\linewidth]{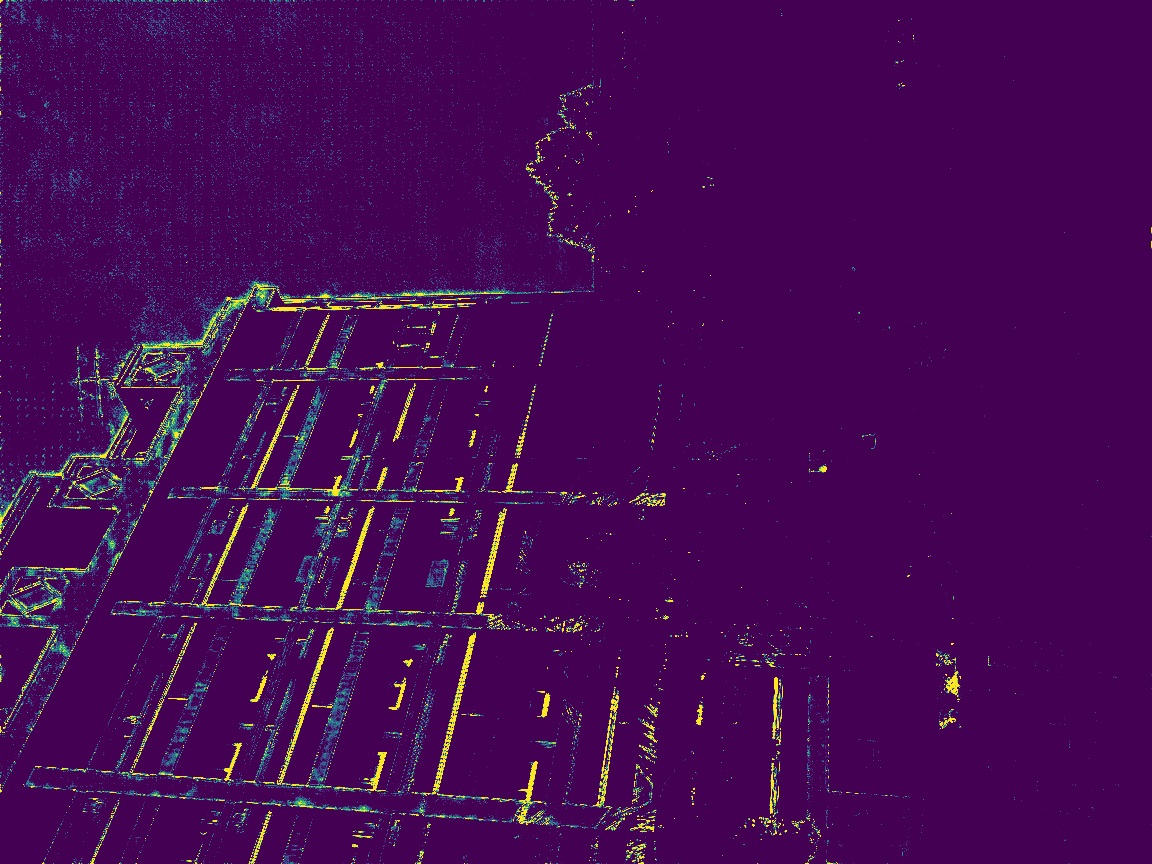}}\includegraphics[width=0.23\linewidth]{figures_arxiv/supp/olympus/0000023_err.png}\llap{\raisebox{\dimen0-7pt}{\setlength{\fboxsep}{2pt}\colorbox{white}{\scriptsize	 PSNR: 51.12dB}}} & \includegraphics[width=0.027\linewidth]{figures_arxiv/colorbar.pdf} \\

        {\small Input} & {\small RIR~\cite{rang}} & {\small SAM~\cite{wacv}} & {\small Ours + fine-tuning} & \\
    \end{tabular}
    \caption{Qualitative comparison on Olympus E-PL6.}
    \label{fig:supp_olympus}
\end{figure*}
\begin{figure*}
    \centering
    \setlength{\tabcolsep}{1pt}
    \begin{tabular}{ccccc}
        \includegraphics[width=0.23\linewidth]{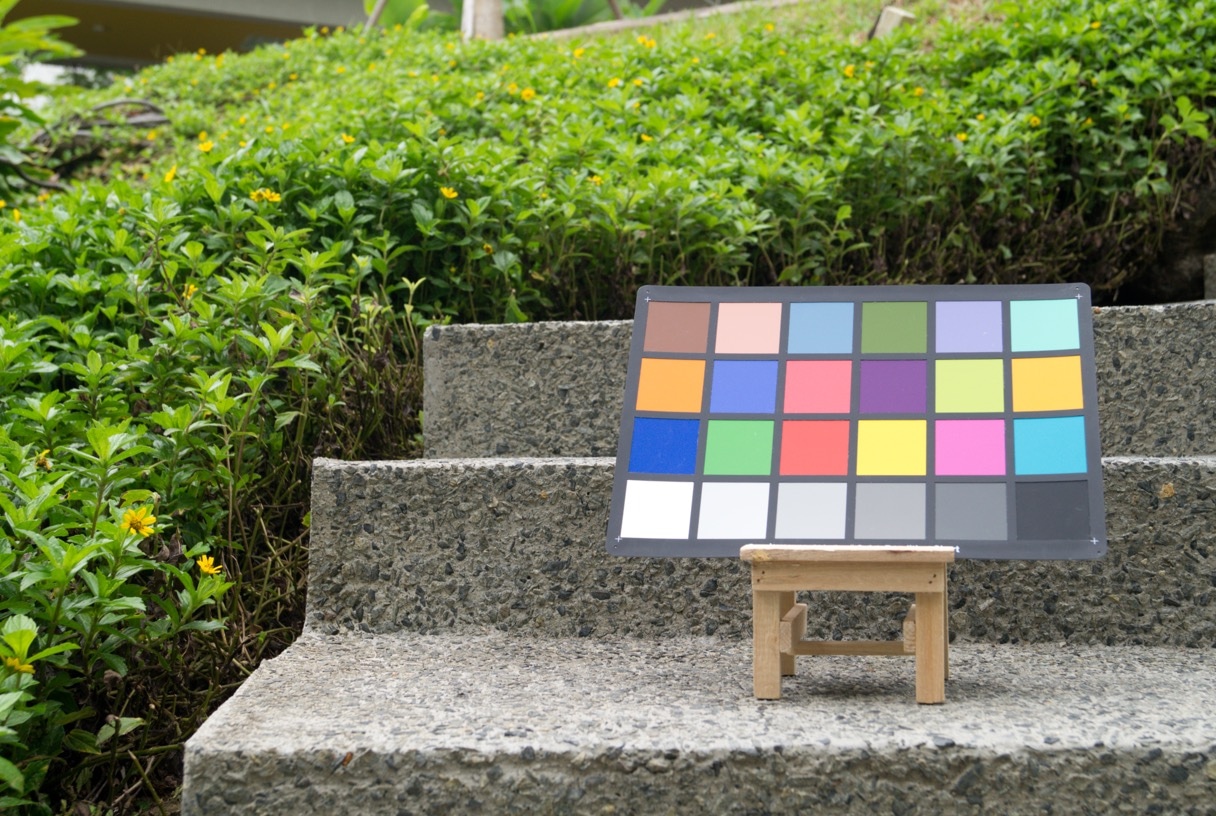} & \settototalheight{\dimen0}{\includegraphics[width=0.23\linewidth]{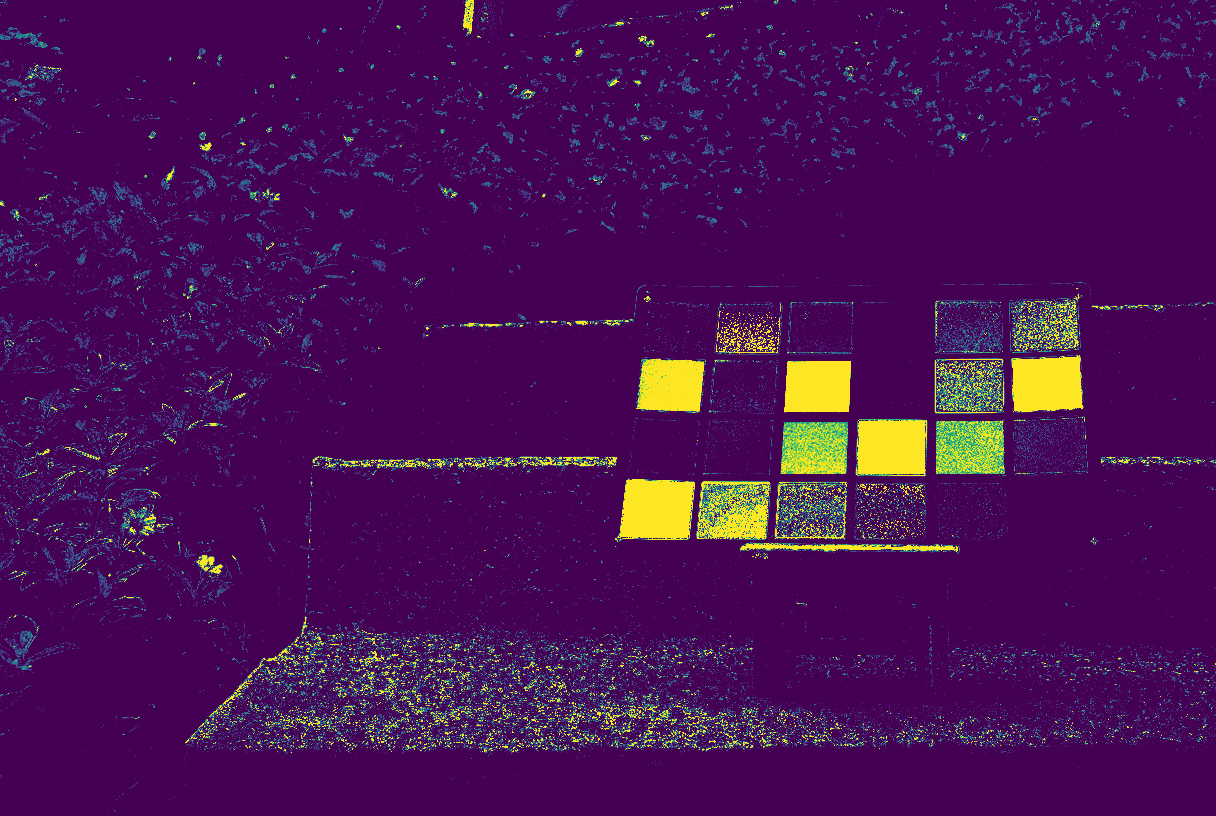}}\includegraphics[width=0.23\linewidth]{figures_arxiv/supp/sony/SonyA57_0060_st4_err_rang.png}\llap{\raisebox{\dimen0-7pt}{\setlength{\fboxsep}{2pt}\colorbox{white}{\scriptsize	 PSNR: 39.94dB}}} & \settototalheight{\dimen0}{\includegraphics[width=0.23\linewidth]{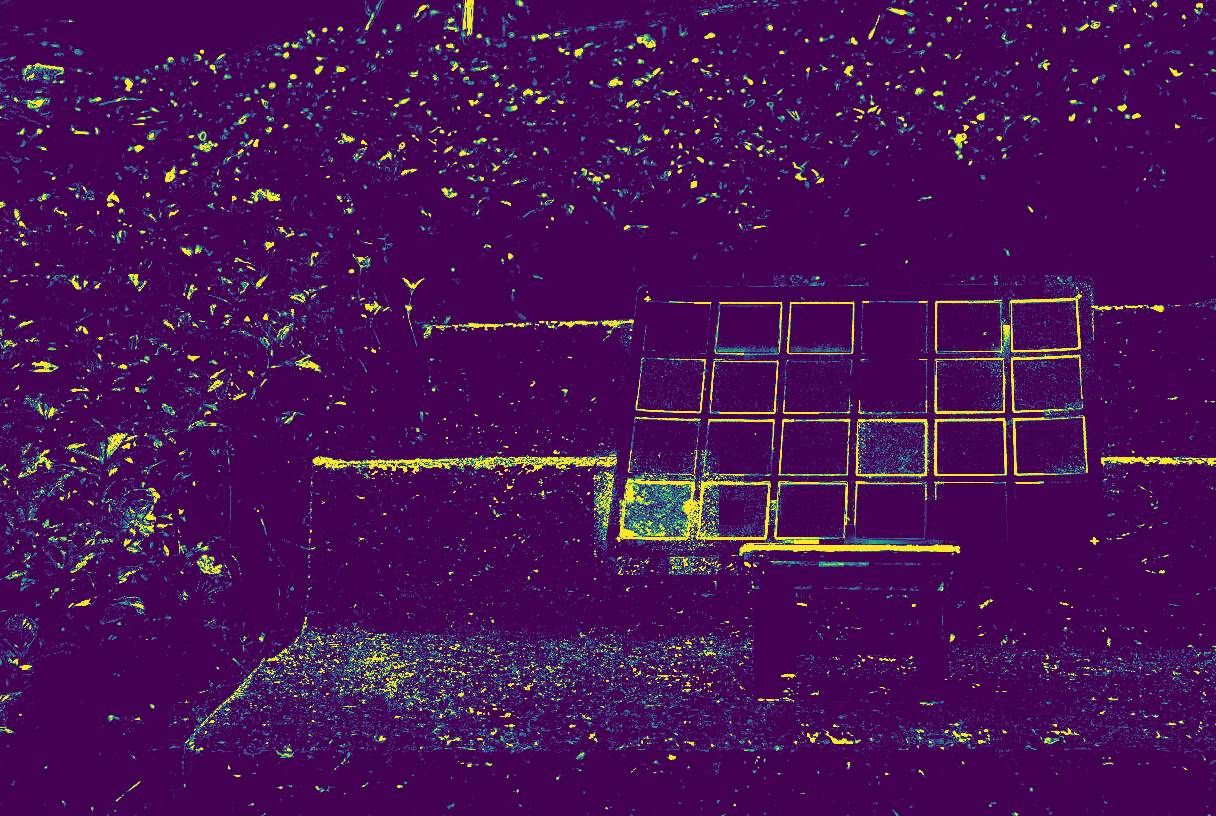}}\includegraphics[width=0.23\linewidth]{figures_arxiv/supp/sony/SonyA57_0060_st4_err_wacv.png}\llap{\raisebox{\dimen0-7pt}{\setlength{\fboxsep}{2pt}\colorbox{white}{\scriptsize	 PSNR: 45.43dB}}} & \settototalheight{\dimen0}{\includegraphics[width=0.23\linewidth]{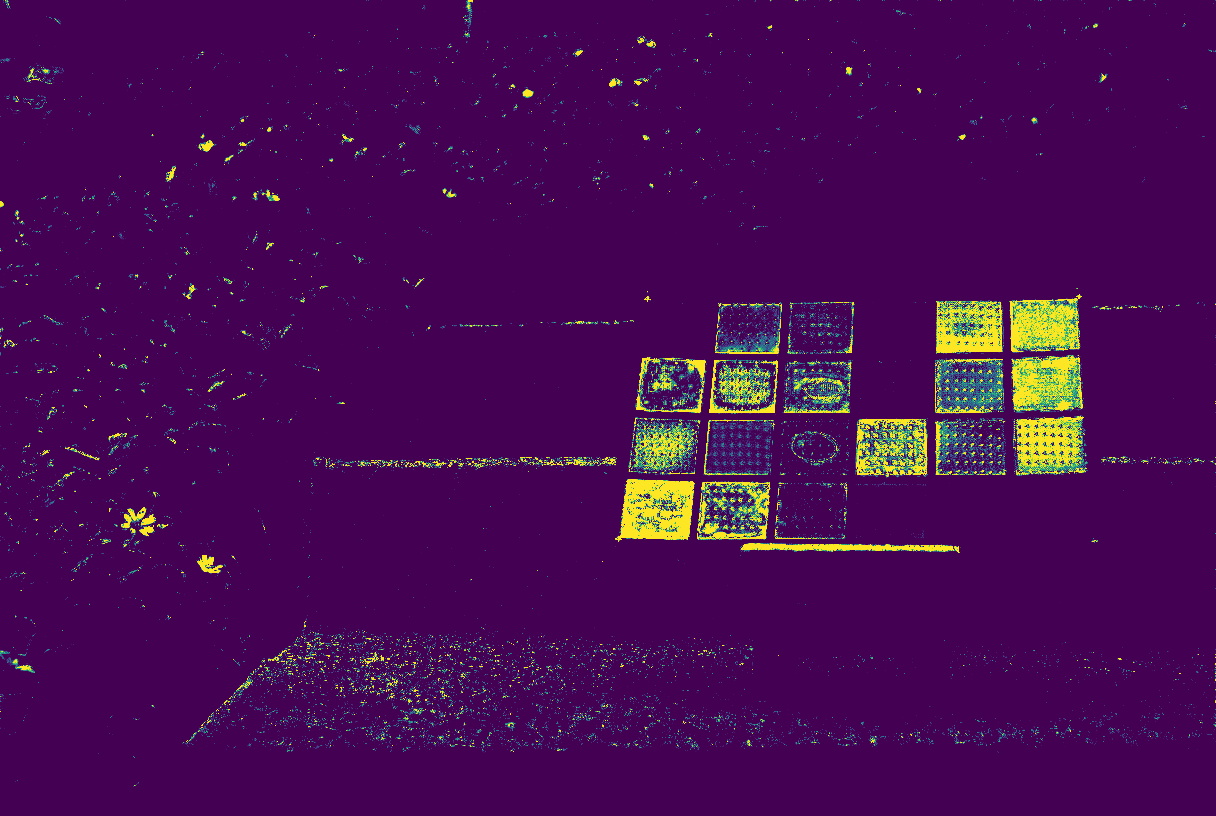}}\includegraphics[width=0.23\linewidth]{figures_arxiv/supp/sony/0000006_err.png}\llap{\raisebox{\dimen0-7pt}{\setlength{\fboxsep}{2pt}\colorbox{white}{\scriptsize	 PSNR: 53.66dB}}} & \includegraphics[width=0.025\linewidth]{figures_arxiv/colorbar.pdf} \\

        \includegraphics[width=0.23\linewidth]{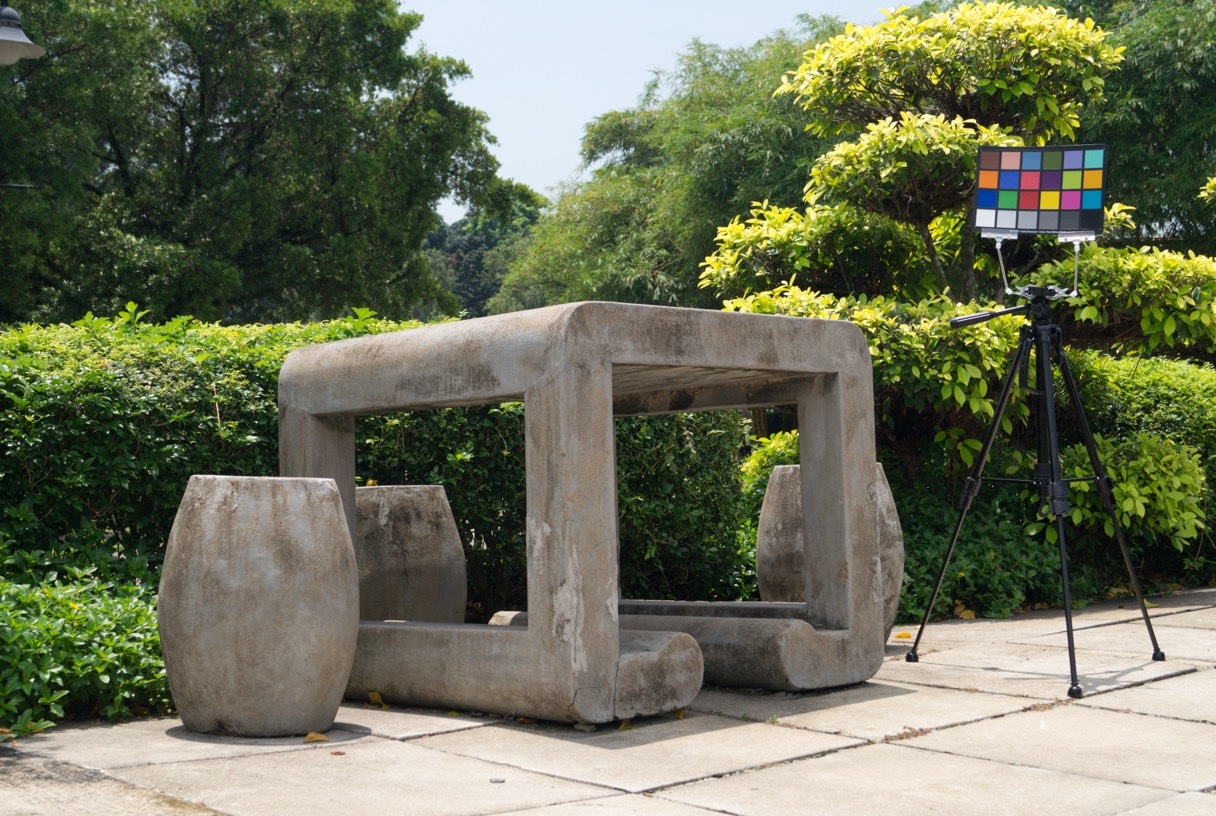} & \settototalheight{\dimen0}{\includegraphics[width=0.23\linewidth]{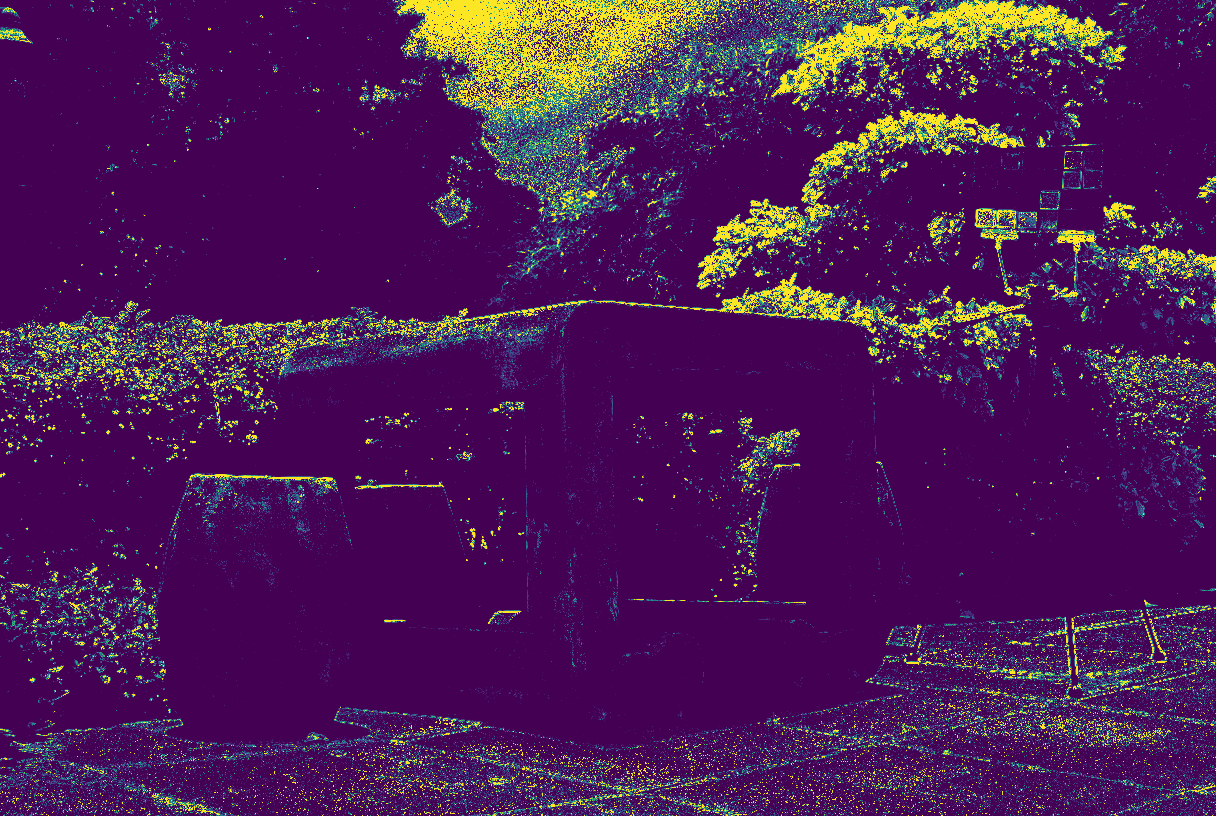}}\includegraphics[width=0.23\linewidth]{figures_arxiv/supp/sony/SonyA57_0076_st4_err_rang.png}\llap{\raisebox{\dimen0-7pt}{\setlength{\fboxsep}{2pt}\colorbox{white}{\scriptsize	 PSNR: 41.38dB}}} & \settototalheight{\dimen0}{\includegraphics[width=0.23\linewidth]{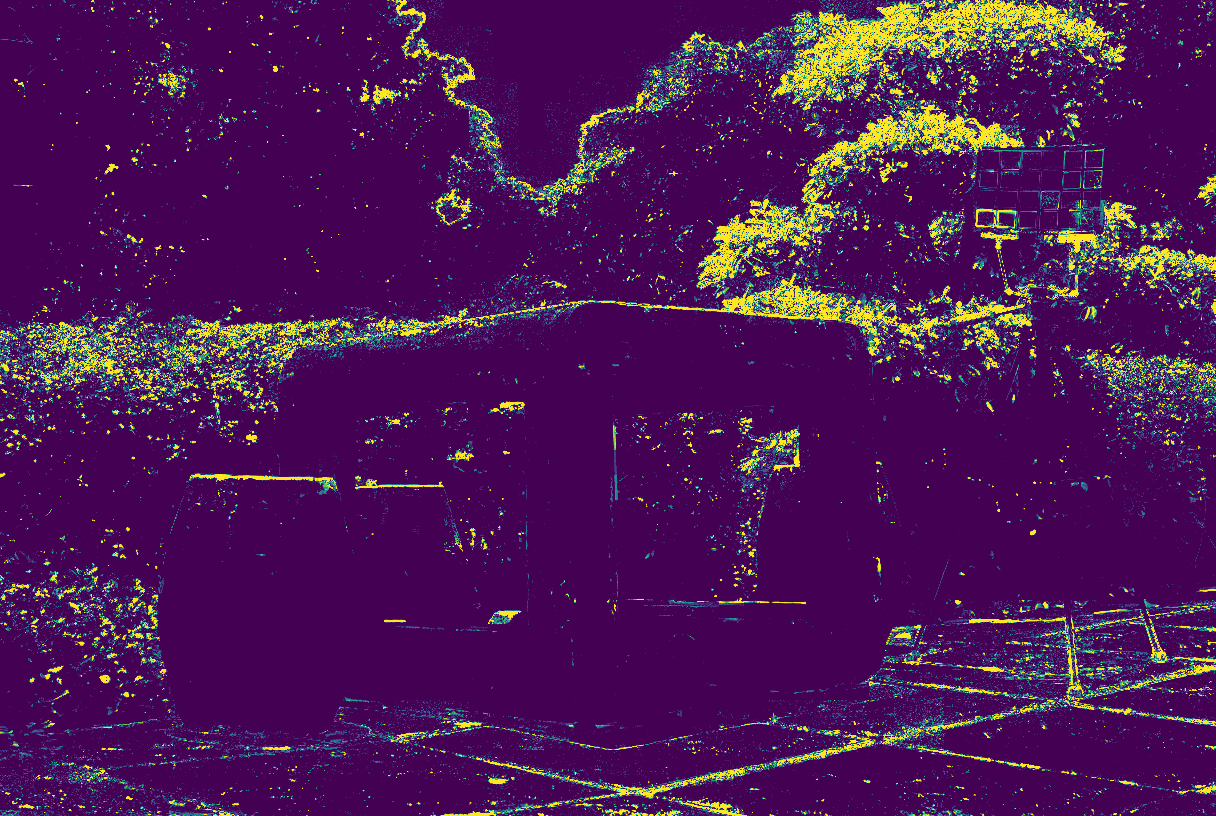}}\includegraphics[width=0.23\linewidth]{figures_arxiv/supp/sony/SonyA57_0076_st4_err_wacv.png}\llap{\raisebox{\dimen0-7pt}{\setlength{\fboxsep}{2pt}\colorbox{white}{\scriptsize	 PSNR: 44.42dB}}} & \settototalheight{\dimen0}{\includegraphics[width=0.23\linewidth]{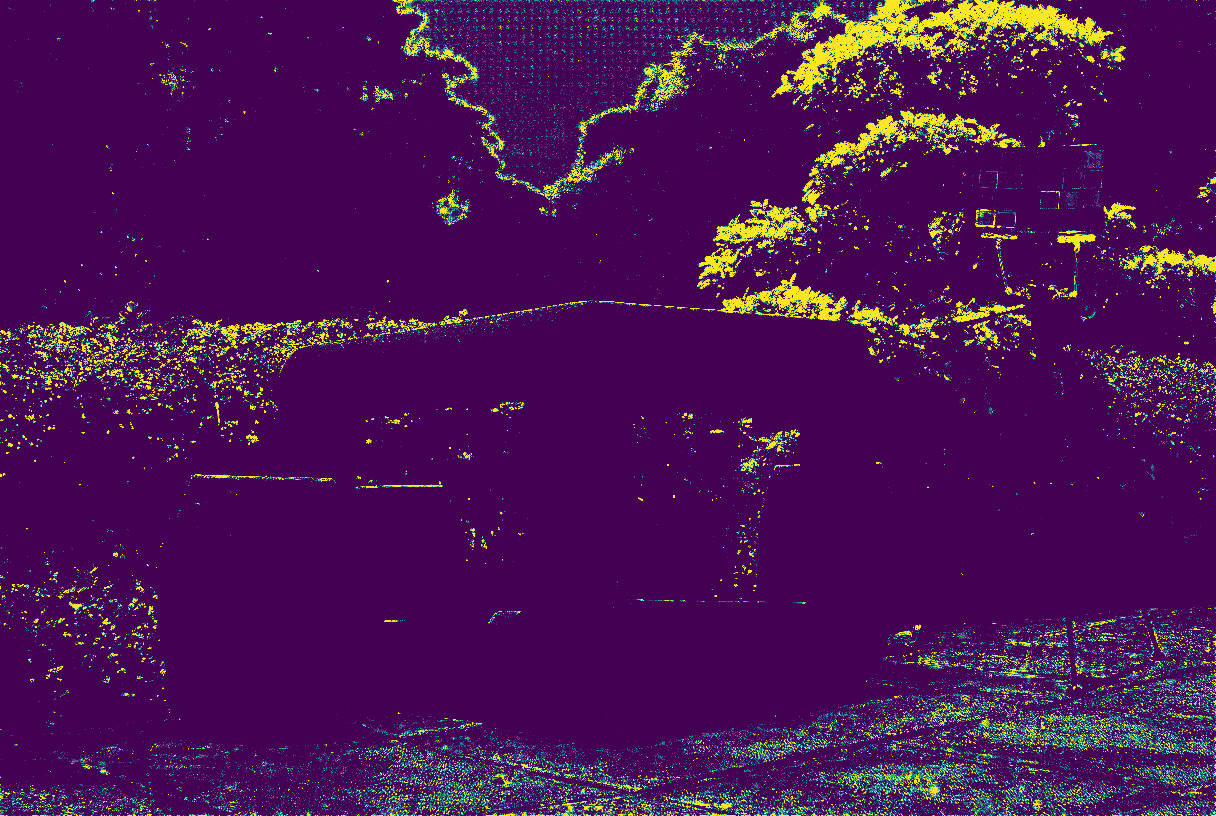}}\includegraphics[width=0.23\linewidth]{figures_arxiv/supp/sony/0000009_err.png}\llap{\raisebox{\dimen0-7pt}{\setlength{\fboxsep}{2pt}\colorbox{white}{\scriptsize	 PSNR: 48.21dB}}} & \includegraphics[width=0.025\linewidth]{figures_arxiv/colorbar.pdf} \\

        \includegraphics[width=0.23\linewidth]{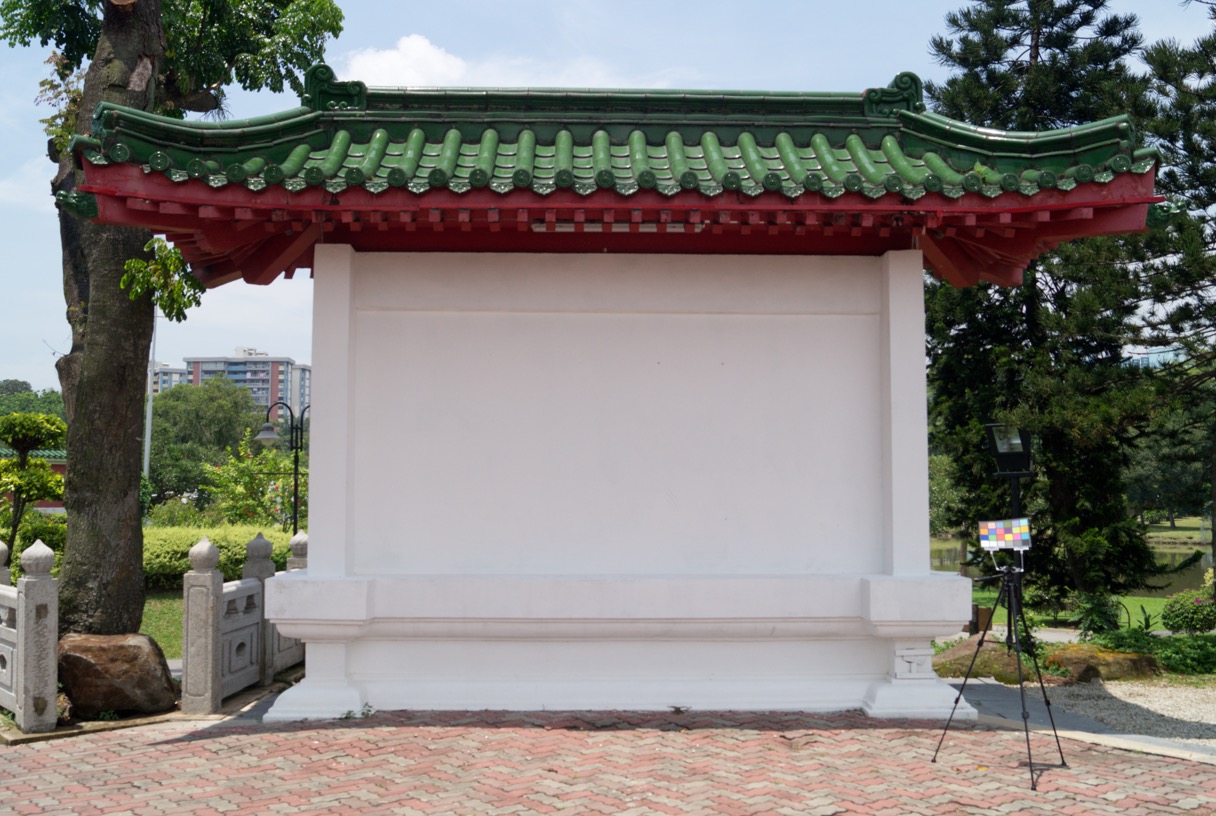} & \settototalheight{\dimen0}{\includegraphics[width=0.23\linewidth]{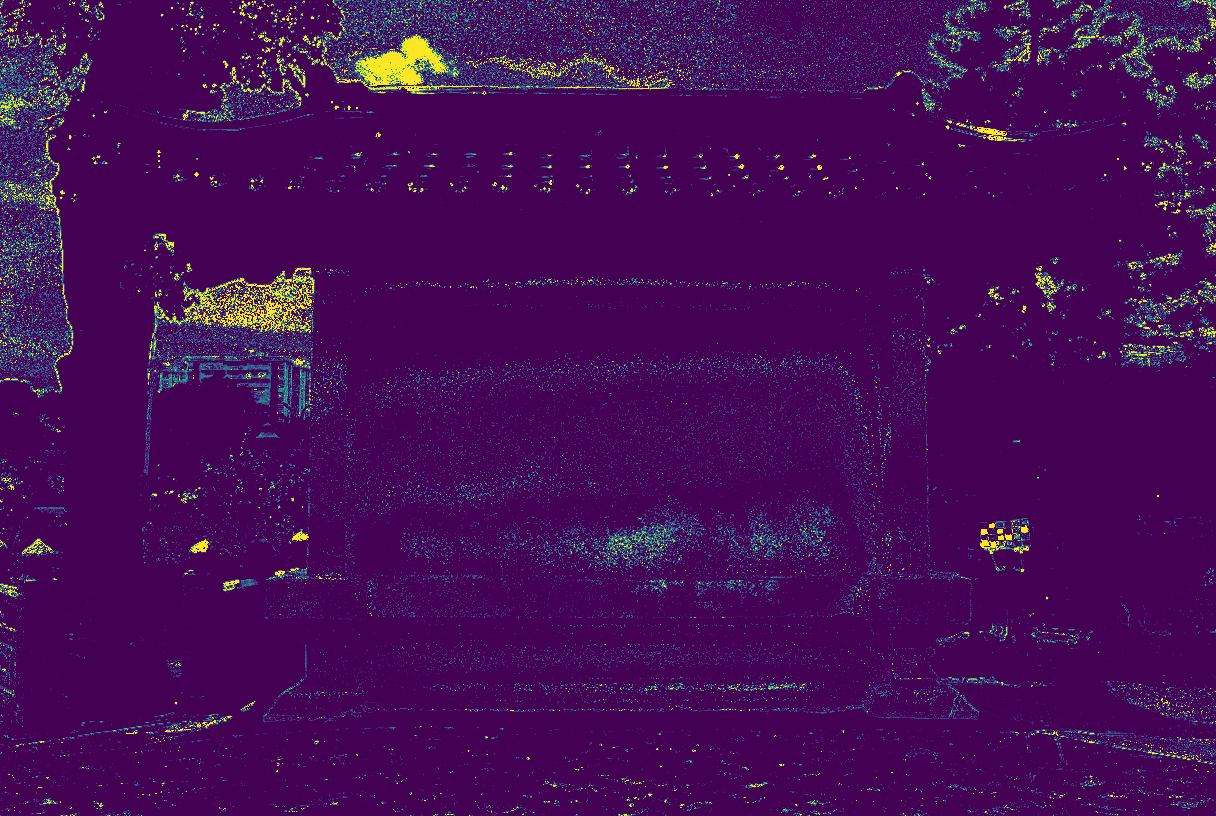}}\includegraphics[width=0.23\linewidth]{figures_arxiv/supp/sony/SonyA57_0077_st4_err_rang.png}\llap{\raisebox{\dimen0-7pt}{\setlength{\fboxsep}{2pt}\colorbox{white}{\scriptsize	 PSNR: 50.14dB}}} & \settototalheight{\dimen0}{\includegraphics[width=0.23\linewidth]{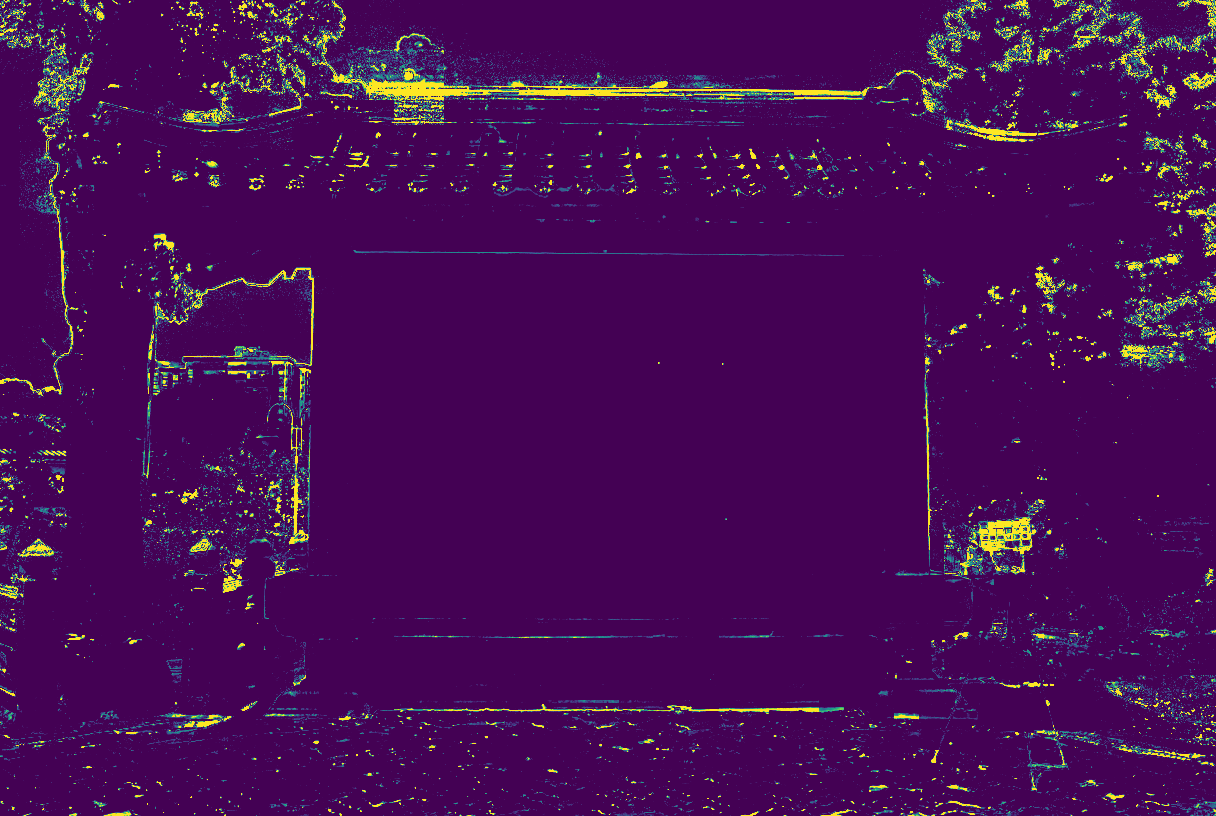}}\includegraphics[width=0.23\linewidth]{figures_arxiv/supp/sony/SonyA57_0077_st4_err_wacv.png}\llap{\raisebox{\dimen0-7pt}{\setlength{\fboxsep}{2pt}\colorbox{white}{\scriptsize	 PSNR: 46.77dB}}} & \settototalheight{\dimen0}{\includegraphics[width=0.23\linewidth]{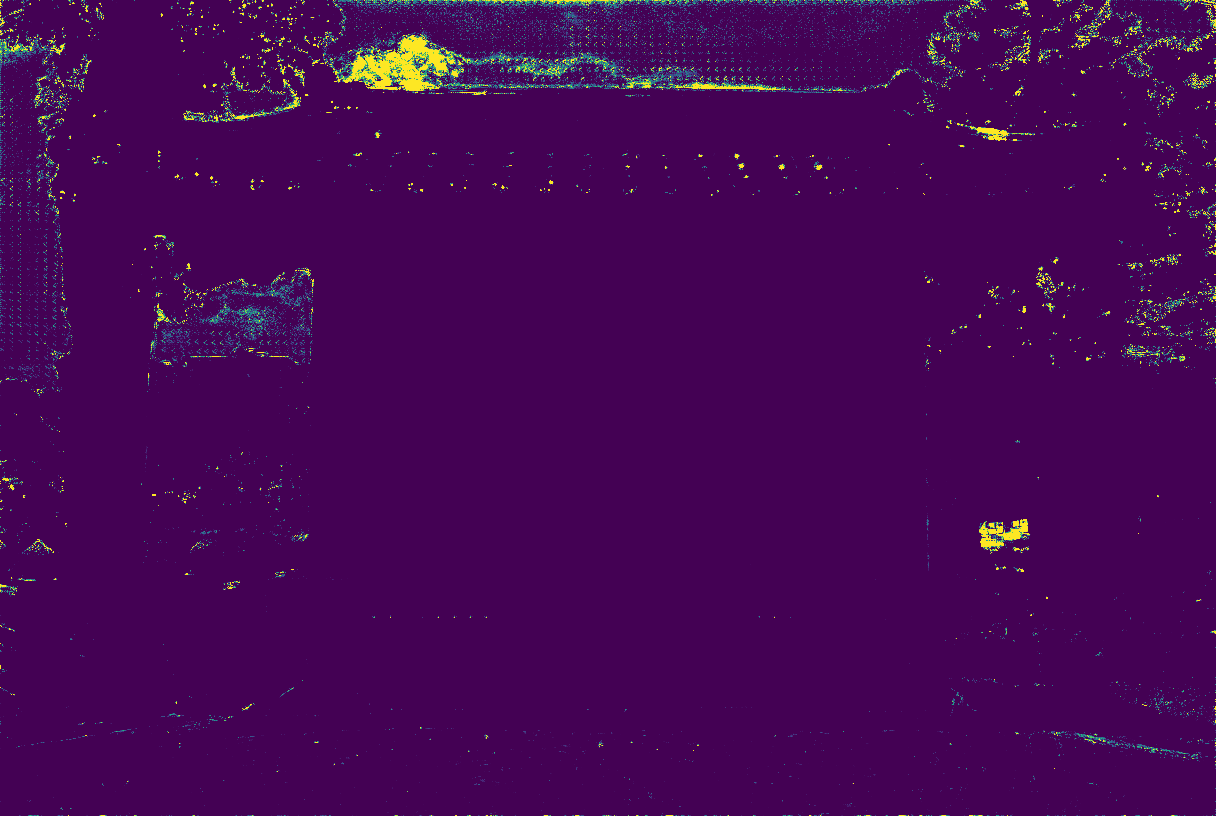}}\includegraphics[width=0.23\linewidth]{figures_arxiv/supp/sony/0000010_err.png}\llap{\raisebox{\dimen0-7pt}{\setlength{\fboxsep}{2pt}\colorbox{white}{\scriptsize	 PSNR: 51.10dB}}} & \includegraphics[width=0.025\linewidth]{figures_arxiv/colorbar.pdf} \\

        \includegraphics[width=0.23\linewidth]{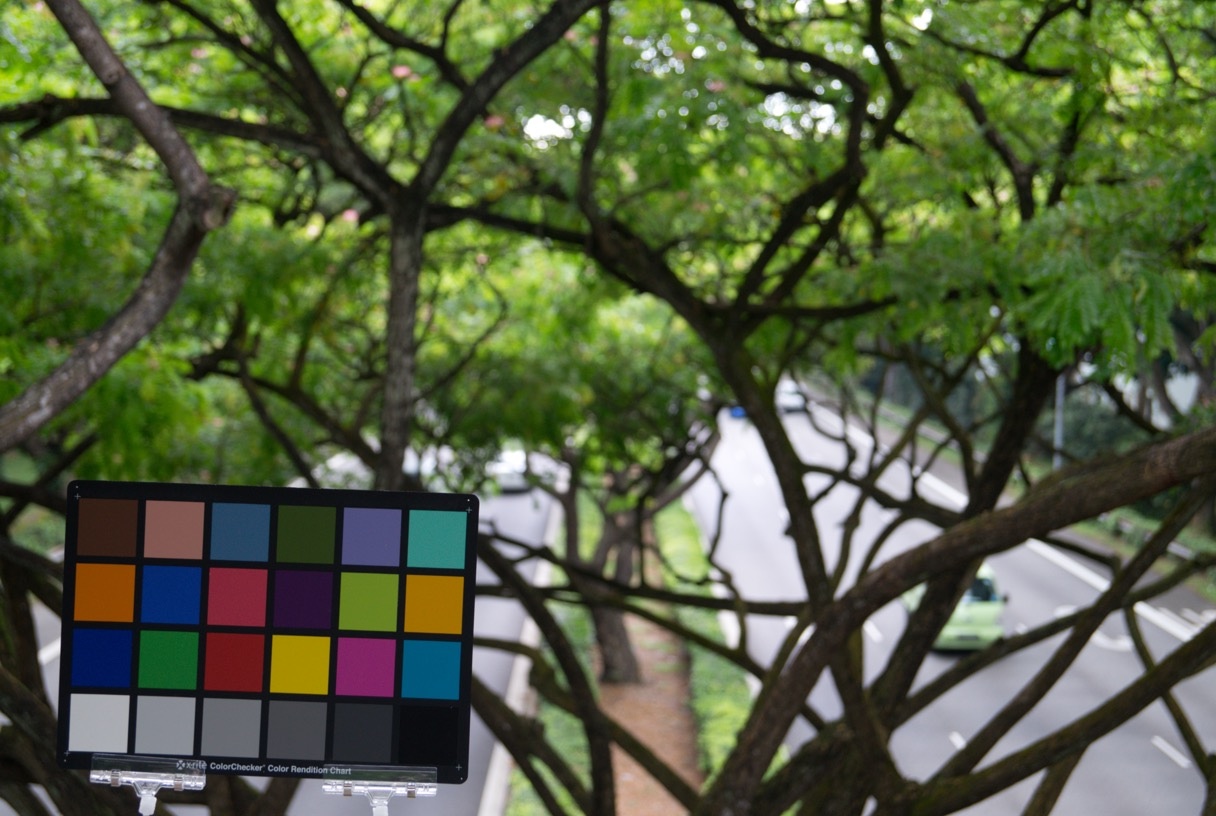} & \settototalheight{\dimen0}{\includegraphics[width=0.23\linewidth]{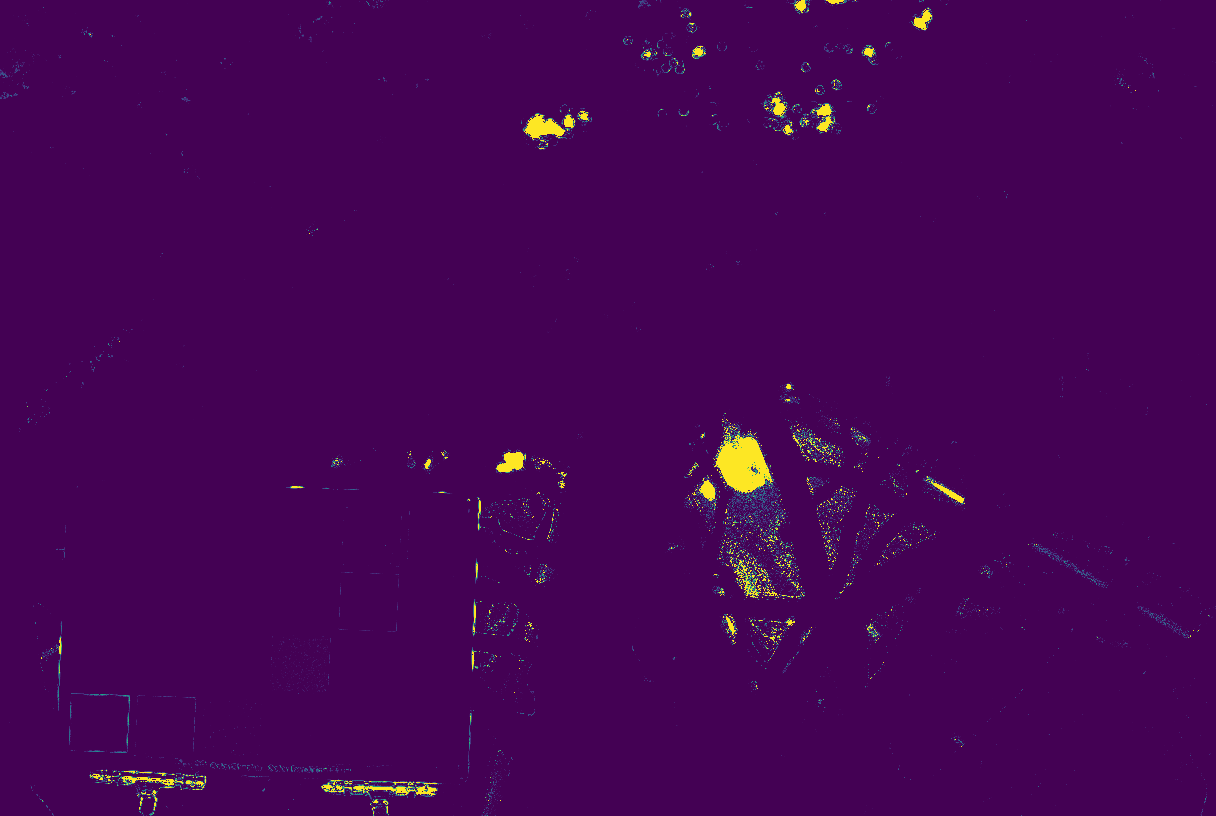}}\includegraphics[width=0.23\linewidth]{figures_arxiv/supp/sony/SonyA57_0137_st4_err_rang.png}\llap{\raisebox{\dimen0-7pt}{\setlength{\fboxsep}{2pt}\colorbox{white}{\scriptsize	 PSNR: 34.07dB}}} & \settototalheight{\dimen0}{\includegraphics[width=0.23\linewidth]{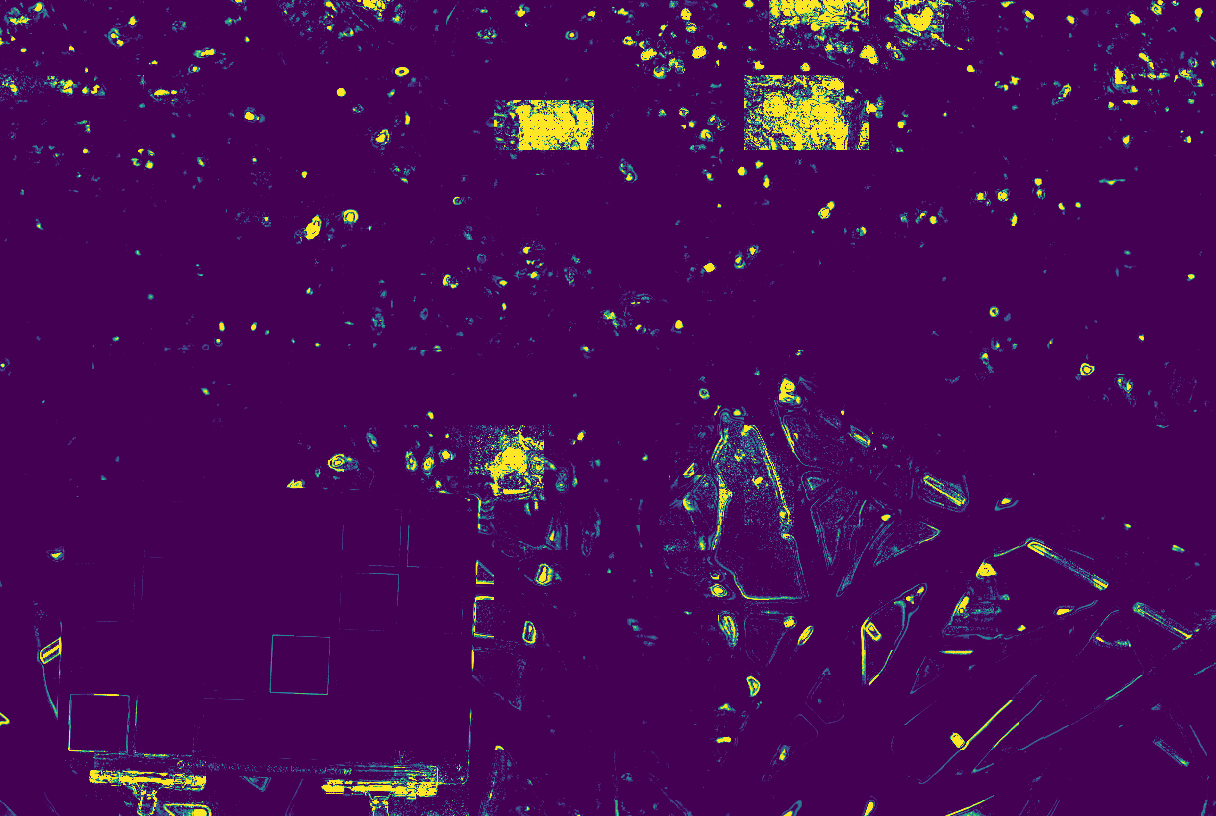}}\includegraphics[width=0.23\linewidth]{figures_arxiv/supp/sony/SonyA57_0137_st4_err_wacv.png}\llap{\raisebox{\dimen0-7pt}{\setlength{\fboxsep}{2pt}\colorbox{white}{\scriptsize	 PSNR: 43.63dB}}} & \settototalheight{\dimen0}{\includegraphics[width=0.23\linewidth]{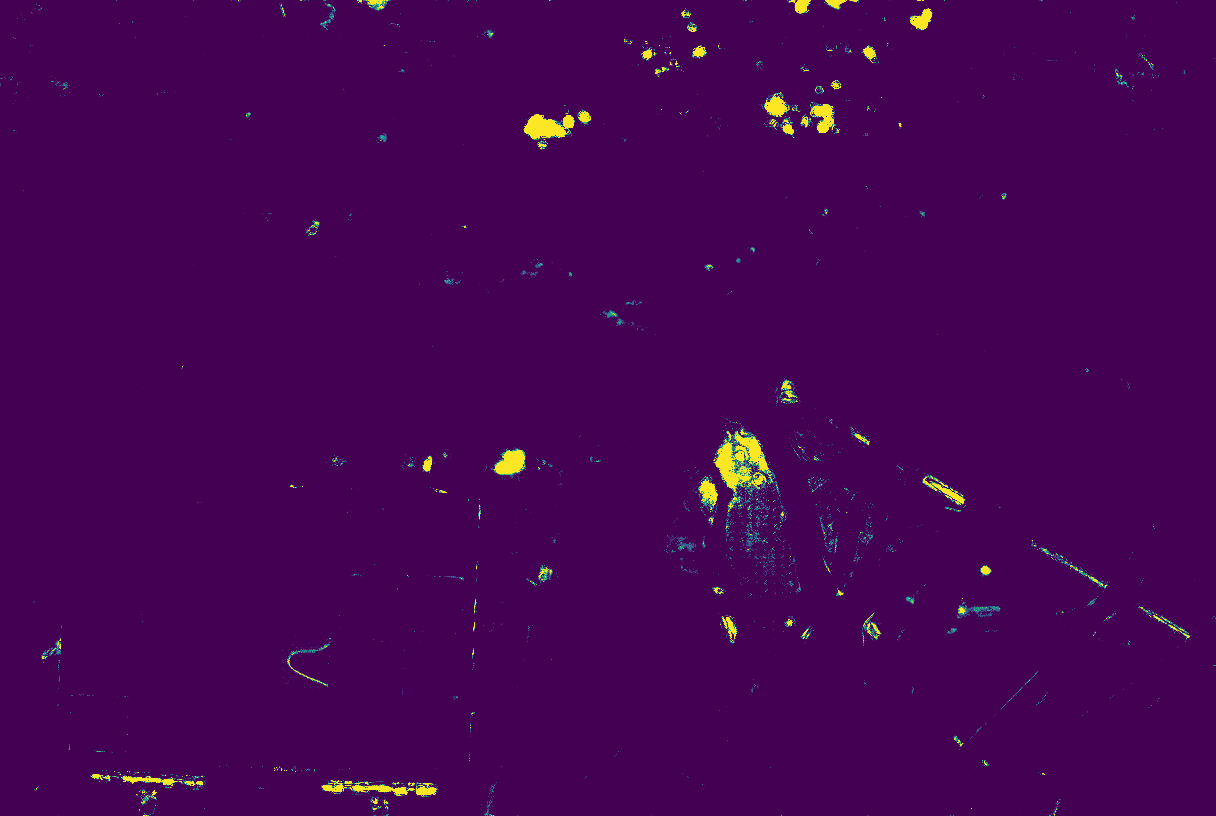}}\includegraphics[width=0.23\linewidth]{figures_arxiv/supp/sony/0000018_err.png}\llap{\raisebox{\dimen0-7pt}{\setlength{\fboxsep}{2pt}\colorbox{white}{\scriptsize	 PSNR: 47.25dB}}} & \includegraphics[width=0.025\linewidth]{figures_arxiv/colorbar.pdf} \\

        \includegraphics[width=0.23\linewidth]{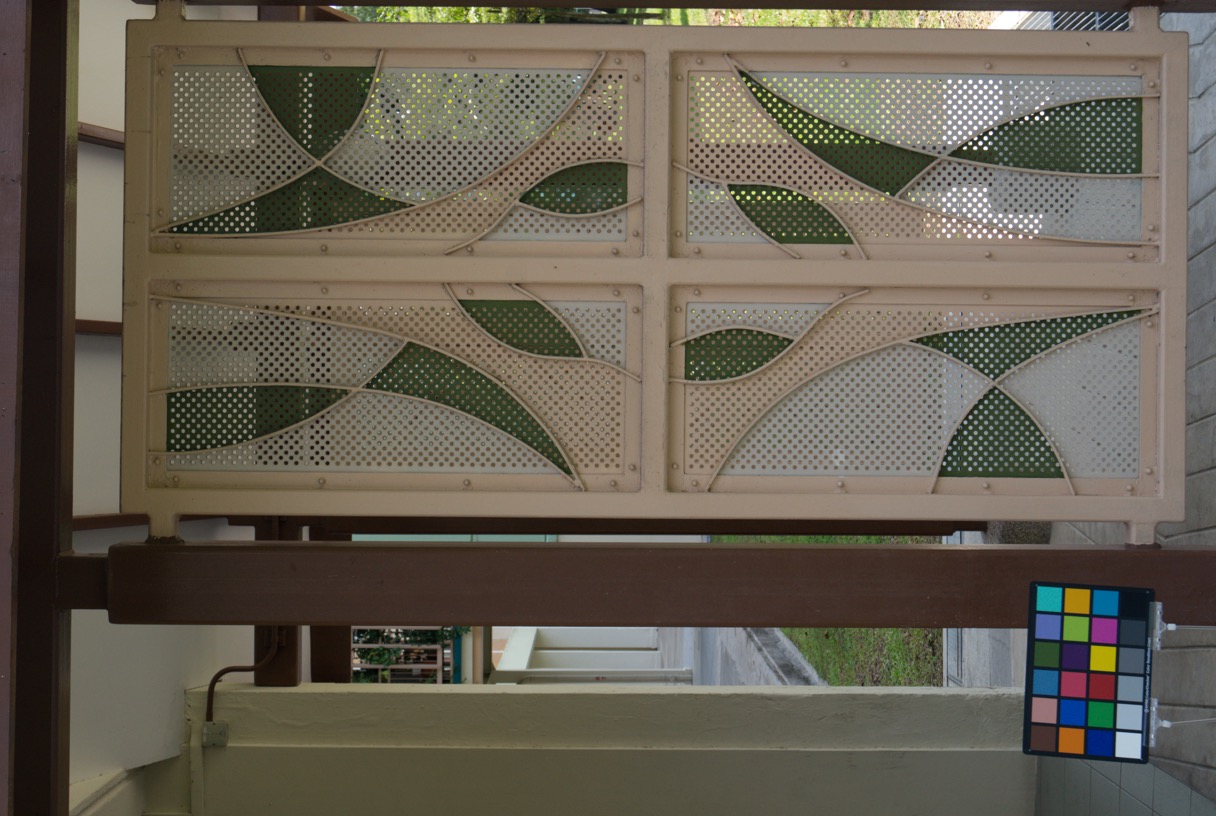} & \settototalheight{\dimen0}{\includegraphics[width=0.23\linewidth]{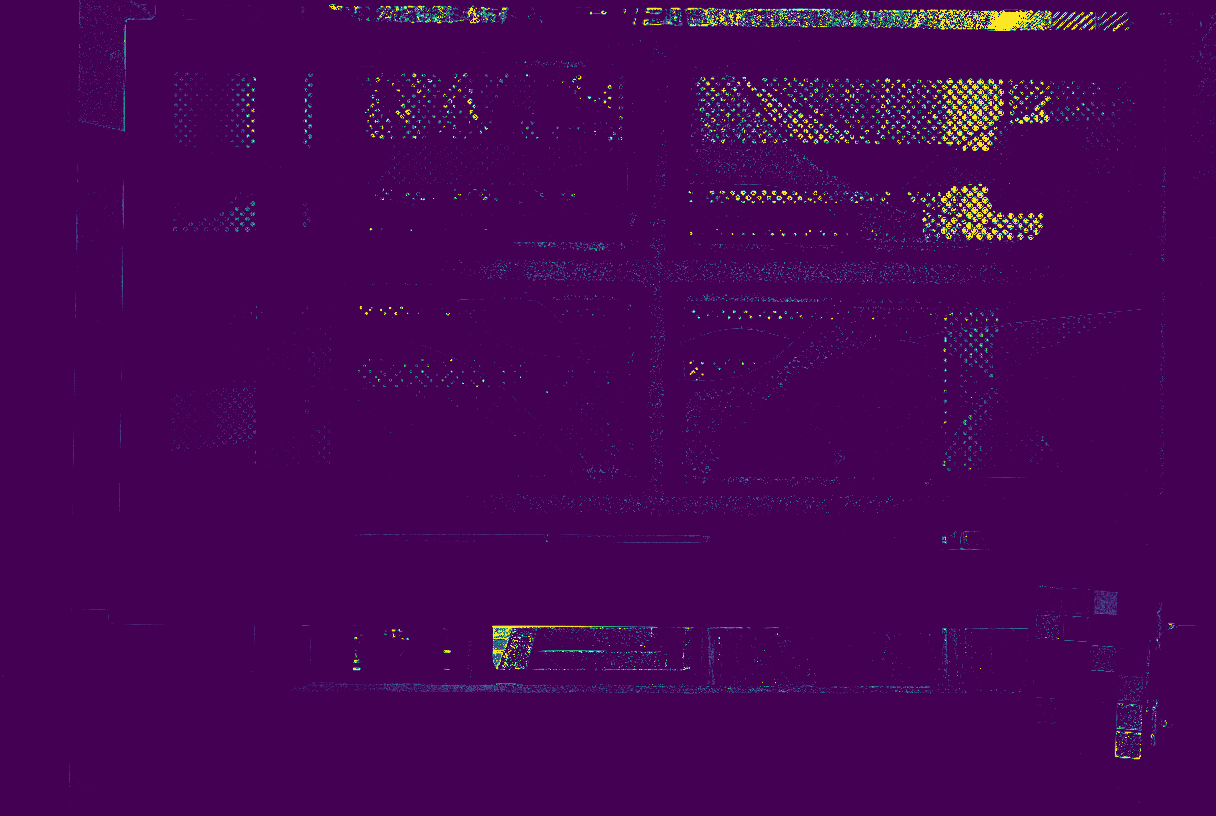}}\includegraphics[width=0.23\linewidth]{figures_arxiv/supp/sony/SonyA57_0141_st4_err_rang.png}\llap{\raisebox{\dimen0-7pt}{\setlength{\fboxsep}{2pt}\colorbox{white}{\scriptsize	 PSNR: 47.67dB}}} & \settototalheight{\dimen0}{\includegraphics[width=0.23\linewidth]{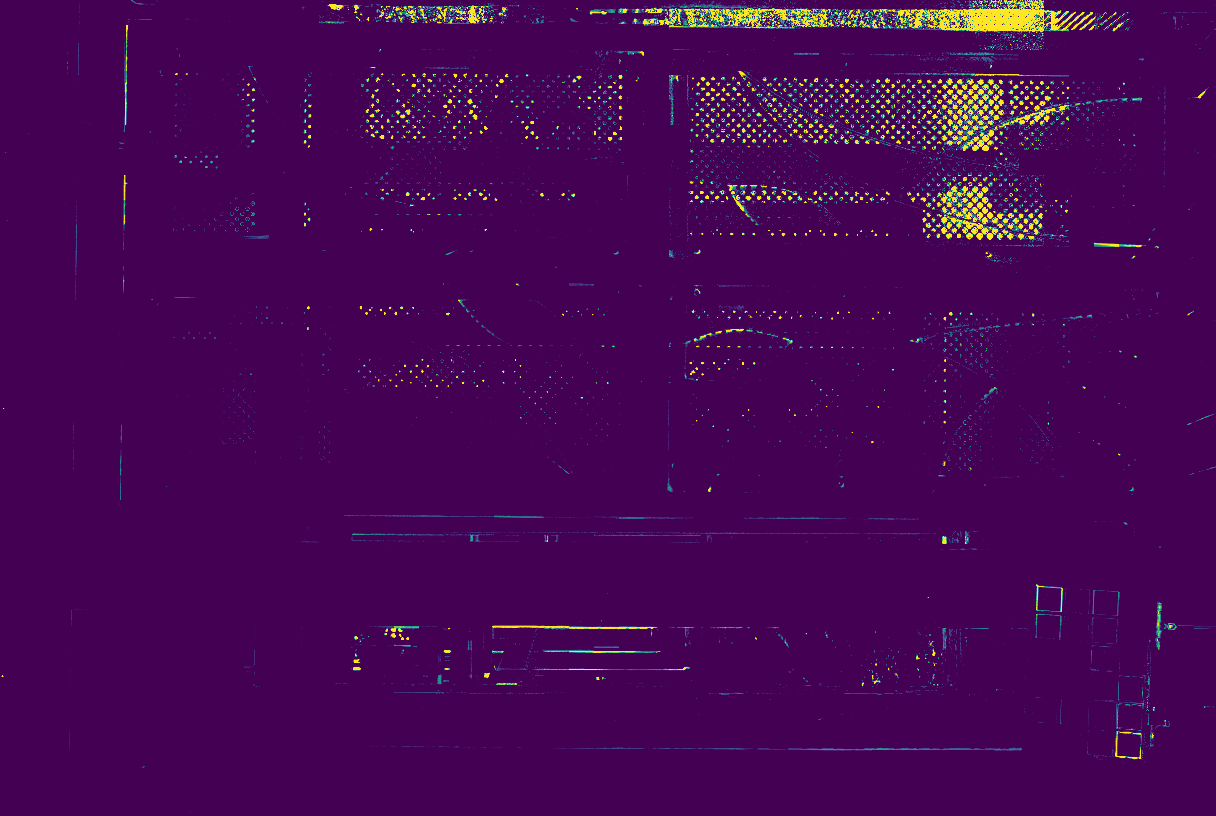}}\includegraphics[width=0.23\linewidth]{figures_arxiv/supp/sony/SonyA57_0141_st4_err_wacv.png}\llap{\raisebox{\dimen0-7pt}{\setlength{\fboxsep}{2pt}\colorbox{white}{\scriptsize	 PSNR: 48.75dB}}} & \settototalheight{\dimen0}{\includegraphics[width=0.23\linewidth]{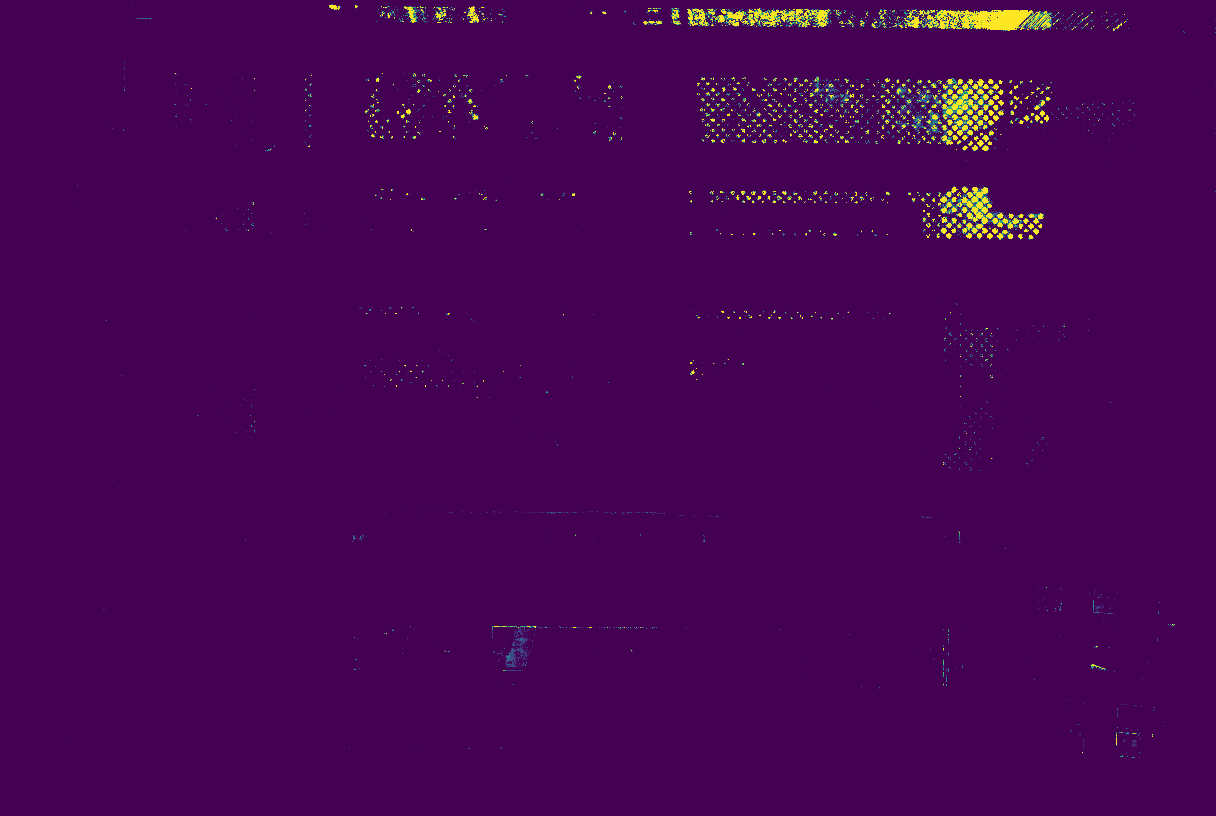}}\includegraphics[width=0.23\linewidth]{figures_arxiv/supp/sony/0000020_err.png}\llap{\raisebox{\dimen0-7pt}{\setlength{\fboxsep}{2pt}\colorbox{white}{\scriptsize	 PSNR: 50.93dB}}} & \includegraphics[width=0.025\linewidth]{figures_arxiv/colorbar.pdf} \\

        {\small Input} & {\small RIR~\cite{rang}} & {\small SAM~\cite{wacv}} & {\small Ours + fine-tuning} & \\
    \end{tabular}
    \caption{Qualitative comparison on Sony SLT-A57.}
    \label{fig:supp_sony}
\end{figure*}
\input{figures_arxiv/visualization_supp}

\end{appendices}

\end{document}